\documentclass[10pt,twocolumn,letterpaper]{article}
\pdfoutput=1

\usepackage{iccv}
\usepackage{times}
\usepackage{epsfig}
\usepackage{graphicx}
\usepackage{amsmath}
\usepackage{amssymb}
\usepackage{enumerate}
\usepackage{docmute}

\usepackage{caption}
\usepackage{subcaption}
\usepackage{duckuments}
\usepackage{booktabs}
\usepackage{tikz}
\usepackage{comment}
\usepackage[inline]{enumitem}

\usepackage{cuted}
\usepackage{capt-of}

\usepackage{multirow}

\usepackage{xcolor}
\usepackage{color, colortbl}
\usepackage{graphbox}

\usepackage[percent]{overpic}

\definecolor{A}{RGB}{255,0,0}
\definecolor{B}{RGB}{0, 0, 255}
\definecolor{star}{RGB}{255, 0, 255}

\definecolor{aabbcc}{rgb}{0.8,0.4,0.8}
\definecolor{aabbdd}{rgb}{0.2,0.2,0}
\definecolor{aabbee}{rgb}{0.2,0.6,0.8}
\definecolor{aabbff}{rgb}{0.2,0.0,0.4}
\definecolor{aabbgg}{rgb}{0.4,0.8,0}
\definecolor{aaccdd}{rgb}{0.4,0.8,1}
\definecolor{aaccee}{rgb}{1,0.6,0.8}
\definecolor{aaccff}{RGB}{255,204,51}
\definecolor{aaccgg}{RGB}{125,125,255}
\definecolor{aaddee}{RGB}{255,127,0}
\definecolor{aaddgg}{RGB}{150,150,150}
\definecolor{aaddff}{RGB}{30,30,30}

\usepackage{tkz-euclide}
\usepackage{tikz}
\usepackage{pgfplots}
\pgfplotsset{compat=1.14}
\usetikzlibrary{positioning,calc,fadings,backgrounds,fit,3d,shapes.misc,shapes.geometric}
\usetikzlibrary{matrix}

\usepackage{algorithm}
\usepackage{algpseudocode}

\usepackage[pagebackref=true,breaklinks=true,letterpaper=true,colorlinks,bookmarks=false]{hyperref}

\usepackage[capitalize]{cleveref}
\crefname{section}{Sec.}{Secs.}
\crefname{section}{Section}{Sections}
\crefname{table}{Table}{Tables}
\crefname{table}{Tab.}{Tabs.}

\iccvfinalcopy 

\def\iccvPaperID{9189} 
\def\httilde{\mbox{\tt\raisebox{-.5ex}{\symbol{126}}}}

\ificcvfinal\fi

\begin{document}

\title{MDP: A Generalized Framework for Text-Guided Image Editing by Manipulating the Diffusion Path}

\author{Qian Wang\\
KAUST\\
{\tt\small qian.wang@kaust.edu.sa}
\and
Biao Zhang\\
KAUST\\
{\tt\small biao.zhang@kaust.edu.sa}
\and
Michael Birsak\\
KAUST\\
{\tt\small michael.birsak@kaust.edu.sa}
\and
Peter Wonka\\
KAUST\\
{\tt\small peter.wonka@kaust.edu.sa}
}

\maketitle
\ificcvfinal\thispagestyle{empty}\fi

\begin{strip}
    \vspace{-20pt}
    \centering
    \scriptsize
    \resizebox{\linewidth}{!}{%
        \setlength{\tabcolsep}{-0pt}
        \begin{tabular}{c@{\hskip 2pt}cc@{\hskip 1pt}cc@{\hskip 1pt}cc@{\hskip 1pt}cc}
            &  \multicolumn{2}{c}{``Cat'' to ``Dog''} & \multicolumn{2}{c}{``Red'' to ``Purple''} & \multicolumn{2}{c}{Remove ``Stuff''} & \multicolumn{2}{c}{Mix ``Teapot'' and ``Shoes''}\\
            Local Edits & \includegraphics[align=c, scale=0.11]{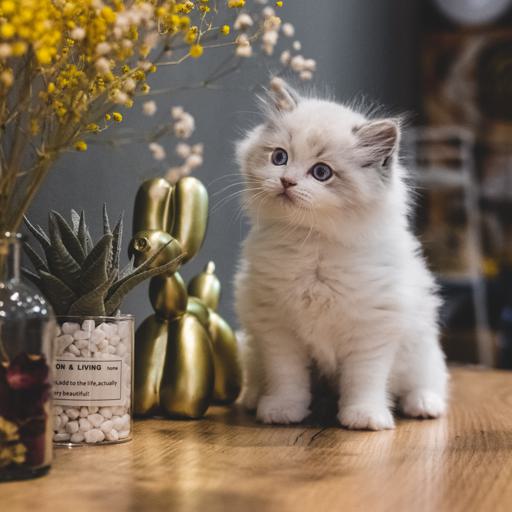} & \includegraphics[align=c, scale=0.11]{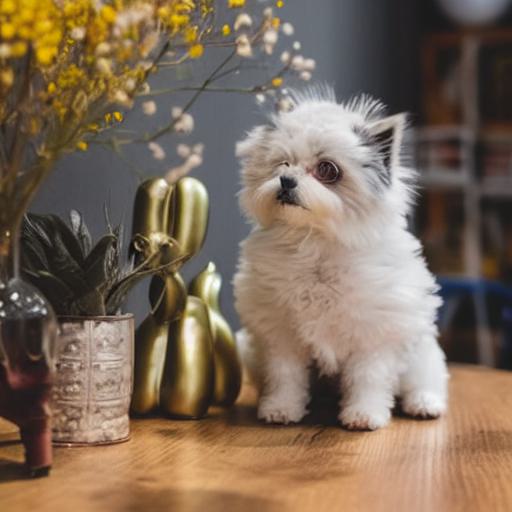} & \includegraphics[align=c, scale=0.11]{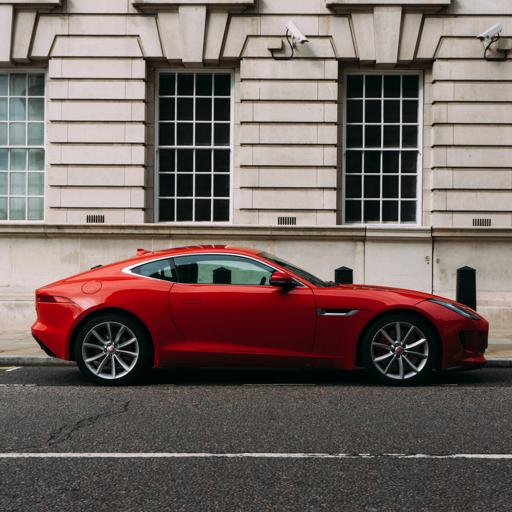} & \includegraphics[align=c, scale=0.11]{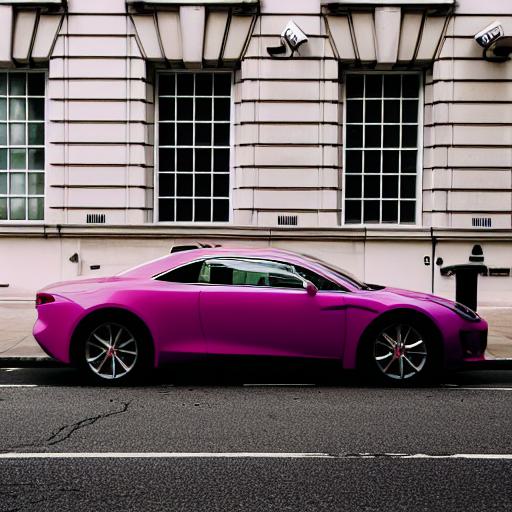} & \includegraphics[align=c, scale=0.11]{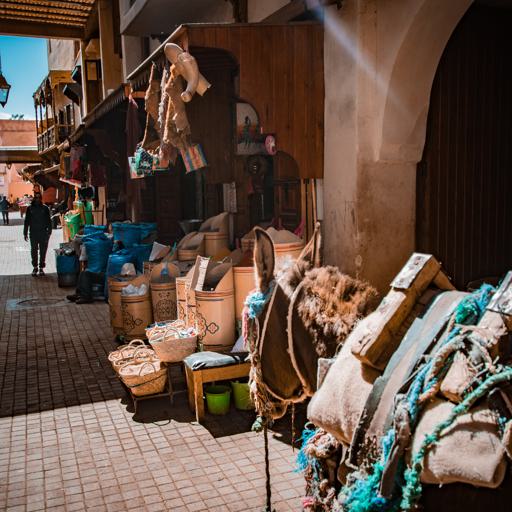} & \includegraphics[align=c, scale=0.11]{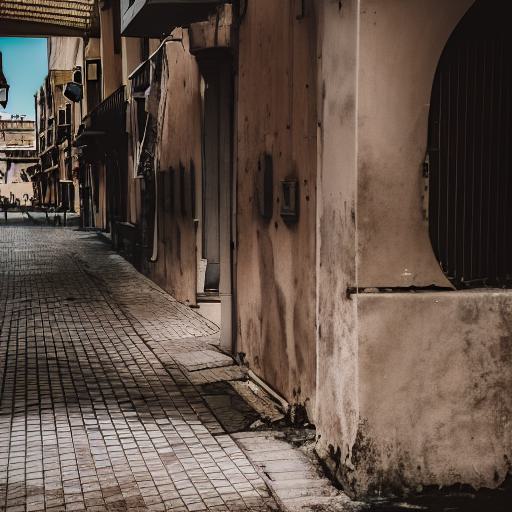} & \includegraphics[align=c, scale=0.11]{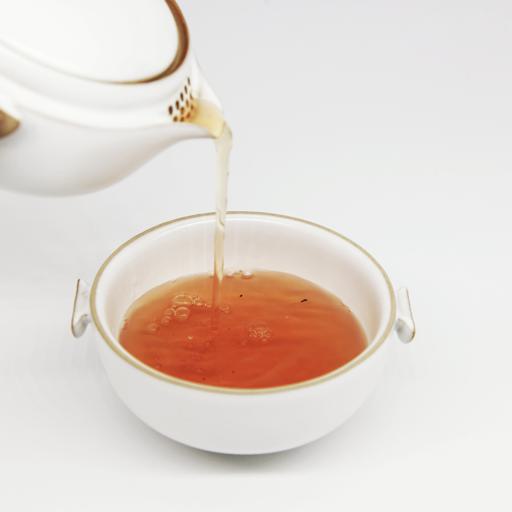} & \includegraphics[align=c, scale=0.11]{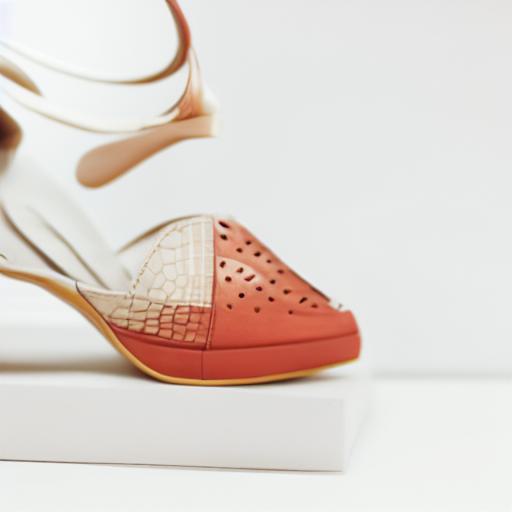}\\ [1cm]
            & \multicolumn{2}{c}{``Table'' to ``Galaxy''} & \multicolumn{2}{c}{``Fall'' to ``Winter''} & \multicolumn{2}{c}{``Painting'' to ``Children's drawing''} & \multicolumn{2}{c}{Stylize ``Blanket''} \\
            Global Edits & \includegraphics[align=c, scale=0.11]{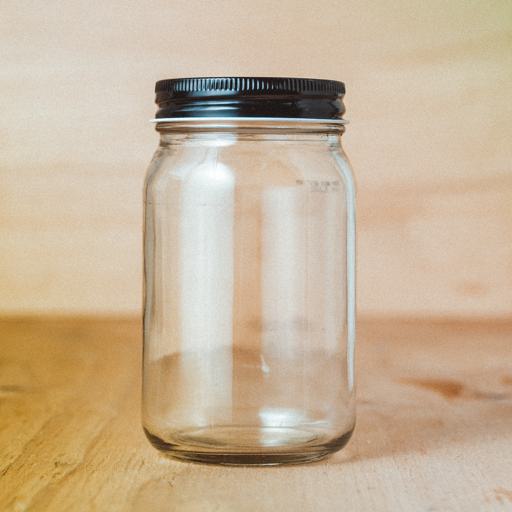} & \includegraphics[align=c, scale=0.11]{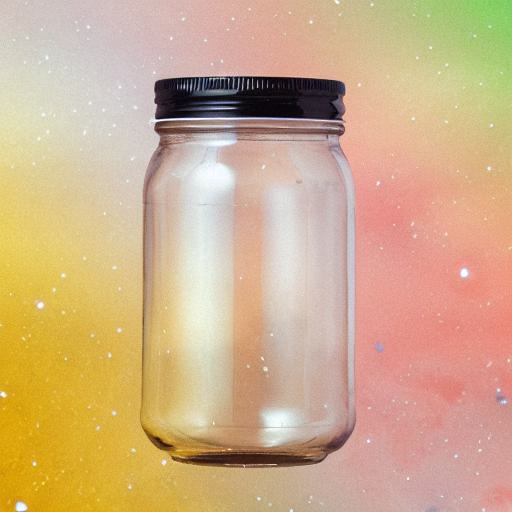} & \includegraphics[align=c, scale=0.11]{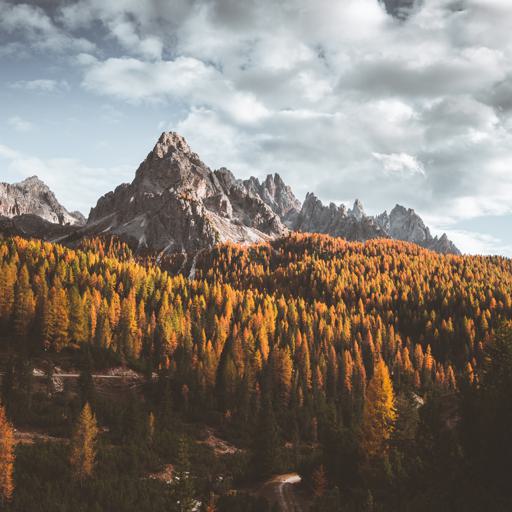} & \includegraphics[align=c, scale=0.11]{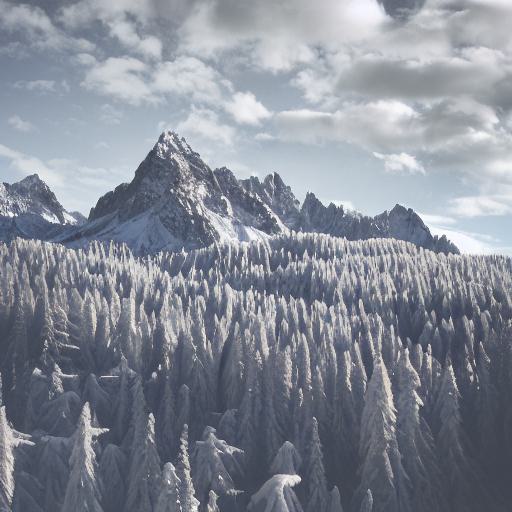} & \includegraphics[align=c, scale=0.11]{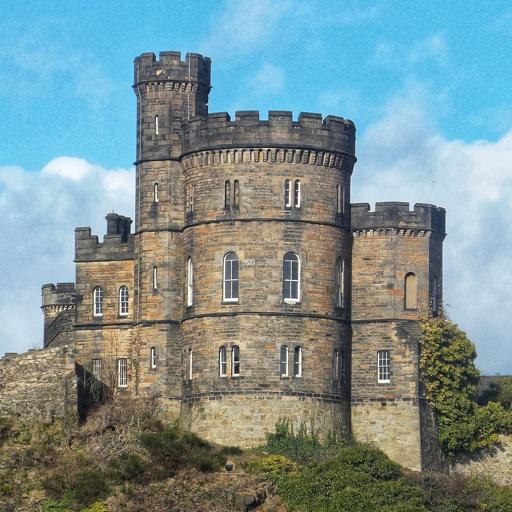} & \includegraphics[align=c, scale=0.11]{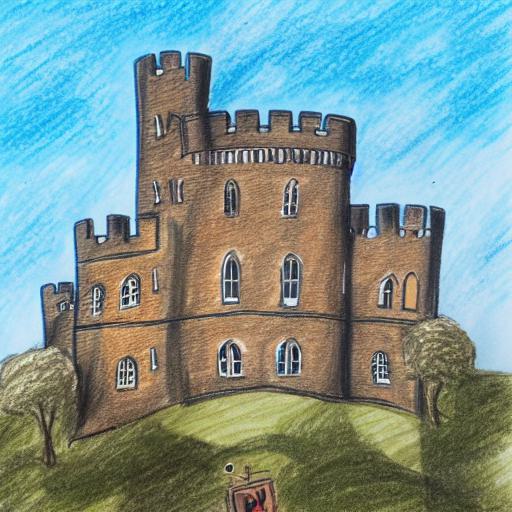} & \includegraphics[align=c, scale=0.11]{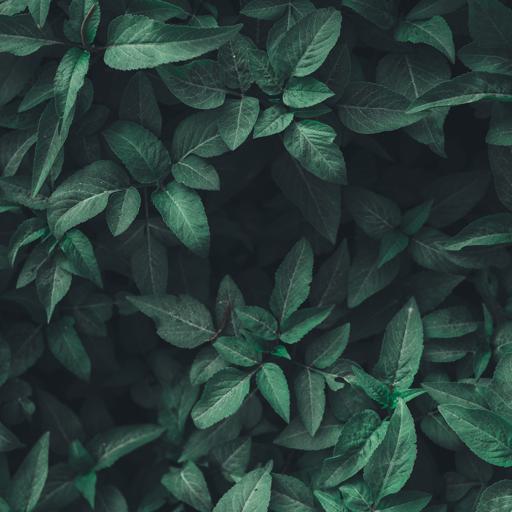} & \includegraphics[align=c, scale=0.11]{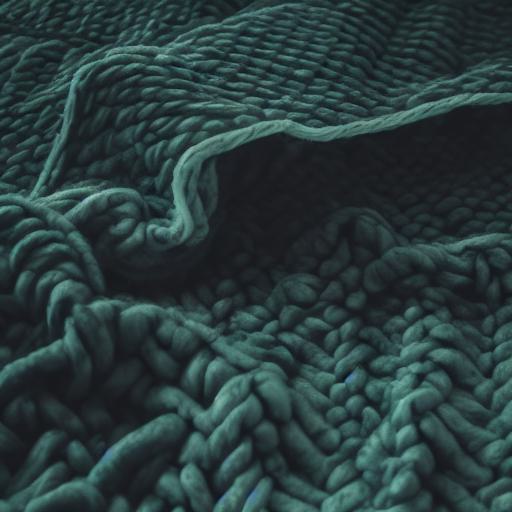}
        \end{tabular}
    }
    \vspace{-8pt}
    \captionof{figure}{Our proposed manipulation of predicted noise, which fits into our proposed editing framework \textbf{MDP}, can do multiple local and global edits without training or finetuning.
\label{fig:teaser}}
\end{strip}

\begin{abstract}
\vspace{-5mm}
Image generation using diffusion can be controlled in multiple ways. 
In this paper, we systematically analyze the equations of modern generative diffusion networks to propose a framework, called MDP, that explains the design space of suitable manipulations. We identify 5 different manipulations, including intermediate latent, conditional embedding, cross attention maps, guidance, and predicted noise. We analyze the corresponding parameters of these manipulations and the manipulation schedule. We show that some previous editing methods fit nicely into our framework. 
Particularly, we identified one specific configuration as a new type of control by manipulating the predicted noise, which can perform higher-quality edits than previous work for a variety of local and global edits. We provide the code in \href{https://github.com/QianWangX/MDP-Diffusion}{https://github.com/QianWangX/MDP-Diffusion}.
\end{abstract}

\section{Introduction}

Previously, image editing using GANs has achieved great success \cite{isola2017imagegan,zhu2017cyclegan,choi2018stargan,wang2018highres,huang2018multimodal,kim2017discogan,bau2020semantic}. As large-scale text-to-image datasets became available, text-guided image synthesis and editing has obtained increasing attention \cite{ramesh2021dalle,crowson2022vqganclip,ding2021cogview,ding2022cogview2,gafni2022make-a-scene}. 
Generative diffusion models ~\cite{ho2020ddpm,rombach2022latentdiffusion,saharia2022imagen,nichol2022glide,ramesh2022dalle2} are also a powerful tool for multiple image processing tasks, such as inpainting ~\cite{lugmayr2022repaint,choi2021ilvr,li2022sdm,xie2022smartbrush}, style transfer~\cite{kwon2022disentangled_style_content,bansal2023universal-guidance}, text-guided image editing\cite{brooks2022instructpix2pix,kawar2022imagic,hertz2022prompt2prompt,tumanyan2022plug-and-play}, map-to-image translation \cite{voynov2022sketch-diffusion,avrahami2022spatext} and segmentation \cite{burgert2022peekaboo,baranchuk2021label-efficient}.

We are interested in simple and effective image editing methods that do not require retraining, fine-tuning, or training of an auxiliary network. Specifically, for text-guided image editing, we can leverage pre-trained diffusion models to perform editing tasks on real images. An arbitrary real image can first be embedded into the diffusion latent space to obtain an initial noise tensor \cite{wallace2022edict,mokady2022null}. Then, by manipulating the diffusion generation path starting from this initial noise tensor, we can edit the input image.

Given an input image, we perform edits guided by a condition (\eg, a text prompt or a class label) while keeping the overall layout from the input image. We can see this editing task as combining the layout from the input image with new semantics from the condition.
Since diffusion models generate images progressively, the layout of an image is generated during the early timesteps of the denoising process, while the semantics, \ie the texture, color, details, are generated during later steps \cite{hoogeboom2023simple-diffusion,liew2022magicmix}. Therefore, it is intuitive that each timestep should be controlled separately.

For each denoising step, there are several components that we can modify during the diffusion process. Intuitively, we want to denoise using information from the input image, for which we want to preserve the layout, in the early stages, and then the components of the new condition in the later stages. At this point, two questions arise: 
\begin{enumerate*}[label=(\arabic*)]
\item \textit{Which component of the generation process should be modified?}
\item \textit{For which denoising steps should the modification take place?}
\end{enumerate*}

In order to tackle the problems, we analyze the equations of modern generative diffusion networks to find out which variables could be manipulated and identified 5 different manipulations that are suitable: intermediate latent, conditional embedding, cross attention maps, guidance and predicted noise. We take the manipulation of attention maps from the seminal paper Prompt-to-Prompt (P2P) \cite{hertz2022prompt2prompt}. As result of the analysis, we introduce a generalized editing framework, \textbf{MDP}, that contains these 5 different manipulations, their corresponding parameters, and a manipulation schedule for them.



In this design space, we identified one particular configuration, MDP-$\epsilon_t$, that yields very good results for a variety of local and global image editing problems. The results from MDP-$\epsilon_t$ are more consistent in quality than previous work (P2P) and other possible configurations. We advocate for the use of MDP-$\epsilon_t$ as a new method for diffusion-based image editing.




Our work makes the following contributions:
\begin{itemize}
\item We present a framework for a generalized design space to manipulate the diffusion path that includes multiple existing methods as special cases.
\item We analyze the design space to find good configurations suitable for editing applications.
\item We propose a new solution by manipulating predicted noise that can solve practical local and global editing problems better than other configurations and previous work.
\end{itemize}

\begin{figure*}[htb]
    \centering
    \resizebox{\linewidth}{!}{%
    \setlength{\tabcolsep}{1pt}
    \begin{tabular}{cccccccccc}
        50 & 40 & 30 & 20 & 10 & 5 & 3 & 2 & 1 & 0\\
        \includegraphics[scale=0.15]{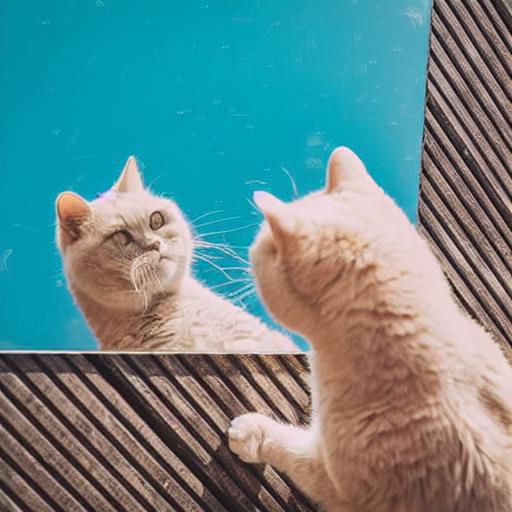} &
        \includegraphics[scale=0.15]{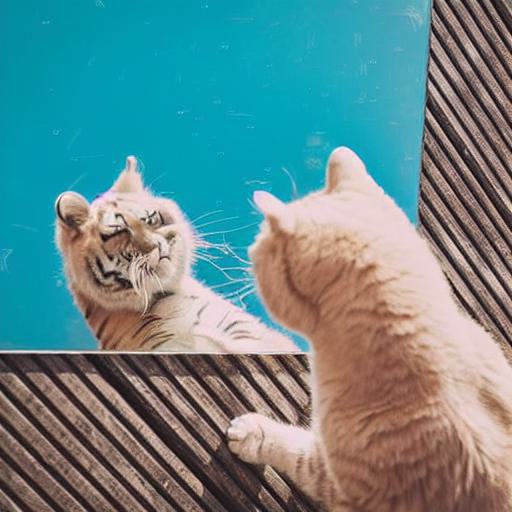} &
        \includegraphics[scale=0.15]{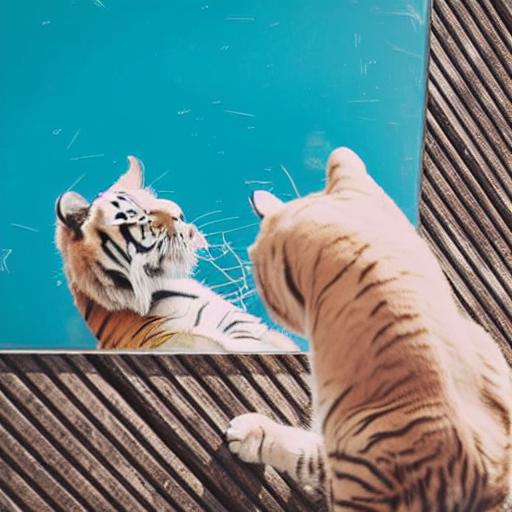} &
        \includegraphics[scale=0.15]{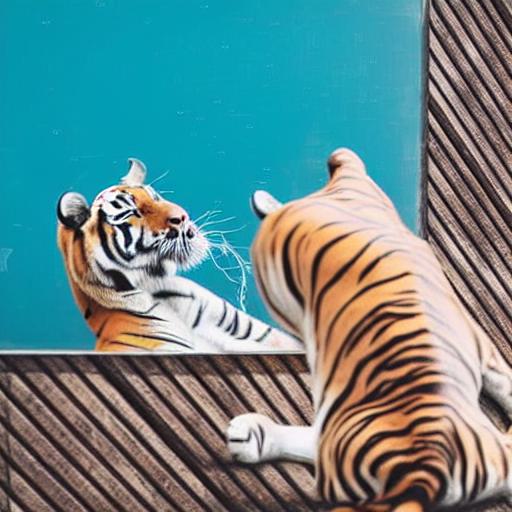} &
        \includegraphics[scale=0.15]{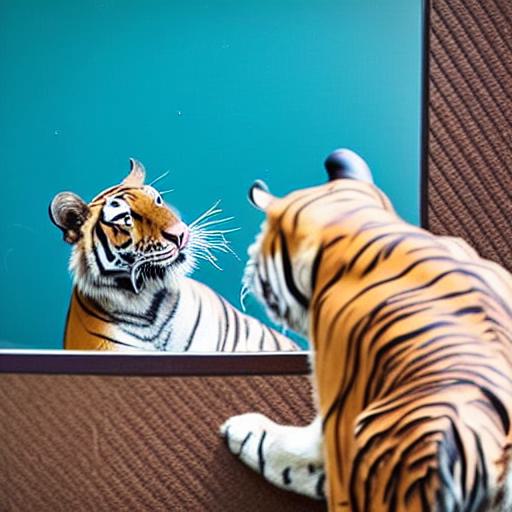} &
        \includegraphics[scale=0.15]{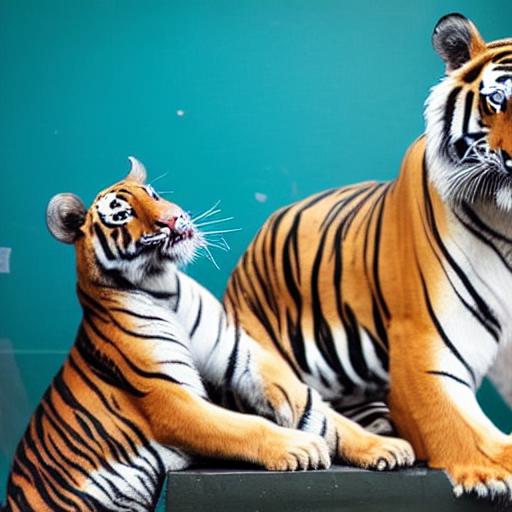} &
        \includegraphics[scale=0.15]{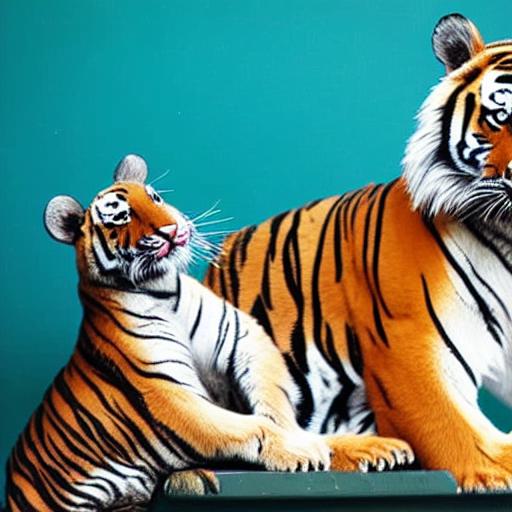} &
        \includegraphics[scale=0.15]{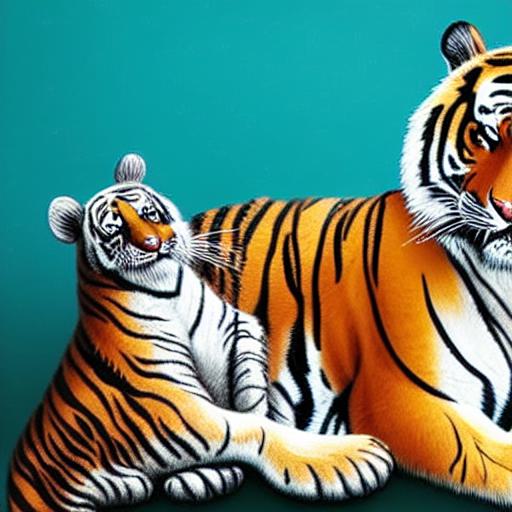} &
        \includegraphics[scale=0.15]{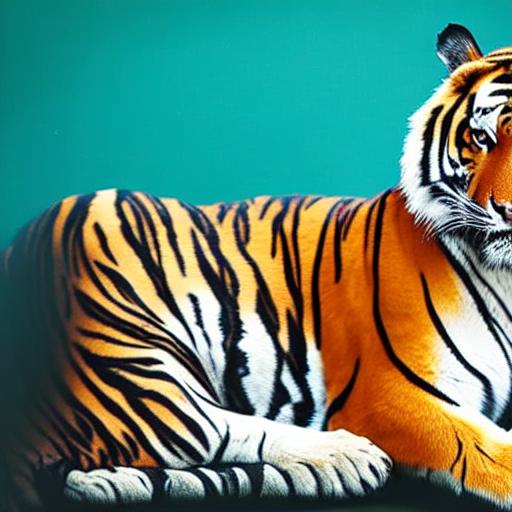} &
        \includegraphics[scale=0.15]{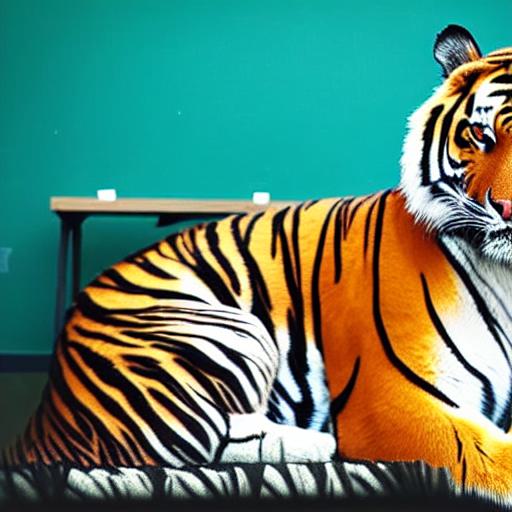} \\
    \end{tabular}
    }
    \vspace{-8pt}
    \caption{Assume the total number of sampling steps $T=50$. The number on top of each image shows the number of diffusion generation steps for which we use $\mathbf c^{(A)}$ ``\textit{Photo of a cat sitting next to a mirror}'' to invert and then denoise. For the remaining diffusion generation steps, we switch to $\mathbf c^{(B)}$ ``\textit{Photo of a tiger sitting next to a mirror}'' to denoise. The leftmost image is equivalent to generating an image using condition A. The rightmost image is equivalent to generating an image using condition B. If we change condition in the early stage (\eg, 30 and 20), we can preserve the layout of $\mathbf c^{(A)}$ while including the semantics from $\mathbf c^{(B)}$. If we switch at a later stage (\eg, 5, 3, 2, 1), the layout of $\mathbf c^{(A)}$ is hardly maintained in the result.}
    \label{fig:diffusion-layout-details}
\end{figure*}
\section{Related Work}
\paragraph{Image diffusion models.}


In recent years, the interest in deep generative models has gained strong momentum and many different methods have been proposed to create high-quality samples from a magnitude of different data domains. One kind of architecture that currently stands out are diffusion models \cite{ho2020ddpm, dhariwal2021beatgans, karras2022edm}.


In order to obtain better control over the denoising process and the generated content, several works proposed text-conditioned image synthesis using diffusion models and either CLIP guidance or classifier-free guidance \cite{nichol2022glide, ramesh2022dalle2, saharia2022imagen}.

Several approaches focus on different optimization strategies and either propose to generalize DDPMs via a class of non-Markovian diffusion processes to trade off computation for sample quality \cite{song2021ddim} or to do the diffusion process in the latent space of a pretrained autoencoder \cite{rombach2022latentdiffusion} instead of the RGB pixel space.

Others proposed to use an ensemble of expert denoising networks \cite{balaji2022ediff-i} or to compose an image using both global and local representative factors \cite{huang2023composer}.


\paragraph{Image editing with diffusion models.}
InstructPix2Pix~\cite{brooks2022instructpix2pix} can edit images by following user instructions. Paint by Example \cite{yang2022paint_by_example} allows users to replace an object with a conditional example image. ControlNet \cite{zhang2023control_net} trains a task-specific condition network on top of a pre-trained diffusion model and supports various edits based on a specific training set. All these works usually require re-designing of the model, collecting training data and a long training time to do image translation.

UniTune \cite{kawar2022imagic} and Imagic \cite{kawar2022imagic} can do realistic text-guided image editing by either just finetuning the diffusion models or the models and text embeddings together. However, because of the finetuning process, it takes several minutes or more to generate a single image. 

Recently, there are many interesting works that only utilize a pretrained diffusion model to do text-guided image editing without any training or finetuning \cite{meng2021sdedit,radford2021clip,couairon2022diffedit,liew2022magicmix,hertz2022prompt2prompt,park2022shape-guided,parmar2023pix2pix_zero,tumanyan2022plug-and-play}. 
DiffEdit \cite{couairon2022diffedit} automatically generates a mask by computing the differences between the noisy latent generated from an input image and a conditional text prompt. However, because of its inpainting nature, DiffEdit cannot perform global editing operations like converting a photo to an oil painting. MagicMix \cite{liew2022magicmix} interpolates the noisy input latent and the denoised latent to mix the objects that have large semantic differences. Prompt-to-Prompt \cite{hertz2022prompt2prompt} proposes to manipulate the cross-attention maps corresponding to the changes between the input and guided text prompt. However, Prompt-to-Prompt always requires a prompt together with an input image. 
Concurrent work pix2pix-zero \cite{parmar2023pix2pix_zero} computes an editing direction to edit an image based on a provided prompt, combined with a cross-attention mechanism to preserve the layout of the input image.

Compared to prior works, we propose a generalization that includes multiple previous approaches as special case.  
We also highlight a novel manipulation, that is suitable to better perform a wide range of local and global edits than previous methods. Our edits do not require masks or prompting for an input image. Our method is purely based on a pre-trained diffusion model and does not require finetuning.

\section{Method}
\subsection{Preliminaries}
\paragraph{Denoising diffusion probabilistic models.}
To train a conditional generative diffusion models, we consider the following objective,
\begin{equation}
    \min_\theta\mathbb{E}_{\mathbf{x}_{0}\sim \mathcal{D}, \boldsymbol\epsilon\sim\mathcal{N}(\mathbf{0}, \mathbf{I}),t\sim U(1, T)}\left\|\boldsymbol\epsilon-\boldsymbol\epsilon_\theta(\mathbf{x}_t, \mathbf{c}, t)\right\|^2,
\end{equation}
where $\mathcal{D}$ is an image dataset (could be raw image pixels or latents obtained from an image autoencoder), $\mathbf{x}_t$ is a noised version of the image $\mathbf{x}_0$, $\mathbf{c}$ is a conditional embedding (\eg, text, class label, image) and $\boldsymbol\epsilon_\theta(\cdot, \cdot, \cdot)$ is a neural network parameterized by $\theta$. After training, we can sample a new image given condition $\mathbf{c}$ with the commonly used DDIM sampler,
\begin{equation}\label{eq:ddim-sampler}
    \begin{aligned}
        \mathbf{x}_{t-1} =& \mathrm{DDIM}(\mathbf{x}_t, \boldsymbol\epsilon_{t}, t) \\
        =& \sqrt{\alpha_{t-1}}\cdot f_\theta(\mathbf{x}_t, \mathbf{c}, t) + \sqrt{1-\alpha_{t-1}}\cdot \boldsymbol\epsilon_\theta(\mathbf{x}_t, \mathbf{c}, t), \\
    \end{aligned}
\end{equation}
where $f_\theta(\mathbf{x}_t, \mathbf{c}, t) = \frac{\mathbf{x}_t - \sqrt{1-\alpha_t}\cdot\boldsymbol\epsilon_\theta(\mathbf{x}_t, \mathbf{c}, t)}{\sqrt{\alpha_t}}$, $\boldsymbol\epsilon_t=\boldsymbol\epsilon_\theta(\mathbf{x}_t, \mathbf{c}, t)$, and $\alpha_t$ is a noise schedule factor as in DDIM. The sampling is to iteratively apply the above equation by giving an initial noise $\mathbf{x}_T\sim\mathcal{N}(\mathbf{0}, \mathbf{I})$. For brevity, we only consider DDIM as our sampler. 

\paragraph{Image editing using pre-trained diffusion models.}
Our generalized editing framework MDP edits an image in its latent space also called noise space. Given an image $\mathbf{x}_0$ and the corresponding initial noise $\mathbf{x}_T$, we modify $\mathbf{x}_T$ or the generating process and our aim is to obtain a different $\mathbf{x}_0^{\star}$ which suits our needs. We summarize the symbols used in our later discussions here. One step of diffusion sampling can be represented as
    $\begin{cases}
        \boldsymbol\epsilon_t = \boldsymbol\epsilon_\theta(\mathbf{x}_t, \mathbf{c}, t), \\
        \mathbf{x}_t = \mathrm{DDIM}(\mathbf{x}_t, \boldsymbol\epsilon_t, t).
    \end{cases}$
We represent the intermediate results of the generating process at timestep $t$ by $\mathbf{x}_t=\mathrm{Gen}(\mathbf{x}_T, \mathbf{c}, t),$ given an initial noise $\mathbf{x}_T$ and a condition $\mathbf{c}$. We also represent all the intermediate outputs by
$\left\{\left(\mathbf{x}_t, \boldsymbol\epsilon_t\right)\right\}_{t=[T,\dots, 0]}=\mathrm{GenPath}\left(\mathbf{x}_T, \mathbf{c}, t\right).$

\subsection{Design Space for Manipulating the Diffusion Path}
Simply switching to another condition during denoising is a very straightforward way to add new semantics to the input image when doing text-guided image editing. We perform a simple experiment in \cref{fig:diffusion-layout-details} to illustrate that the early timesteps in the diffusion process contribute to the layout, while the later timesteps are adding semantics and details to the image. Our goal is to systematically analyze the design space for manipulating the diffusion path over time to perform these types of edits. As result of our analysis, we propose \textbf{MDP} as generalized framework for text-guided image editing and show how concurrent and previous methods fit our framework as special cases. A summary can be found in Table~\ref{tab:manip-design}. Each manipulating operation is formalized as an operator, including 
$$
\begin{aligned}
\mathcal{O}=
\{
&
\mathcal{O}_{\mathrm{IDI}}, \mathcal{O}_{\mathrm{IDM}}, \mathcal{O}_{\mathrm{CEI}}, 
\mathcal{O}_{\mathrm{CAM}}, \\&\mathcal{O}_{\mathrm{G}}, \mathcal{O}_{\mathrm{PNI}}, \mathcal{O}_{\mathrm{PNM}}\}.
\end{aligned}
$$
All manipulations in our framework are time-dependent and have parameters, \eg interpolation parameters. MDP allows the user to specify different types of manipulations and different manipulation parameters at each timestep, which we call the manipulation schedule.

Assume we have an image $\mathbf{x}_0^{(A)}$ as input, along with a condition $\mathbf{c}^{(A)}$ that is used to generate $\mathbf{x}_0^{(A)}$ and a new condition $\mathbf{c}^{(B)}$. We first invert the image $\mathbf{x}_0^{(A)}$ to get an initial noise $\mathbf{x}_T$. We identify the following types of manipulations:

\begin{figure}[htb]
    \centering
    \includegraphics[width=\linewidth]{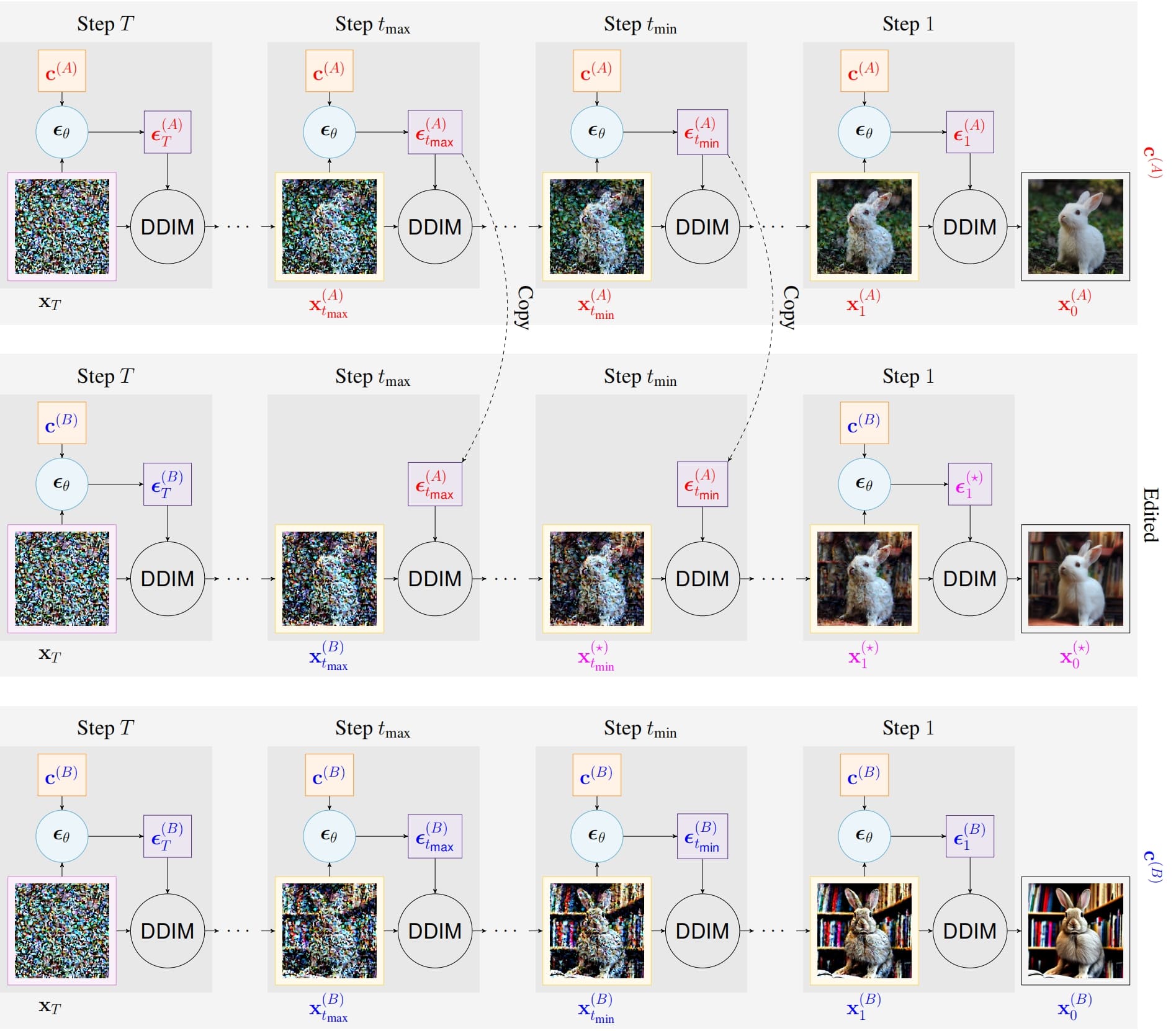}
    \caption{Predicted Noise Manipulation. The top branch is inverted from a real image given condition $\textcolor{A}{\mathbf{c}^{(A)}}$ ``\textit{Photo of a rabbit on the grass}''. The bottom branch is generated using condition $\textcolor{B}{\mathbf{c}^{(B)}}$ ``\textit{Photo of a rabbit in a library}''. We copy the predicted noise from step $t_{max}$ to $t_{min}$ of the top branch, then use $\textcolor{B}{\mathbf{c}^{(B)}}$ to denoise and generate the middle images.}
    \label{fig:pred-noise-manip}
\end{figure}
\paragraph{Intermediate denoised output.}
Starting from $\mathbf{x}_T$, we can generate image $\mathbf{x}_0^{(A)}$ using the new condition $\mathbf{x}_0^{(B)}$.
Assume we have all the intermediate latents for the generation of $\mathbf{x}_t^{(A)}$,
    $\left\{\left(\mathbf{x}_t^{(A)}\right)\right\}_{t=[T,\dots, 0]}=\mathrm{GenPath}\left(\mathbf{x}_T, \mathbf{c}^{(A)}, t\right).$
We denote the intermediate latents in the new path as $\mathbf{x}_t^{(\star)}$. We can modify the intermediate outputs $\mathbf{x}_t^{(A)}$ and $\mathbf{x}_t^{(\star)}$ with either linear interpolation or masking.
We can control the resulting image by changing the interpolation factor $w$ (or the binary mask $\mathbf{M}$),
\begin{equation}
\label{inter_denoised}
    \mathbf{x}^{(\star)}_t=\mathrm{lerp}\left(\mathbf{x}_t^{(A)}, \mathbf{x}_t^{(\star)}, \omega\right)
\end{equation}
or
\begin{equation}
    \mathbf{x}^{(\star)}_t=\mathbf{M} \odot \mathbf{x}_t^{(A)} + (1-\mathbf{M})\odot \mathbf{x}_t^{(\star)},
\end{equation}
where when we start this manipulation, $\mathbf{x}_t^{(\star)}=\mathbf{x}_t^{(B)}$. The strategy is used in MagicMix~\cite{liew2022magicmix} and DiffEdit~\cite{couairon2022diffedit}. 

\paragraph{Condition embeddings.}
Given two conditions $\mathbf{c}^{(A)}$ and $\mathbf{c}^{(B)}$, we can change $\mathbf{c}^{(A)}$ at timestep $t$,
\begin{equation}\label{eq:lerp-c}
    \mathbf{c}_t^{(\star)}=\mathrm{lerp}\left(\mathbf{c}^{(A)}, \mathbf{c}^{(B)}, \omega\right),
\end{equation}
where $\mathbf{c}_t^{(\star)}$ will be used as the new condition that embeds the information from both condition $\mathbf{c}^{(A)}$ and $\mathbf{c}^{(B)}$ .

\paragraph{Cross attention.}
Recent condition diffusion models inject condition information $\mathbf{c}$ by cross attending between $\mathbf{c}$ and image features. Prompt-to-Prompt ~\cite{hertz2022prompt2prompt} and Attend-and-Excite \cite{chefer2023attend} find that modifying the cross attention maps can give interesting image editing effects. Here, we do not dig into details of how to manipulate the cross-attention maps. Instead, we simply write the manipulation in functional form $\mathcal{M}$,
\begin{equation}\label{eq:p2p}
    \boldsymbol\epsilon_t^{(\star)}=\mathcal{M}\left(\boldsymbol\epsilon_\theta\left(\mathbf{x}^{(A)}, \cdot, t\right), \mathbf{c}^{(A)}, \mathbf{c}^{(B)}\right).
\end{equation}

\paragraph{Guidance.}
Classifier-free guidance~\cite{ho2022classifier} proposed 
\begin{equation}\label{eq:c.f.g.}
    \begin{aligned}
    \boldsymbol\epsilon_t^{(\star)} =& \boldsymbol\epsilon_\theta\left(\mathbf{x}_t, \mathbf{c}^{(A)}, t\right) + \\ 
    & \beta\left(\boldsymbol\epsilon_\theta\left(\mathbf{x}_t, \mathbf{c}^{(A)}, t\right) - \boldsymbol\epsilon_\theta\left(\mathbf{x}_t, \varnothing\phantom{^(}, t\right)\right),
    \end{aligned}
\end{equation}
where $\beta$ is a real number and $\varnothing$ denotes an empty condition. The $\beta$ is often called ``guidance scale''. The term $\boldsymbol\epsilon_\theta\left(\mathbf{x}_t, \varnothing, t\right)$ can be seen as unconditional output of $\boldsymbol\epsilon_\theta$. 
Similar ideas are also used in \cite{bansal2023universal-guidance,liu2023more-control,brack2022stable-artist} for image editing.
This can be generalized to
\begin{equation}\label{eq:guidance}
\begin{aligned}
    \boldsymbol\epsilon_t^{(\star)} = & \boldsymbol\epsilon_\theta\left(\mathbf{x}_t, \mathbf{c}^{(A)}, t\right) + \\
    & \beta\left(\boldsymbol\epsilon_\theta\left(\mathbf{x}_t, \mathbf{c}^{(A)}, t\right) - \boldsymbol\epsilon_\theta\left(\mathbf{x}_t, \mathbf{c}^{(B)}, t\right)\right).
\end{aligned}
\end{equation}
Intuitively, with Eq.~\eqref{eq:guidance}, we want the output to have more characteristics from $\mathbf{c}^{(A)}$ when generating using the new condition $\mathbf{c}^{(B)}$ to preserve the layout.
If $\beta$ is a number between $[-1, 0]$, we obtain this linear interpolation,
\begin{equation}\label{eq:guidance-lerp}
\begin{aligned}
    \boldsymbol\epsilon_t^{(\star)} = \omega\cdot\boldsymbol\epsilon_\theta\left(\mathbf{x}_t, \mathbf{c}^{(A)}, t\right) + (1-\omega)\cdot\boldsymbol\epsilon_\theta\left(\mathbf{x}_t, \mathbf{c}^{(B)}, t\right).
\end{aligned}
\end{equation}

\paragraph{Predicted noise.}
Both the Eq.~\eqref{eq:p2p} and Eq.~\eqref{eq:guidance} are modifying predicted noise $\boldsymbol\epsilon_t$. Inspired by this, we investigate another manipulation. 
In timestep $t$, we interpolate the predicted noise $\mathbf{x}_t^{(A)}$ with the predicted noise that is using condition $\mathbf{c}^{(B)}$. Assume we have all the predicted noises for the generation of $\mathbf{x}_t^{(A)}$:
\begin{equation}
\begin{aligned}
    \left\{\left(\mathbf{x}_t^{(A)}, \boldsymbol\epsilon_t^{(A)}\right)\right\}_{t=[T,\dots, 0]}=\mathrm{GenPath}\left(\mathbf{x}_T, \mathbf{c}^{(A)}, t\right),\\
\end{aligned}
\end{equation}
we can mix the two predicted noises,
\begin{equation}\label{eq:eps-lerp}
\begin{aligned}
    \boldsymbol\epsilon_t^{(\star)} & = \omega\cdot\boldsymbol\epsilon_\theta\left(\mathbf{x}_t^{(A)}, \mathbf{c}^{(A)}, t\right) + (1-\omega)\cdot\boldsymbol\epsilon_\theta\left(\mathbf{x}_t^{(\star)}, \mathbf{c}^{(B)}, t\right)\\
    & = \mathrm{lerp}\left(
        \boldsymbol\epsilon_\theta\left(\mathbf{x}_t^{(A)}, \mathbf{c}^{(A)}, t\right), \boldsymbol\epsilon_\theta\left(\mathbf{x}_t^{(\star)}, \mathbf{c}^{(B)}, t\right), \omega
    \right),
\end{aligned}
\end{equation}
where when we start this manipulation, $\mathbf{x}_t^{(\star)}=\mathbf{x}_t^{(B)}$. We show the pipeline of this method in \cref{fig:pred-noise-manip}.

\begin{table*}[!htbp]
    \centering
    \resizebox{\linewidth}{!}{%
    \def\arraystretch{1.15}\tabcolsep=0.01em
    \begin{tabular}{cccc|c|c}
        \specialrule{1pt}{1pt}{1pt}
        & Pre & Changes & Post &  Operator & Exist. Methods\\
        \hline\hline
        \cellcolor{aaccff!10}Inter. Denoised Interp. & 
        \multirow{2}{*}{$
        \begin{cases}
        \textcolor{A}{\mathbf{x}_t^{(A)}}=\mathrm{GenPath}\left(\mathbf{x}_T, \textcolor{A}{\mathbf{c}^{(A)}}, t\right)\\
        \end{cases}
        $}
        & $\textcolor{star}{\mathbf{x}^{(\star)}_t}=\mathrm{lerp}\left(\textcolor{A}{\mathbf{x}_t^{(A)}}, \textcolor{star}{\mathbf{x}_t^{(\star)}}, \omega\right)$ & 
            \multirow{2}{*}{$\begin{cases}
                \textcolor{star}{\boldsymbol\epsilon_t^{(\star)}} =\boldsymbol\epsilon_\theta\left(\textcolor{star}{\mathbf{x}^{(\star)}_t}, \textcolor{B}{\mathbf{c}^{(B)}}, t\right) \\
                \textcolor{star}{\mathbf{x}^{(\star)}_{t-1}}=\mathrm{DDIM}\left(\textcolor{star}{\textcolor{star}{\mathbf{x}^{(\star)}_t},\boldsymbol\epsilon_t^{(\star)}},t\right)\\
            \end{cases}$}
        & 
        $\textcolor{star}{\mathbf{x}^{(\star)}_{t-1}}=\mathcal{O}_{\mathrm{IDI}}\left(\textcolor{A}{\mathbf{x}_t^{(A)}}, \textcolor{star}{\mathbf{x}_t^{(\star)}}, \textcolor{A}{\mathbf{c}_t^{(A)}}, \textcolor{B}{\mathbf{c}_t^{(B)}}, \omega, t\right)$
        & \eg, MagicMix~\cite{liew2022magicmix}\\ 
        \cellcolor{aaccff!10}Inter. Denoised Masking & & $\textcolor{star}{\mathbf{x}^{(\star)}_t}=\mathbf{M} \odot \textcolor{A}{\mathbf{x}_t^{(A)}} + (1-\mathbf{M})\odot \textcolor{star}{\mathbf{x}_t^{(\star)}}$ & & 
        $\textcolor{star}{\mathbf{x}^{(\star)}_{t-1}}=\mathcal{O}_{\mathrm{IDM}}\left(\textcolor{A}{\mathbf{x}_t^{(A)}}, \textcolor{star}{\mathbf{x}_t^{(\star)}}, \textcolor{A}{\mathbf{c}_t^{(A)}}, \textcolor{B}{\mathbf{c}_t^{(B)}}, \mathbf{M}, t\right)$
        &\eg, DiffEdit~\cite{couairon2022diffedit} \\ \hline\hline
        \cellcolor{aaddee!10}Condition Emb. Interp. & 
            $\begin{cases}
                \textcolor{A}{\mathbf{x}_t^{(A)}} = \mathrm{Gen}(\mathbf{x}_T, \textcolor{A}{\mathbf{c}^{(A)}}, t)
            \end{cases}$
        & $\textcolor{star}{\mathbf{c}^{(\star)}}=\mathrm{lerp}\left(\textcolor{A}{\mathbf{c}^{(A)}}, \textcolor{B}{\mathbf{c}^{(B)}}, \omega\right)$ & 
        $\begin{cases}
            \textcolor{star}{\boldsymbol\epsilon_t^{(\star)}} =\boldsymbol\epsilon_\theta\left(\textcolor{A}{\mathbf{x}^{(A)}_t}, \textcolor{B}{\textcolor{star}{\mathbf{c}^{(\star)}}}, t\right) \\
            \textcolor{star}{\mathbf{x}^{(\star)}_{t-1}}=\mathrm{DDIM}\left(\textcolor{star}{\textcolor{A}{\mathbf{x}^{(A)}_t},\boldsymbol\epsilon_t^{(\star)}},t\right)\\
        \end{cases}$
        & 
        $\textcolor{star}{\mathbf{x}^{(\star)}_{t-1}}=\mathcal{O}_{\mathrm{CEI}}\left(\textcolor{A}{\mathbf{x}^{(A)}_{t}}, \textcolor{A}{\mathbf{c}_t^{(A)}}, \textcolor{B}{\mathbf{c}_t^{(B)}}, \omega,t\right)$
        &\\ \hline\hline
        \cellcolor{aabbee!10}Cross Attn. Manip. & 
        $\begin{cases}
            \textcolor{A}{\mathbf{x}_t^{(A)}} = \mathrm{Gen}(\mathbf{x}_T, \textcolor{A}{\mathbf{c}^{(A)}}, t)
        \end{cases}$ 
        & $\textcolor{star}{\boldsymbol\epsilon_t^{(\star)}}=\mathcal{M}\left(\boldsymbol\epsilon_\theta(\textcolor{A}{\mathbf{x}_t^{(A)}}, \cdot, t), \textcolor{A}{\mathbf{c}^{(A)}}, \textcolor{B}{\mathbf{c}^{(B)}}, t\right)$ & 
        $\begin{cases}
            \textcolor{star}{\mathbf{x}^{(\star)}_{t-1}}=\mathrm{DDIM}\left(\textcolor{star}{\textcolor{A}{\mathbf{x}^{(A)}_t},\boldsymbol\epsilon_t^{(\star)}},t\right)\\
        \end{cases}$
        & 
        $\textcolor{star}{\mathbf{x}^{(\star)}_{t-1}}=\mathcal{O}_{\mathrm{CAM}}\left(\textcolor{A}{\mathbf{x}^{(A)}_{t}}, \textcolor{A}{\mathbf{c}_t^{(A)}}, \textcolor{B}{\mathbf{c}_t^{(B)}}, t\right)$
        &\eg, P2P~\cite{hertz2022prompt2prompt}, A\&E~\cite{chefer2023attend} \\ \hline\hline
        \cellcolor{aabbff!10}Guidance & 
        $\begin{cases}
        \mathbf{x}_T\sim\mathcal{N}(\mathbf{0}, \mathbf{I})\\
        \end{cases}$
        & 
        $\begin{cases}
            \textcolor{star}{\boldsymbol\epsilon_t^{A\star}}=\boldsymbol\epsilon_\theta\left(\textcolor{star}{\mathbf{x}_t^{(\star)}}, \textcolor{A}{\mathbf{c}^{(A)}}, t\right)\\
            \textcolor{star}{\boldsymbol\epsilon_t^{B\star}}=\boldsymbol\epsilon_\theta\left(\textcolor{star}{\mathbf{x}_t^{(\star)}}, \textcolor{B}{\mathbf{c}^{(B)}}, t\right)\\
            \textcolor{star}{\boldsymbol\epsilon_t^{\star}}=(1+\beta)\cdot\textcolor{star}{\boldsymbol\epsilon_t^{A\star}} -\beta \cdot \textcolor{star}{\boldsymbol\epsilon_t^{B\star}}
        \end{cases}$
        &         
        $\begin{cases}
            \textcolor{star}{\mathbf{x}^{(\star)}_{t-1}}=\mathrm{DDIM}\left(\textcolor{star}{\textcolor{star}{\mathbf{x}^{(\star)}_t},\boldsymbol\epsilon_t^{(\star)}},t\right)\\
        \end{cases}$
        & 
        $\textcolor{star}{\mathbf{x}^{(\star)}_{t-1}}=\mathcal{O}_{\mathrm{G}}\left(\textcolor{star}{\mathbf{x}^{(\star)}_{t}}, \textcolor{A}{\mathbf{c}_t^{(A)}}, \textcolor{B}{\mathbf{c}_t^{(B)}}, t\right)$
        &\eg, C.F.G.~\cite{ho2022classifier}\\
        \hline\hline
        \cellcolor{aabbff!10}Pred. Noise Interp.& 
        \multirow{2}{*}{
        $\begin{cases}
        \{
        (\textcolor{A}{\mathbf{x}_t^{(A)}}, \textcolor{A}{\boldsymbol\epsilon_t^{(A)}}
        )\}_{t=[T,\dots, 0]}=\mathrm{GenPath}\left(\mathbf{x}_T, \textcolor{A}{\mathbf{c}^{(A)}}, t\right)\\
        \end{cases}$
        }
        & $\textcolor{star}{\boldsymbol\epsilon_t^{(\star)}}=\mathrm{lerp}\left(\textcolor{A}{\boldsymbol\epsilon_t^{(A)}}, \textcolor{star}{\boldsymbol\epsilon_t^{(\star)}}, \omega\right)$ & 
        \multirow{2}{*}{$\begin{cases}
            \textcolor{star}{\mathbf{x}^{(\star)}_{t-1}}=\mathrm{DDIM}\left(\textcolor{A}{\mathbf{x}^{(A)}_t},\textcolor{star}{\boldsymbol\epsilon_t^{(\star)}},t\right)\\
        \end{cases}$}
        & 
        $\textcolor{star}{\mathbf{x}^{(\star)}_{t-1}}=\mathcal{O}_{\mathrm{PNI}}\left(\textcolor{A}{\mathbf{x}^{(A)}_{t}}, \textcolor{A}{\mathbf{c}_t^{(A)}}, \textcolor{B}{\mathbf{c}_t^{(B)}}, \omega, t\right)$
        &\\ 
        \cellcolor{aabbff!10}Pred. Noise Masking & & $\textcolor{star}{\boldsymbol\epsilon_t^{(\star)}}=\mathbf{M} \odot \textcolor{A}{\boldsymbol\epsilon_t^{(A)}} + (1-\mathbf{M})\odot \textcolor{star}{\boldsymbol\epsilon_t^{(\star)}}$ & & 
        $\textcolor{star}{\mathbf{x}^{(\star)}_{t-1}}=\mathcal{O}_{\mathrm{PNM}}\left(\textcolor{A}{\mathbf{x}^{(A)}_{t}}, \textcolor{A}{\mathbf{c}_t^{(A)}}, \textcolor{B}{\mathbf{c}_t^{(B)}}, \mathbf{M}, t\right)$
        &\\ 
        \specialrule{1pt}{1pt}{1pt}
    \end{tabular}
    }
    \vspace{-8pt}
    \caption{We use the superscript $\textcolor{A}{^{(A)}}$ to represent outputs obtained from condition $\textcolor{A}{\mathbf{c}^{(A)}}$ and $\textcolor{B}{^{(B)}}$ for condition $\textcolor{B}{\mathbf{c}^{(B)}}$. The edited outputs are represented with superscript $\textcolor{star}{^{(\star)}}$. $\omega$ is a number between $0$ and $1$. $\beta$ is usually a positive real number.}
    \label{tab:manip-design}
\end{table*}

\section{Experiments and Analysis}
\subsection{Settings}
\paragraph{Editing tasks.}
We describe a small taxonomy for image editing applications that we use to analyze our framework and compare it to previous work. We divide all common image editing operations into two categories: local editing and global editing. Local edits include: changing object, adding object, removing object, changing attribute and mixing objects. Global edits include: changing background, in-domain transfer, out-domain transfer, and stylization. We provide the details in the Supplementary Materials.
\paragraph{Our manipulations and baselines.}
Given a real input image $\mathbf{x}_0^{(A)}$, we first use Null-text Inversion \cite{mokady2022null} together with a text-condition $\mathbf{c}^{(A)}$ to obtain an initial noise tensor $\mathbf{x}_T$. 
Another given condition $\mathbf{c}^{(B)}$ that is used to guide the editing process would induce the generation of output image $\mathbf{x}_0^{(B)}$ starting from the same initial noise tensor $\mathbf{x}_T$. 
As we do not use a mask as an input, we only consider linear operations that do not involve a mask:
\begin{itemize}
\item MDP-$x_t$: intermediate latent interpolation ($\mathcal{O}_{\mathrm{IDI}}$). We linearly interpolate the intermediate latents $\mathbf{x}_t^{(A)}$ and $\mathbf{x}_t^{(B)}$ using Eq. \eqref{inter_denoised}. This manipulation differs from MagicMix \cite{liew2022magicmix} by using Null-text Inversion to obtain the initial noise tensor $\mathbf{x}_T$ rather than directly adding noise. 
\item MDP-$c$: conditional embedding interpolation ($\mathcal{O}_{\mathrm{CEI}}$). We interpolate the condition $\mathbf{c}^{(A)}$ and $\mathbf{c}^{(B)}$ by using Eq. \eqref{eq:lerp-c}.
\item P2P (Prompt-to-Prompt) \cite{hertz2022prompt2prompt}: cross attention manipulation ($\mathcal{O}_{\mathrm{CAM}}$). P2P manipulates the cross attention maps and self attention maps according to the changes in the new text prompt compared to the input prompt.
\item MDP-$\beta$: guidance ($\mathcal{O}_{\mathrm{G}}$). We use condition $\mathbf{c}^{(A)}$ as a guidance to inject the layout while generating new semantics by using Eq. \eqref{eq:guidance-lerp}.
\item MDP-$\epsilon_t$: predicted noise interpolation ($\mathcal{O}_{\mathrm{PNI}}$). We mix the predicted noises when generating input image $\mathbf{x}_0^{(A)}$ with the noises using condition $\mathbf{c}^{(B)}$ by following Eq. \eqref{eq:eps-lerp}. 
\end{itemize}
\paragraph{Manipulation schedule.} We investigate how the methods work under a simple schedule using default settings. 
We start the manipulation at step $t_{max}$ and end at step $t_{min}$. The total number of timesteps of a manipulation is denoted as $T_M = t_{max} - t_{min}$. We perform edits only during these $T_M$ steps. For the remaining steps, we simply use $\mathbf{c}^{(B)}$ to denoise the noisy latent. We vary $t_{max}$ and $t_{min}$ and manually select the best result for each method. 
We fix the interpolation factors (guiding scale for MDP-$\beta$) for the following concerns: 
\begin{enumerate*}[label=(\arabic*)]
\item We observe that setting the manipulation schedule and interpolation factor can both adjust the degree to which the layout is maintained in the edited image to some extent;
\item We want to find a good default setting of interpolation factors for each manipulation;
\item When setting the interpolation factor $\beta = 0$ for MDP-$\beta$, this manipulation is equal to MDP-$c$, which we want to avoid in order to show the characteristic of each manipulation.
\end{enumerate*}
We therefore empirically fix the interpolation factor of MDP-$x_t$, MDP-$c$, P2P, MDP-$\beta$ and MDP-$\epsilon_t$ to be 0.7, 1, 1, -0.3, and 1, respectively. In general, $t_{max}$ is ranging from 0 to 5 while $T_M$ is set to be around 20 for each manipulation. Individual settings for each manipulation differ to obtain the desired editing result. We analyze how the methods work under various manipulation schedules in the Supplementary Materials.
\vspace{-3mm}
\paragraph{Implementation.} For text-guided editing, we test all the methods using the publicly available latent diffusion model Stable Diffusion \footnote{\href{https://huggingface.co/CompVis/stable-diffusion-v1-4}{https://huggingface.co/CompVis/stable-diffusion-v1-4}}. For class-guided editing, we use the conditional latent diffusion model \cite{rombach2022latentdiffusion} trained on ImageNet \cite{deng2009imagenet}. We test our manipulations on one NVIDIA A100 GPU. As no training and finetuning is required, each manipulation can be generally done within 10 seconds. As the inversion method we use is built on top of the DDIM sampler \cite{mokady2022null}, we also use DDIM sampler during the sampling process. However, our method can adopt other deterministic samplers.
\subsection{Editing results}
\begin{figure}
  \begin{subfigure}[t]{0.19\linewidth}
    \includegraphics[width=\linewidth]{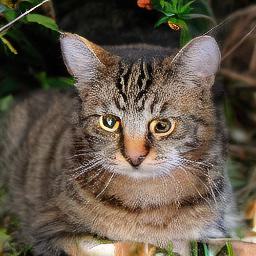}
    \caption*{Input}
\end{subfigure}
  \begin{subfigure}[t]{0.19\linewidth}
    \includegraphics[width=\linewidth]{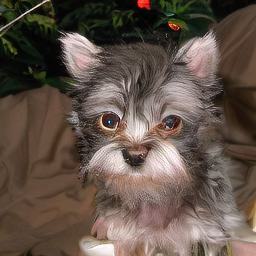}
    \caption*{\textit{Dog}}
\end{subfigure}
  \begin{subfigure}[t]{0.19\linewidth}
    \includegraphics[width=\linewidth]{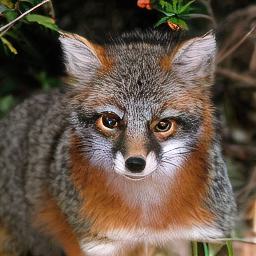}
    \caption*{\textit{Fox}}
\end{subfigure}
  \begin{subfigure}[t]{0.19\linewidth}
    \includegraphics[width=\linewidth]{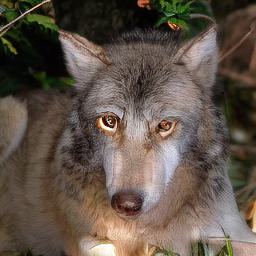}
    \caption*{\textit{Wolf}}
\end{subfigure}
  \begin{subfigure}[t]{0.19\linewidth}
    \includegraphics[width=\linewidth]{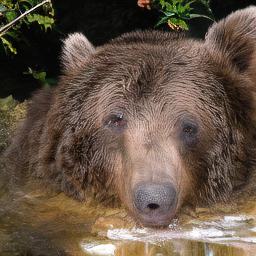}
    \caption*{\textit{Bear}}
\end{subfigure}
\quad
  \begin{subfigure}[t]{0.19\linewidth}
    \includegraphics[width=\linewidth]{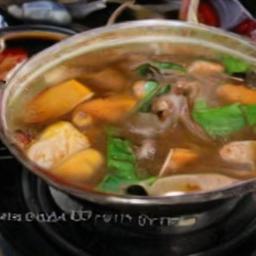}
    \caption*{Input}
\end{subfigure}
  \begin{subfigure}[t]{0.19\linewidth}
    \includegraphics[width=\linewidth]{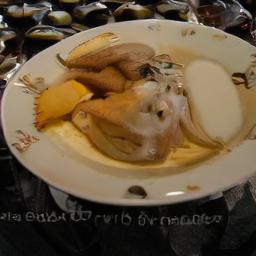}
    \caption*{\textit{Plate}}
\end{subfigure}
  \begin{subfigure}[t]{0.19\linewidth}
    \includegraphics[width=\linewidth]{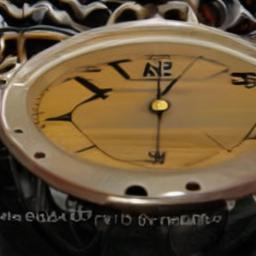}
    \caption*{\textit{Clock}}
\end{subfigure}
  \begin{subfigure}[t]{0.19\linewidth}
    \includegraphics[width=\linewidth]{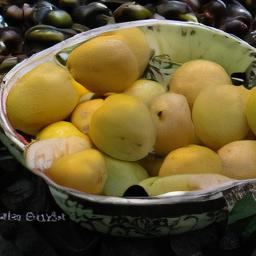}
    \caption*{\textit{Lemon}}
\end{subfigure}
  \begin{subfigure}[t]{0.19\linewidth}
    \includegraphics[width=\linewidth]{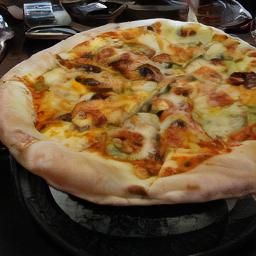}
    \caption*{\textit{Pizza}}
\end{subfigure}
\quad
  \begin{subfigure}[t]{0.19\linewidth}
    \includegraphics[width=\linewidth]{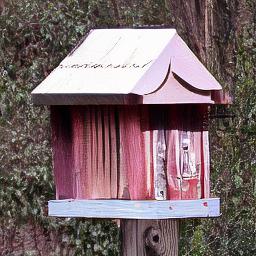}
    \caption*{Input}
\end{subfigure}
  \begin{subfigure}[t]{0.19\linewidth}
    \includegraphics[width=\linewidth]{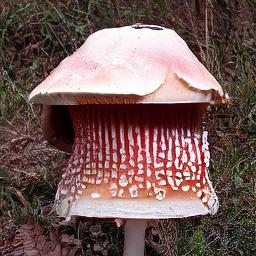}
    \caption*{\textit{Agaric}}
\end{subfigure}
  \begin{subfigure}[t]{0.19\linewidth}
    \includegraphics[width=\linewidth]{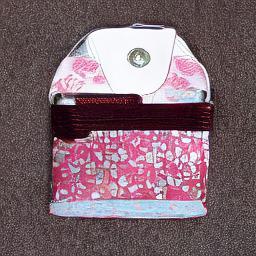}
    \caption*{\textit{Purse}}
\end{subfigure}
  \begin{subfigure}[t]{0.19\linewidth}
    \includegraphics[width=\linewidth]{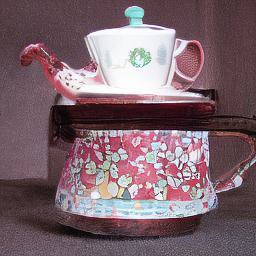}
    \caption*{\textit{Teapot}}
\end{subfigure}
  \begin{subfigure}[t]{0.19\linewidth}
    \includegraphics[width=\linewidth]{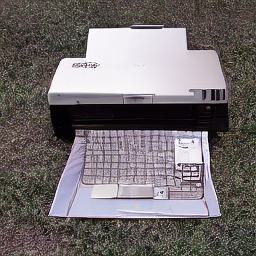}
    \caption*{\textit{Printer}}
\end{subfigure}
\vspace{-8pt}
\caption{Class-guided image editing. We synthesize the input image using one class label, then we use another category from ImageNet to edit the input image. We use MDP-$\epsilon_t$ to demonstrate the ability to do the class-guided editing.}
\label{fig:class}
\end{figure}

\begin{figure}
    \centering
    \begin{tikzpicture}
        \node (small) {
            \includegraphics[width=0.1\linewidth]{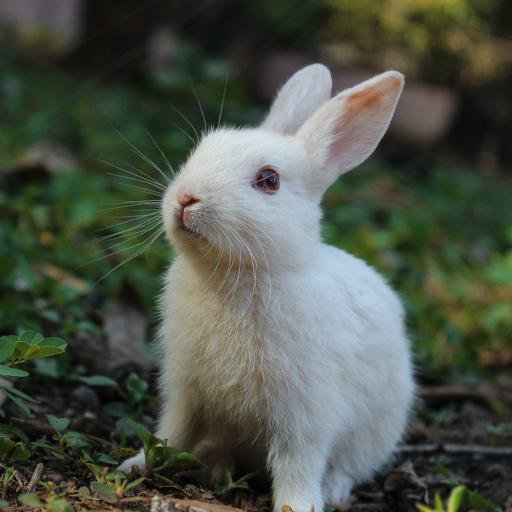}
            \put(-23,28){\rotatebox[]{0}{Input}}
        };

        \node [right=0.1cm of small] (large) {
            \begin{overpic}[width=0.8\linewidth]{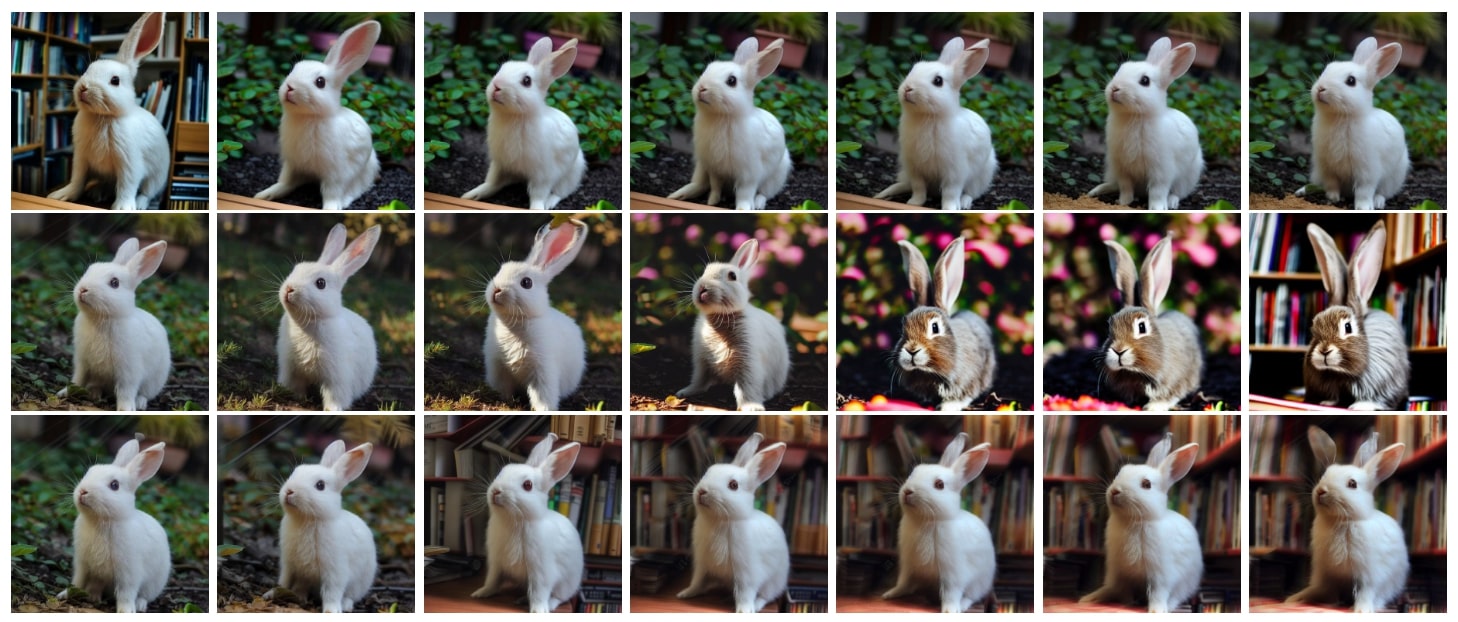}
                \put(0,43){\color{black}\vector(1,0){100}}
                \put(0,-1){\color{black}\vector(1,0){100}}
                \put(102,33){\rotatebox[]{0}{$x_t$}}
                \put(102,20){\rotatebox[]{0}{$c$}}
                \put(102,8){\rotatebox[]{0}{$\epsilon_t$}}
                \put(10,-7){\rotatebox[]{0}{Manipulation starts from early to late}}
            \end{overpic} 
        };
    \end{tikzpicture}
    \vspace{2mm}
    \caption{We change the background of the input image from ``grass'' to ``library''. We apply MDP-$x_t$, MDP-$c$, and MDP-$\epsilon_t$. The manipulation range $T_M$ is set to be 10, 20, and 20 respectively, which are the best settings for each method. From left to right, $t_{max}$ is progressing as $50, 49, ..., 44$. Manipulation $\epsilon_t$ can do faithful edits across different manipulation ranges while the other two methods fail.}
    \label{fig:comparison}
    \vspace{-5mm}
\end{figure}
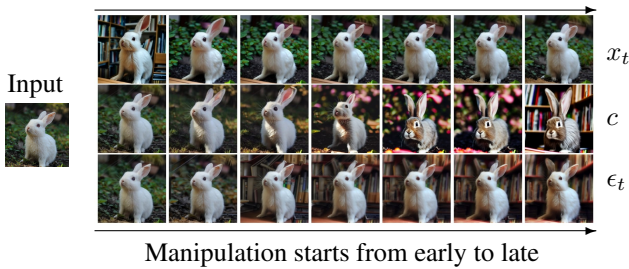

\subsubsection{Local editing}
\begin{figure*}[!htbp]
  \centering
  \begin{tikzpicture}
    \draw (0, 0) node[inner sep=0] {\includegraphics[width=0.13\linewidth]{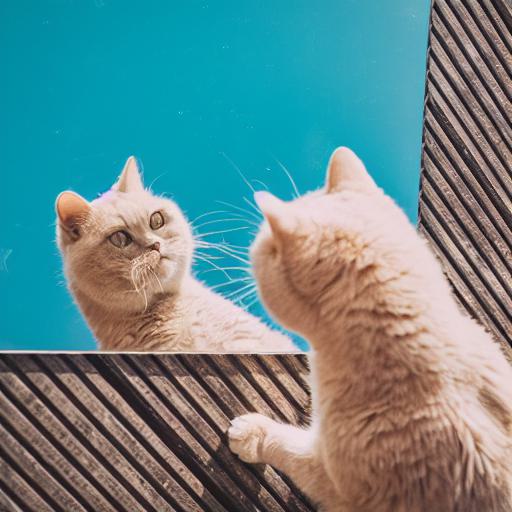}};
    \draw (0, 1.55) node {Input};
    \node (A) at (1.4, -0.8) {};
    \node (B) at (3, -0.8) {};
    \draw[->] (A) edge (B);
    \node[align=center] (C) at (2.2, 0.2) {``Cat''\\to\\``Tiger''};
  \end{tikzpicture}
  \begin{tikzpicture}
    \draw (0, 0) node[inner sep=0] {\includegraphics[width=0.13\linewidth]{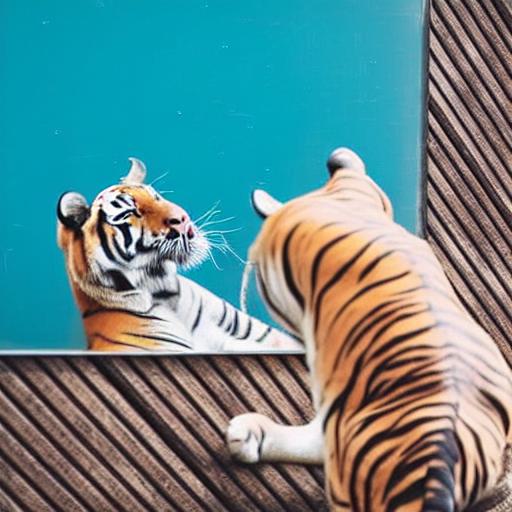}};
    \draw (0, 1.55) node {MDP-$x_t$};
  \end{tikzpicture}
  \begin{tikzpicture}
    \draw (0, 0) node[inner sep=0] {\includegraphics[width=0.13\linewidth]{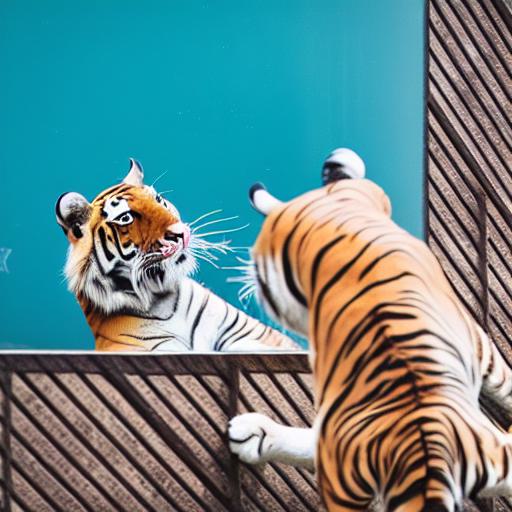}};
    \draw (0, 1.55) node {MDP-$c$};
  \end{tikzpicture}
  \begin{tikzpicture}
    \draw (0, 0) node[inner sep=0] {\includegraphics[width=0.13\linewidth]{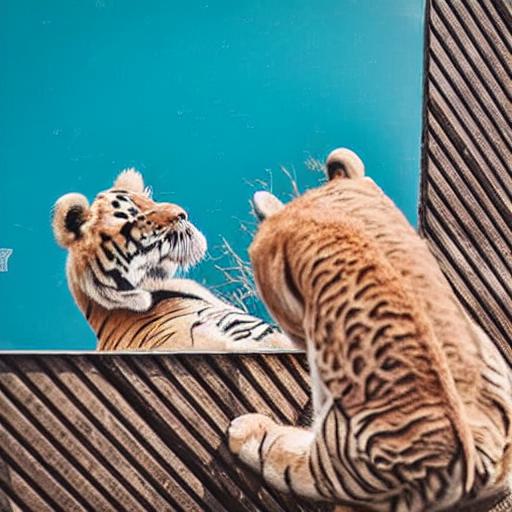}};
    \draw (0, 1.55) node {P2P};
  \end{tikzpicture}
  \begin{tikzpicture}
    \draw (0, 0) node[inner sep=0] {\includegraphics[width=0.13\linewidth]{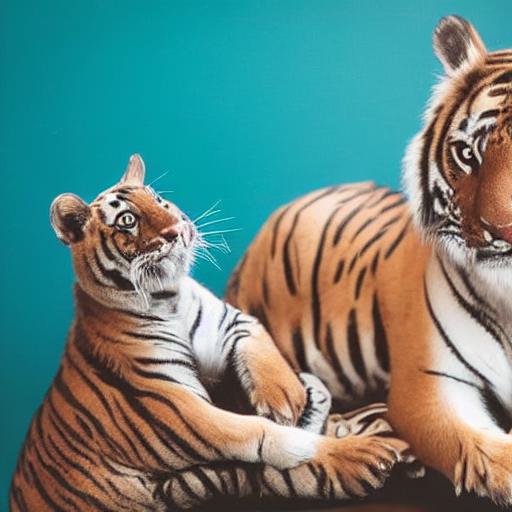}};
    \draw (0, 1.55) node {MDP-$\beta$};
  \end{tikzpicture}
  \begin{tikzpicture}
    \draw (0, 0) node[inner sep=0] {\includegraphics[width=0.13\linewidth]{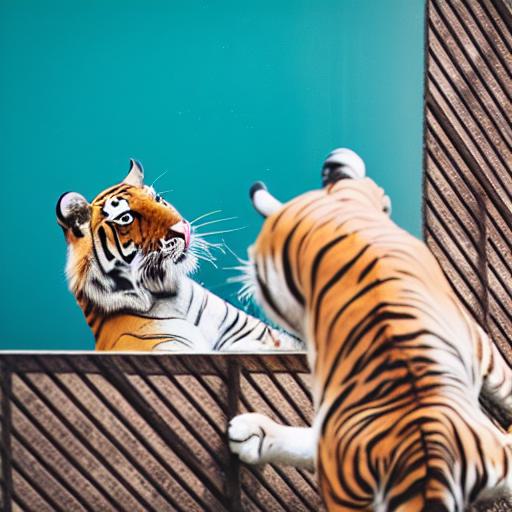}};
    \draw (0, 1.55) node {\textbf{MDP-$\epsilon_t$}};
  \end{tikzpicture}
\quad
\\
  \begin{tikzpicture}
  \draw (0, 0) node[inner sep=0] {\includegraphics[width=0.13\linewidth]{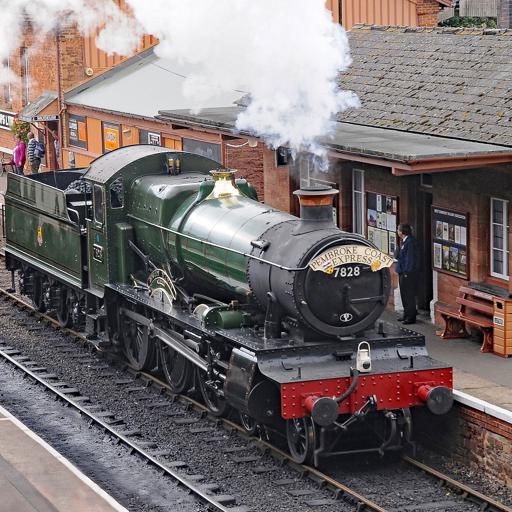}};
    \node (A) at (1.4, -0.8) {};
    \node (B) at (3, -0.8) {};
    \draw[->] (A) edge (B);
    \node[align=center] (C) at (2.2, 0.2) {``Train''\\to\\``Coach''};
  \end{tikzpicture}
  \begin{subfigure}[t]{0.13\linewidth}
    \includegraphics[width=\linewidth]{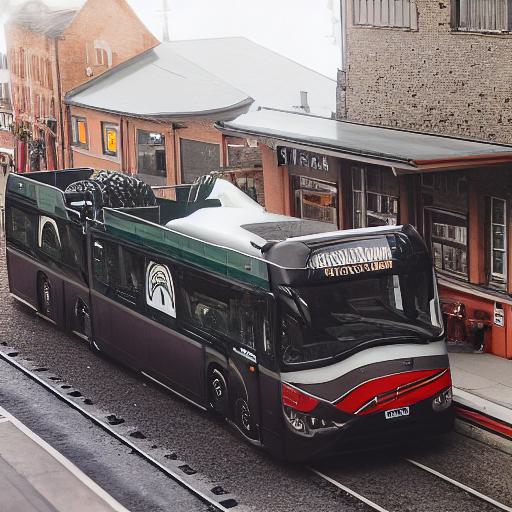}
\end{subfigure}
  \begin{subfigure}[t]{0.13\linewidth}
    \includegraphics[width=\linewidth]{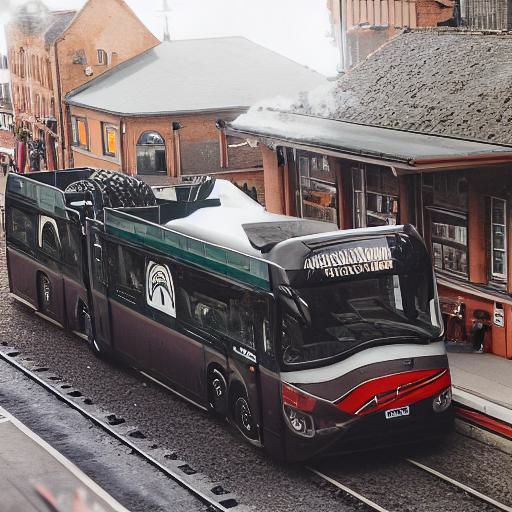}
\end{subfigure}
  \begin{subfigure}[t]{0.13\linewidth}
    \includegraphics[width=\linewidth]{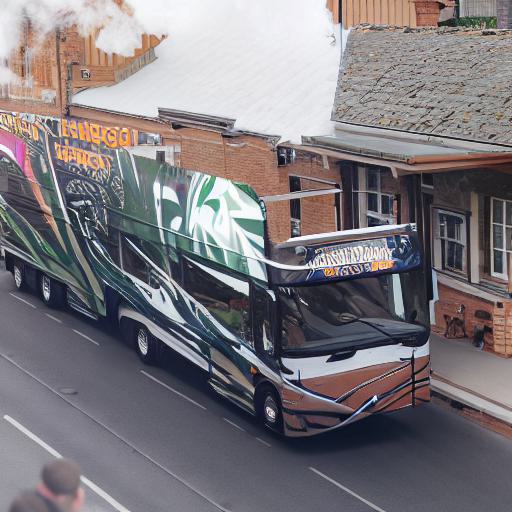}
\end{subfigure}
  \begin{subfigure}[t]{0.13\linewidth}
    \includegraphics[width=\linewidth]{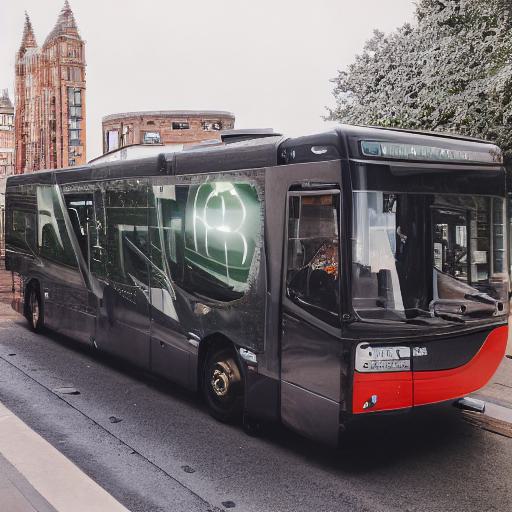}
\end{subfigure}
  \begin{subfigure}[t]{0.13\linewidth}
    \includegraphics[width=\linewidth]{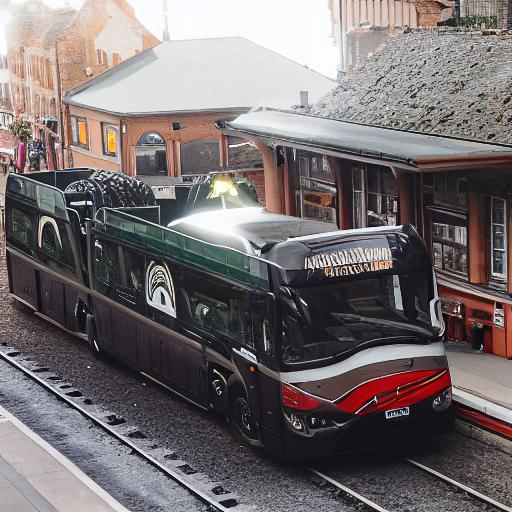}
\end{subfigure}
  \begin{tikzpicture}
  \draw (0, 0) node[inner sep=0] {\includegraphics[width=0.13\linewidth]{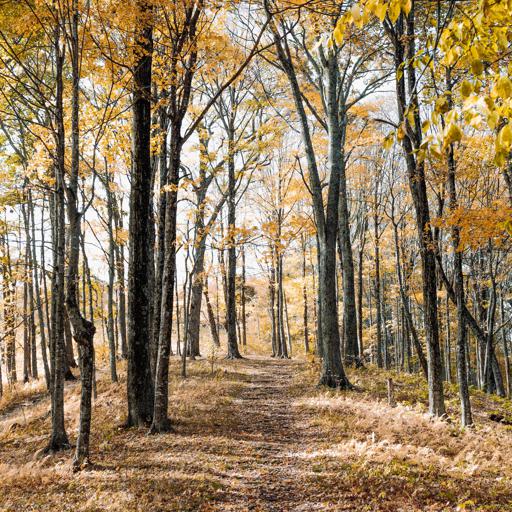}};
    \node (A) at (1.4, -0.8) {};
    \node (B) at (3, -0.8) {};
    \draw[->] (A) edge (B);
    \node[align=center] (C) at (2.2, 0.2) {Add\\``Car''};
  \end{tikzpicture}
  \begin{subfigure}[t]{0.13\linewidth}
    \includegraphics[width=\linewidth]{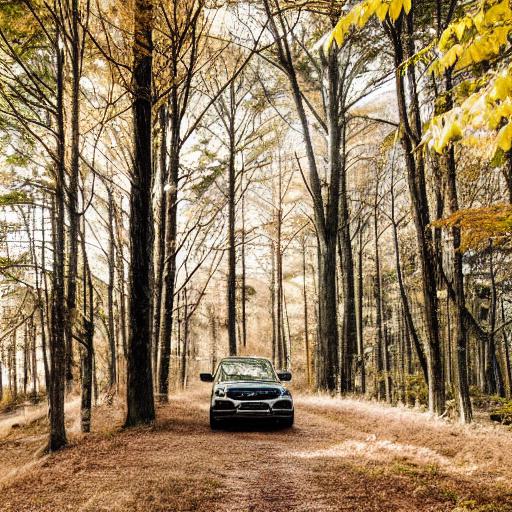}
\end{subfigure}
  \begin{subfigure}[t]{0.13\linewidth}
    \includegraphics[width=\linewidth]{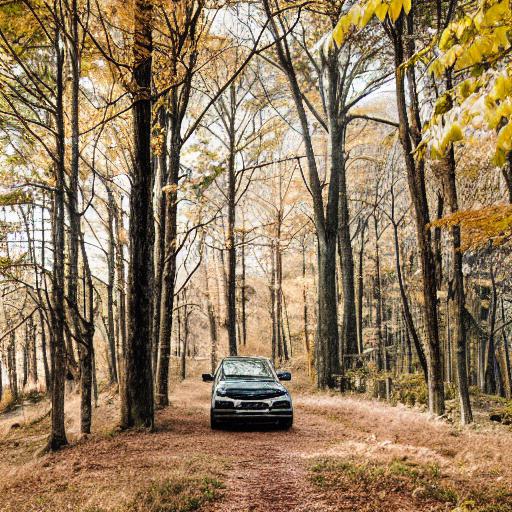}
\end{subfigure}
  \begin{subfigure}[t]{0.13\linewidth}
    \includegraphics[width=\linewidth]{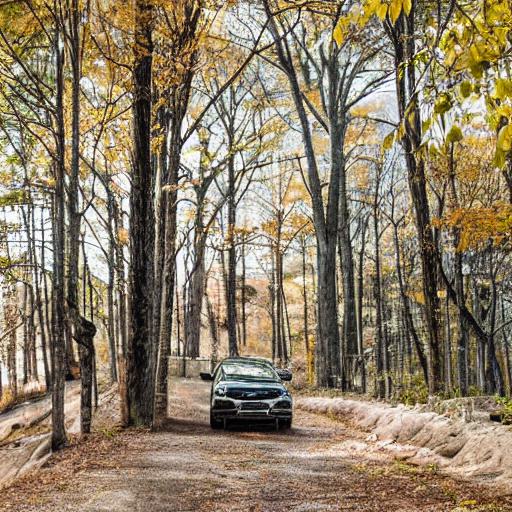}
\end{subfigure}
  \begin{subfigure}[t]{0.13\linewidth}
    \includegraphics[width=\linewidth]{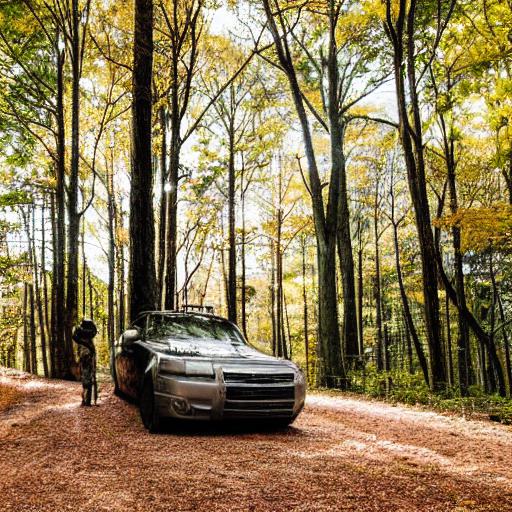}
\end{subfigure}
  \begin{subfigure}[t]{0.13\linewidth}
    \includegraphics[width=\linewidth]{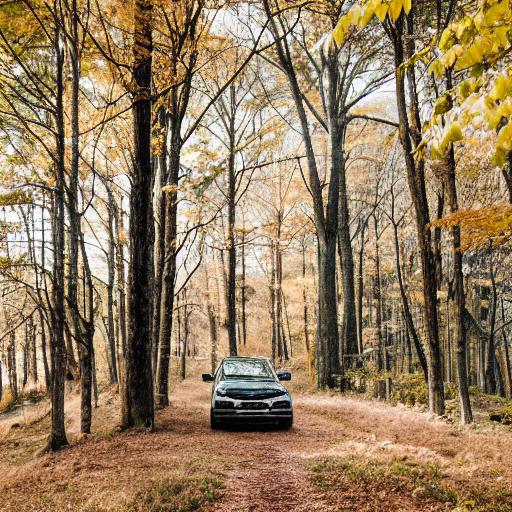}
\end{subfigure}
  \begin{tikzpicture}
  \draw (0, 0) node[inner sep=0] {\includegraphics[width=0.13\linewidth]{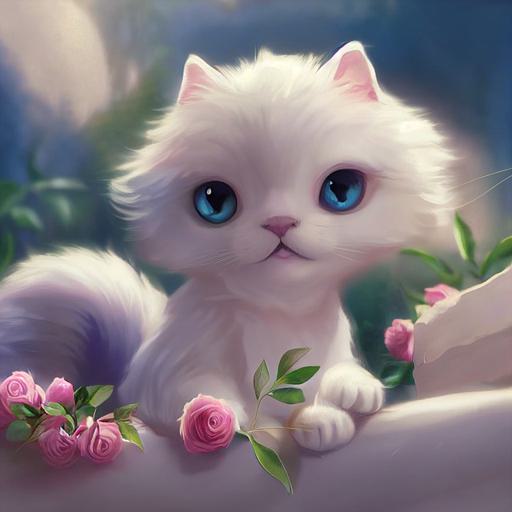}};
    \node (A) at (1.4, -0.8) {};
    \node (B) at (3, -0.8) {};
    \draw[->] (A) edge (B);
    \node[align=center] (C) at (2.2, 0.2) {Add\\``Glasses''};
  \end{tikzpicture}
  \begin{subfigure}[t]{0.13\linewidth}
    \includegraphics[width=\linewidth]{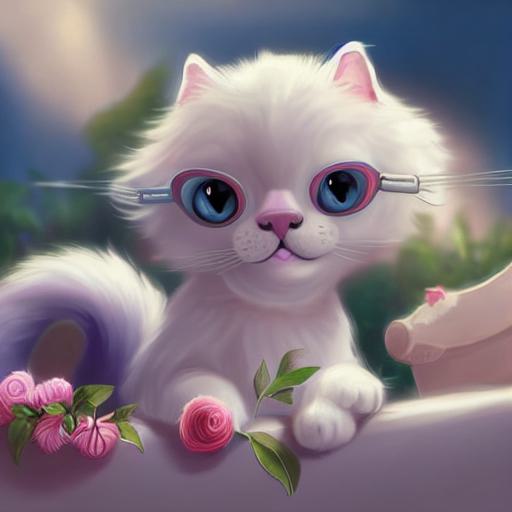}
\end{subfigure}
  \begin{subfigure}[t]{0.13\linewidth}
    \includegraphics[width=\linewidth]{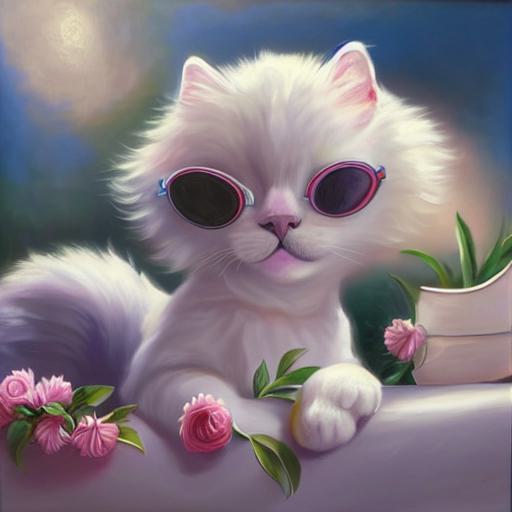}
\end{subfigure}
  \begin{subfigure}[t]{0.13\linewidth}
    \includegraphics[width=\linewidth]{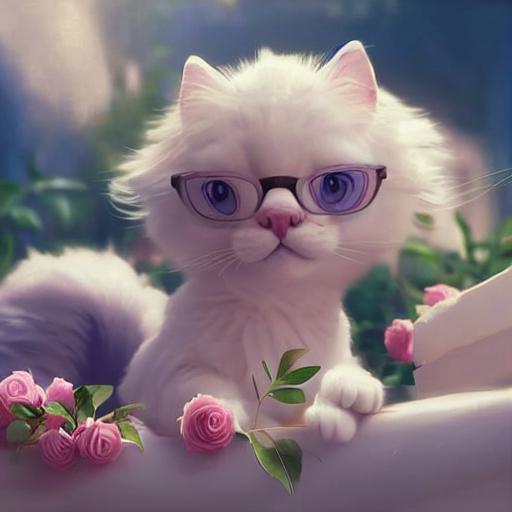}
\end{subfigure}
  \begin{subfigure}[t]{0.13\linewidth}
    \includegraphics[width=\linewidth]{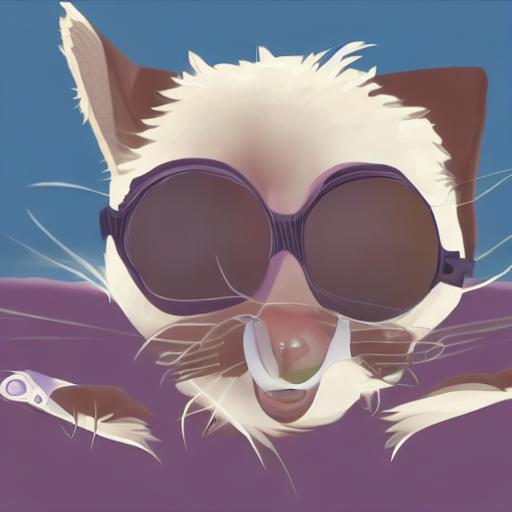}
\end{subfigure}
  \begin{subfigure}[t]{0.13\linewidth}
    \includegraphics[width=\linewidth]{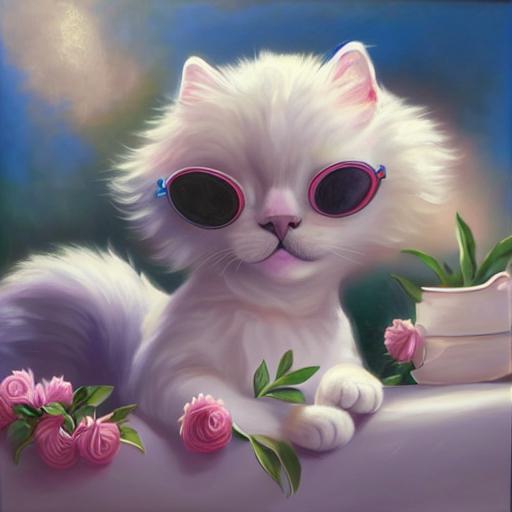}
\end{subfigure}
  \begin{tikzpicture}
  \draw (0, 0) node[inner sep=0] {\includegraphics[width=0.13\linewidth]{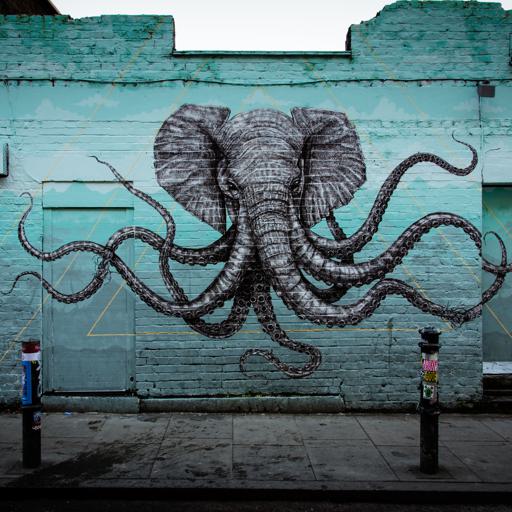}};
    \node (A) at (1.4, -0.8) {};
    \node (B) at (3, -0.8) {};
    \draw[->] (A) edge (B);
    \node[align=center] (C) at (2.2, 0.2) {Remove\\``Graffiti''};
  \end{tikzpicture}
  \begin{subfigure}[t]{0.13\linewidth}
    \includegraphics[width=\linewidth]{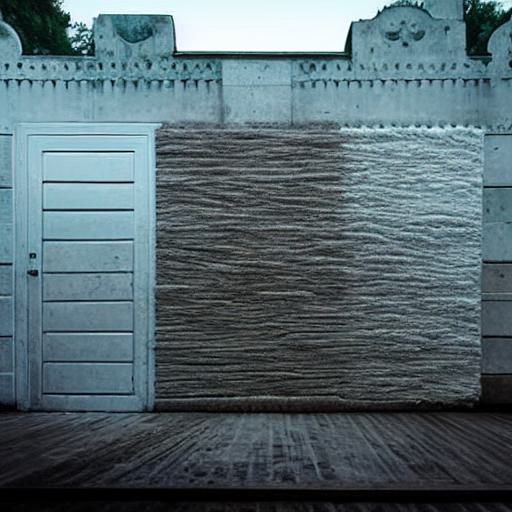}
\end{subfigure}
  \begin{subfigure}[t]{0.13\linewidth}
    \includegraphics[width=\linewidth]{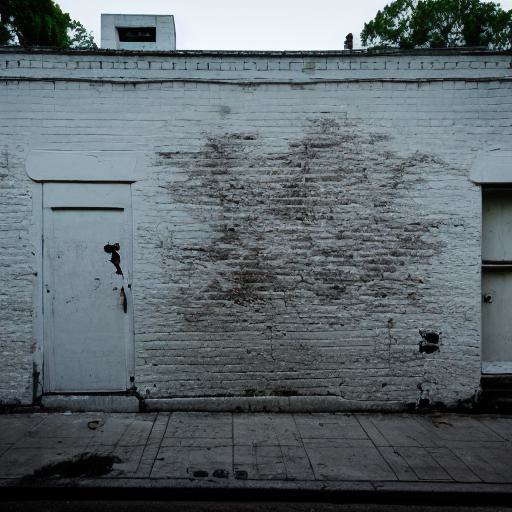}
\end{subfigure}
  \begin{subfigure}[t]{0.13\linewidth}
    \includegraphics[width=\linewidth]{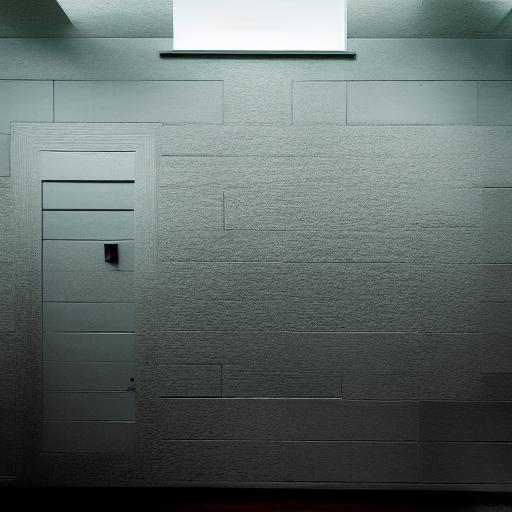}
\end{subfigure}
  \begin{subfigure}[t]{0.13\linewidth}
    \includegraphics[width=\linewidth]{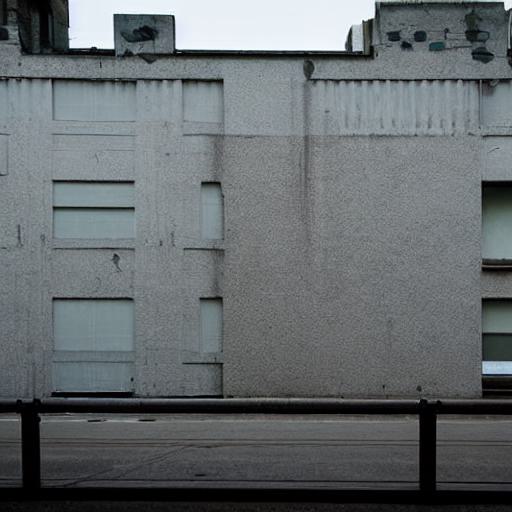}
\end{subfigure}
  \begin{subfigure}[t]{0.13\linewidth}
    \includegraphics[width=\linewidth]{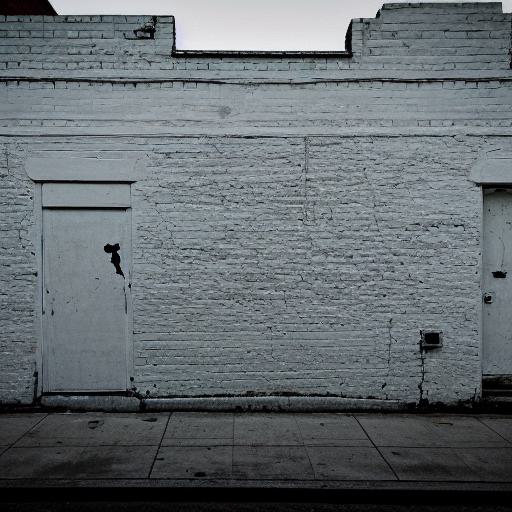}
\end{subfigure}
  \begin{tikzpicture}
  \draw (0, 0) node[inner sep=0] {\includegraphics[width=0.13\linewidth]{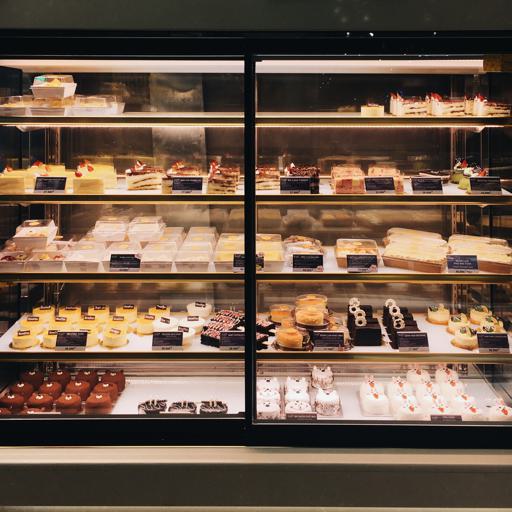}};
    \node (A) at (1.4, -0.8) {};
    \node (B) at (3, -0.8) {};
    \draw[->] (A) edge (B);
    \node[align=center] (C) at (2.2, 0.2) {Remove\\``Cakes''};
  \end{tikzpicture}
  \begin{subfigure}[t]{0.13\linewidth}
    \includegraphics[width=\linewidth]{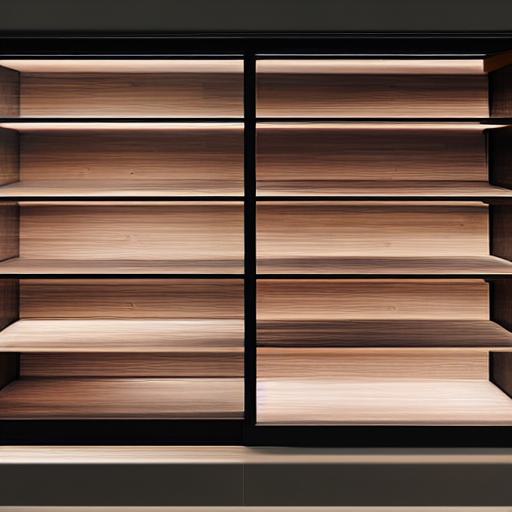}
\end{subfigure}
  \begin{subfigure}[t]{0.13\linewidth}
    \includegraphics[width=\linewidth]{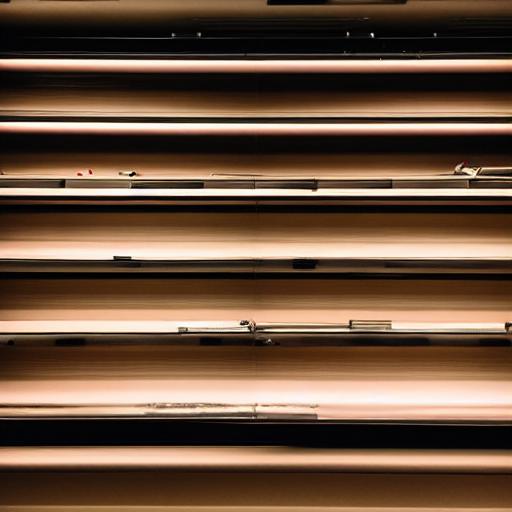}
\end{subfigure}
  \begin{subfigure}[t]{0.13\linewidth}
    \includegraphics[width=\linewidth]{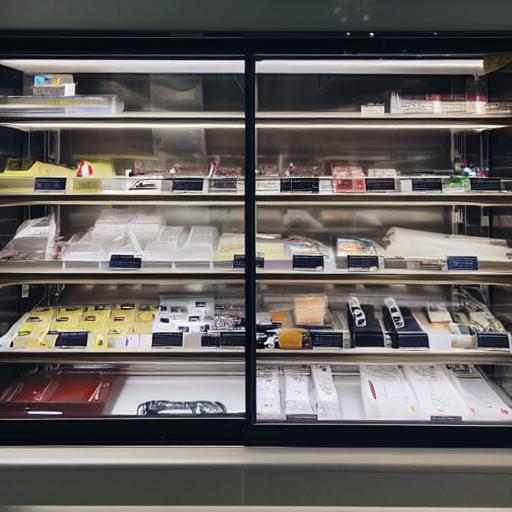}
\end{subfigure}
  \begin{subfigure}[t]{0.13\linewidth}
    \includegraphics[width=\linewidth]{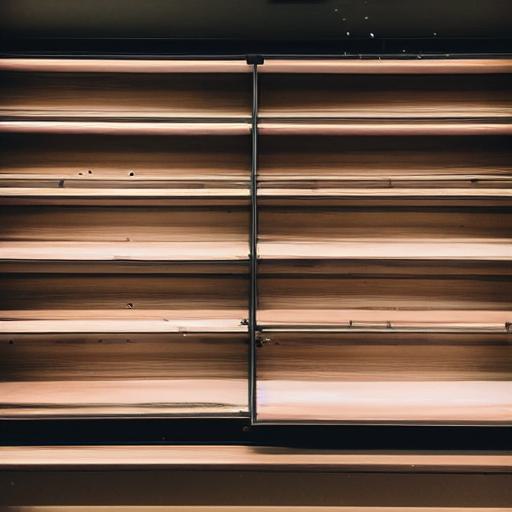}
\end{subfigure}
  \begin{subfigure}[t]{0.13\linewidth}
    \includegraphics[width=\linewidth]{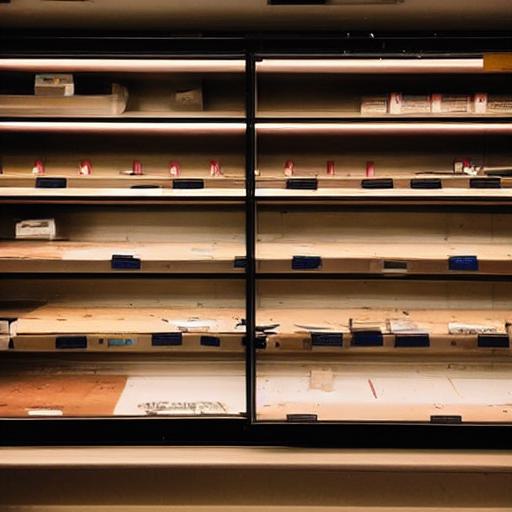}
\end{subfigure}
  \begin{tikzpicture}
  \draw (0, 0) node[inner sep=0] {\includegraphics[width=0.13\linewidth]{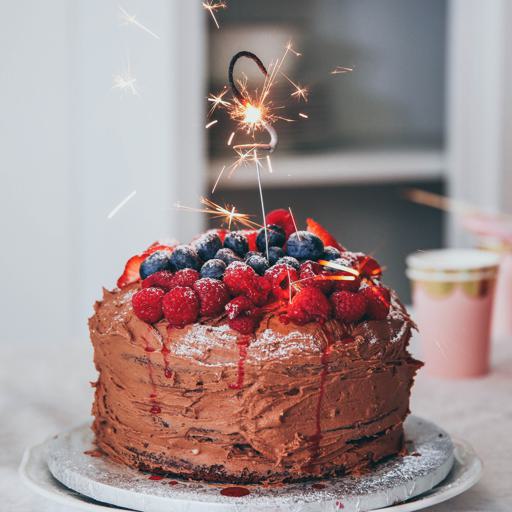}};
    \node (A) at (1.4, -0.8) {};
    \node (B) at (3, -0.8) {};
    \draw[->] (A) edge (B);
    \node[align=center] (C) at (2.2, 0.2) {``Chocolate''\\to\\``Straw\\-berry''};
  \end{tikzpicture}
  \begin{subfigure}[t]{0.13\linewidth}
    \includegraphics[width=\linewidth]{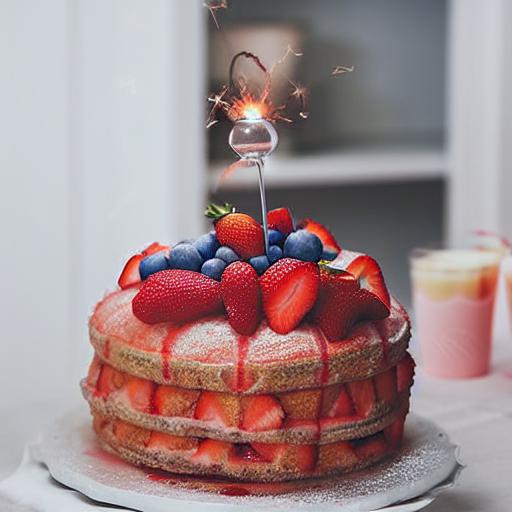}
\end{subfigure}
  \begin{subfigure}[t]{0.13\linewidth}
    \includegraphics[width=\linewidth]{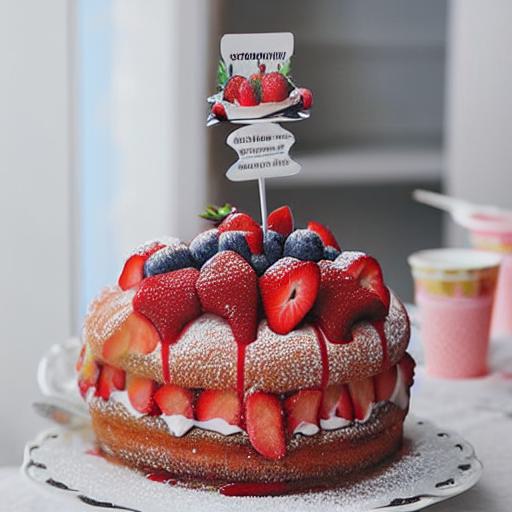}
\end{subfigure}
  \begin{subfigure}[t]{0.13\linewidth}
    \includegraphics[width=\linewidth]{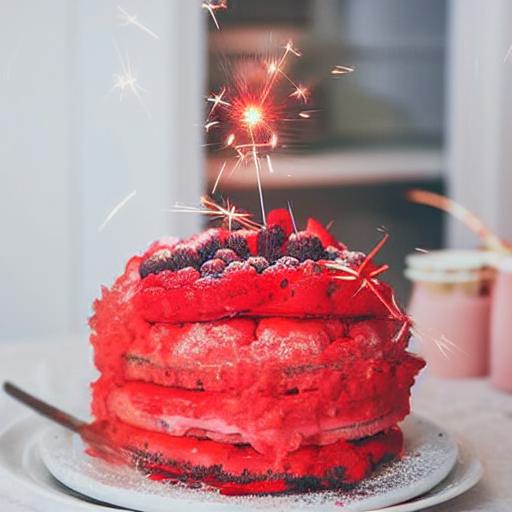}
\end{subfigure}
  \begin{subfigure}[t]{0.13\linewidth}
    \includegraphics[width=\linewidth]{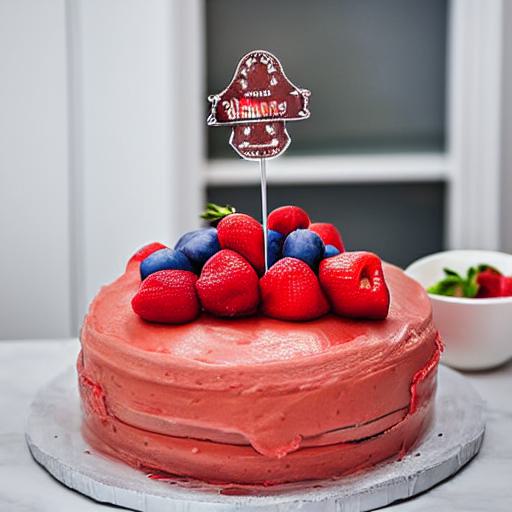}
\end{subfigure}
  \begin{subfigure}[t]{0.13\linewidth}
    \includegraphics[width=\linewidth]{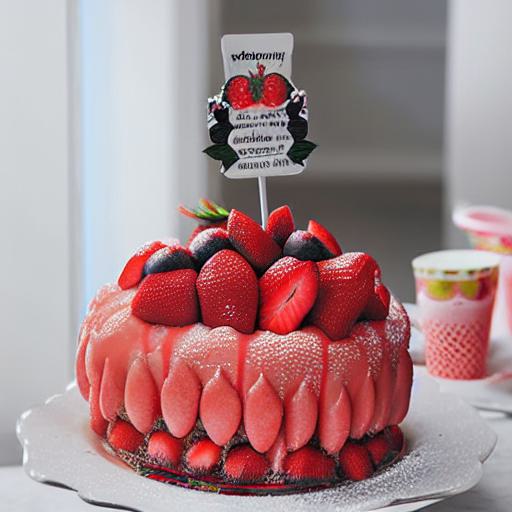}
\end{subfigure}
  \begin{tikzpicture}
  \draw (0, 0) node[inner sep=0] {\includegraphics[width=0.13\linewidth]{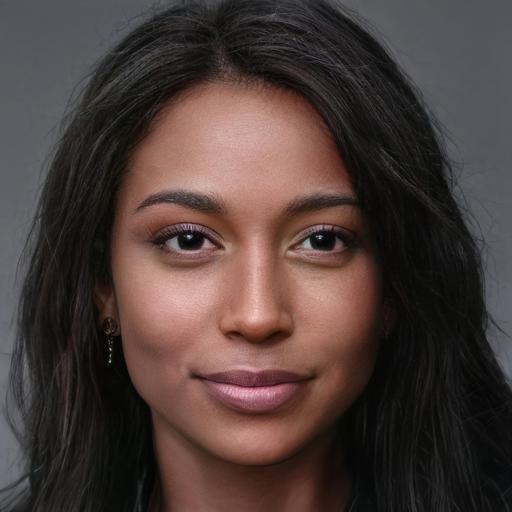}};
    \node (A) at (1.4, -0.8) {};
    \node (B) at (3, -0.8) {};
    \draw[->] (A) edge (B);
    \node[align=center] (C) at (2.2, 0.2) {``Woman''\\to\\``Kid''};
  \end{tikzpicture}
  \begin{subfigure}[t]{0.13\linewidth}
    \includegraphics[width=\linewidth]{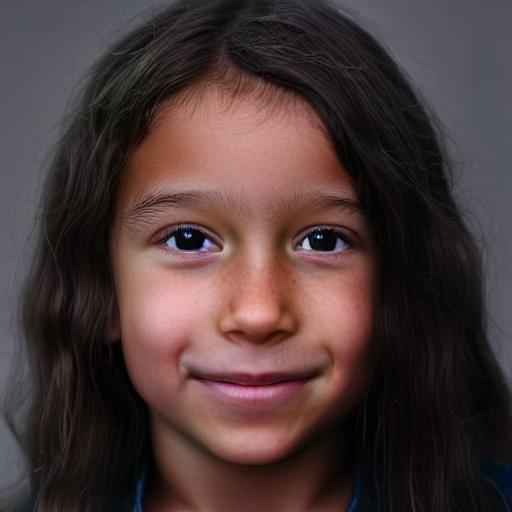}
\end{subfigure}
  \begin{subfigure}[t]{0.13\linewidth}
    \includegraphics[width=\linewidth]{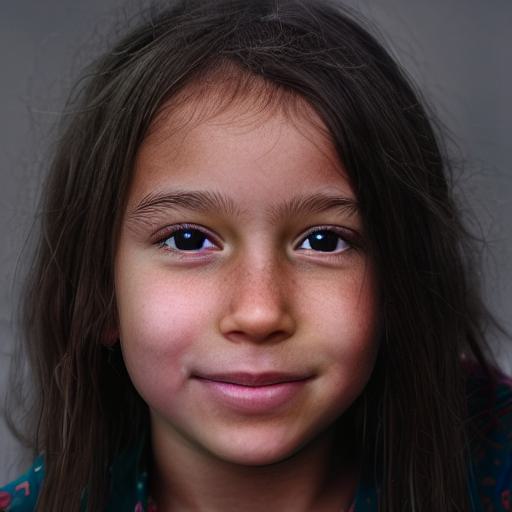}
\end{subfigure}
  \begin{subfigure}[t]{0.13\linewidth}
    \includegraphics[width=\linewidth]{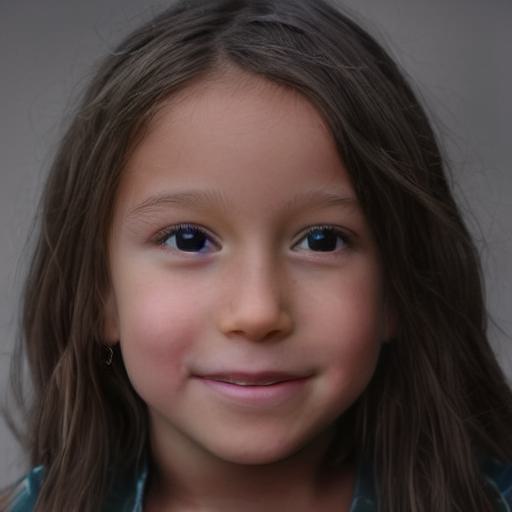}
\end{subfigure}
  \begin{subfigure}[t]{0.13\linewidth}
    \includegraphics[width=\linewidth]{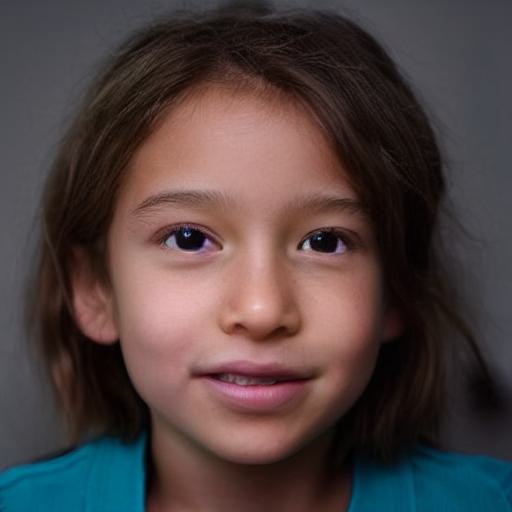}
\end{subfigure}
  \begin{subfigure}[t]{0.13\linewidth}
    \includegraphics[width=\linewidth]{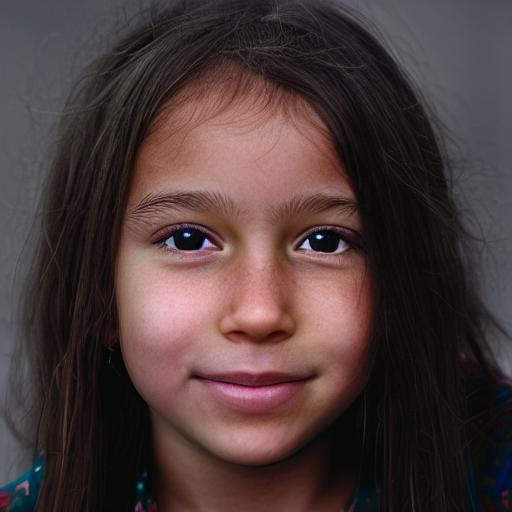}
\end{subfigure}
  \begin{tikzpicture}
  \draw (0, 0) node[inner sep=0] {\includegraphics[width=0.13\linewidth]{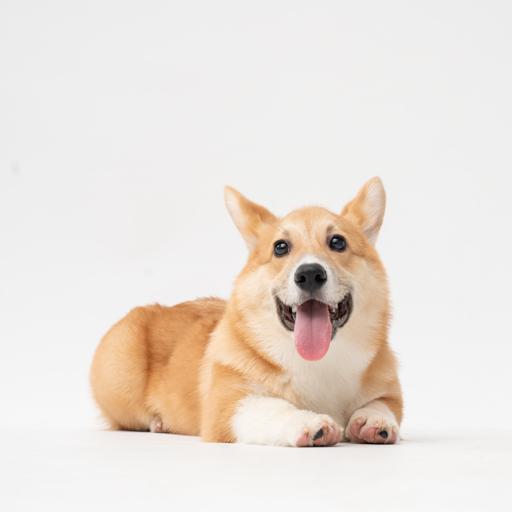}};
    \node (A) at (1.4, -0.8) {};
    \node (B) at (3, -0.8) {};
    \draw[->] (A) edge (B);
    \node[align=center] (C) at (2.2, 0.2) {``Corgi''\\mixed with\\``Coffee \\Machine''};
  \end{tikzpicture}
  \begin{subfigure}[t]{0.13\linewidth}
    \includegraphics[width=\linewidth]{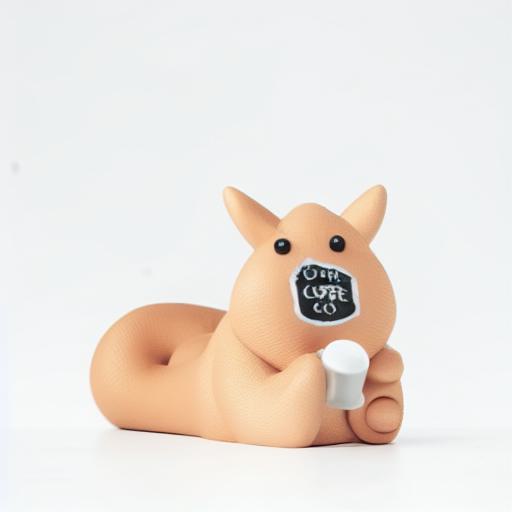}
\end{subfigure}
  \begin{subfigure}[t]{0.13\linewidth}
    \includegraphics[width=\linewidth]{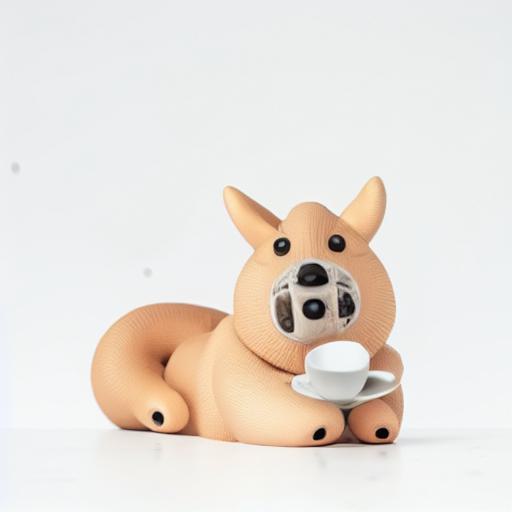}
\end{subfigure}
  \begin{subfigure}[t]{0.13\linewidth}
    \includegraphics[width=\linewidth]{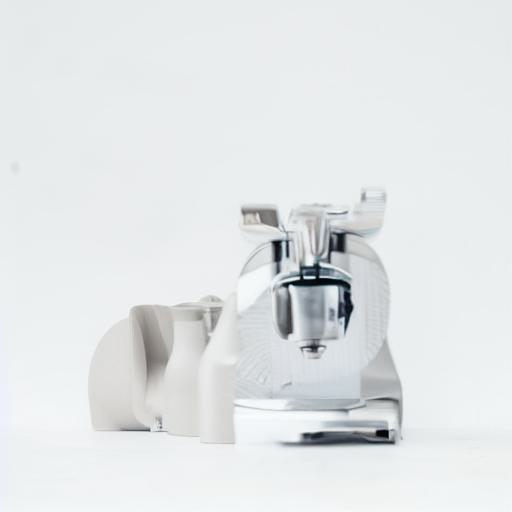}
\end{subfigure}
  \begin{subfigure}[t]{0.13\linewidth}
    \includegraphics[width=\linewidth]{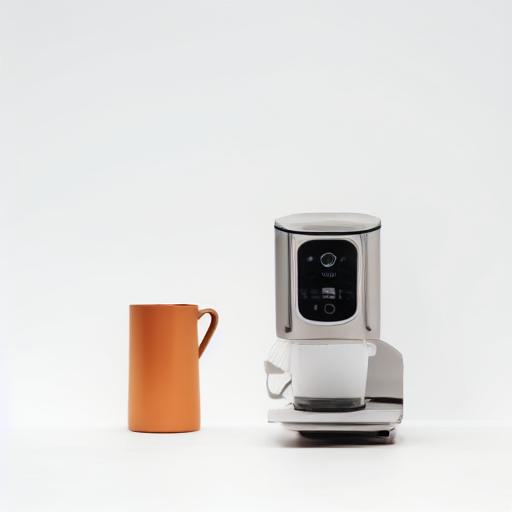}
\end{subfigure}
  \begin{subfigure}[t]{0.13\linewidth}
    \includegraphics[width=\linewidth]{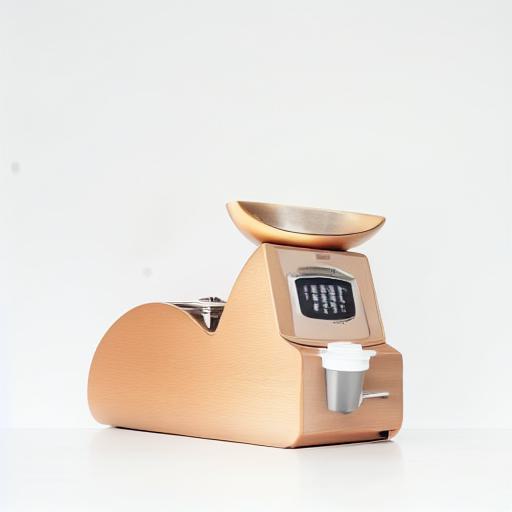}
\end{subfigure}
\caption{Text-guided local editing results.}
\label{fig:local-editing}
\end{figure*}
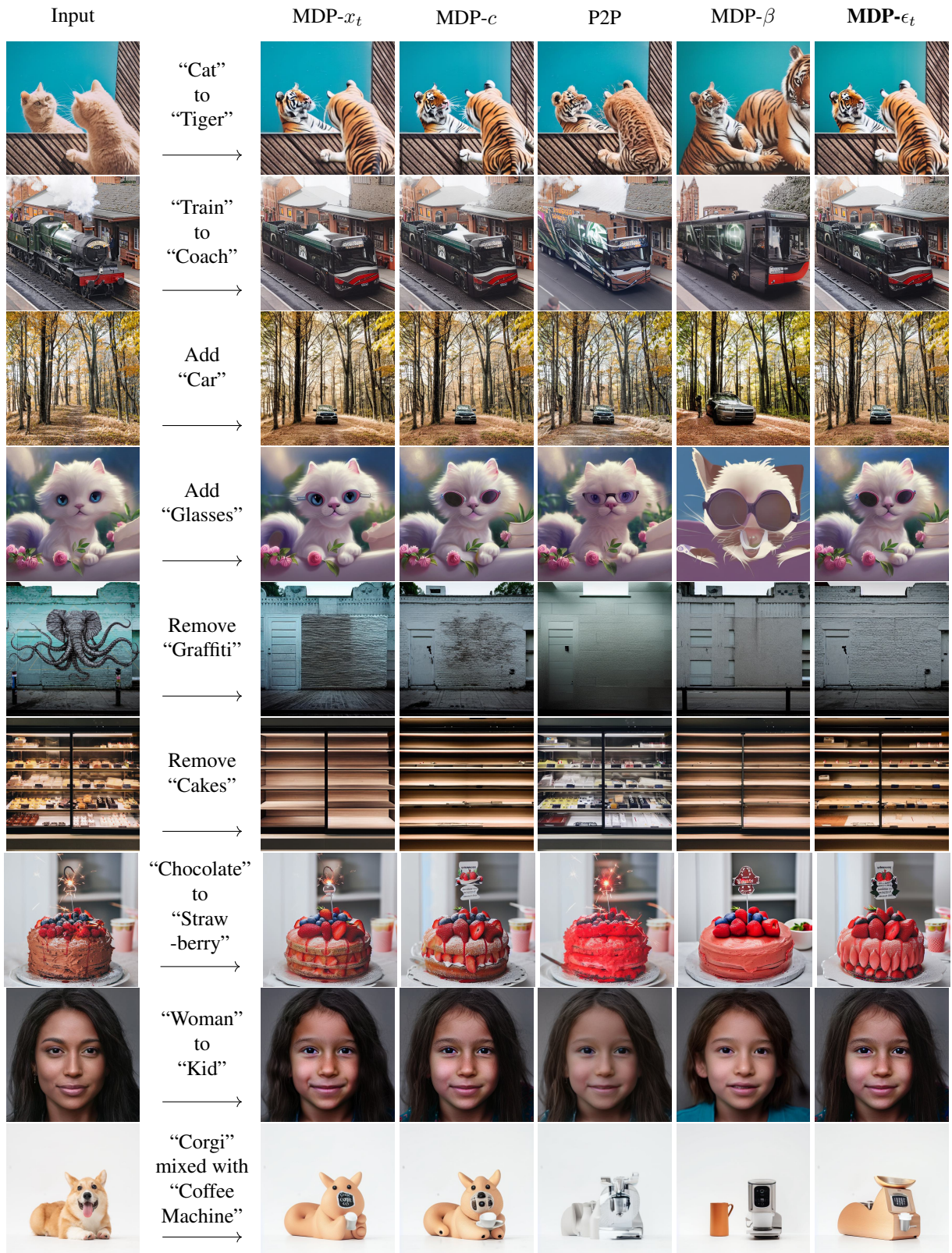

We show examples of local edits for changing object, adding object, removing object, changing attribute, and mixing objects in \cref{fig:local-editing}. The edits by MDP-$\beta$ are somewhat reasonable, but the overall layout is not well preserved. In general, all the other manipulations can do the edits guided by the text prompt while preserving the background of the input image. As the initial diffusion generation steps contribute to the layout of the generated image, we thus recommend to start the manipulation at the early diffusion stage, usually $t_s$ can range from 50 to 45, and $T_M$ can be ranging from 15 to 25. For larger $T_M$ the editing effect is stronger. However, for MDP-$\beta$, even if we do the manipulation under $T_M = 50$, the layout is still not well-preserved. We actually observe that when setting $\beta = 0$ and $t_{max} = 0$, MDP-$\beta$ can do the desired edits; however as we discussed before, MDP-$\beta$ is equal to MDP-$c$ with interpolation factor $w=1$. This is not the case we want to showcase for MDP-$\beta$. 
For MDP-$x_t$ and MDP-$\epsilon_t$, editing can start at later steps, \ie $t_{max}$ can be chosen to be smaller, as more layout is preserved during editing. 
Additionally, we observe some interesting findings that Prompt-to-Prompt fails for edits that remove objects. 
For the application of mixing objects, as there is no universal standard of what the mixed object should look like, we provide more examples generated by each method in the Supplementary Materials. 
\subsubsection{Global editing}
\begin{figure*}[!htbp]
  \centering
  \begin{tikzpicture}
    \draw (0, 0) node[inner sep=0] {\includegraphics[width=0.13\linewidth]{figures/global_editing_jpg/changing_background/001_input.jpg}};
    \draw (0, 1.6) node {Input};
    \node (A) at (1.4, -0.8) {};
    \node (B) at (3, -0.8) {};
    \draw[->] (A) edge (B);
    \node[align=center] (C) at (2.2, 0.2) {``Grass''\\to\\``Library''};
  \end{tikzpicture}
  \begin{tikzpicture}
    \draw (0, 0) node[inner sep=0] {\includegraphics[width=0.13\linewidth]{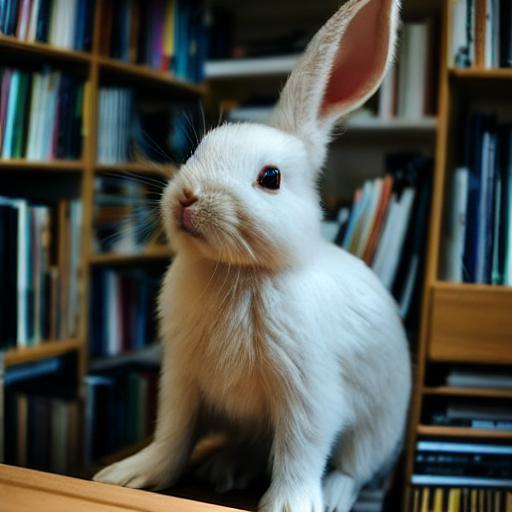}};
    \draw (0, 1.6) node {MDP-$x_t$};
  \end{tikzpicture}
  \begin{tikzpicture}
    \draw (0, 0) node[inner sep=0] {\includegraphics[width=0.13\linewidth]{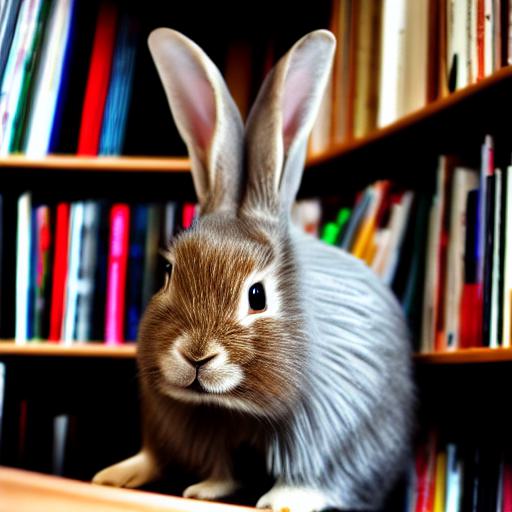}};
    \draw (0, 1.6) node {MDP-$c$};
  \end{tikzpicture}
  \begin{tikzpicture}
    \draw (0, 0) node[inner sep=0] {\includegraphics[width=0.13\linewidth]{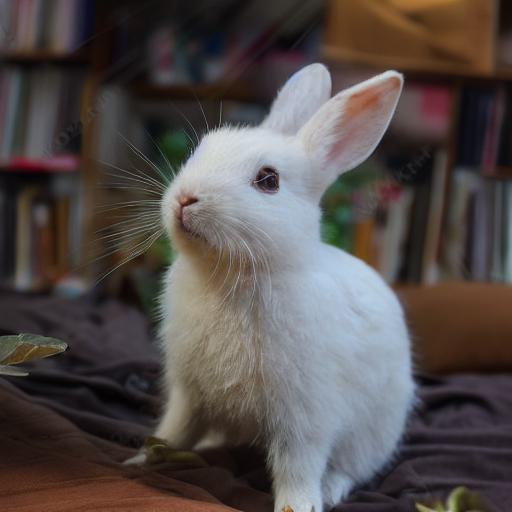}};
    \draw (0, 1.6) node {P2P};
  \end{tikzpicture}
  \begin{tikzpicture}
    \draw (0, 0) node[inner sep=0] {\includegraphics[width=0.13\linewidth]{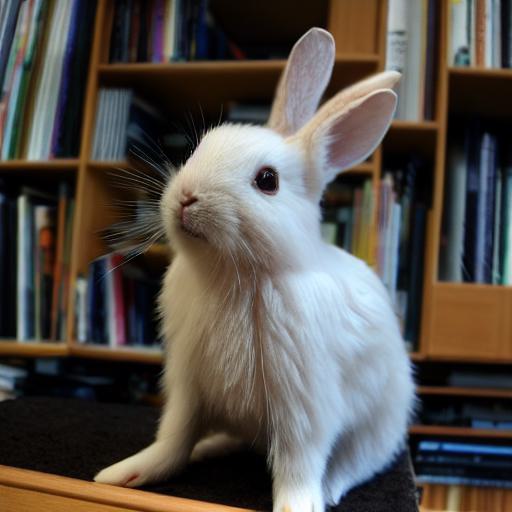}};
    \draw (0, 1.6) node {MDP-$\beta$};
  \end{tikzpicture}
  \begin{tikzpicture}
    \draw (0, 0) node[inner sep=0] {\includegraphics[width=0.13\linewidth]{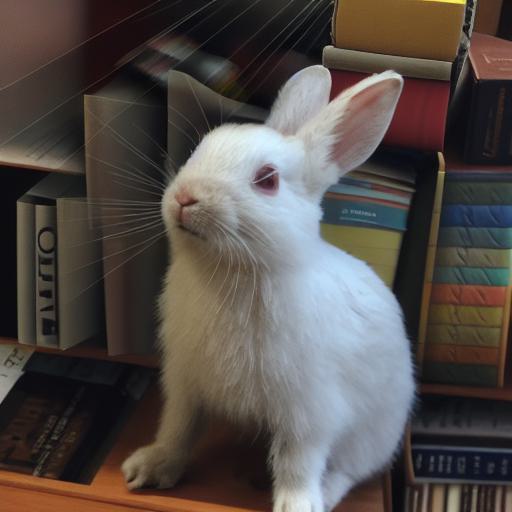}};
    \draw (0, 1.6) node {\textbf{MDP-$\epsilon_t$}};
  \end{tikzpicture}
\quad
  \begin{tikzpicture}
    \draw (0, 0) node[inner sep=0] {\includegraphics[width=0.13\linewidth]{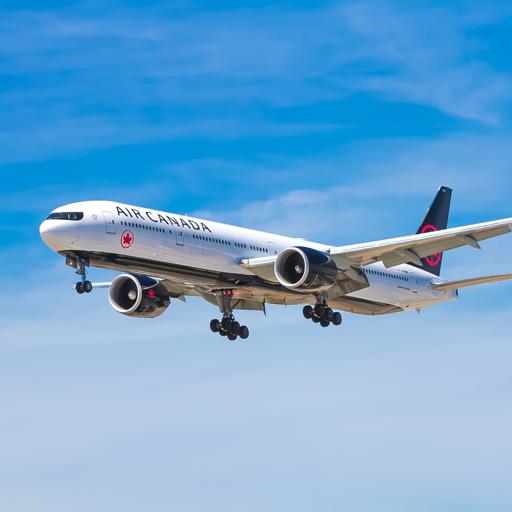}};
    \node (A) at (1.4, -0.8) {};
    \node (B) at (3, -0.8) {};
    \draw[->] (A) edge (B);
    \node[align=center] (C) at (2.2, 0.2) {``Sky''\\to\\``Runway''};
  \end{tikzpicture}
  \begin{subfigure}[t]{0.13\linewidth}
    \includegraphics[width=\linewidth]{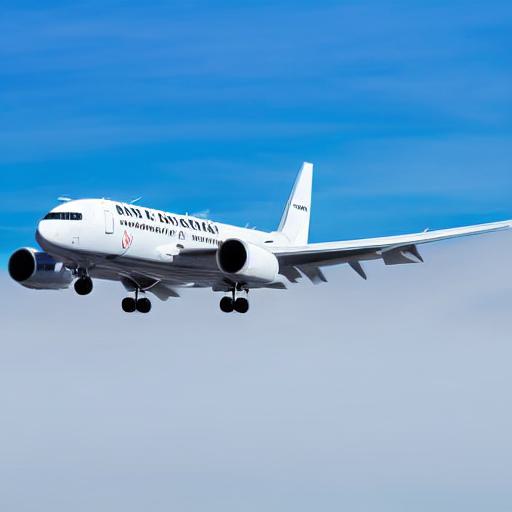}
\end{subfigure}
  \begin{subfigure}[t]{0.13\linewidth}
    \includegraphics[width=\linewidth]{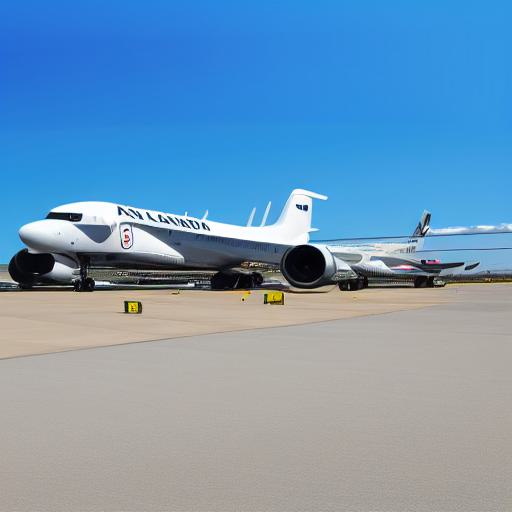}
\end{subfigure}
  \begin{subfigure}[t]{0.13\linewidth}
    \includegraphics[width=\linewidth]{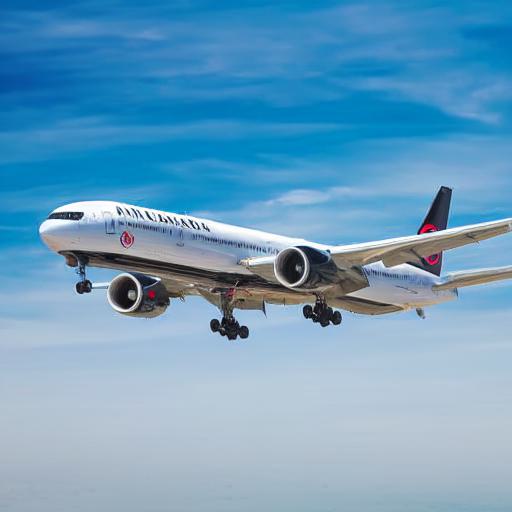}
\end{subfigure}
  \begin{subfigure}[t]{0.13\linewidth}
    \includegraphics[width=\linewidth]{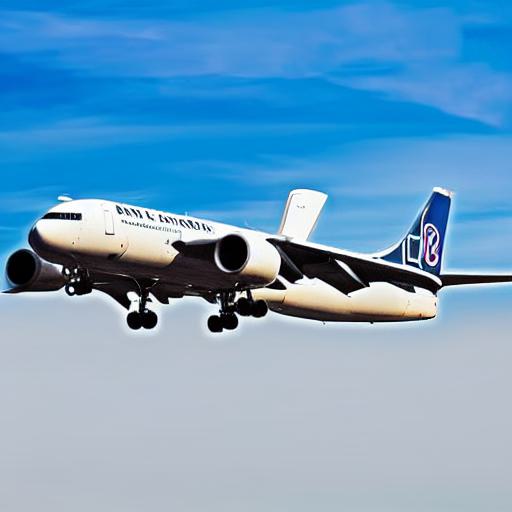}
\end{subfigure}
  \begin{subfigure}[t]{0.13\linewidth}
    \includegraphics[width=\linewidth]{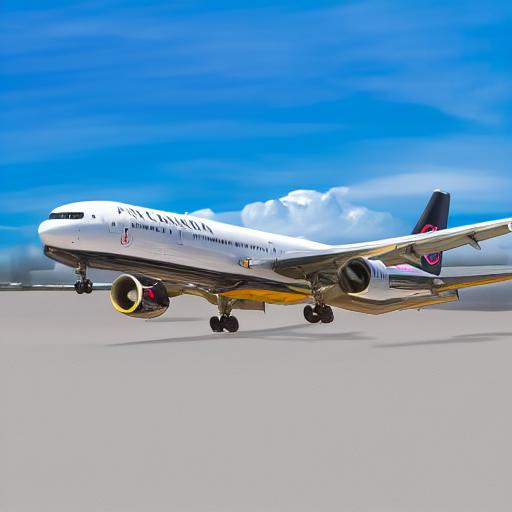}
\end{subfigure}
  \begin{tikzpicture}
    \draw (0, 0) node[inner sep=0] {\includegraphics[width=0.13\linewidth]{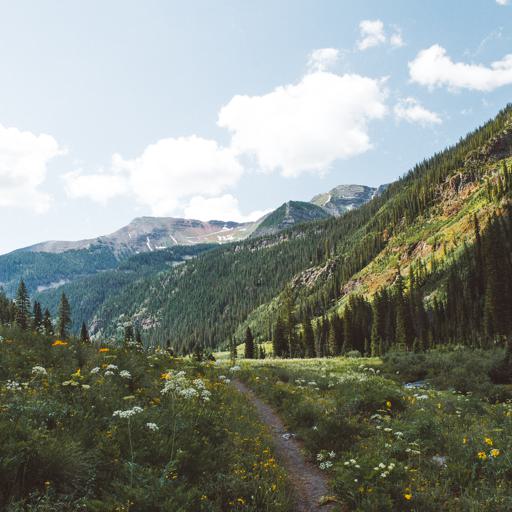}};
    \node (A) at (1.4, -0.8) {};
    \node (B) at (3, -0.8) {};
    \draw[->] (A) edge (B);
    \node[align=center] (C) at (2.2, 0.2) {``Spring''\\to\\``Winter''};
  \end{tikzpicture}
  \begin{subfigure}[t]{0.13\linewidth}
    \includegraphics[width=\linewidth]{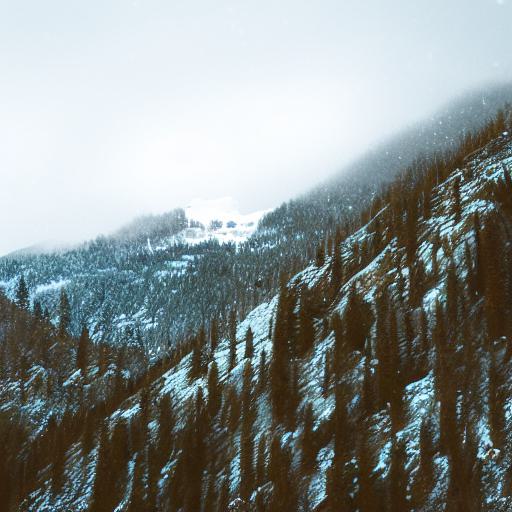}
\end{subfigure}
  \begin{subfigure}[t]{0.13\linewidth}
    \includegraphics[width=\linewidth]{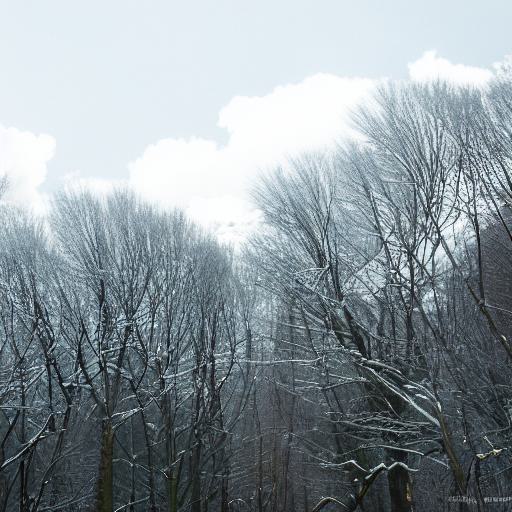}
\end{subfigure}
  \begin{subfigure}[t]{0.13\linewidth}
    \includegraphics[width=\linewidth]{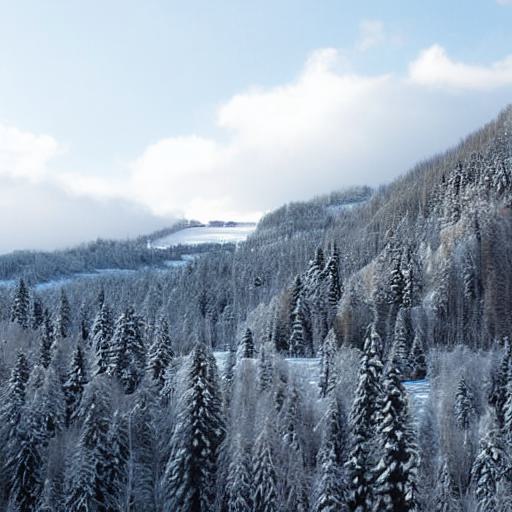}
\end{subfigure}
  \begin{subfigure}[t]{0.13\linewidth}
    \includegraphics[width=\linewidth]{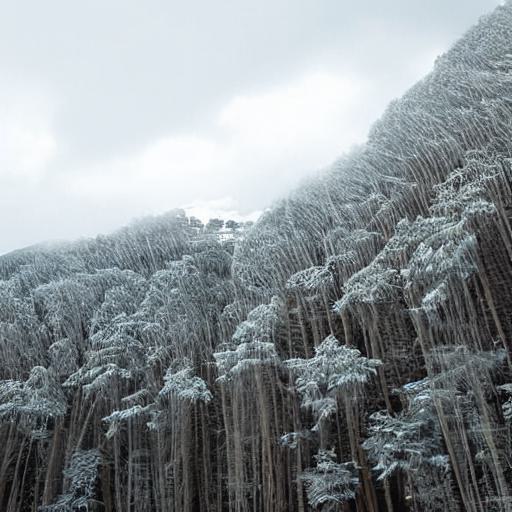}
\end{subfigure}
  \begin{subfigure}[t]{0.13\linewidth}
    \includegraphics[width=\linewidth]{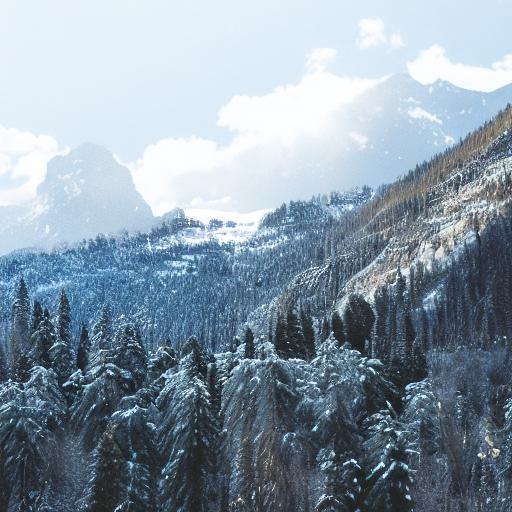}
\end{subfigure}
  \begin{tikzpicture}
    \draw (0, 0) node[inner sep=0] {\includegraphics[width=0.13\linewidth]{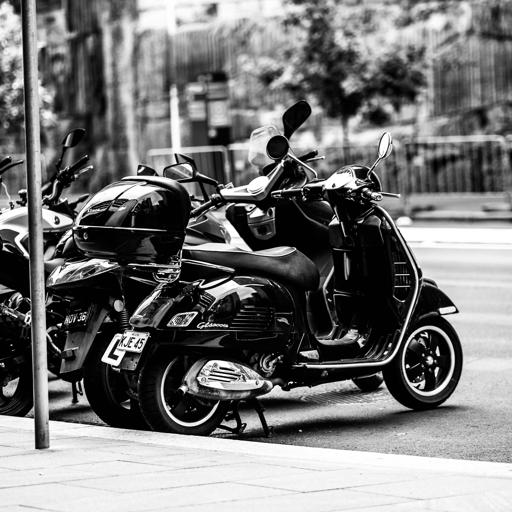}};
    \node (A) at (1.4, -0.8) {};
    \node (B) at (3, -0.8) {};
    \draw[->] (A) edge (B);
    \node[align=center] (C) at (2.2, 0.2) {``Black\\-White''\\to\\``Colored''};
  \end{tikzpicture}
  \begin{subfigure}[t]{0.13\linewidth}
    \includegraphics[width=\linewidth]{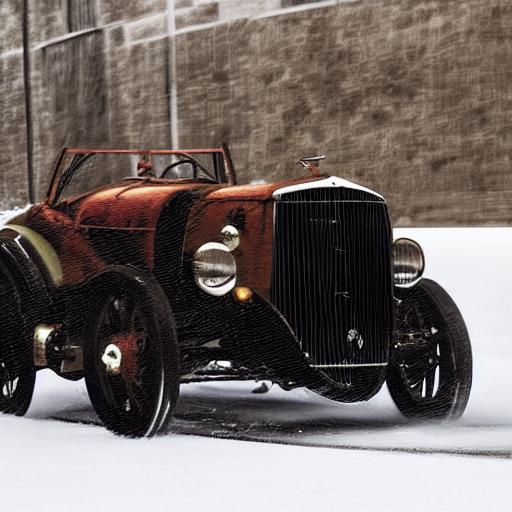}
\end{subfigure}
  \begin{subfigure}[t]{0.13\linewidth}
    \includegraphics[width=\linewidth]{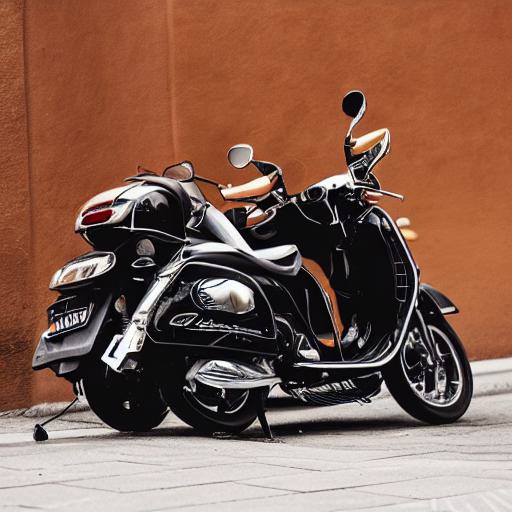}
\end{subfigure}
  \begin{subfigure}[t]{0.13\linewidth}
    \includegraphics[width=\linewidth]{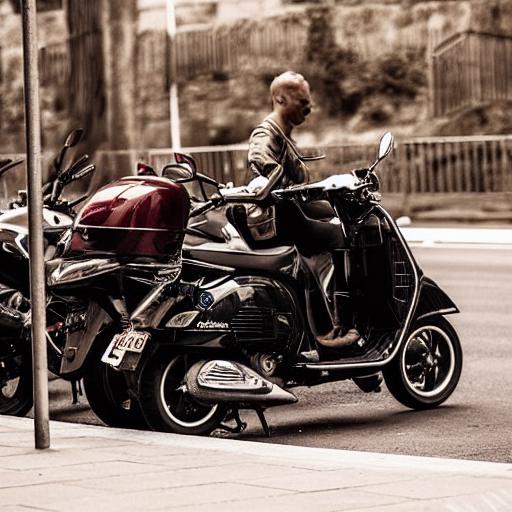}
\end{subfigure}
  \begin{subfigure}[t]{0.13\linewidth}
    \includegraphics[width=\linewidth]{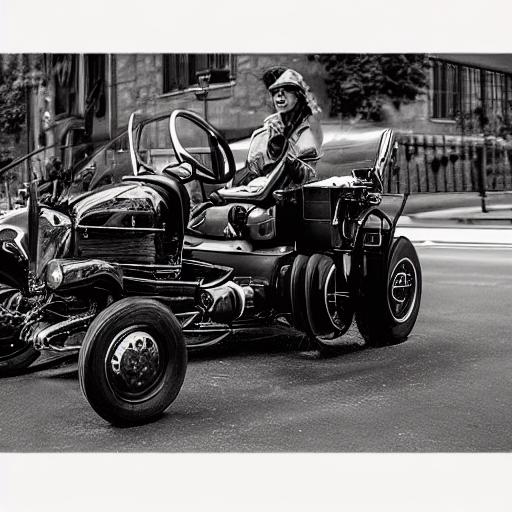}
\end{subfigure}
  \begin{subfigure}[t]{0.13\linewidth}
    \includegraphics[width=\linewidth]{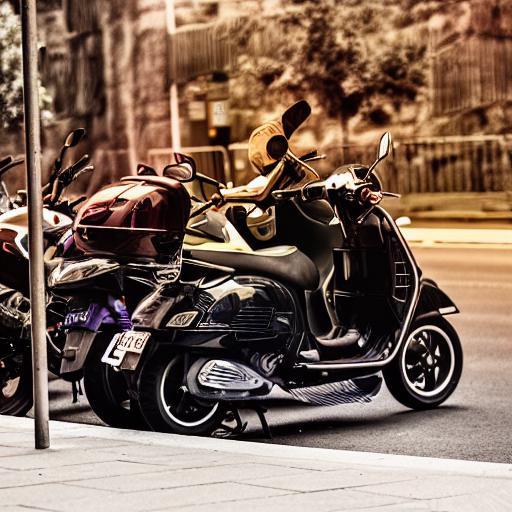}
\end{subfigure}
  \begin{tikzpicture}
    \draw (0, 0) node[inner sep=0] {\includegraphics[width=0.13\linewidth]{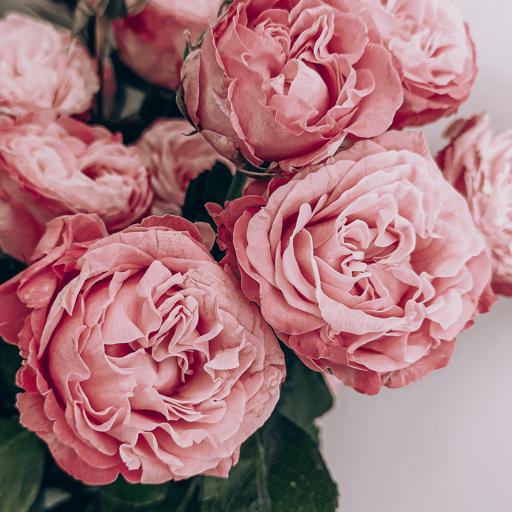}};
    \node (A) at (1.4, -0.8) {};
    \node (B) at (3, -0.8) {};
    \draw[->] (A) edge (B);
    \node[align=center] (C) at (2.2, 0.2) {``Photo''\\to\\``Oil\\Painting''};
  \end{tikzpicture}
  \begin{subfigure}[t]{0.13\linewidth}
    \includegraphics[width=\linewidth]{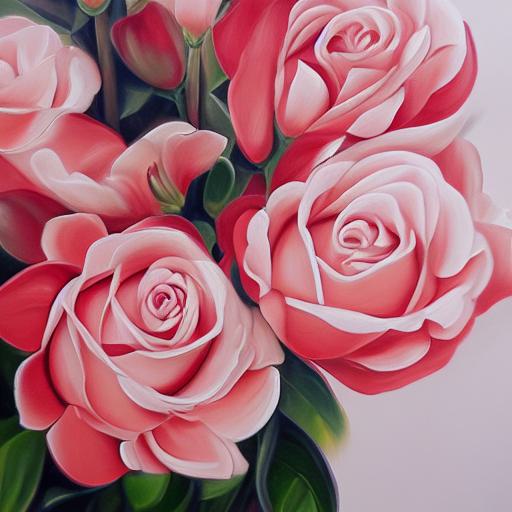}
\end{subfigure}
  \begin{subfigure}[t]{0.13\linewidth}
    \includegraphics[width=\linewidth]{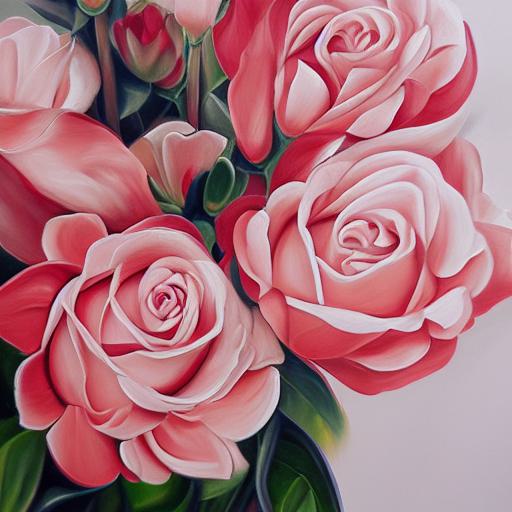}
\end{subfigure}
  \begin{subfigure}[t]{0.13\linewidth}
    \includegraphics[width=\linewidth]{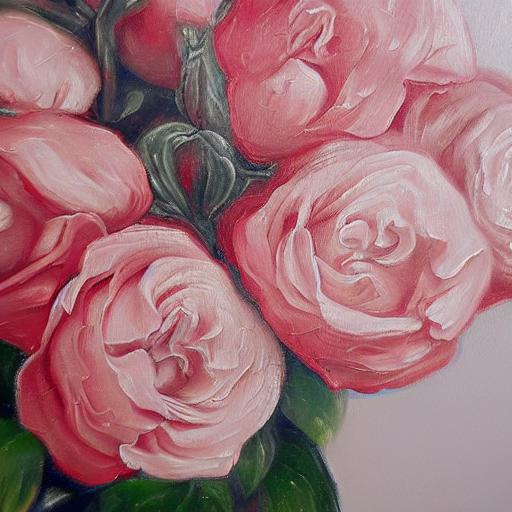}
\end{subfigure}
  \begin{subfigure}[t]{0.13\linewidth}
    \includegraphics[width=\linewidth]{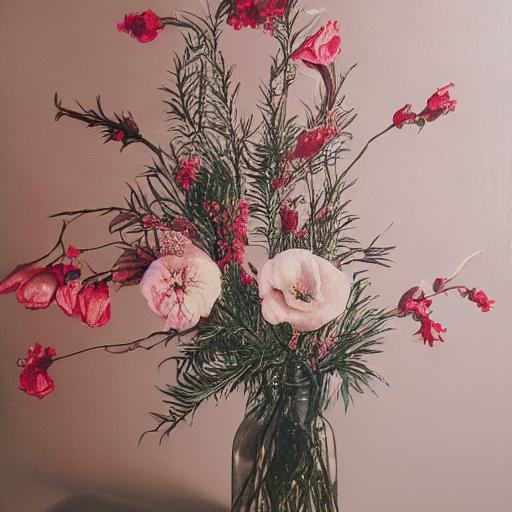}
\end{subfigure}
  \begin{subfigure}[t]{0.13\linewidth}
    \includegraphics[width=\linewidth]{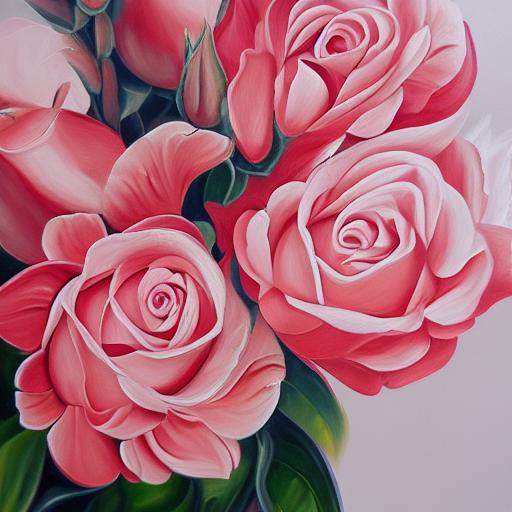}
\end{subfigure}
  \begin{tikzpicture}
    \draw (0, 0) node[inner sep=0] {\includegraphics[width=0.13\linewidth]{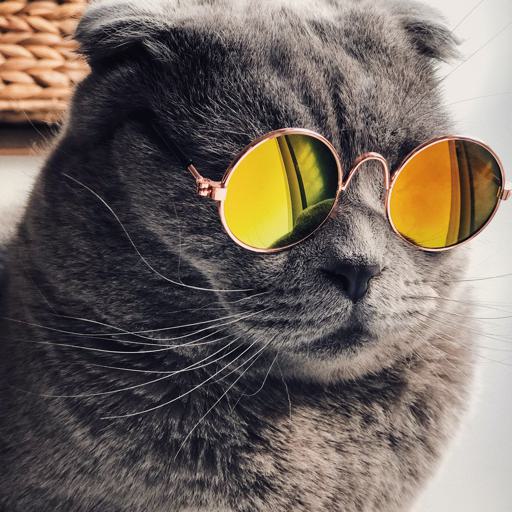}};
    \node (A) at (1.4, -0.8) {};
    \node (B) at (3, -0.8) {};
    \draw[->] (A) edge (B);
    \node[align=center] (C) at (2.2, 0.2) {``Photo''\\to\\``Pencil\\Sketch''};
  \end{tikzpicture}
  \begin{subfigure}[t]{0.13\linewidth}
    \includegraphics[width=\linewidth]{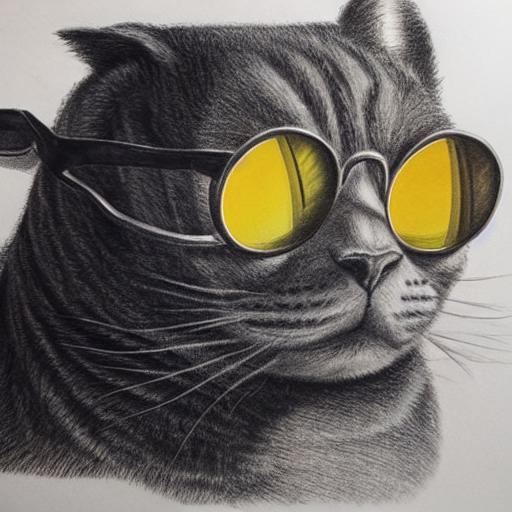}
\end{subfigure}
  \begin{subfigure}[t]{0.13\linewidth}
    \includegraphics[width=\linewidth]{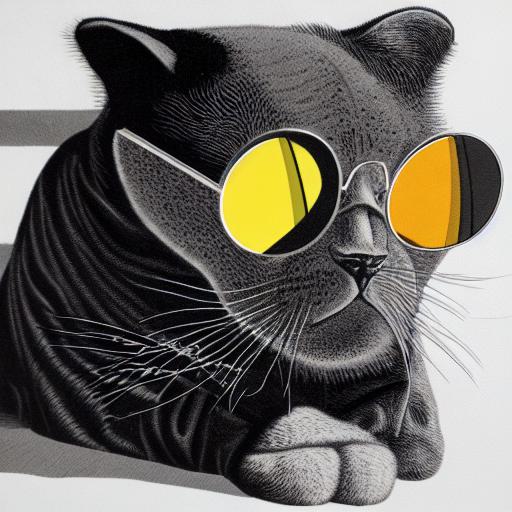}
\end{subfigure}
  \begin{subfigure}[t]{0.13\linewidth}
    \includegraphics[width=\linewidth]{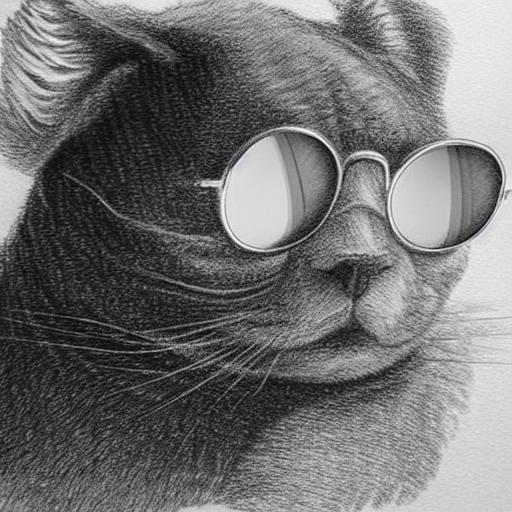}
\end{subfigure}
  \begin{subfigure}[t]{0.13\linewidth}
    \includegraphics[width=\linewidth]{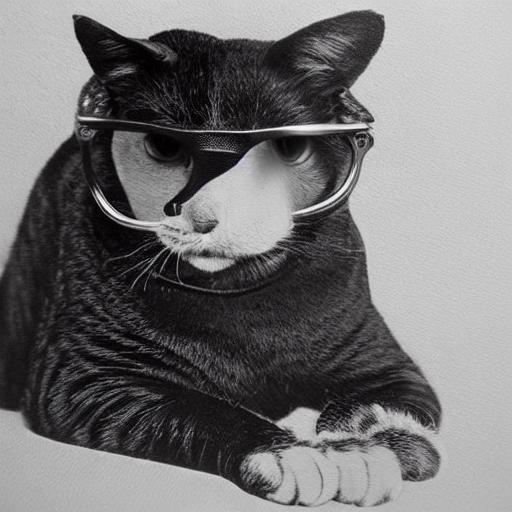}
\end{subfigure}
  \begin{subfigure}[t]{0.13\linewidth}
    \includegraphics[width=\linewidth]{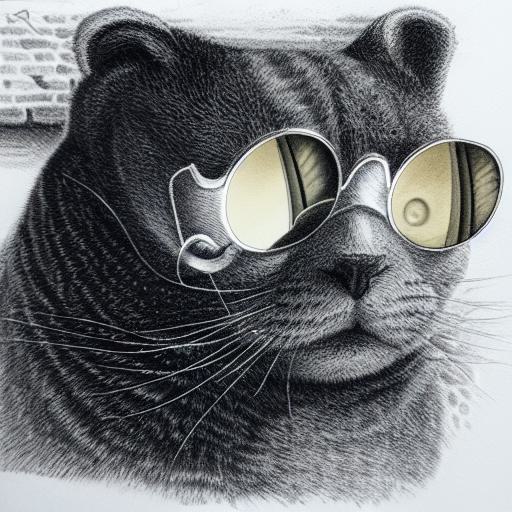}
\end{subfigure}
  \begin{tikzpicture}
    \draw (0, 0) node[inner sep=0] {\includegraphics[width=0.13\linewidth]{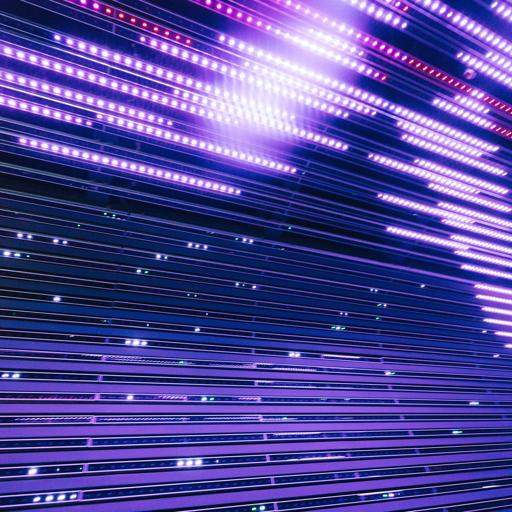}};
    \node (A) at (1.4, -0.8) {};
    \node (B) at (3, -0.8) {};
    \draw[->] (A) edge (B);
    \node[align=center] (C) at (2.2, 0.2) {Stylize\\``City''};
  \end{tikzpicture}
  \begin{subfigure}[t]{0.13\linewidth}
    \includegraphics[width=\linewidth]{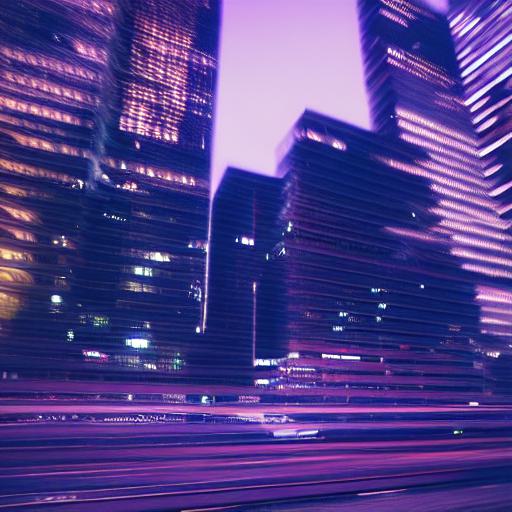}
\end{subfigure}
  \begin{subfigure}[t]{0.13\linewidth}
    \includegraphics[width=\linewidth]{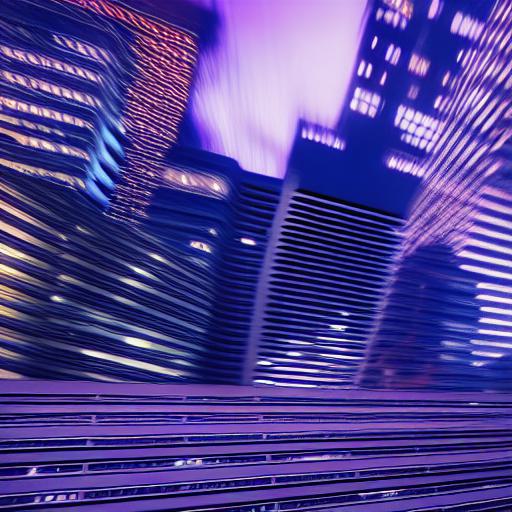}
\end{subfigure}
  \begin{subfigure}[t]{0.13\linewidth}
    \includegraphics[width=\linewidth]{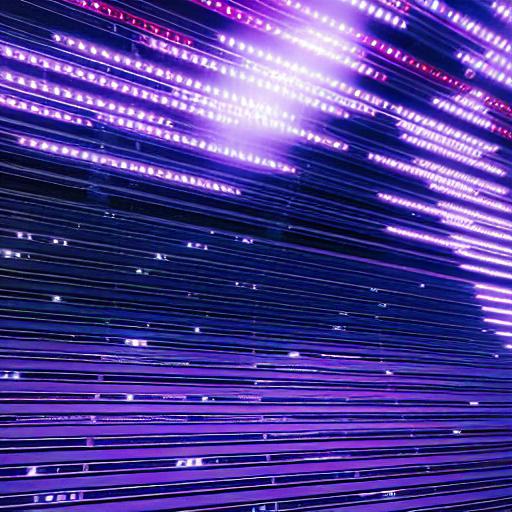}
\end{subfigure}
  \begin{subfigure}[t]{0.13\linewidth}
    \includegraphics[width=\linewidth]{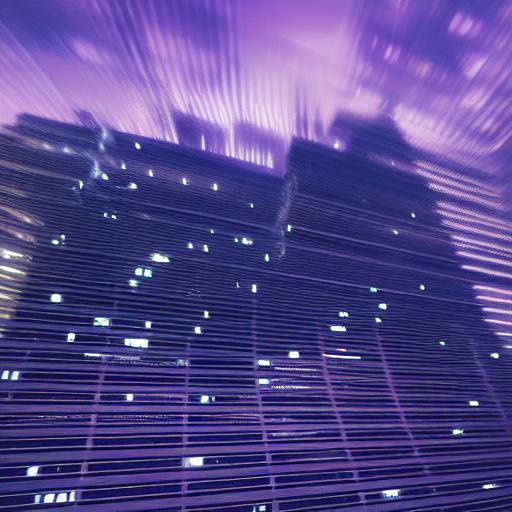}
\end{subfigure}
  \begin{subfigure}[t]{0.13\linewidth}
    \includegraphics[width=\linewidth]{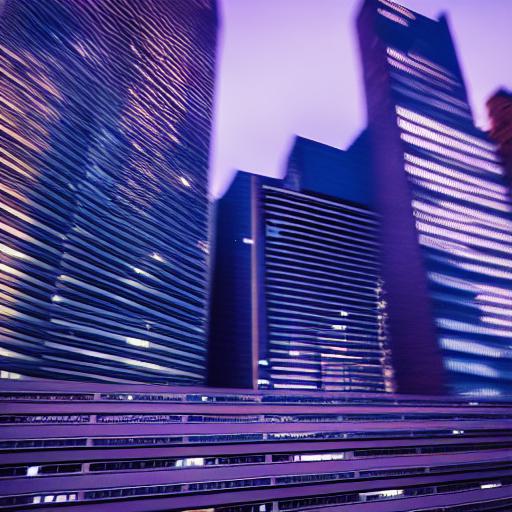}
\end{subfigure}
  \begin{tikzpicture}
    \draw (0, 0) node[inner sep=0] {\includegraphics[width=0.13\linewidth]{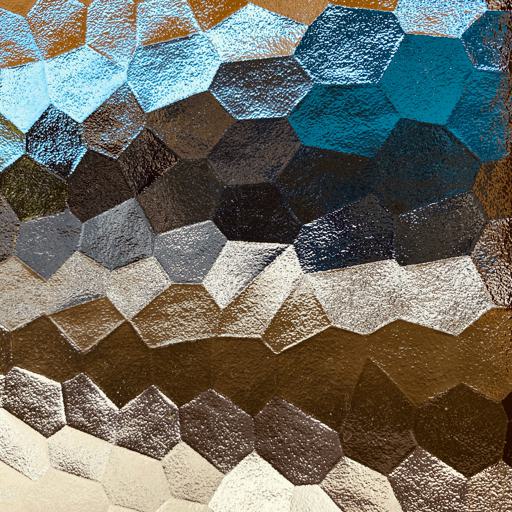}};
    \node (A) at (1.4, -0.8) {};
    \node (B) at (3, -0.8) {};
    \draw[->] (A) edge (B);
    \node[align=center] (C) at (2.2, 0.2) {Stylize\\``Chair''};
  \end{tikzpicture}
  \begin{subfigure}[t]{0.13\linewidth}
    \includegraphics[width=\linewidth]{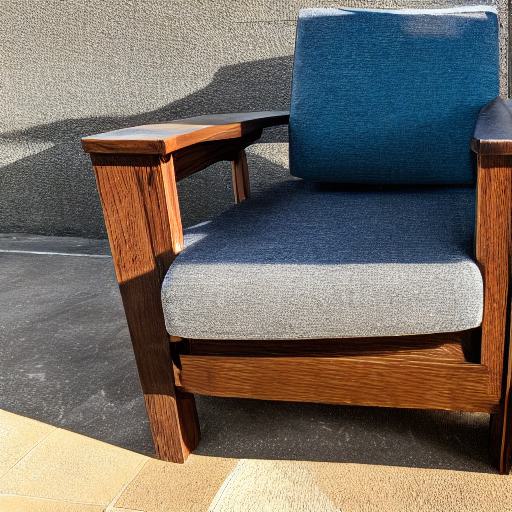}
\end{subfigure}
  \begin{subfigure}[t]{0.13\linewidth}
    \includegraphics[width=\linewidth]{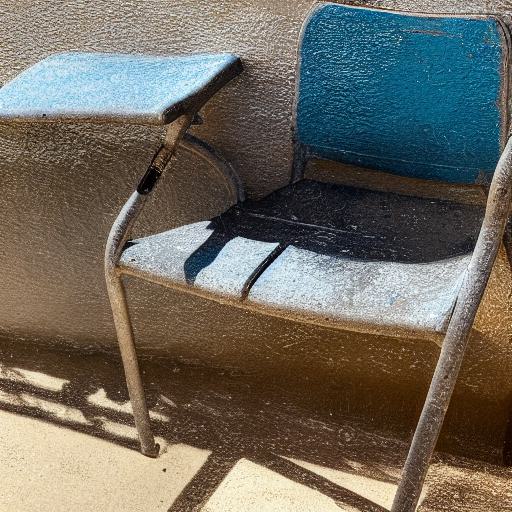}
\end{subfigure}
  \begin{subfigure}[t]{0.13\linewidth}
    \includegraphics[width=\linewidth]{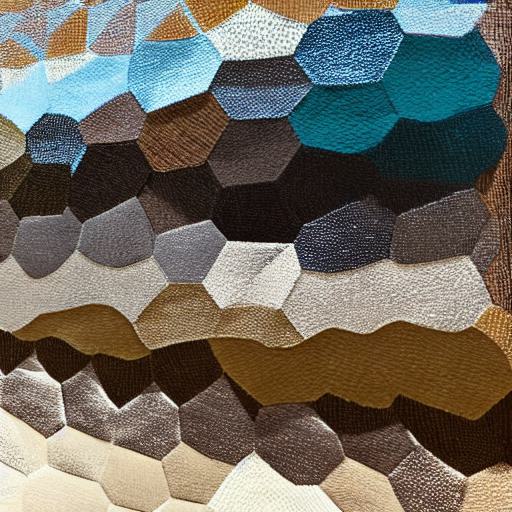}
\end{subfigure}
  \begin{subfigure}[t]{0.13\linewidth}
    \includegraphics[width=\linewidth]{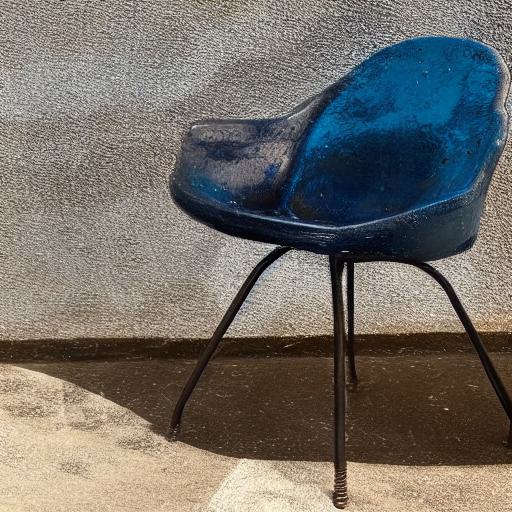}
\end{subfigure}
  \begin{subfigure}[t]{0.13\linewidth}
    \includegraphics[width=\linewidth]{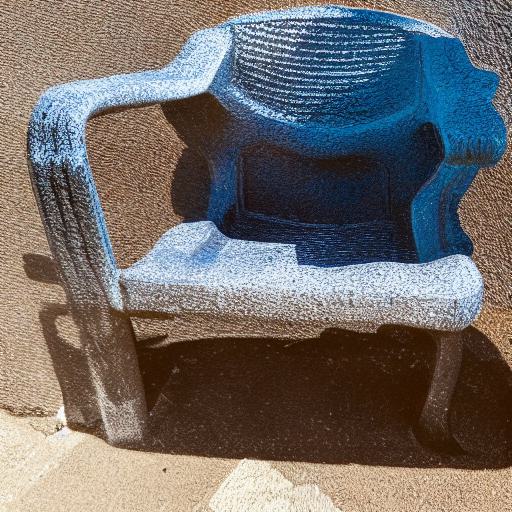}
\end{subfigure}
\vspace{-0pt}
\caption{Text-guided global editing results.}
\label{global-editing}
\end{figure*}

We visualize the results in \cref{global-editing}. While in local editing MDP-$x_t$ and MDP-$c$ can yield good results, in global editing they are either not able to do the edits, or the overall layout is changed too much. In contrast, we observe that MDP-$\epsilon_t$ can produce very good results most of the time even when the strong baseline method Prompt-to-Prompt (P2P) fails. Further, we compare the editing results for MDP-$x_t$ MDP-$c$ and MDP-$\epsilon_t$ under different ranges to inject layout from the input image~\cref{fig:comparison}. We observe that MDP-$\epsilon_t$ can create meaningful edits when the manipulation starts at early or later stages, while MDP-$x_t$ only works when starting the manipulation in the initial timestep and manipulation MDP-$c$ cannot faithfully preserve the layout for any range. Also, MDP-$\beta$ sometimes fails. We therefore strongly recommend using MDP-$\epsilon_t$ when doing global editing, as it is more stable and can provide a variety of meaningful results. 
Note that all of manipulations can do stylization while we failed to find a proper setting of Prompt-to-Prompt for this application. We conjecture that Prompt-to-Prompt's manipulation of the attention maps, is good for fine-grained control, while it constrains edits to be local and it also inherits some limitations of the language model. While in our manipulation, especially for MDP-$\epsilon_t$, we do not directly manipulate the attention maps, but keep the editing ability by manipulating the diffusion paths.
\subsubsection{Class condition models}
We provide class-guided editing results using class-condition diffusion model instead of the common text-condition model. We show the images edited by MDP-$\epsilon_t$ in \cref{fig:class}.

\section{Conclusion and Future work}
We introduce a generalized editing framework, MDP,
which contains 5 different manipulations that are suitable, their parameters, and the manipulation schedule. We highlight a new manipulation by editing the predicted noise. 
\clearpage
The results show that this manipulation can achieve high quality edits in challenging cases where previous work may fail. 
Our work also has multiple limitations. 
We cannot claim that our framework is exhaustive. 
We did not analyze editing operations that become possible using fine-tuning or modifications of the network architecture. Also, we rely on the reconstruction ability of inversion for real images. In some cases the inversion may fail to faithfully reconstruct the input. An interesting topic for future work is video editing which we plan to tackle next.

{\small
\bibliographystyle{ieee_fullname}

\bibliography{egbib}
}

\clearpage

\iccvfinalcopy 

\def\iccvPaperID{9189} 
\def\httilde{\mbox{\tt\raisebox{-.5ex}{\symbol{126}}}}

\ificcvfinal\pagestyle{empty}\fi
\appendix
\section{More details}
\subsection{Diffusion Inversion}
We can edit an image using its latent representation. Diffusion inversion tries to revert the generating process, \ie, given an image $\mathbf{x}_0$, to find the initial noise $\mathbf{x}_T$ (and the intermediate noised images $\mathbf{x}_t$) which generates $\mathbf{x}_0$. The first approach works by adding noise to $\mathbf{x}_0$. The noise schedule is exactly the same as the training noise schedule of DDPM. However, this process is stochastic and we cannot guarantee the obtained initial noise $\mathbf{x}_T$ can reconstruct $\mathbf{x}_0$ faithfully. The second approach is DDIM inversion based on the DDIM sampler in Eq.~\eqref{eq:ddim-sampler}. We can calculate the initial noise by applying the formula iteratively, $\mathbf{x}_{t+1}=\sqrt{\alpha_{t+1}}\cdot f_\theta(\mathbf{x}_t, \mathbf{c}, t) + \sqrt{1-\alpha_{t+1}}\cdot\boldsymbol\epsilon_\theta(\mathbf{x}_t, \mathbf{c}, t)$. But when working with classifier-free guidance, the reconstruction quality is not satisfying. The third inversion method is Null-text Inversion. It is based on the DDIM inversion and optimized for high reconstruction quality. In most cases, we will use the Null-text Inversion as our diffusion inversion method.

\subsection{Algorithms}
We show the algorithm MDP-$\boldsymbol\epsilon_t$ in \cref{alg:eps}, MDP-$\mathbf{c}$ in \cref{alg:c}, MDP-$\mathbf{x}_t$ in \cref{alg:x}, and MDP-$\boldsymbol \beta$ in \cref{alg:beta}. For real image editing, we use Null-text inversion to obtain $\mathbf{x}_T$. For synthetic image editing, $\mathbf{x}_T$ is sampled from the standard Gaussian distribution.

\begin{algorithm}[H]
\caption{MDP-$\boldsymbol\epsilon_t$. }\label{alg:eps}
\begin{algorithmic}
\Require $\{\omega_t\}_{t=[T,\dots, 0]}, \mathbf{x}_{T}$, $\mathbf{c}^{(A)}$, $\mathbf{c}^{(B)}$, $\left\{\left(\mathbf{x}_t^{(A)}, \boldsymbol\epsilon_t^{(A)}\right)\right\}_{t=[T,\dots, 0]}=\mathrm{GenPath}\left(\mathbf{x}_T, \mathbf{c}^{(A)}, t\right)$
\State $\mathbf{x}_T^{(\star)} = \mathbf{x}_T$
\For{$t$ in $\{T, T-1, \dots, 1\}$}
    \State $\boldsymbol\epsilon_t^{(B)}=\boldsymbol\epsilon_\theta\left(\mathbf{x}_t^{(\star)}, \mathbf{c}^{(B)}, t\right)$
    \State $ \boldsymbol\epsilon_t^{(\star)} = (1-\omega_t) \boldsymbol\epsilon_t^{(B)} + \omega_t \boldsymbol\epsilon_t^{(A)} $
    \State $\mathbf{x}_{t-1}^{(\star)} = \mathrm{DDIM}\left( \mathbf{x}_t^{(\star)}, \boldsymbol\epsilon_t^{(\star)}, t\right)$
\EndFor
\end{algorithmic}
\end{algorithm}
\begin{algorithm}[H]
\caption{MDP-$\mathbf{c}$. }\label{alg:c}
\begin{algorithmic}
\Require $\{\omega_t\}_{t=[T,\dots, 0]}, \mathbf{x}_{T}$, $\mathbf{c}^{(A)}$, $\mathbf{c}^{(B)}$
\State $\mathbf{x}_T^{(\star)} = \mathbf{x}_T$
\For{$t$ in $\{T, T-1, \dots, 1\}$}
    \State $\mathbf{c}_t^{(\star)} = (1-\omega_t) \mathbf{c}^{(B)} + \omega_t \mathbf{c}^{(A)} $
    \State $\boldsymbol\epsilon_t^{(\star)}=\boldsymbol\epsilon_\theta\left(\mathbf{x}_t^{(\star)}, \mathbf{c}_t^{(\star)}, t\right)$
    \State $\mathbf{x}_{t-1}^{(\star)} = \mathrm{DDIM}\left( \mathbf{x}_t^{(\star)}, \boldsymbol\epsilon_t^{(\star)}, t\right)$
\EndFor
\end{algorithmic}
\end{algorithm}
\begin{algorithm}[H]
\caption{MDP-$\mathbf{x}_t$. }\label{alg:x}
\begin{algorithmic}
\Require $\{\omega_t\}_{t=[T,\dots, 0]}, \mathbf{x}_{T}$, $\mathbf{c}^{(A)}$, $\mathbf{c}^{(B)}$, $\left\{\left(\mathbf{x}_t^{(A)}, \boldsymbol\epsilon_t^{(A)}\right)\right\}_{t=[T,\dots, 0]}=\mathrm{GenPath}\left(\mathbf{x}_T, \mathbf{c}^{(A)}, t\right)$
\State $\mathbf{x}_T^{(\star)} = \mathbf{x}_T$
\For{$t$ in $\{T, T-1, \dots, 1\}$}
    \State $\boldsymbol\epsilon_t^{(B\star)}=\boldsymbol\epsilon_\theta\left(\mathbf{x}_t^{(\star)}, \mathbf{c}_t^{(B)}, t\right)$
    \State $\mathbf{x}_{t-1}^{(B\star)} = \mathrm{DDIM}\left( \mathbf{x}_t^{(\star)}, \boldsymbol\epsilon_t^{(B\star)}, t\right)$
    \State $\mathbf{x}_{t-1}^{(\star)} = (1-\omega_t) \mathbf{x}^{(B\star)}_{t-1} + \omega_t \mathbf{x}^{(A)}_{t-1} $
\EndFor
\end{algorithmic}
\end{algorithm}
\begin{algorithm}[H]
\caption{MDP-$\boldsymbol \beta$. }\label{alg:beta}
\begin{algorithmic}
\Require $\{\omega_t\}_{t=[T,\dots, 0]}, \mathbf{x}_{T}$, $\mathbf{c}^{(A)}$, $\mathbf{c}^{(B)}$
\State $\mathbf{x}_T^{(\star)} = \mathbf{x}_T$
\For{$t$ in $\{T, T-1, \dots, 1\}$}
    \State $\boldsymbol\epsilon_t^{(A\star)}=\boldsymbol\epsilon_\theta\left(\mathbf{x}_t^{(\star)}, \mathbf{c}^{(A)}, t\right)$
    \State $\boldsymbol\epsilon_t^{(B\star)}=\boldsymbol\epsilon_\theta\left(\mathbf{x}_t^{(\star)}, \mathbf{c}^{(B)}, t\right)$
    \State $\boldsymbol\epsilon_t^{(\star)} = (1-\omega_t) \boldsymbol\epsilon_t^{(B\star)} + \omega_t \boldsymbol\epsilon_t^{(A\star)}$
    \State $\mathbf{x}_{t-1}^{(\star)} = \mathrm{DDIM}\left( \mathbf{x}_t^{(\star)}, \boldsymbol\epsilon_t^{(\star)}, t\right)$
\EndFor
\end{algorithmic}
\end{algorithm}

\section{Taxonomy of Image Editing Applications}
\label{sec:taxonomy}
In this subsection we describe a small taxonomy for text-guided image editing applications that we use to analyze our framework and compare it to previous work. To be noted that this taxonomy only includes some major image editing applications but is not an exhaustive one. Given either a real or synthetic image as input, we edit the image following a condition such as a text prompt or a class label. We divide all common image editing operations into two categories: local editing and global editing. We further identify sub-categories for both local and global editing.
\subsection{Local editing}
Typically, we want to perform edits while keeping the overall image layout, \eg background or the shape of the chosen object in the input image.
\begin{itemize}
    \item Changing object: change an object (or objects) in the image to another one. For example, if there is a basket of \textit{apples}, we may change it to a basket of \textit{oranges}. Also, we may change the image of a \textit{dog} to an image of a \textit{cat}.
    \item Adding object: add an object that does not exist in the original image. For example, given an image of a forest, we add a \textit{car} in that forest. Given a cat face, we can add a pair of \textit{sunglasses} on it.
    \item Removing object: remove an object from the image. For example, we can remove the eggs in a basket so that the basket becomes \textit{empty}. 
    \item Changing attribute: change the attribute such as color and texture of an object. Given a \textit{red} bird, we may change it to a \textit{blue} bird. We may change a human portrait from a \textit{young} person to an \textit{old} person.
    \item Mixing objects: combine an object in the input image with another object. The two objects we may want to mix can have different semantics, \eg we can mix a \textit{corgi} and a \textit{coffee machine}, or a \textit{chocolate bar} and a \textit{purse}.  
\end{itemize}
\subsection{Global editing}
For global editing, the overall texture and style of the image can be changed while the layout and semantics should be the same.
\begin{itemize}
    \item Changing background: change the background while keeping a foreground object untouched. For example, we change a rabbit on \textit{grass} to a rabbit on the \textit{moon}. We can also convert a black-and-white photo to be \textit{colored}. 
    \item Stylization: stylize an object or a scene using an input style image. For example, we give an \textit{futuristic} image as a style input and we want the a generated image to have the same style.
    \item In-domain transfer: edit the input image by performing in-domain changes. For example, we transfer a photo of a valley in \textit{summer} to a photo of a valley in \textit{winter}, or we transfer a city during the \textit{day} to a city at \textit{night}. 
    \item Out-of-domain transfer: edit the input image with out-of-domain changes. For example, we transfer a \textit{photo} of a valley to an \textit{oil painting} of a valley or we change a \textit{portrait} of a human to a \textit{cartoon} character. We can also change a \textit{black-and-white} photo to a \textit{colored} photo. 
\end{itemize}

\section{More Analysis in the Design Space}
\begin{figure*} [h]
\centering
\begin{subfigure}[t]{0.4\linewidth}
\includegraphics[width=\linewidth]{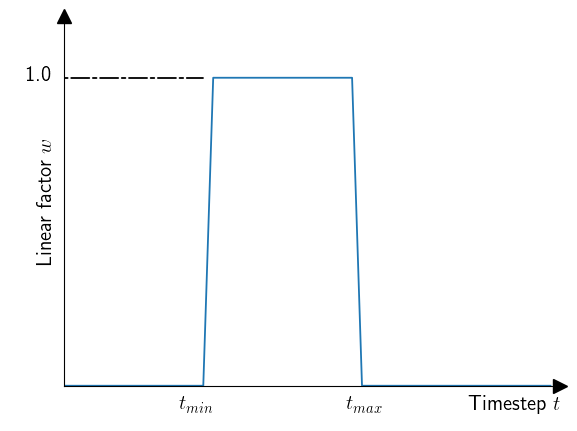} 
\caption{}
\label{fig:tri}
\end{subfigure}
\begin{subfigure}[t]{0.4\linewidth}
\includegraphics[width=\linewidth]{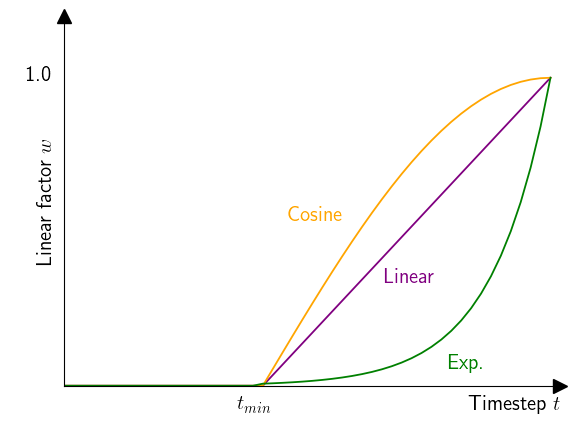} 
\caption{}
\label{fig:linear}
\end{subfigure}
\caption{Linear schedule is shown in (a), while linear, cosine, and exponential schedule are shown in purple, orange, and green, respectively, in (b). For linear schedule, we fix the linear factor as 1.0 while varying $t_{\text{max}}$ and $t_{\text{min}}$ by varying $T_M$. For the other three schedules, we fix the $t_{\text{max}} = 50$ then vary scale factors and $t_{\text{min}}$.}
\label{fig:schedules}
\end{figure*}

\begin{figure*}
\centering
\setlength{\tabcolsep}{1pt}
\begin{tabular}{cccccc}
 $t_{\text{max}}$/$T_M$ & 5 & 10 & 15 & 20 & 25 \\
 50 & \includegraphics[align=c,width=0.14\linewidth]{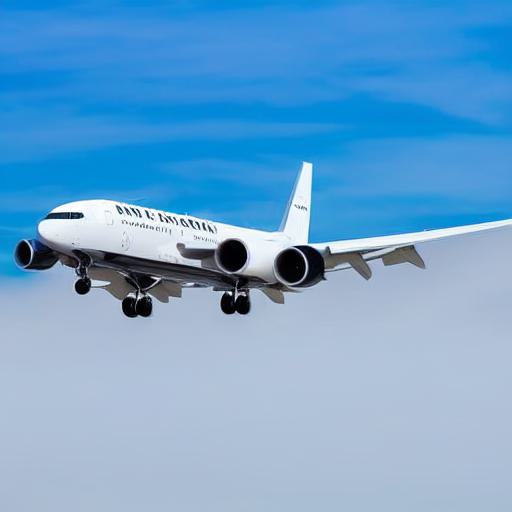}  & \includegraphics[align=c,width=0.14\linewidth]{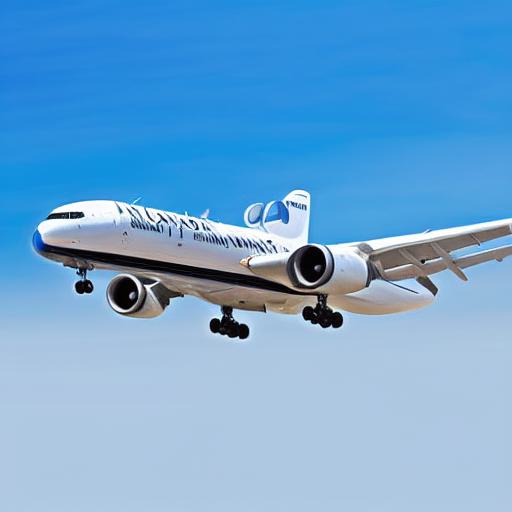}  & \includegraphics[align=c,width=0.14\linewidth]{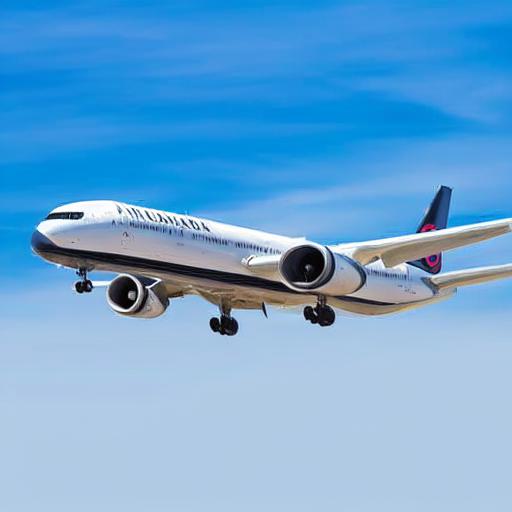}  & \includegraphics[align=c,width=0.14\linewidth]{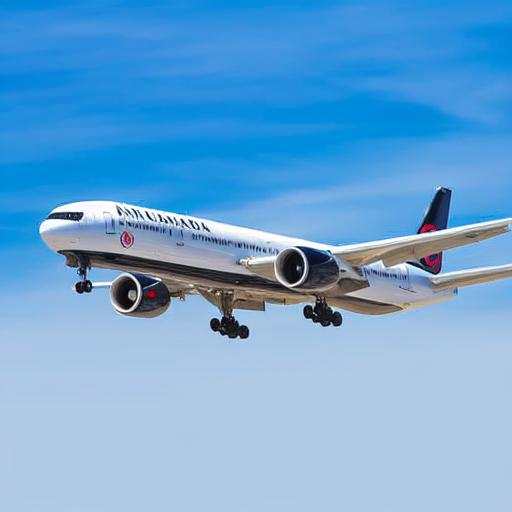}  & \includegraphics[align=c,width=0.14\linewidth]{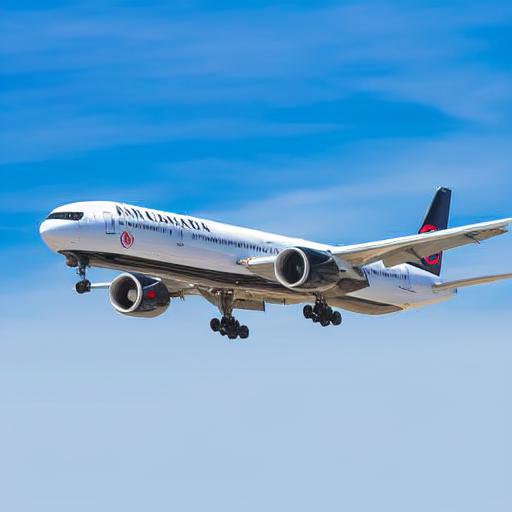}  \\
 48 & \includegraphics[align=c,width=0.14\linewidth]{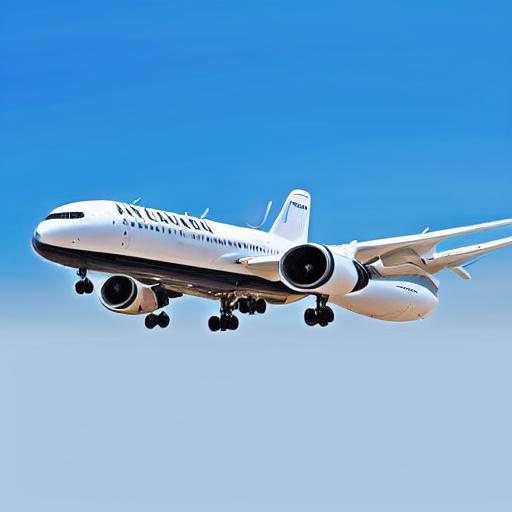}  & \includegraphics[align=c,width=0.14\linewidth]{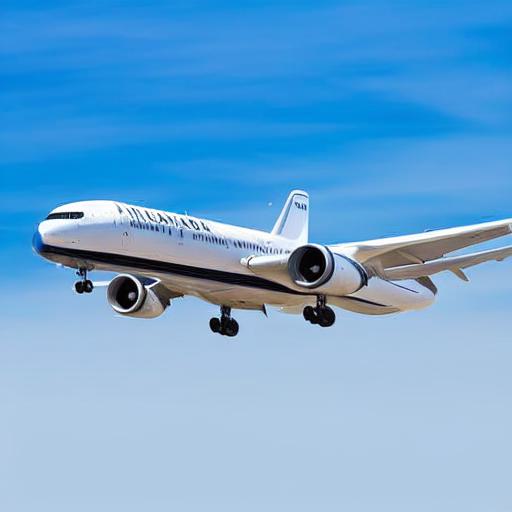}  & \includegraphics[align=c,width=0.14\linewidth]{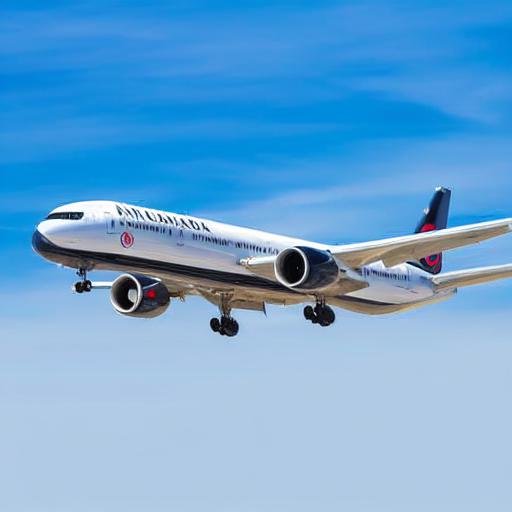}  & \includegraphics[align=c,width=0.14\linewidth]{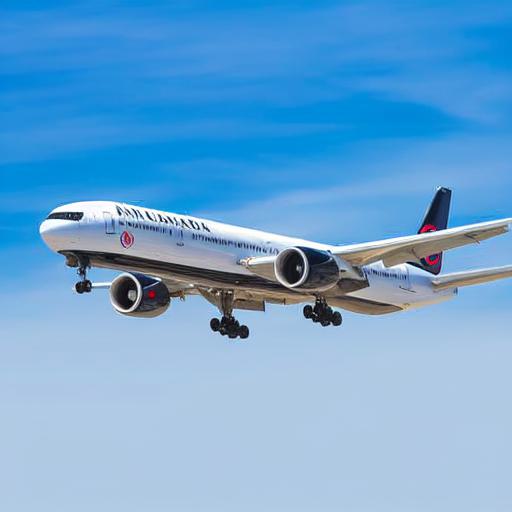}  & \includegraphics[align=c,width=0.14\linewidth]{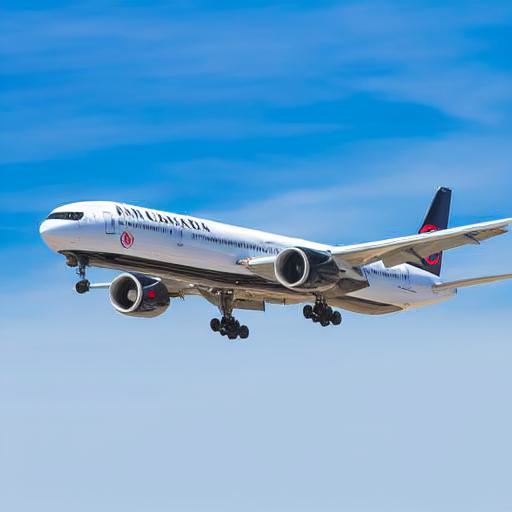}  \\
 46 & \includegraphics[align=c,width=0.14\linewidth]{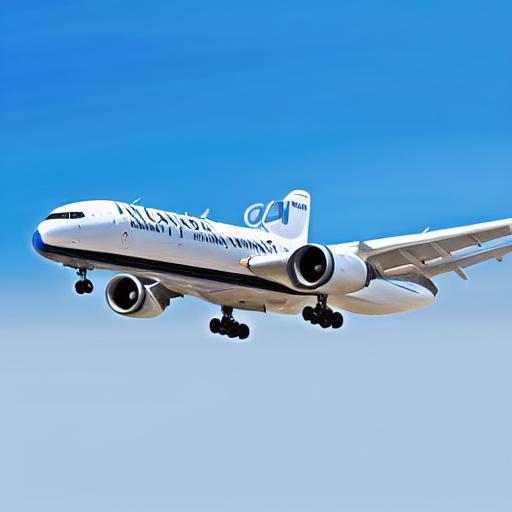}  & \includegraphics[align=c,width=0.14\linewidth]{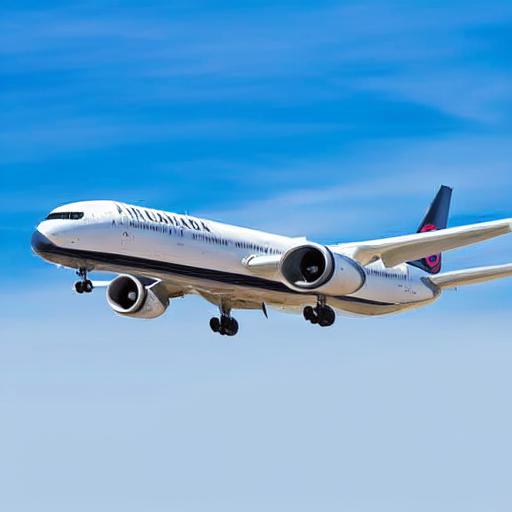}  & \includegraphics[align=c,width=0.14\linewidth]{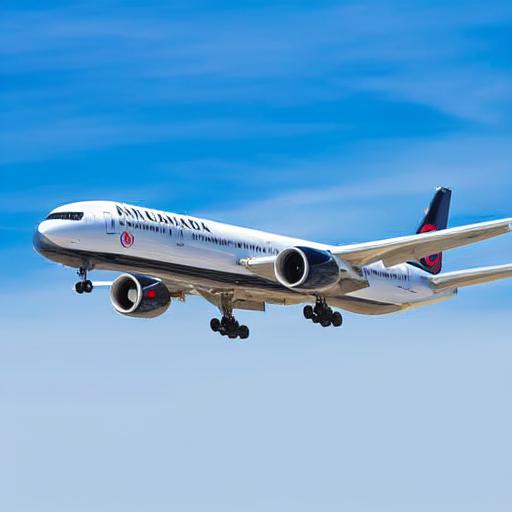}  & \includegraphics[align=c,width=0.14\linewidth]{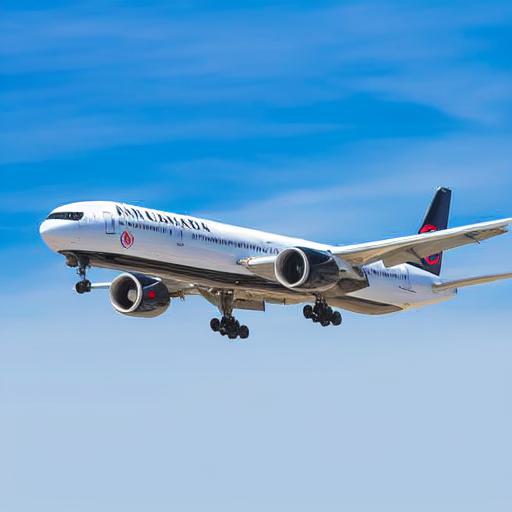}  & \includegraphics[align=c,width=0.14\linewidth]{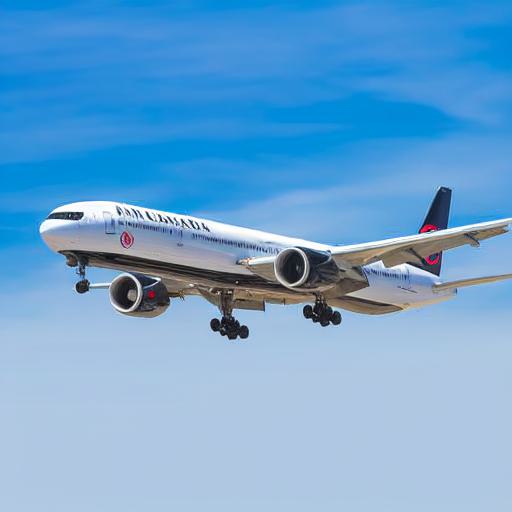}  \\
 44 & \includegraphics[align=c,width=0.14\linewidth]{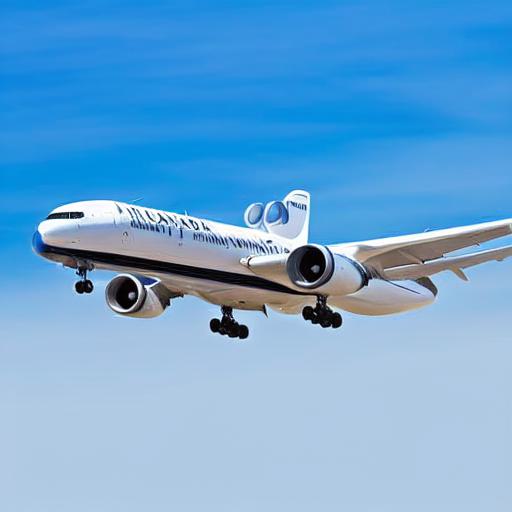}  & 
 \includegraphics[align=c,width=0.14\linewidth]{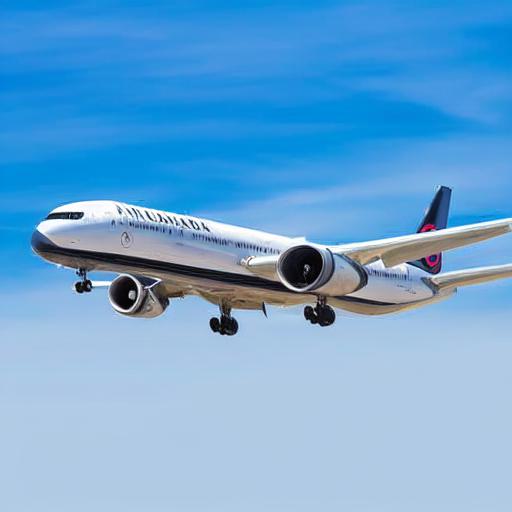}  & \includegraphics[align=c,width=0.14\linewidth]{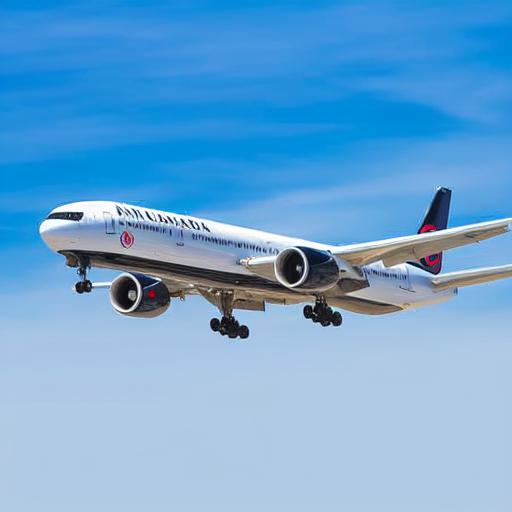}  & \includegraphics[align=c,width=0.14\linewidth]{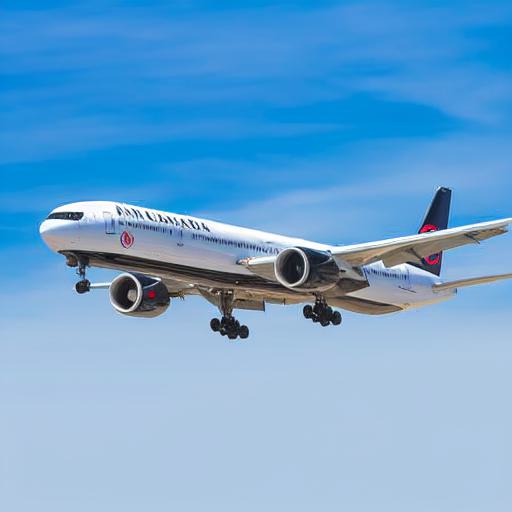}  & \includegraphics[align=c,width=0.14\linewidth]{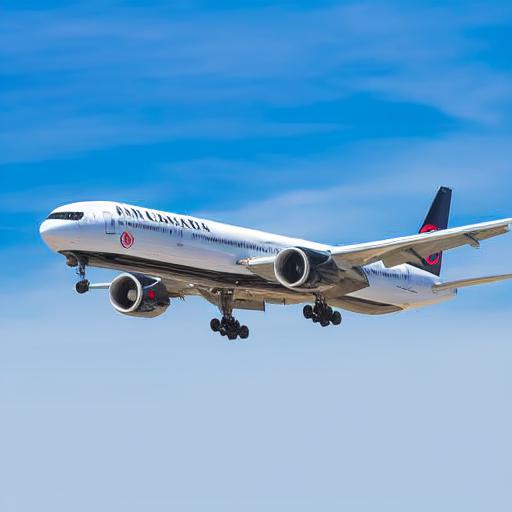}  \\
 42 & \includegraphics[align=c,width=0.14\linewidth]{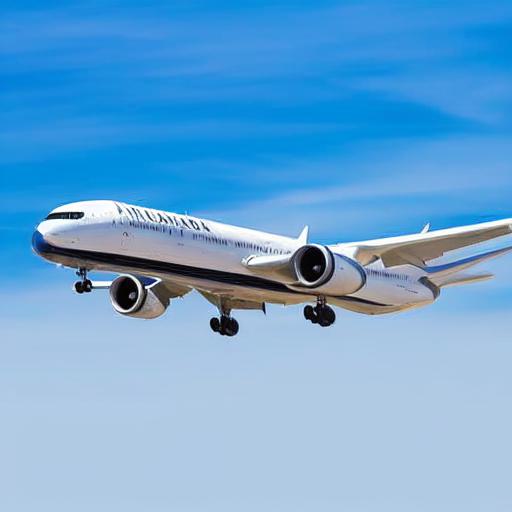} & \includegraphics[align=c,width=0.14\linewidth]{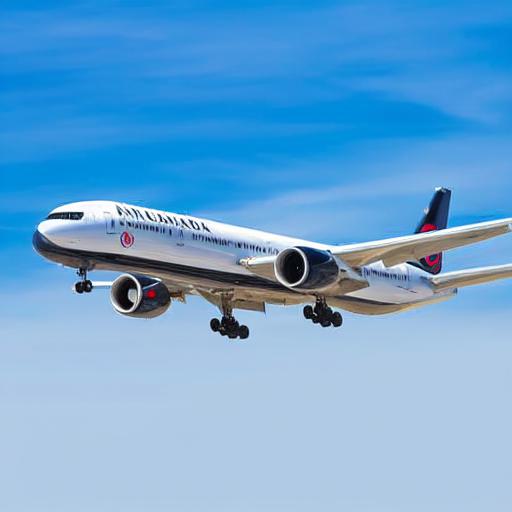} & \includegraphics[align=c,width=0.14\linewidth]{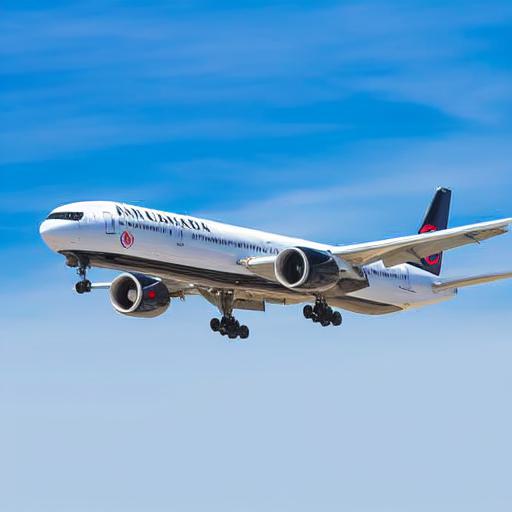} & \includegraphics[align=c,width=0.14\linewidth]{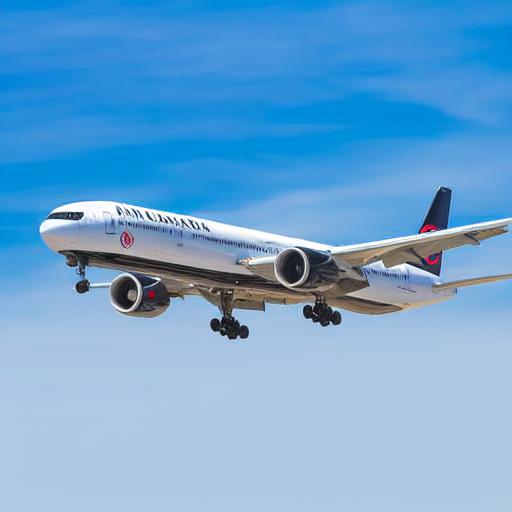} & \includegraphics[align=c,width=0.14\linewidth]{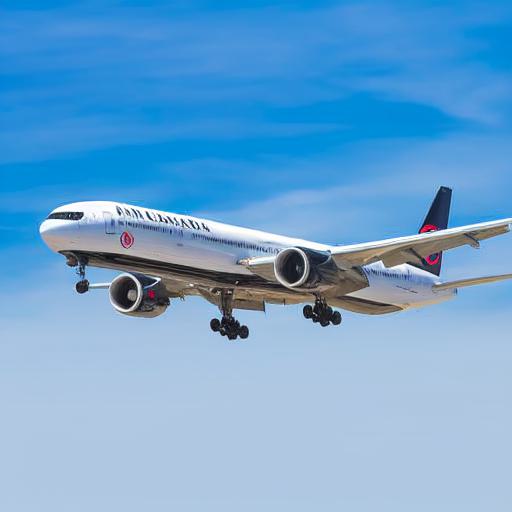}
\end{tabular}
\caption{Results of \textbf{MDP-$\mathbf{x}_t$} using constant schedule.}
\label{fig:xt_tri}
\end{figure*}

\begin{figure*}
\centering
\setlength{\tabcolsep}{1pt}
\begin{tabular}{cccccccc}
Type/$T_M$ & 20 & 25 & 30 & 35 & 40 & 45 & 50 \\
Linear & \includegraphics[align=c,width=0.13\linewidth]{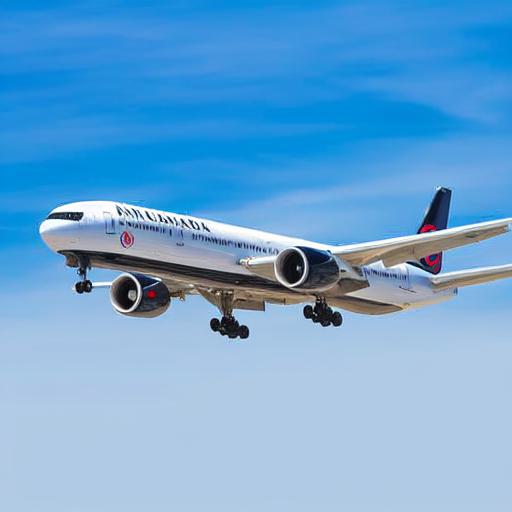} & \includegraphics[align=c,width=0.13\linewidth]{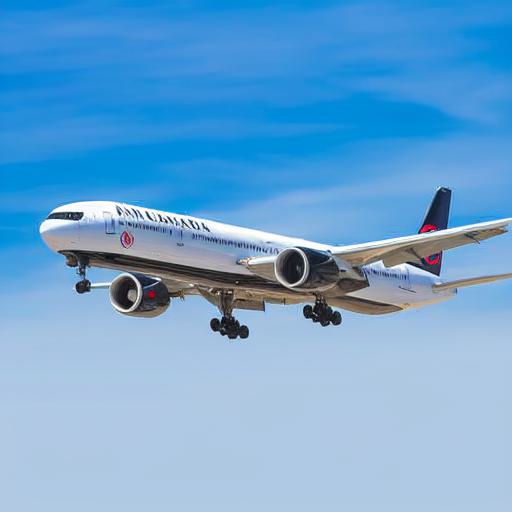} & \includegraphics[align=c,width=0.13\linewidth]{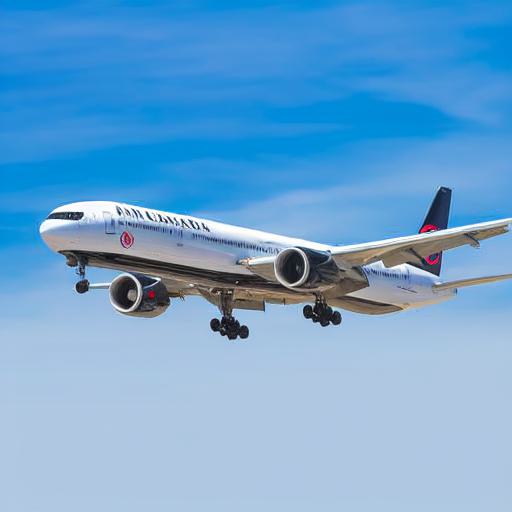} & \includegraphics[align=c,width=0.13\linewidth]{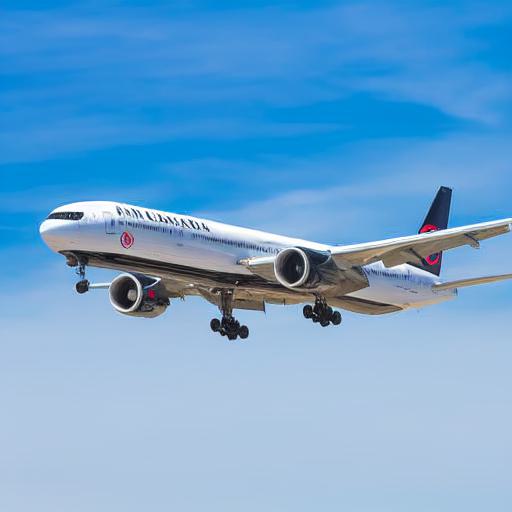} & \includegraphics[align=c,width=0.13\linewidth]{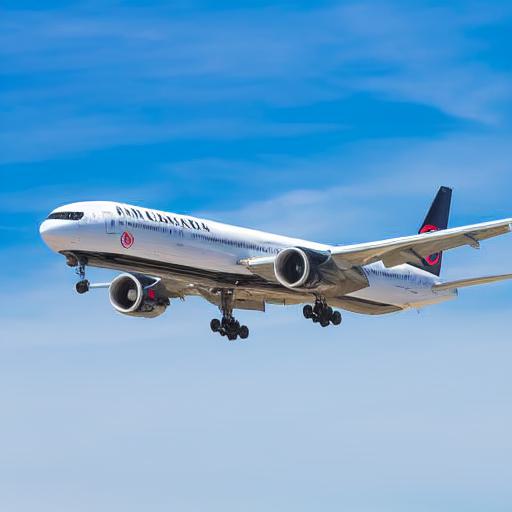} & \includegraphics[align=c,width=0.13\linewidth]{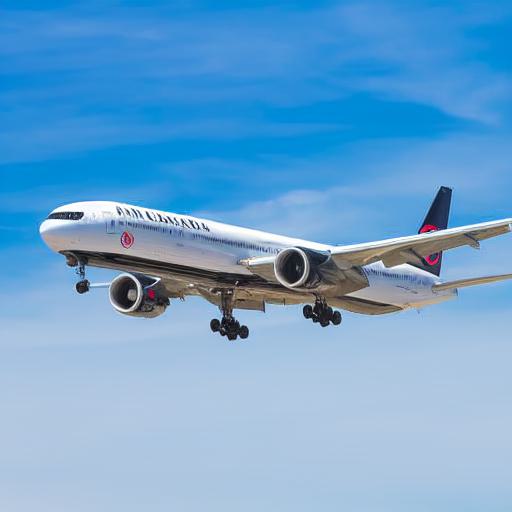} & \includegraphics[align=c,width=0.13\linewidth]{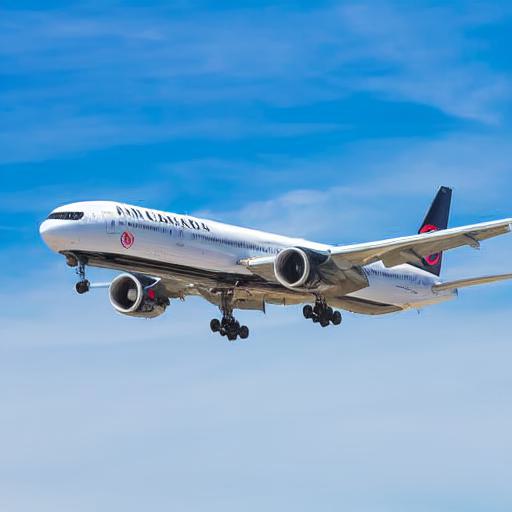} \\
Cosine & 
\includegraphics[align=c,width=0.13\linewidth]{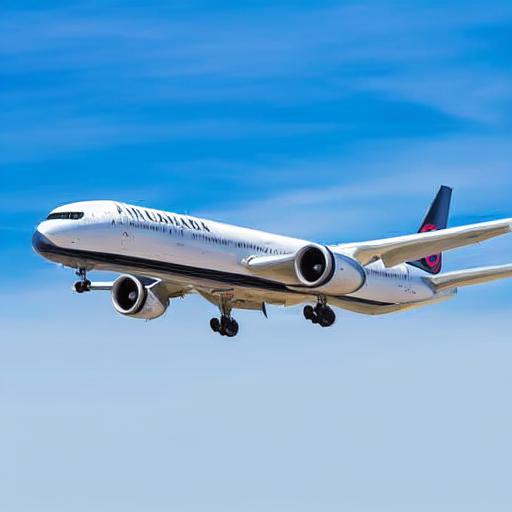} & \includegraphics[align=c,width=0.13\linewidth]{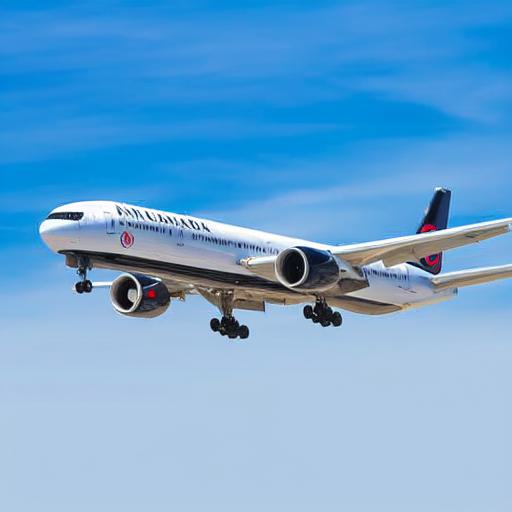} & \includegraphics[align=c,width=0.13\linewidth]{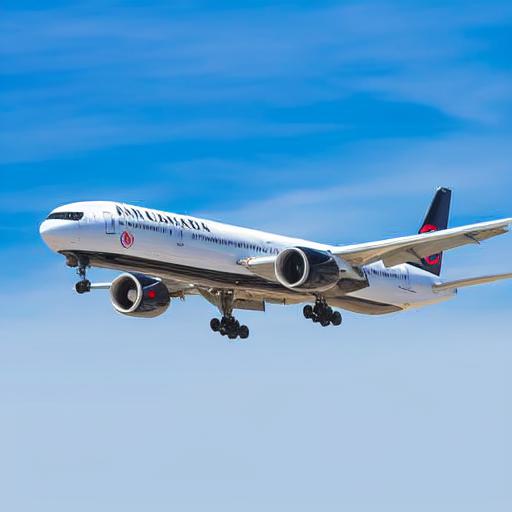} & \includegraphics[align=c,width=0.13\linewidth]{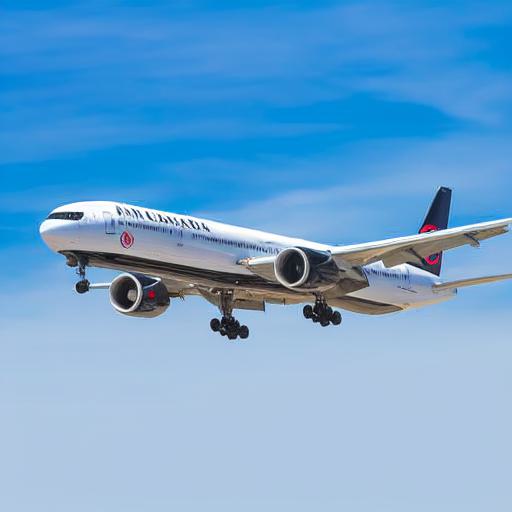} & \includegraphics[align=c,width=0.13\linewidth]{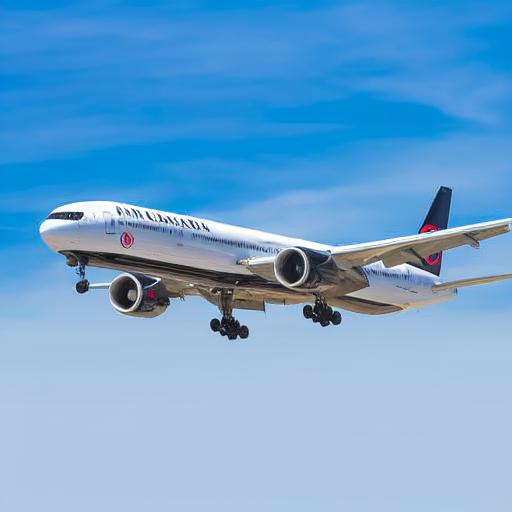} & \includegraphics[align=c,width=0.13\linewidth]{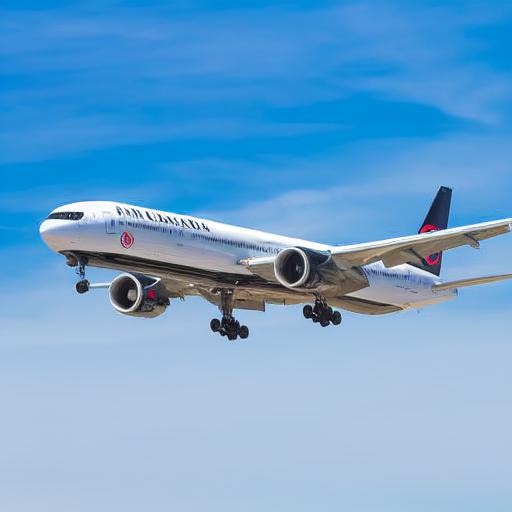} & \includegraphics[align=c,width=0.13\linewidth]{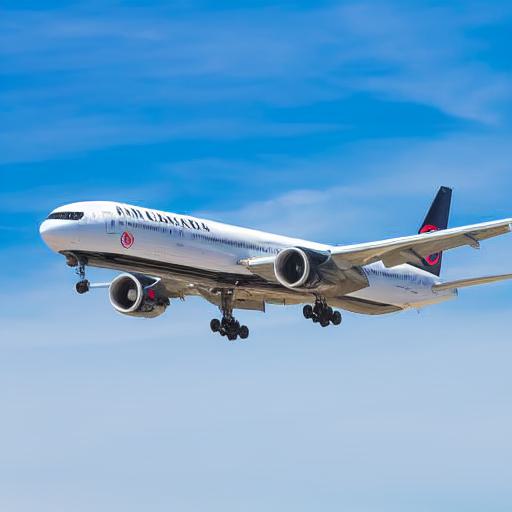} \\
Exponential & 
\includegraphics[align=c,width=0.13\linewidth]{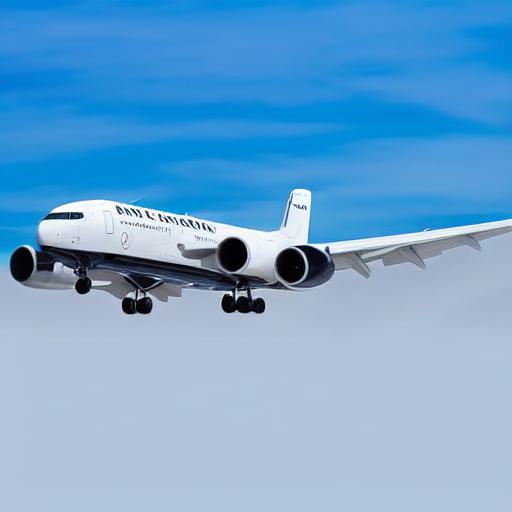} & \includegraphics[align=c,width=0.13\linewidth]{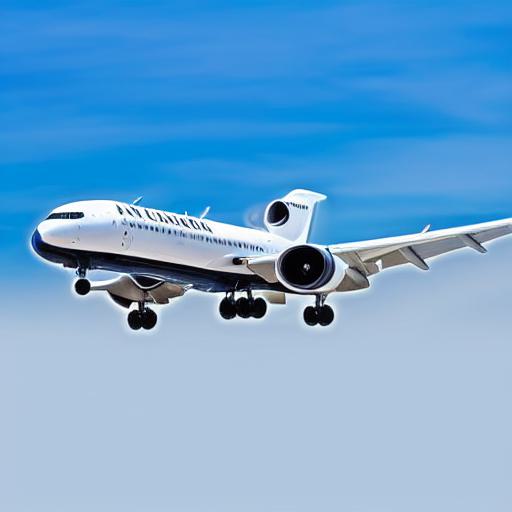} & \includegraphics[align=c,width=0.13\linewidth]{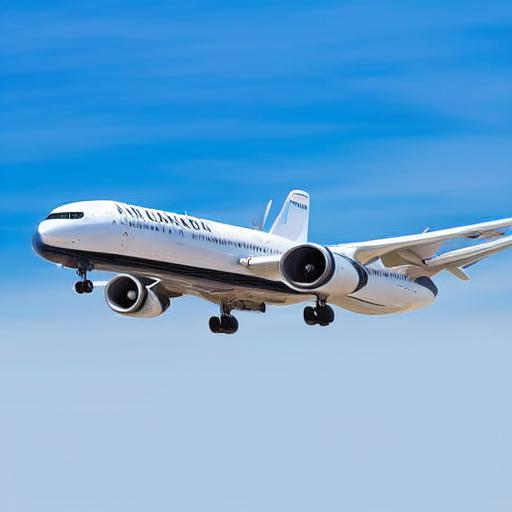} & \includegraphics[align=c,width=0.13\linewidth]{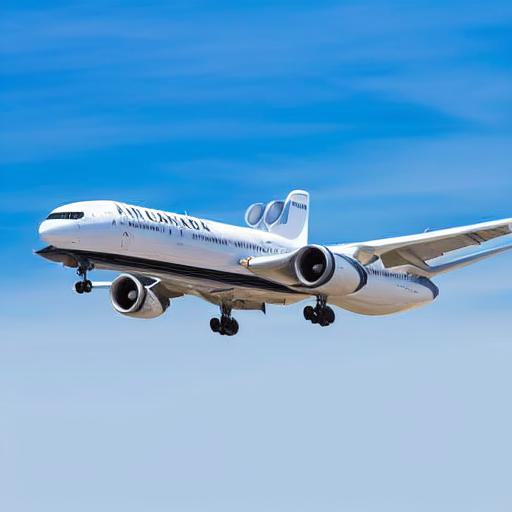} & \includegraphics[align=c,width=0.13\linewidth]{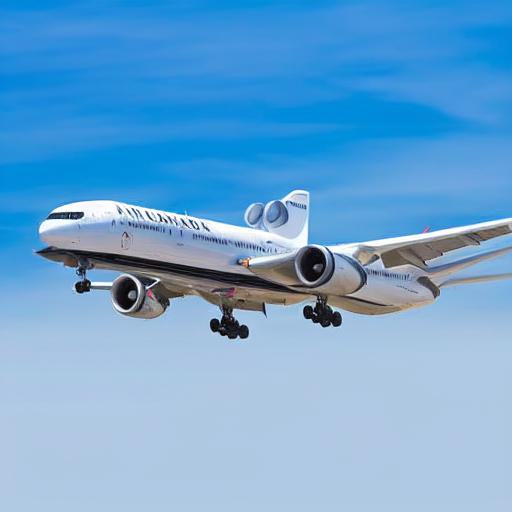} & \includegraphics[align=c,width=0.13\linewidth]{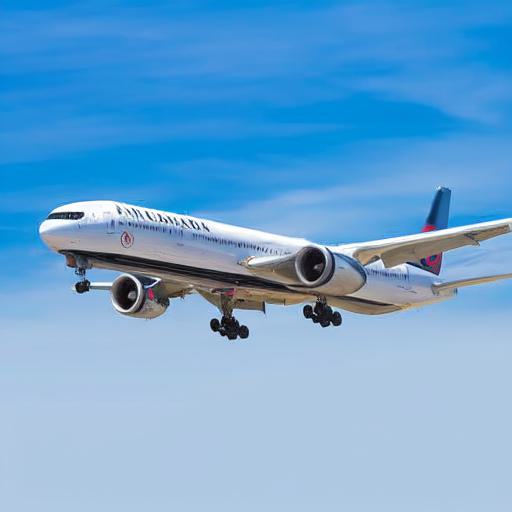} & \includegraphics[align=c,width=0.13\linewidth]{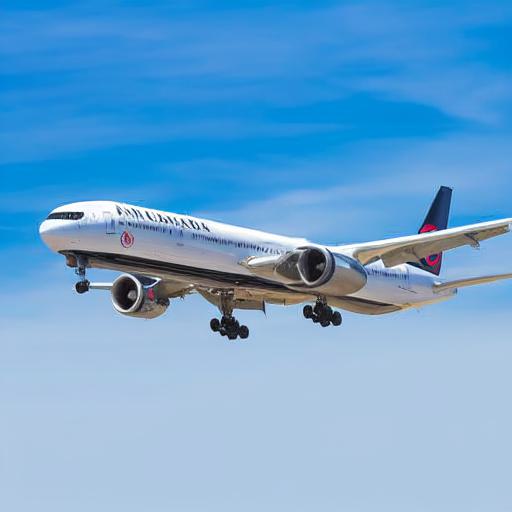} \\
\end{tabular}
\caption{Results of \textbf{MDP-$\mathbf{x}_t$} using linear, cosine and exponential schedule.}
\label{fig:xt_linear}
\end{figure*}

\begin{figure*}
\centering
\setlength{\tabcolsep}{1pt}
\begin{tabular}{cccccc}
 $t_{\text{max}}$/$T_M$ & 5 & 10 & 15 & 20 & 25 \\
 50 & \includegraphics[align=c,width=0.14\linewidth]{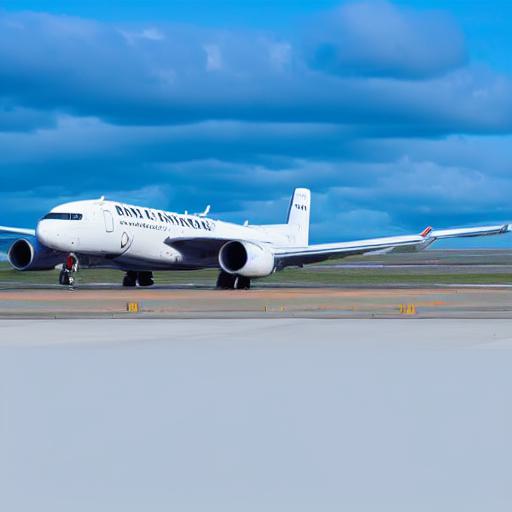}  & \includegraphics[align=c,width=0.14\linewidth]{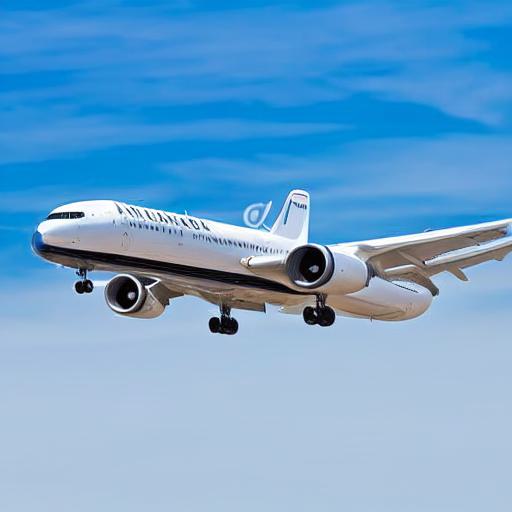}  & \includegraphics[align=c,width=0.14\linewidth]{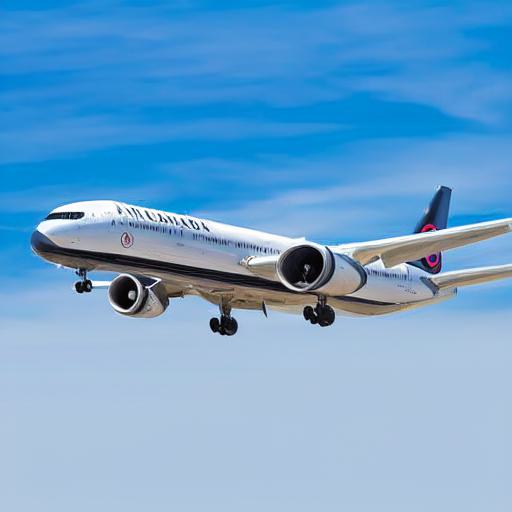}  & \includegraphics[align=c,width=0.14\linewidth]{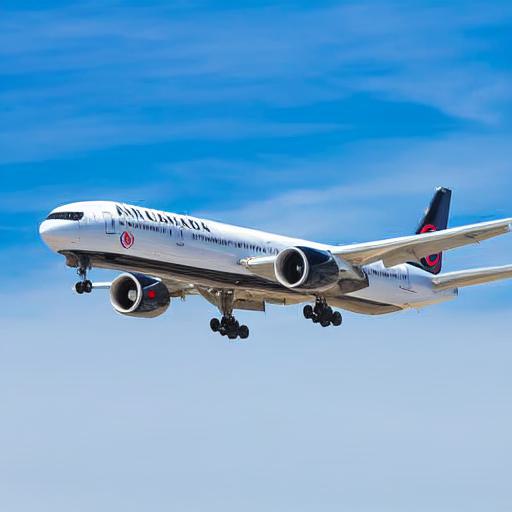}  & \includegraphics[align=c,width=0.14\linewidth]{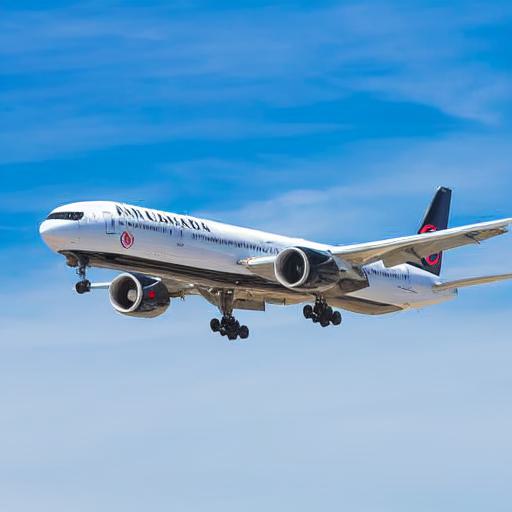}  \\
 48 & \includegraphics[align=c,width=0.14\linewidth]{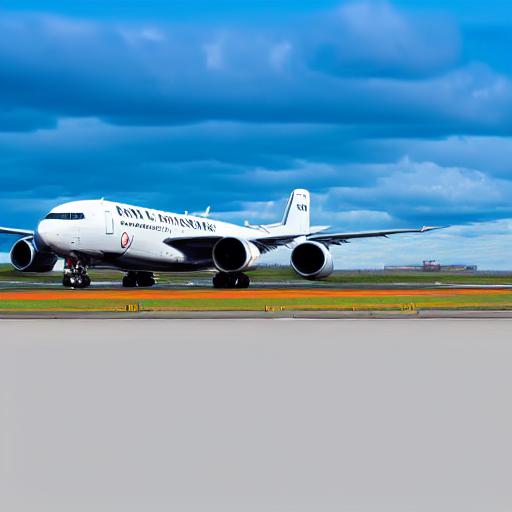}  & \includegraphics[align=c,width=0.14\linewidth]{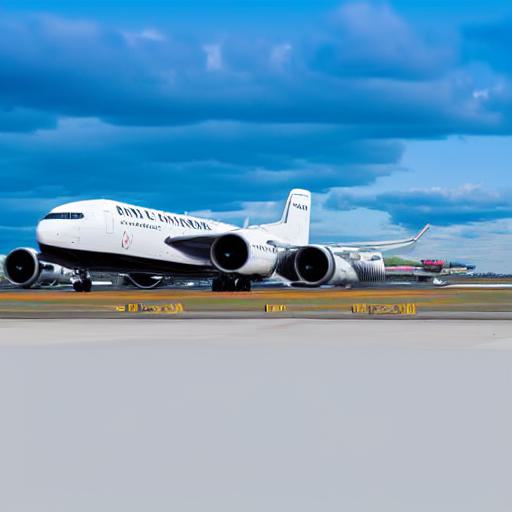}  & \includegraphics[align=c,width=0.14\linewidth]{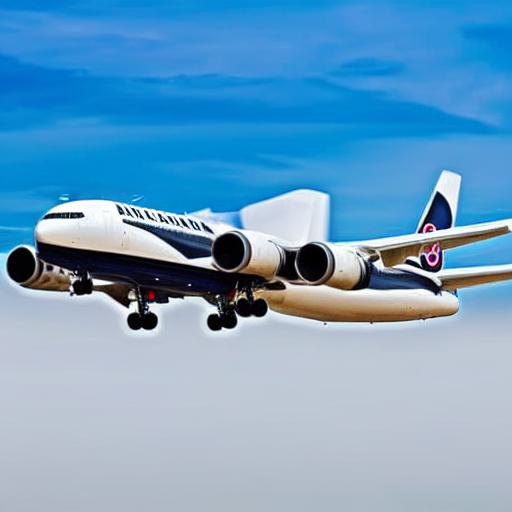}  & \includegraphics[align=c,width=0.14\linewidth]{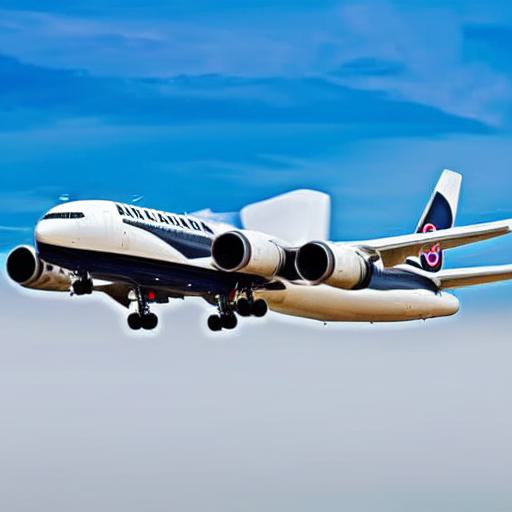}  & \includegraphics[align=c,width=0.14\linewidth]{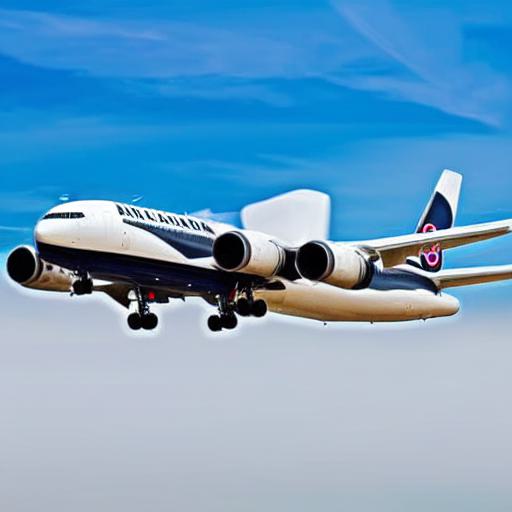}  \\
 46 & \includegraphics[align=c,width=0.14\linewidth]{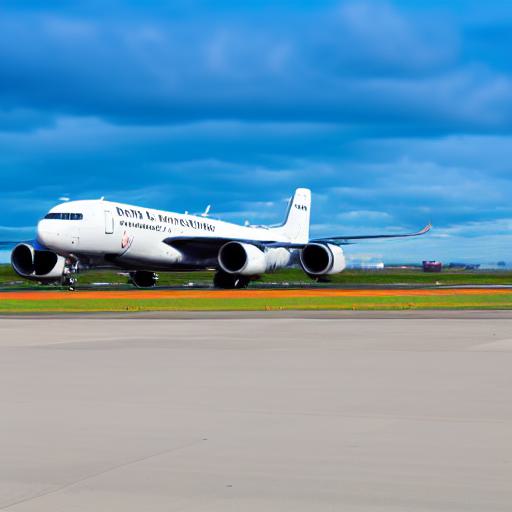}  & \includegraphics[align=c,width=0.14\linewidth]{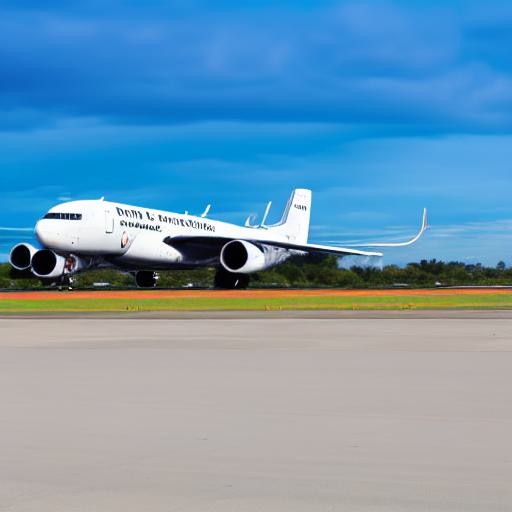}  & \includegraphics[align=c,width=0.14\linewidth]{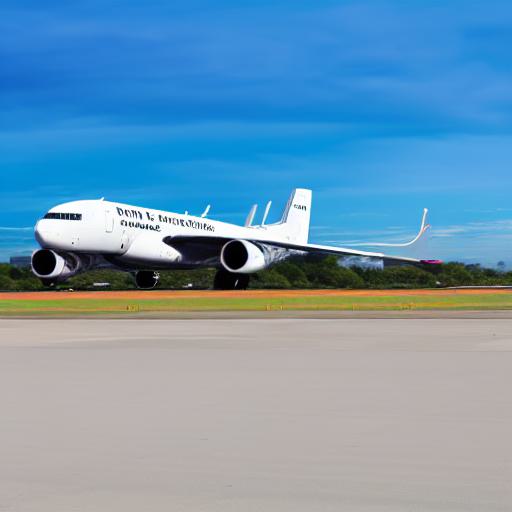}  & \includegraphics[align=c,width=0.14\linewidth]{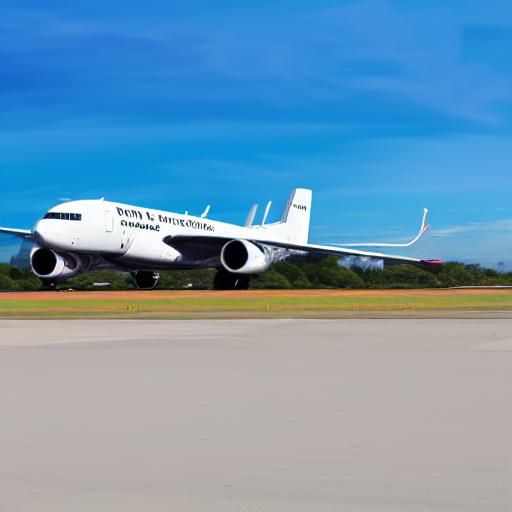}  & \includegraphics[align=c,width=0.14\linewidth]{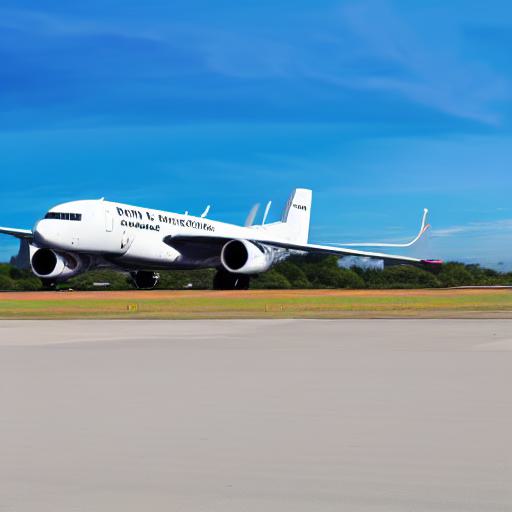}  \\
 44 & \includegraphics[align=c,width=0.14\linewidth]{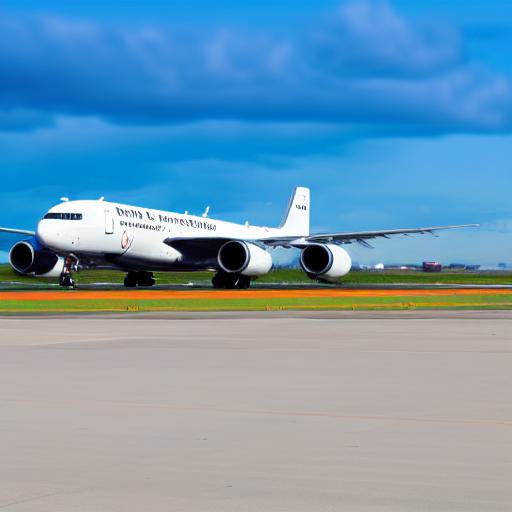}  & 
 \includegraphics[align=c,width=0.14\linewidth]{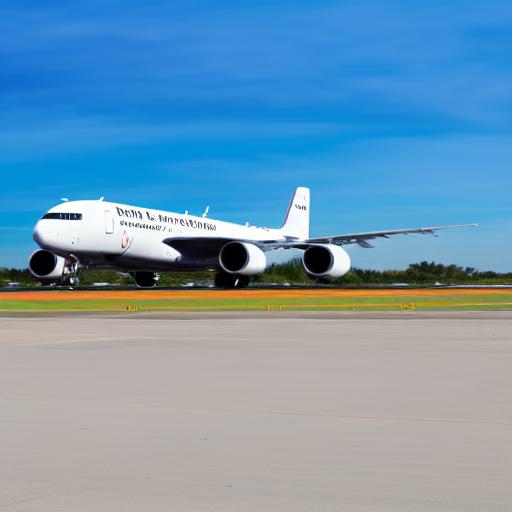}  & \includegraphics[align=c,width=0.14\linewidth]{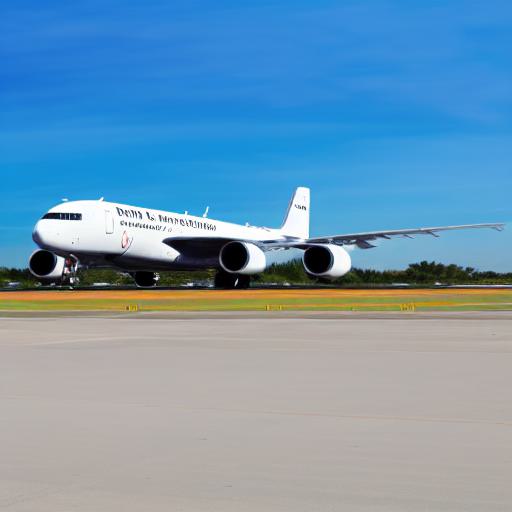}  & \includegraphics[align=c,width=0.14\linewidth]{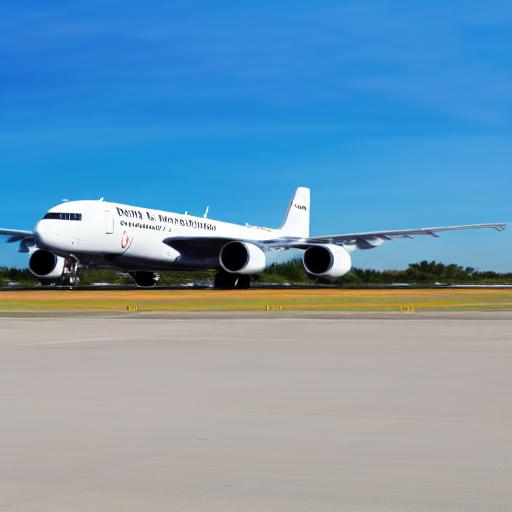}  & \includegraphics[align=c,width=0.14\linewidth]{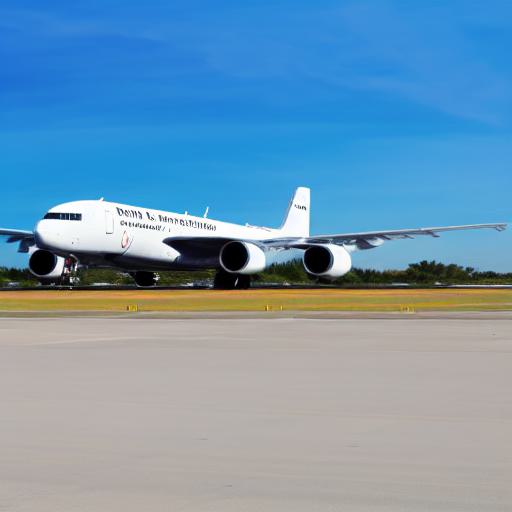}  \\
 42 & \includegraphics[align=c,width=0.14\linewidth]{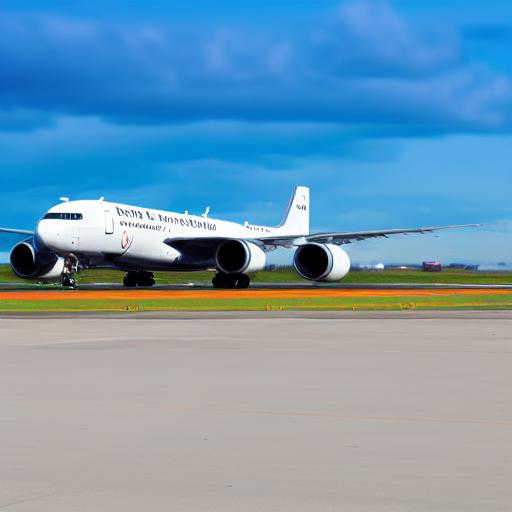} & \includegraphics[align=c,width=0.14\linewidth]{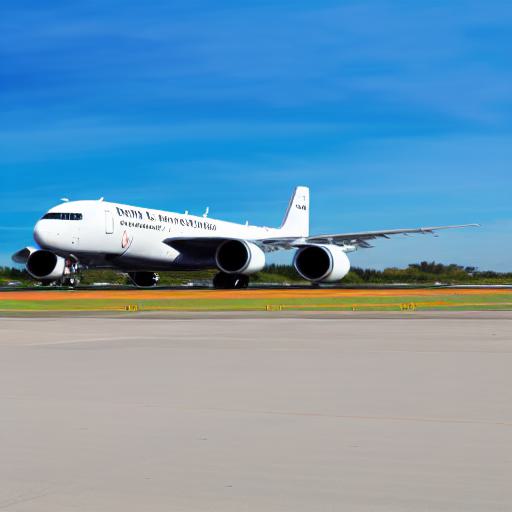} & \includegraphics[align=c,width=0.14\linewidth]{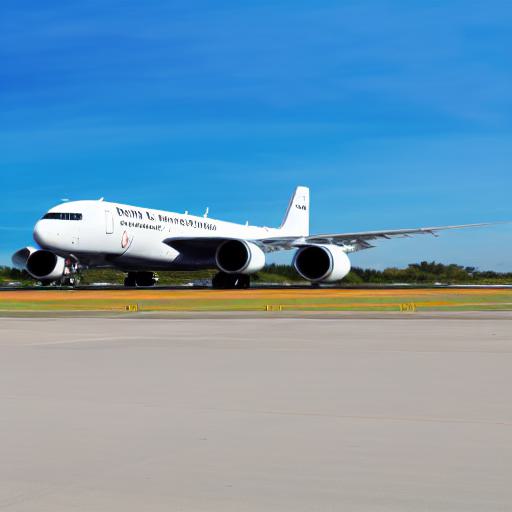} & \includegraphics[align=c,width=0.14\linewidth]{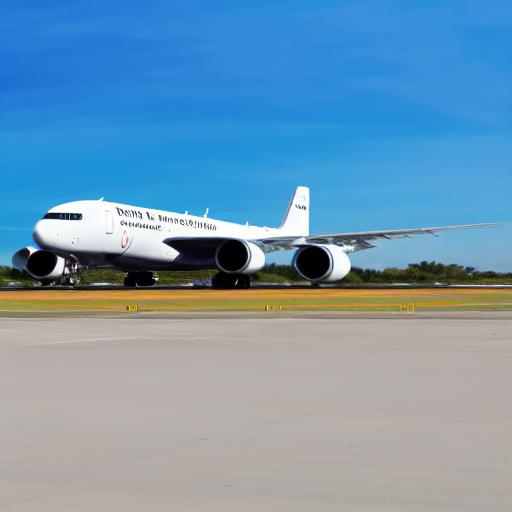} & \includegraphics[align=c,width=0.14\linewidth]{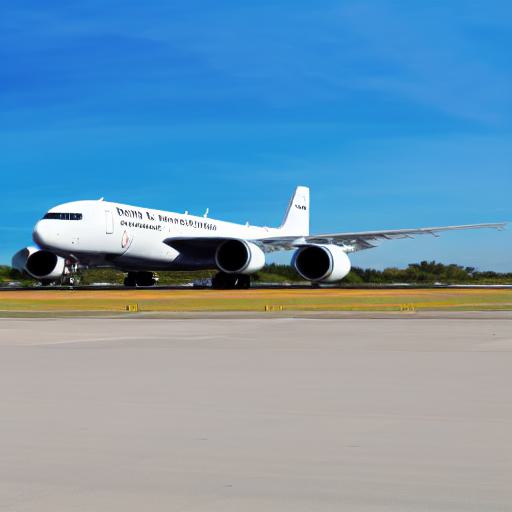}
\end{tabular}
\caption{Results of \textbf{MDP-$\mathbf{c}$} using constant schedule.}
\label{fig:c_tri}
\end{figure*}

\begin{figure*}
\centering
\setlength{\tabcolsep}{1pt}
\begin{tabular}{cccccccc}
Type/$T_M$ & 20 & 25 & 30 & 35 & 40 & 45 & 50 \\
Linear & \includegraphics[align=c,width=0.13\linewidth]{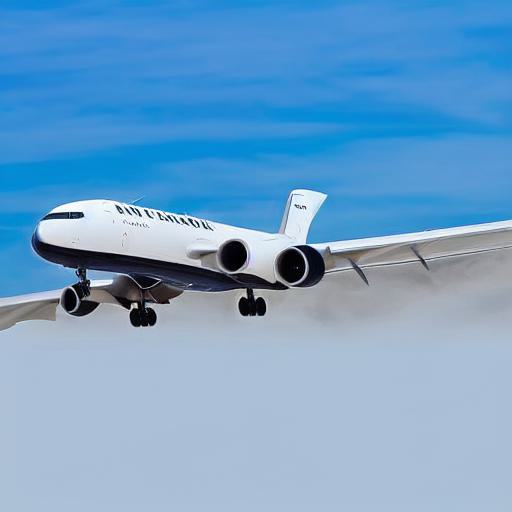} & \includegraphics[align=c,width=0.13\linewidth]{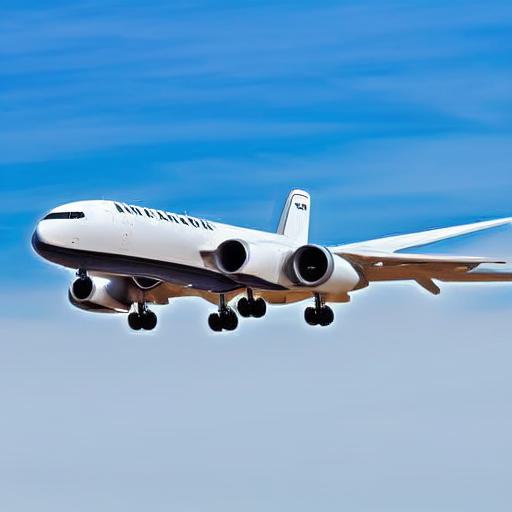} & \includegraphics[align=c,width=0.13\linewidth]{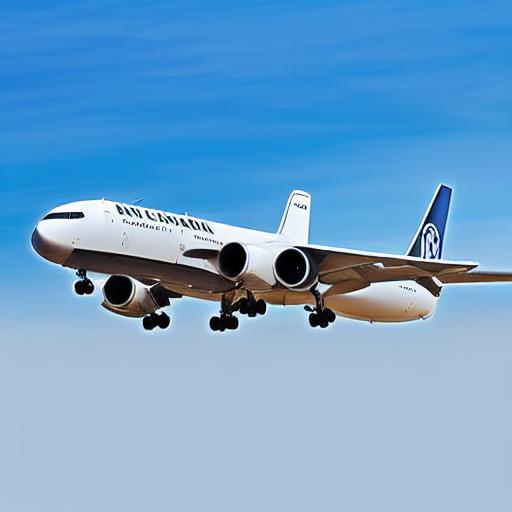} & \includegraphics[align=c,width=0.13\linewidth]{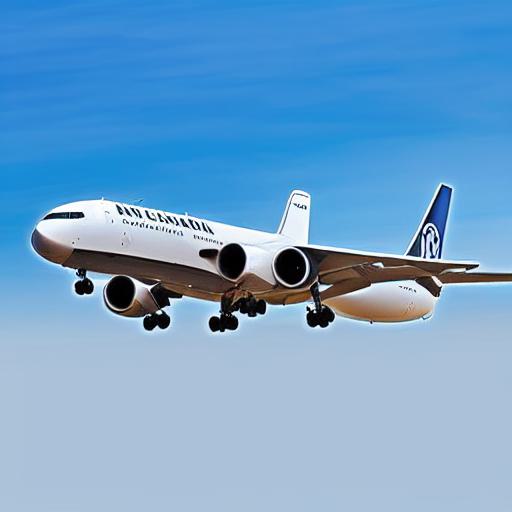} & \includegraphics[align=c,width=0.13\linewidth]{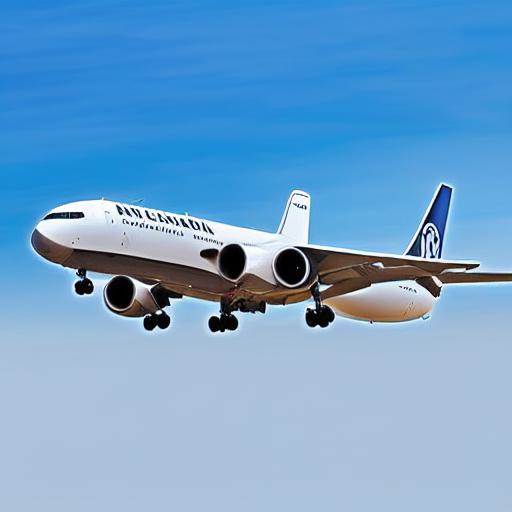} & \includegraphics[align=c,width=0.13\linewidth]{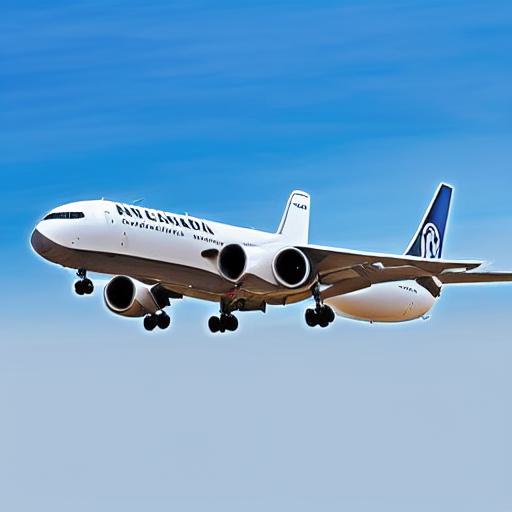} & \includegraphics[align=c,width=0.13\linewidth]{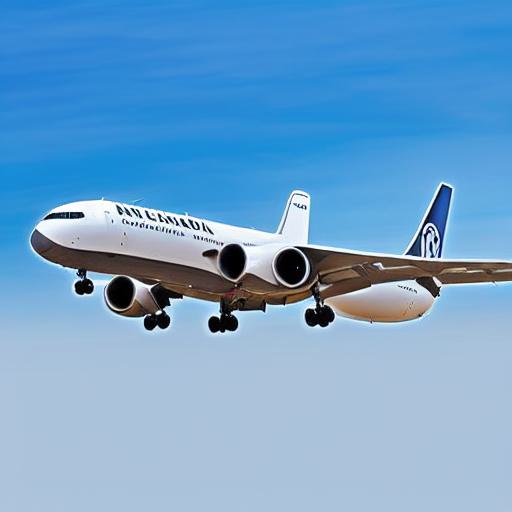} \\
Cosine & 
\includegraphics[align=c,width=0.13\linewidth]{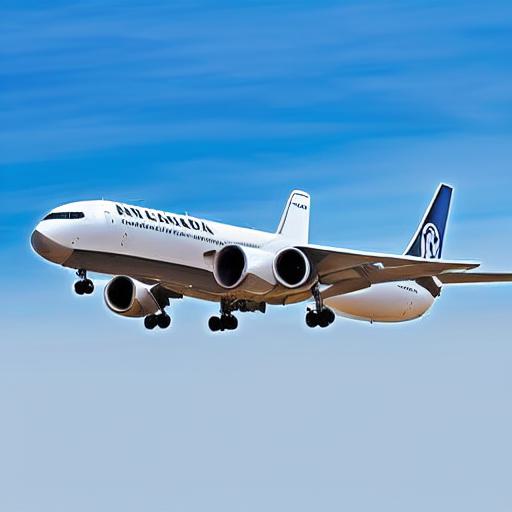} & \includegraphics[align=c,width=0.13\linewidth]{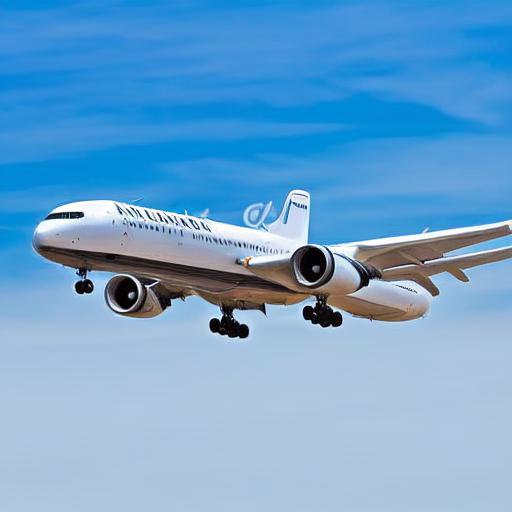} & \includegraphics[align=c,width=0.13\linewidth]{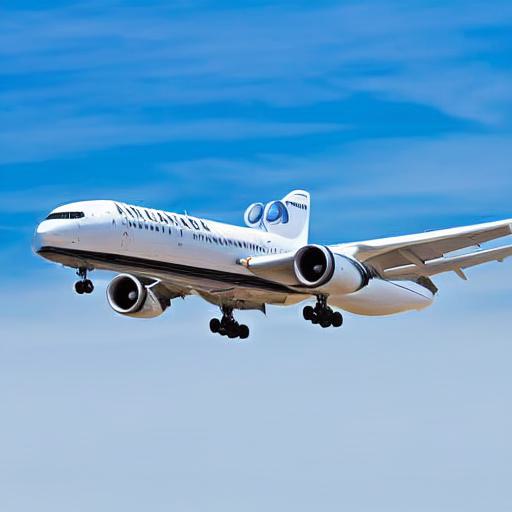} & \includegraphics[align=c,width=0.13\linewidth]{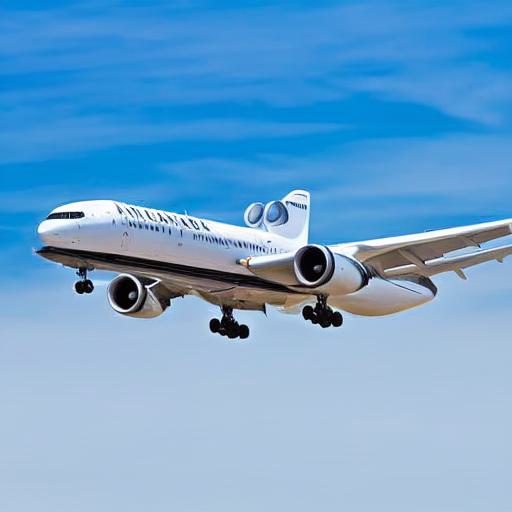} & \includegraphics[align=c,width=0.13\linewidth]{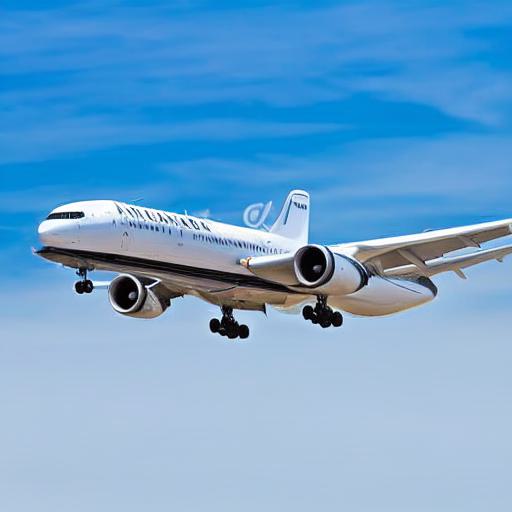} & \includegraphics[align=c,width=0.13\linewidth]{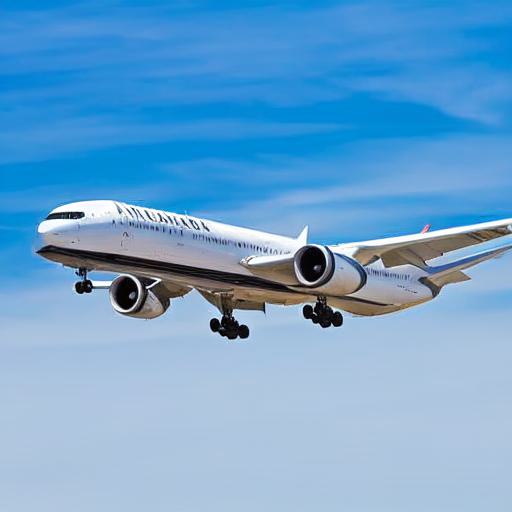} & \includegraphics[align=c,width=0.13\linewidth]{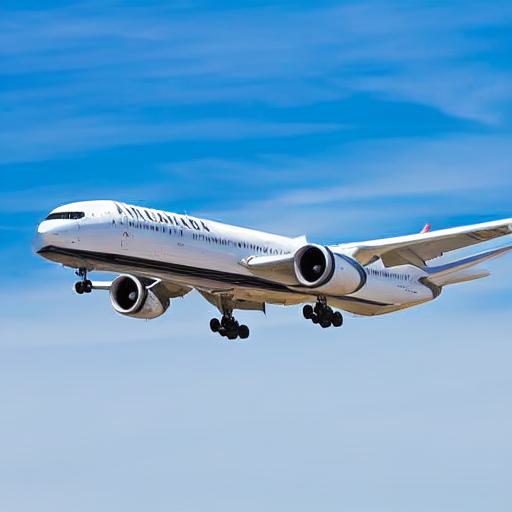} \\
Exponential & 
\includegraphics[align=c,width=0.13\linewidth]{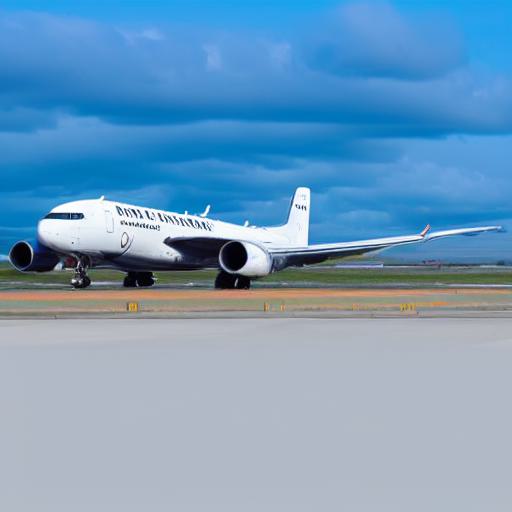} & \includegraphics[align=c,width=0.13\linewidth]{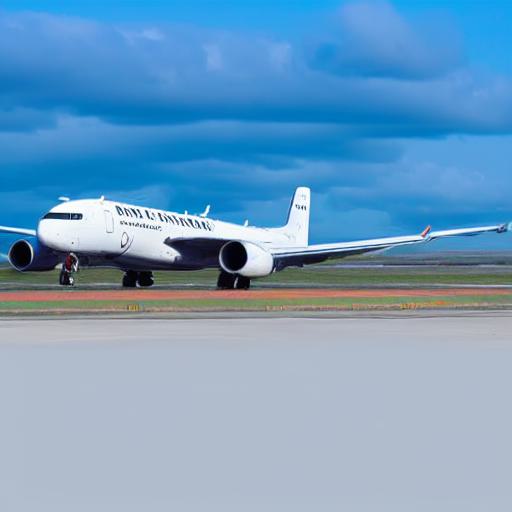} & \includegraphics[align=c,width=0.13\linewidth]{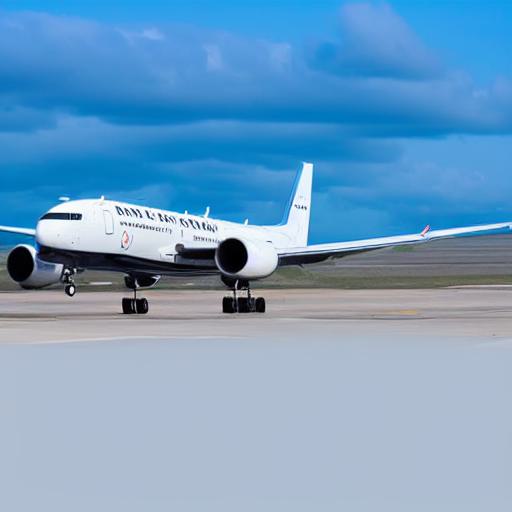} & \includegraphics[align=c,width=0.13\linewidth]{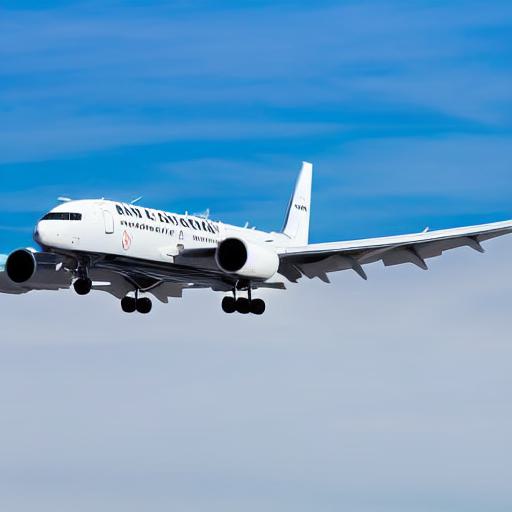} & \includegraphics[align=c,width=0.13\linewidth]{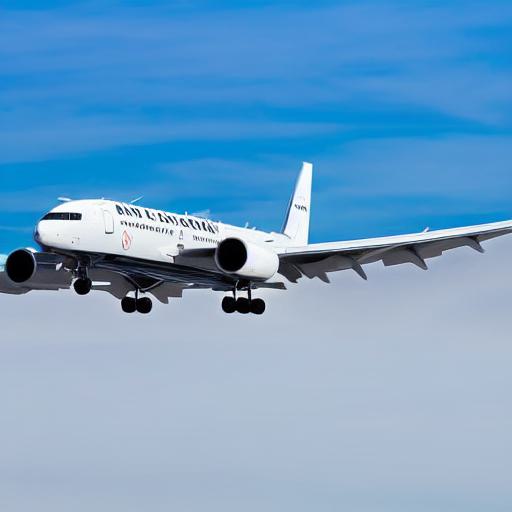} & \includegraphics[align=c,width=0.13\linewidth]{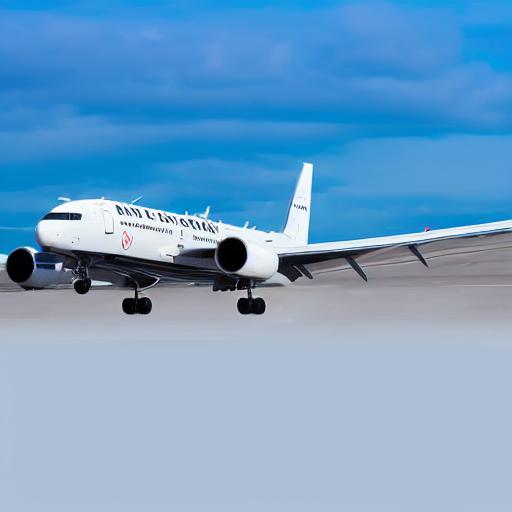} & \includegraphics[align=c,width=0.13\linewidth]{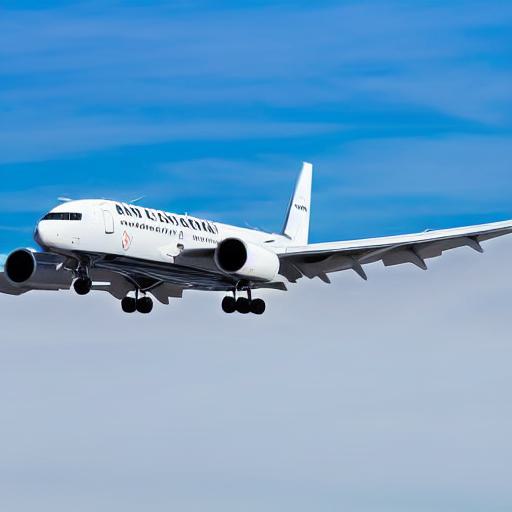} \\
\end{tabular}
\caption{Results of \textbf{MDP-$\mathbf{c}$} using linear, cosine and exponential schedule.}
\label{fig:c_linear}
\end{figure*}

\begin{figure*}
\centering
\setlength{\tabcolsep}{1pt}
\begin{tabular}{cccccc}
 $t_{\text{max}}$/$T_M$ & 5 & 10 & 15 & 20 & 25 \\
 50 & \includegraphics[align=c,width=0.14\linewidth]{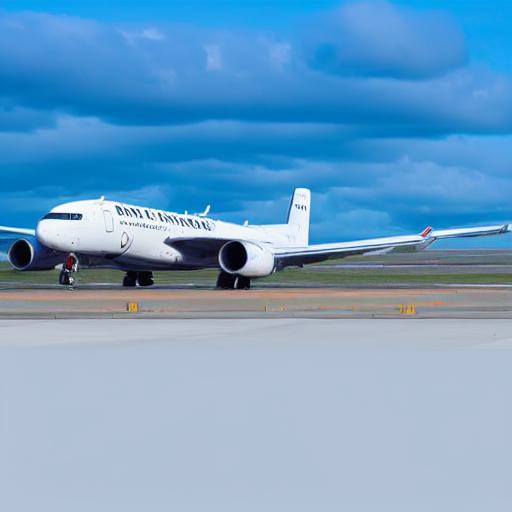}  & \includegraphics[align=c,width=0.14\linewidth]{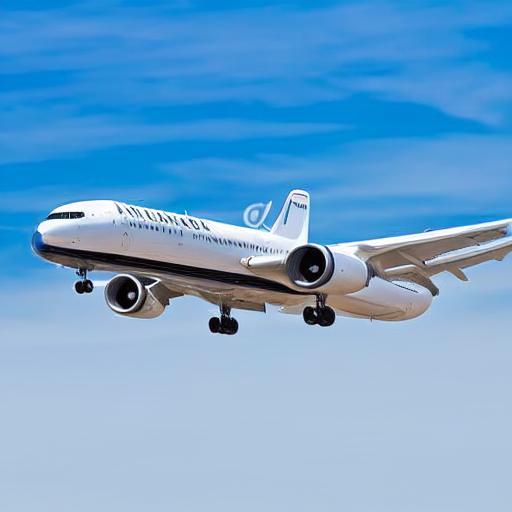}  & \includegraphics[align=c,width=0.14\linewidth]{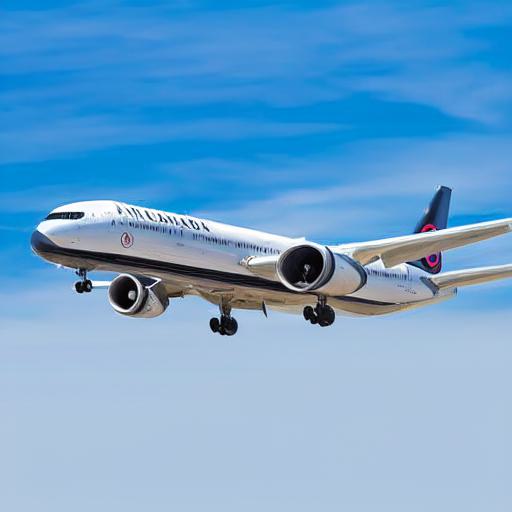}  & \includegraphics[align=c,width=0.14\linewidth]{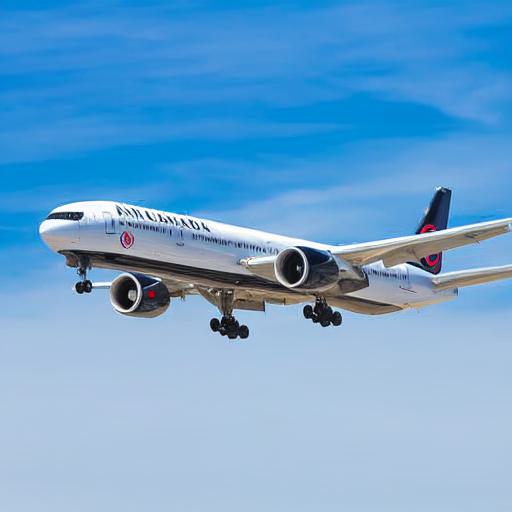}  & \includegraphics[align=c,width=0.14\linewidth]{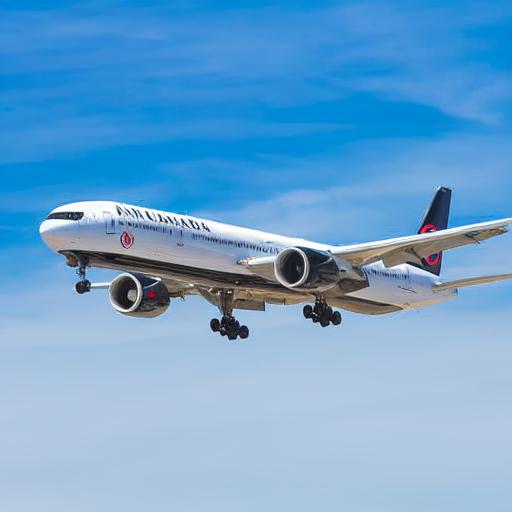}  \\
 48 & \includegraphics[align=c,width=0.14\linewidth]{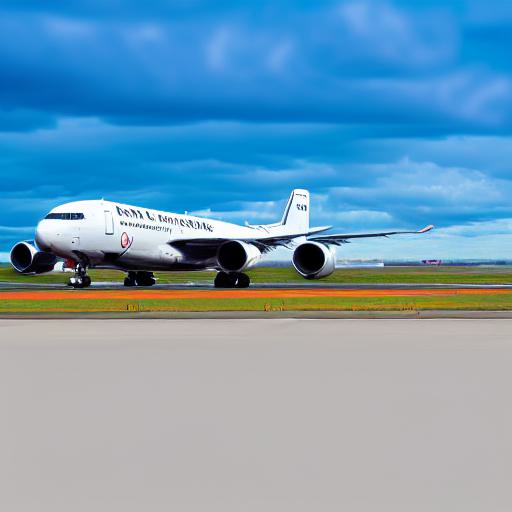}  & \includegraphics[align=c,width=0.14\linewidth]{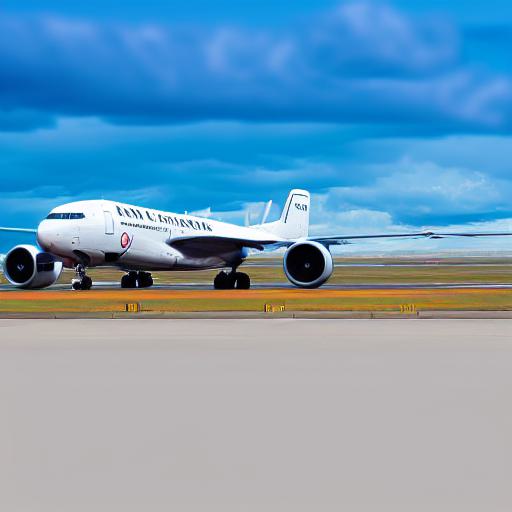}  & \includegraphics[align=c,width=0.14\linewidth]{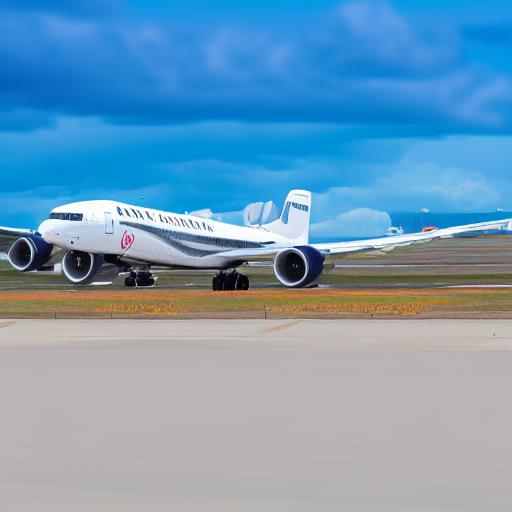}  & \includegraphics[align=c,width=0.14\linewidth]{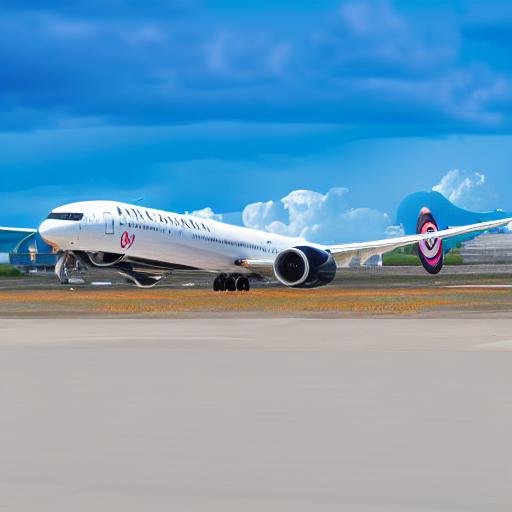}  & \includegraphics[align=c,width=0.14\linewidth]{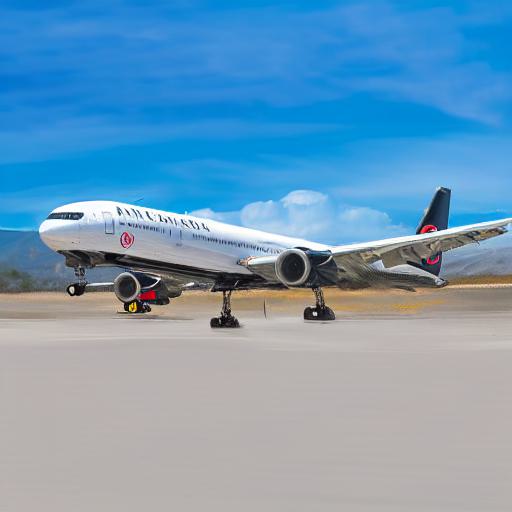}  \\
 46 & \includegraphics[align=c,width=0.14\linewidth]{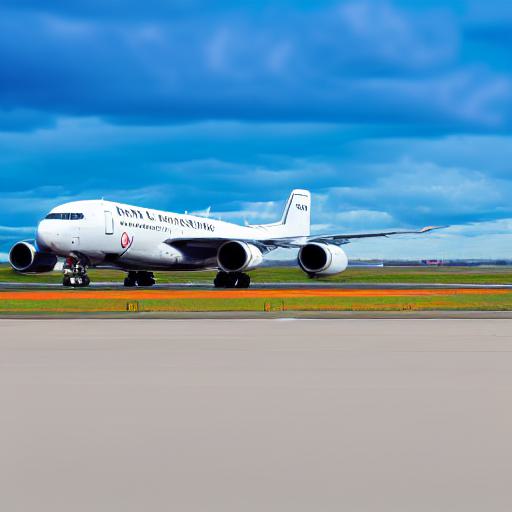}  & \includegraphics[align=c,width=0.14\linewidth]{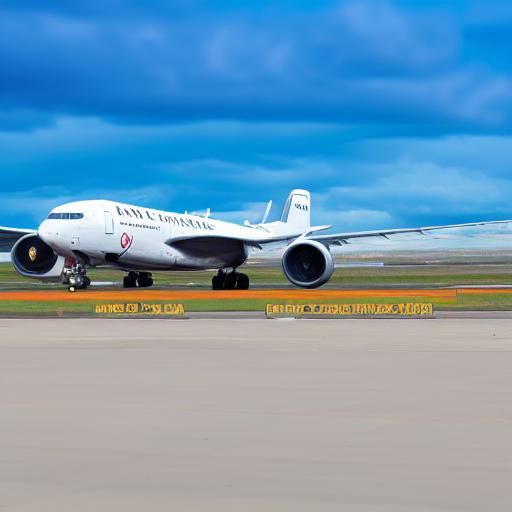}  & \includegraphics[align=c,width=0.14\linewidth]{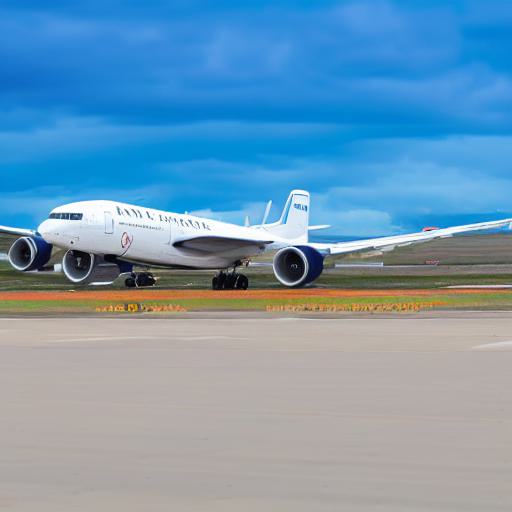}  & \includegraphics[align=c,width=0.14\linewidth]{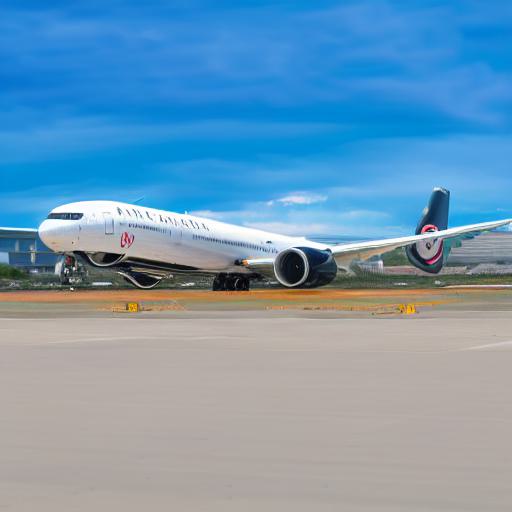}  & \includegraphics[align=c,width=0.14\linewidth]{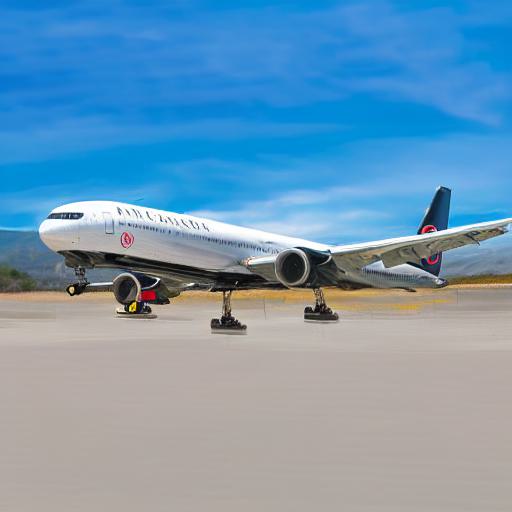}  \\
 44 & \includegraphics[align=c,width=0.14\linewidth]{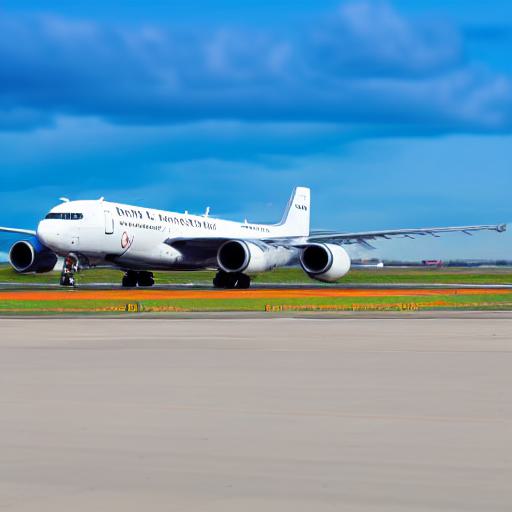}  & 
 \includegraphics[align=c,width=0.14\linewidth]{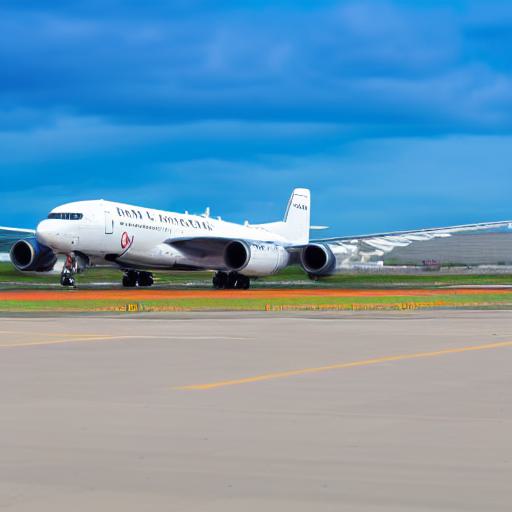}  & \includegraphics[align=c,width=0.14\linewidth]{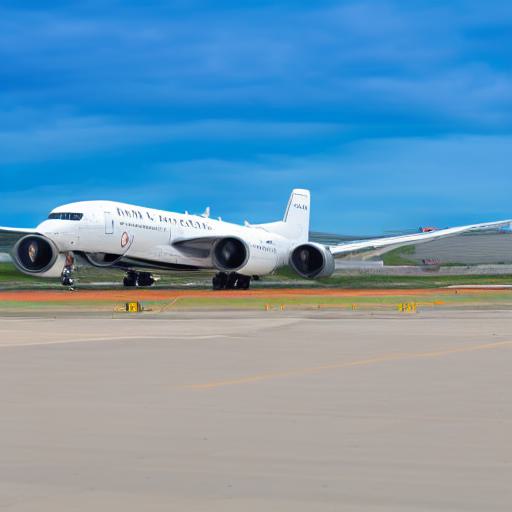}  & \includegraphics[align=c,width=0.14\linewidth]{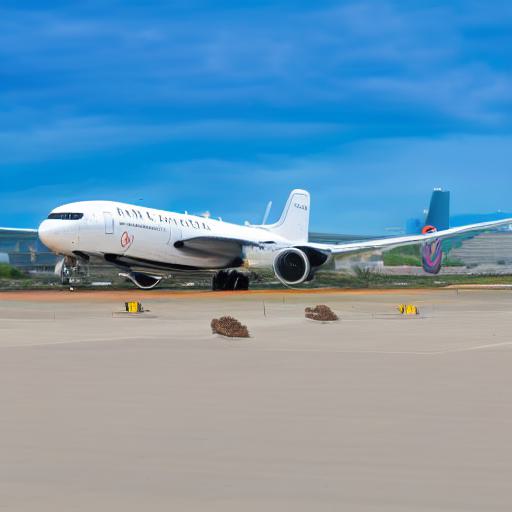}  & \includegraphics[align=c,width=0.14\linewidth]{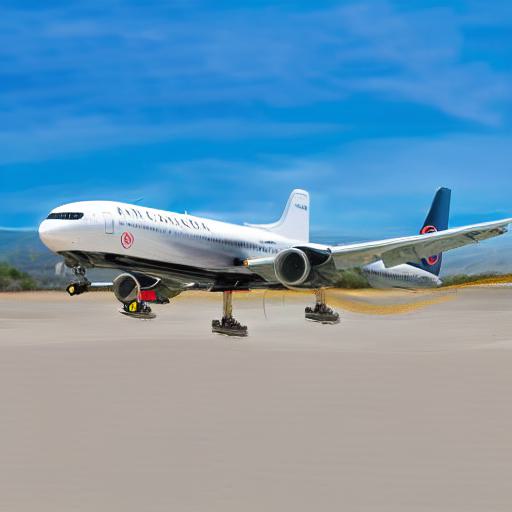}  \\
 42 & \includegraphics[align=c,width=0.14\linewidth]{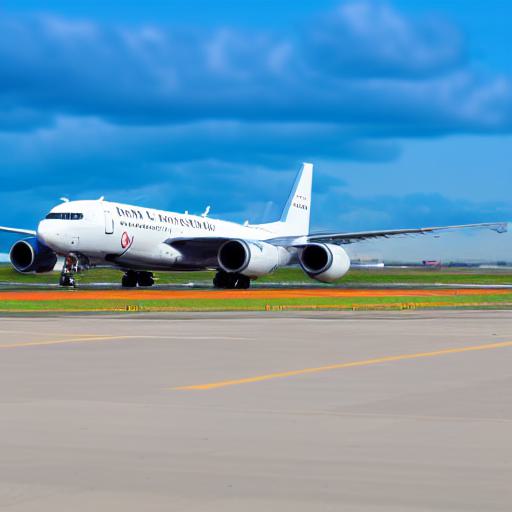} & \includegraphics[align=c,width=0.14\linewidth]{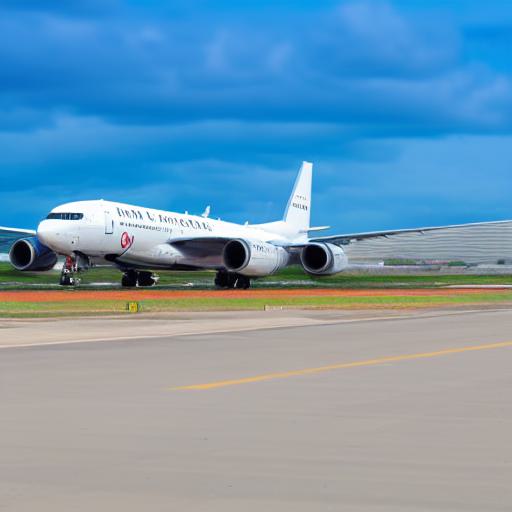} & \includegraphics[align=c,width=0.14\linewidth]{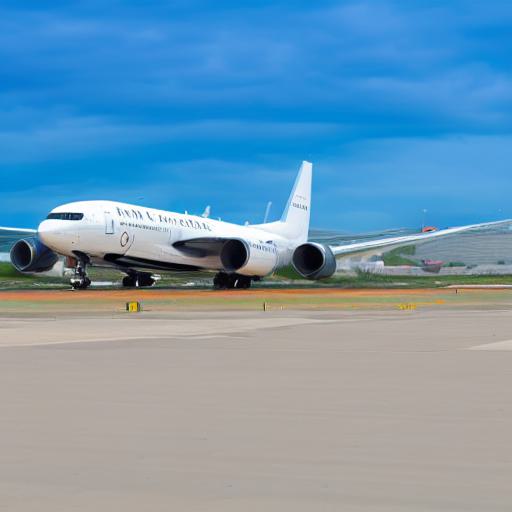} & \includegraphics[align=c,width=0.14\linewidth]{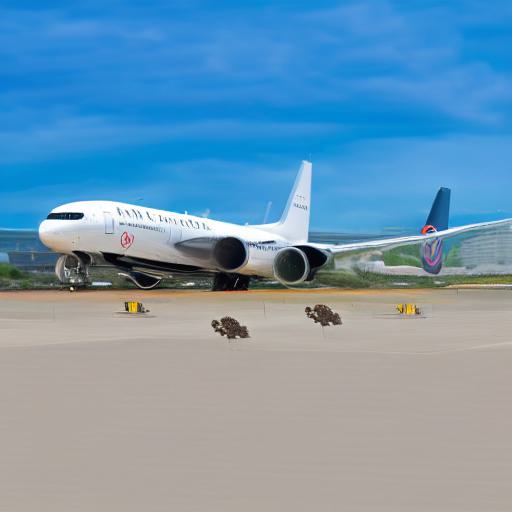} & \includegraphics[align=c,width=0.14\linewidth]{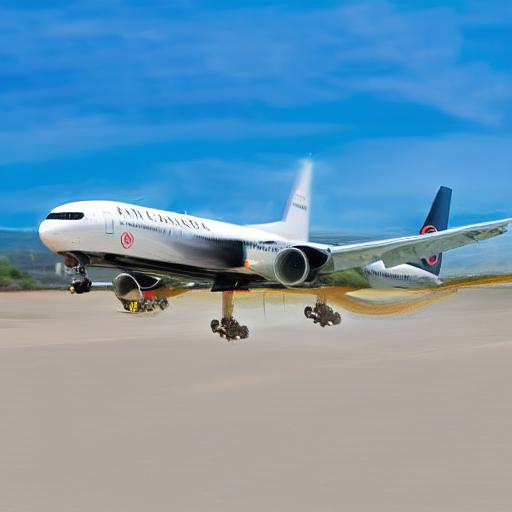}
\end{tabular}
\caption{Results of \textbf{MDP-$\boldsymbol\epsilon_t$} using constant schedule.}
\label{fig:epsilon_tri}
\end{figure*}

\begin{figure*}
\centering
\setlength{\tabcolsep}{1pt}
\begin{tabular}{cccccccc}
Type/$T_M$ & 20 & 25 & 30 & 35 & 40 & 45 & 50 \\
Linear & \includegraphics[align=c,width=0.13\linewidth]{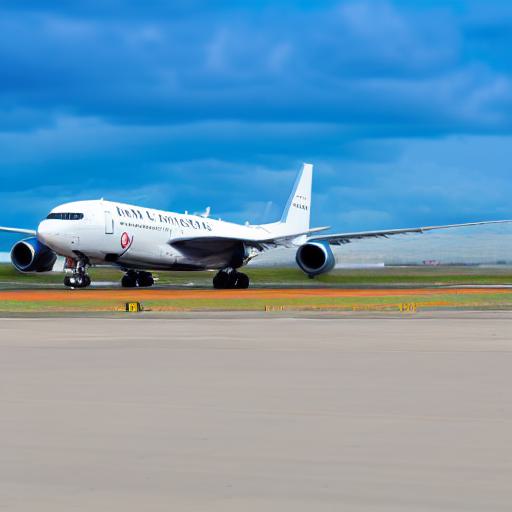} & \includegraphics[align=c,width=0.13\linewidth]{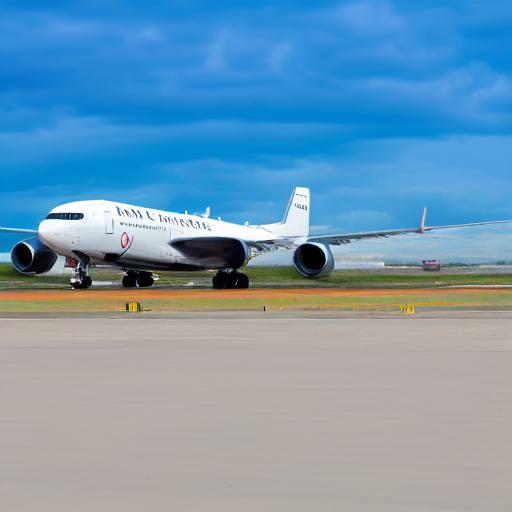} & \includegraphics[align=c,width=0.13\linewidth]{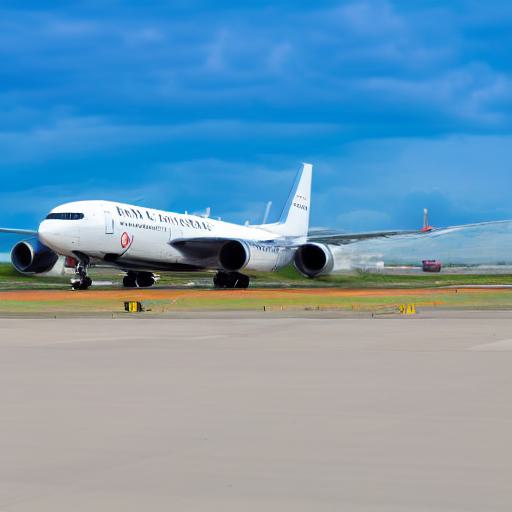} & \includegraphics[align=c,width=0.13\linewidth]{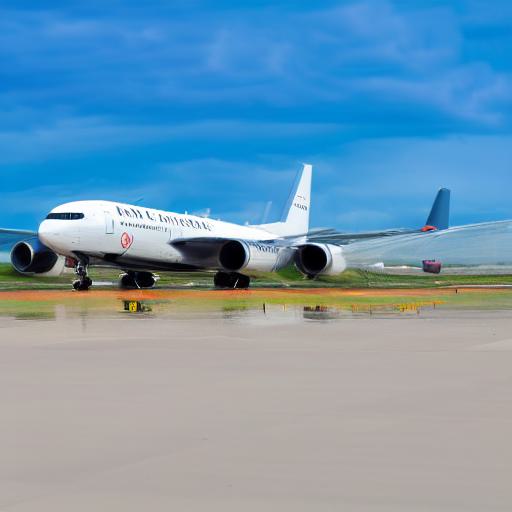} & \includegraphics[align=c,width=0.13\linewidth]{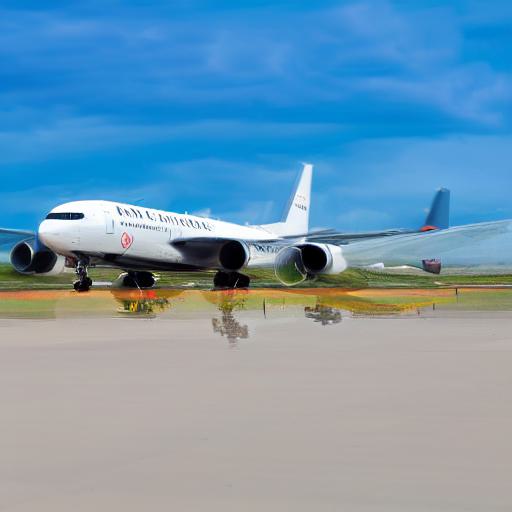} & \includegraphics[align=c,width=0.13\linewidth]{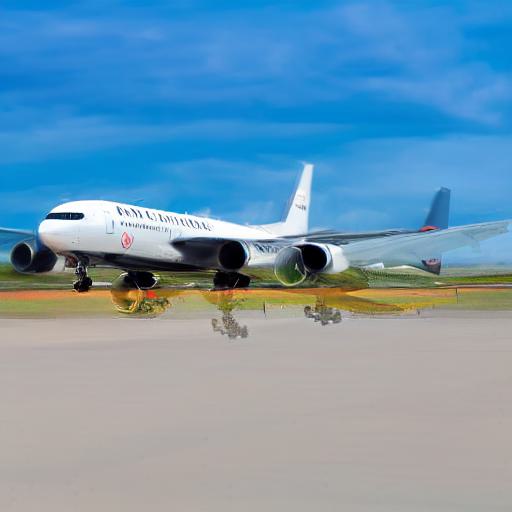} & \includegraphics[align=c,width=0.13\linewidth]{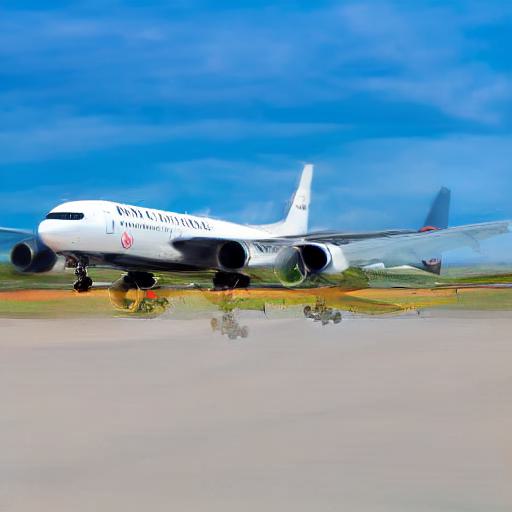} \\
Cosine & 
\includegraphics[align=c,width=0.13\linewidth]{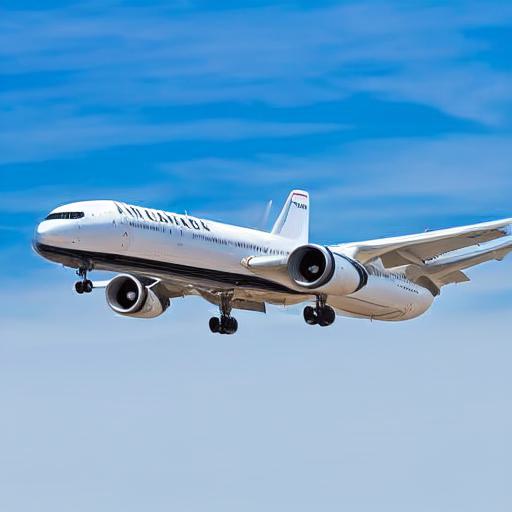} & \includegraphics[align=c,width=0.13\linewidth]{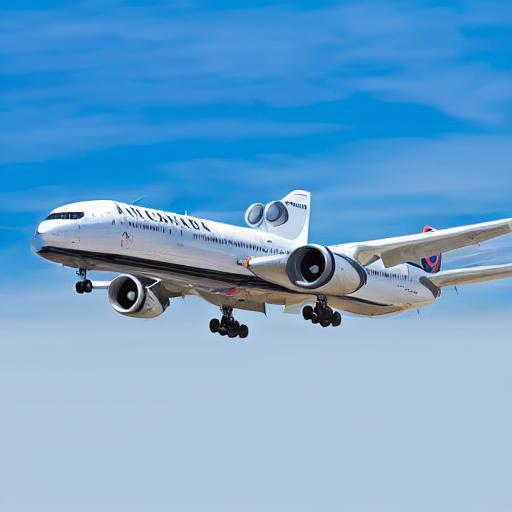} & \includegraphics[align=c,width=0.13\linewidth]{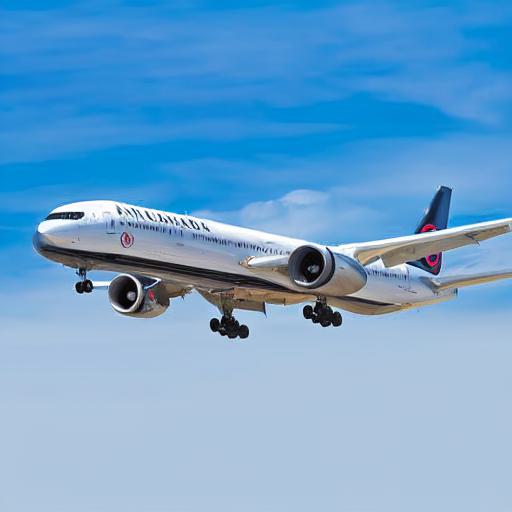} & \includegraphics[align=c,width=0.13\linewidth]{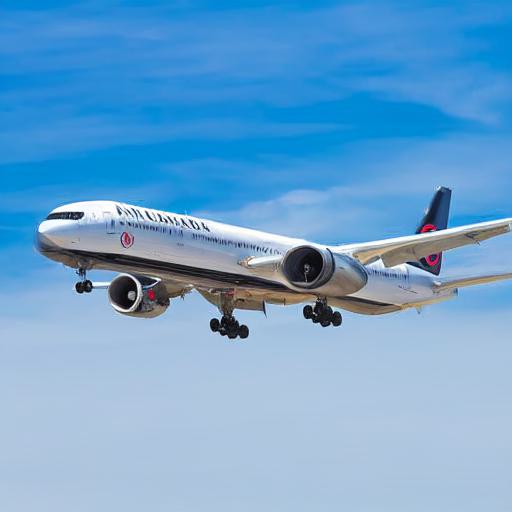} & \includegraphics[align=c,width=0.13\linewidth]{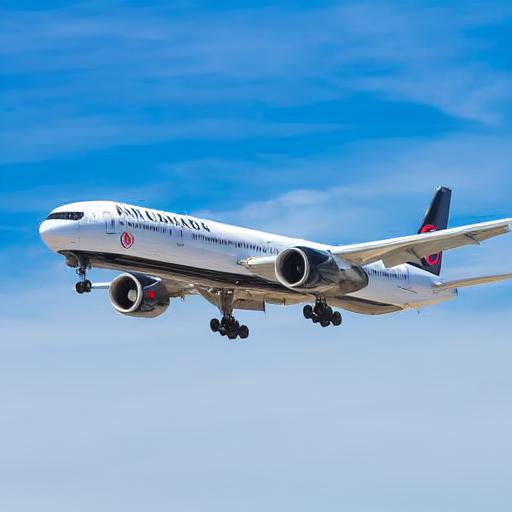} & \includegraphics[align=c,width=0.13\linewidth]{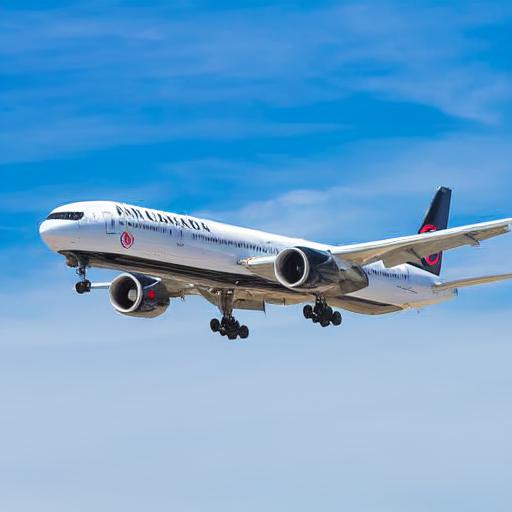} & \includegraphics[align=c,width=0.13\linewidth]{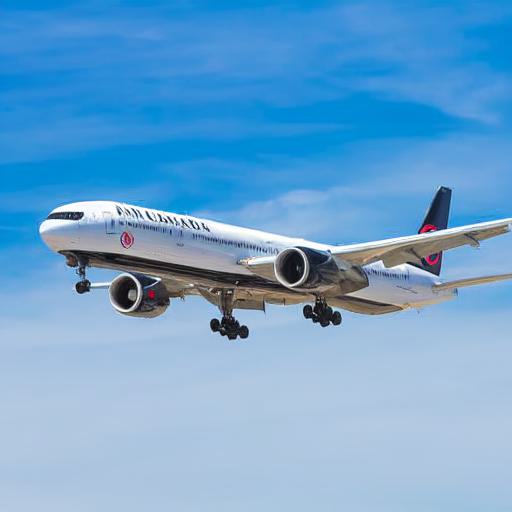} \\
Exponential & 
\includegraphics[align=c,width=0.13\linewidth]{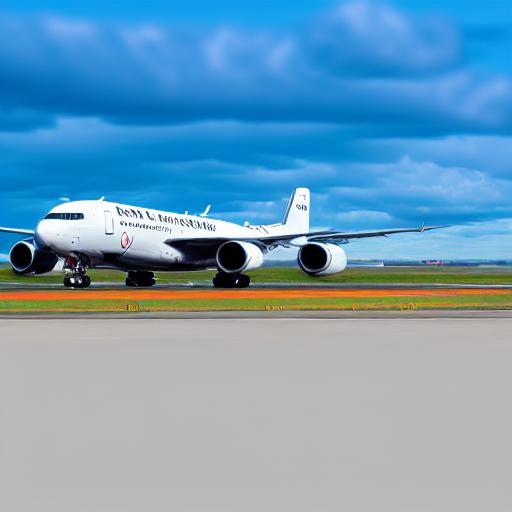} & \includegraphics[align=c,width=0.13\linewidth]{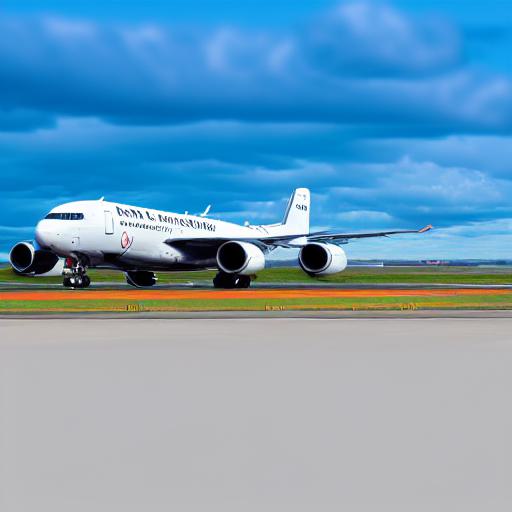} & \includegraphics[align=c,width=0.13\linewidth]{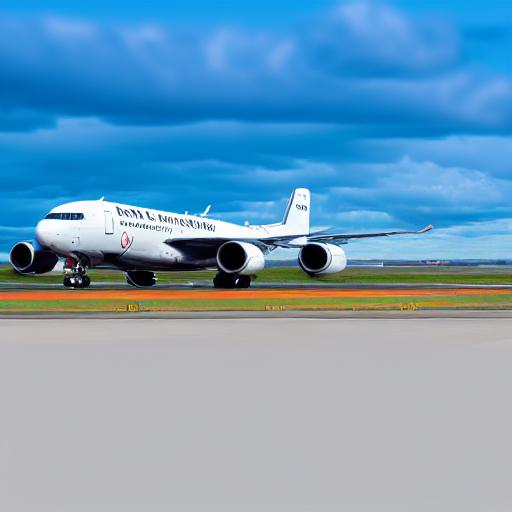} & \includegraphics[align=c,width=0.13\linewidth]{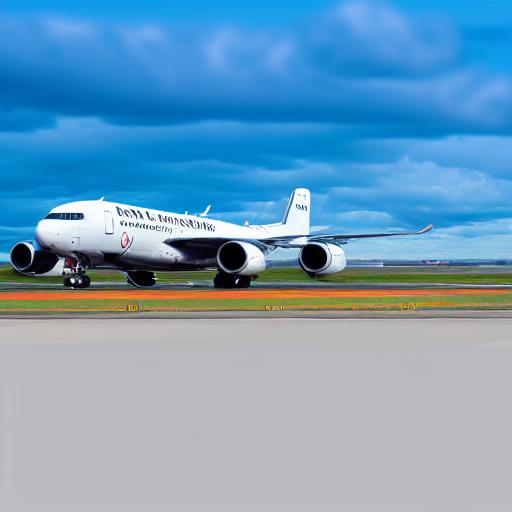} & \includegraphics[align=c,width=0.13\linewidth]{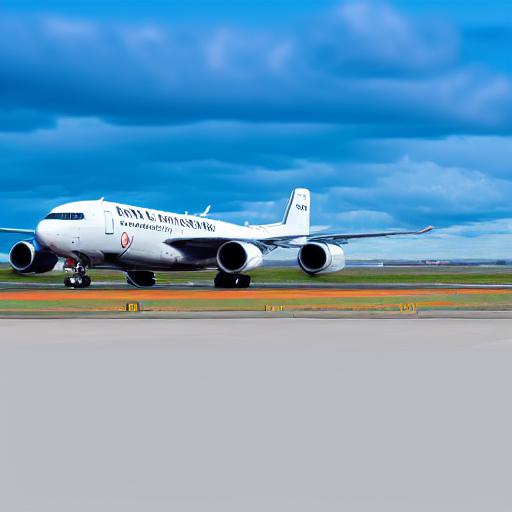} & \includegraphics[align=c,width=0.13\linewidth]{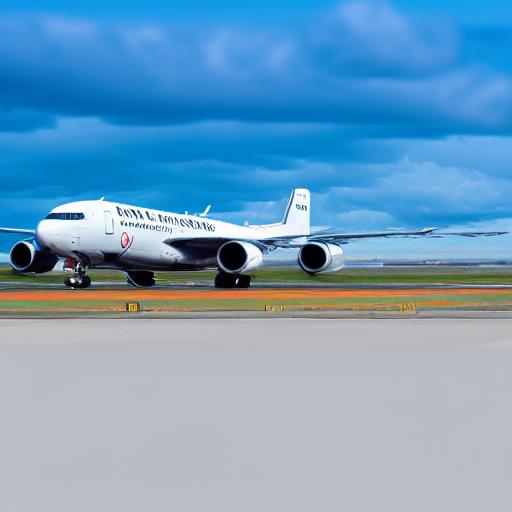} & \includegraphics[align=c,width=0.13\linewidth]{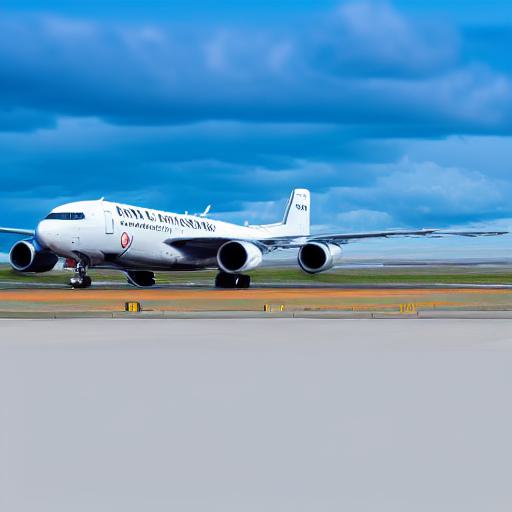} \\
\end{tabular}
\caption{Results of \textbf{MDP-$\boldsymbol\epsilon_t$} using linear, cosine and exponential schedule.}
\label{fig:epsilon_linear}
\end{figure*}

We observe that the manipulation schedule has a relatively larger influence on MDP-$\mathbf{x}_t$, MDP-$\mathbf{c}$, and MDP-$\boldsymbol\epsilon_t$. For MDP-$\boldsymbol \beta$ the linear factfor (guidance scale) has a much more important effect towards the controllability of the edited image. For Prompt-to-Prompt (a manipulation for attention maps) the range of time steps to inject the self-attention maps strongly determines what the edited image will look like. Therefore, for MDP-$\mathbf{x}_t$, MDP-$\mathbf{c}$, and MDP-$\boldsymbol\epsilon_t$, we explore four different manipulation schedules: constant, linear, cosine, and exponential (\cref{fig:schedules}). Given an integer timestep $t \in [0, 50]$, we give the equations for the linear factor for constant schedule ($\omega_t^{\text{const}}$), linear schedule ($\omega_t^{\text{linear}}$), cosine schedule ($\omega_t^{\text{cos}}$) and exponential schedule linear schedule ($\omega_t^{\text{exp}}$):
\begin{equation}
\begin{split}
  \omega_t^{\text{const}} &=
    \begin{cases}
      1 & \text{if $t_{\text{min}}\leq t \leq t_{\text{max}}$,}\\
      0 & \text{else.}\\
    \end{cases}  
    \\
    \omega_t^{\text{linear}} &=
    \begin{cases}
      \frac{t - t_{\text{min}}}{50 - t_{\text{min}}} & \text{if $t_{\text{min}}\leq t \leq 50$,}\\
      0 & \text{else.}\\
    \end{cases}  
    \\
    \omega_t^{\text{cosine}} &=
    \begin{cases}
      \cos \left(\frac{\pi}{2} \frac{50 - t}{50 - t_{\text{min}}} \right) & \text{if $t_{\text{min}}\leq t \leq 50$,}\\
      0 & \text{else.}\\
    \end{cases}  
    \\
    \omega_t^{\text{exp}} &=
    \begin{cases}
      \exp{\left(-5 \left( \frac{50 - t}{50 - t_{\text{min}}} \right) \right)} & \text{if $t_{\text{min}}\leq t \leq 50$,}\\
      0 & \text{else.}\\
    \end{cases}  
\end{split}
\end{equation}
The results are shown in \cref{fig:xt_tri,fig:xt_linear,fig:c_tri,fig:c_linear,fig:epsilon_tri,fig:epsilon_linear}. For MDP-$\boldsymbol \beta$, we vary the guidance scale from -0.7 to 0.7 and show the results in \cref{fig:guidance}. For Prompt-to-Prompt, we use the constant schedule to inject the attention maps and show the results in \cref{fig:p2p_global} and \cref{fig:p2p_local}. We only visualize one example for global editing as we show in the paper: given an airplane image input, we change its background from ``sky'' to ``airport runway''. We only show this as an example because this example is challenging so we can identify the ability of different methods. For Prompt-to-Prompt, we show an additional example from local editing for a better understanding of the method. We discuss the results in the following subsections.

For the parameters in the design space for each manipulation, we show a general recommendation in \cref{table:local} and \cref{table:global}, which is summarized based on the examples we have tried. Note that this is only considered a recommendation for general cases. For specific applications (and input image), it improved results can still be attained by fine tuning the parameters.

\begin{table*}[h]
\centering
\begin{tabular}{ccccccc}
\toprule
             & \begin{tabular}[c]{@{}c@{}}Suitable\\ or not\end{tabular} & $t_{\text{max}}$ & $T_M$ & \begin{tabular}[c]{@{}c@{}}Manipulation\\ schedule\end{tabular} & Linear factor & Other \\ \midrule
$x_t$        &    Partly                                                               &      [42, 48]     &   [15, 20]    &     Constant or others                                                         &     $< 1.0$          &    --   \\ 
$c$          &    Yes                                                               &      [47, 50]     &   [15, 25]    &        Constant or others                                                         &      Can be 1.0 or smaller         &   --    \\ 
P2P          &     Yes                                                              &     Usually 50      &   40    &         Can be 1.0 or smaller                                                        &   Can be 1.0 or smaller            &      \begin{tabular}[c]{@{}c@{}}The timesteps to inject\\ self-attention maps are critical\end{tabular}     \\ 
$\beta$      &     Partly                                                              &     50      &   50    &          Does not matter                                                       &      [-0.8, -0.5]         &   The guidance scale is very sensitive    \\ 
$\epsilon_t$ &     Yes                                                              &     [44, 50]      &   [15, 25]    &     Constant or others                                                            &     Can be 1.0 or smaller          &    --   \\ \bottomrule
\end{tabular}
\caption{The recommended parameters in the design space for local edits.}
\label{table:local}
\end{table*}

\begin{table*}[h]
\centering
\begin{tabular}{ccccccc}
\toprule
             & \begin{tabular}[c]{@{}c@{}}Suitable\\ or not\end{tabular} & $t_{\text{max}}$ & $T_M$ & \begin{tabular}[c]{@{}c@{}}Manipulation\\ schedule\end{tabular} & Linear factor & Other \\ \midrule
$x_t$        &    Partly                                                               &     [40, 50]      &   [10, 15]    &         Constant or others                                                        &      $<1.0$         &   --    \\ 
$c$          &      Partly                                                             &    50       &   [20, 25]    &     Constant or others                                                            &        Can be 1.0 or smaller       &   --    \\ 
P2P          &     Yes                                                              &   Usually 50        &  40     &          Constant or others                                                       &    Can be 1.0 or smaller           &  \begin{tabular}[c]{@{}c@{}}The timesteps to inject\\ self-attention maps are critical\end{tabular}     \\ 
$\beta$      &    Partly                                                               &    50       &  50     &          Does not matter                                                       &     [-0.8, 0.5]          &   The guidance scale is very sensitive    \\ 
$\epsilon_t$ &     Yes                                                              &     [44, 48]      &   [15, 25]    &        Constant or others                                                         &     Can be 1.0 or smaller          &   --    \\ \bottomrule
\end{tabular}
\caption{The recommended parameters in the design space for global edits.}
\label{table:global}
\end{table*}

\subsection{MDP-$\mathbf{x}_t$, MDP-$\mathbf{c}$ and MDP-$\boldsymbol\epsilon_t$}
While in most of the local editing applications and some of the global editing applications, MDP-$\mathbf{x}_t$ can faithfully do the edits, for the example we show in the figures, all the examples generated by MDP-$\mathbf{x}_t$ under different schedules fail. The edited image preserves the overall layout of the input image, however, the new semantics fail to be injected. We conjecture that keeping the intermediate latent from the path when generating the input image will impose a very strong condition to preserve the information from the input image. For the result in the main paper, we fix the linear factors for MDP-$\mathbf{x}_t$ to be 0.7 under all the edits. However, for a clear comparison with other two similar methods MDP-$\mathbf{c}$ and MDP-$\boldsymbol\epsilon_t$, we fix the linear factors as 1.0, which results in that the new condition has no influence when $t_{\text{min}} \leq t \leq t_{\text{max}}$. That is also the reason why we empirically set the linear factor of MDP-$\mathbf{x}_t$ to be 0.7 for all the results in the main paper, rather than 1.0 as the other two manipulations. 

On the other hand, MDP-$\mathbf{c}$ and MDP-$\boldsymbol\epsilon_t$ can inject the new semantics under the same schedules when linear factors are all set to be 1.0 when $t_{\text{max}}$. For MDP-$\mathbf{c}$, typically when $t_{\text{max}}$ is smaller than 50 or $T_M$ is smaller than 15, the new semantics can be injected into the edited image. However, the layout information from the input image is also lost. For MDP-$\boldsymbol\epsilon_t$, in terms of constant schedule, when $T_M$ is 25 and $t_{\text{max}}$ is smaller than 50, the edited image can both preserve the layout and incorporate the new semantics. 

Empirically, we observe that MDP-$\mathbf{x}_t$ is the strongest one to inject layout information, then is MDP-$\boldsymbol\epsilon_t$, while MDP-$\mathbf{c}$ is the weakest one when using the same manipulation schedule. Theoretically, this observation is aligned with the diffusion generation formula:
\begin{equation}\label{eq:ddim-sampler}
    \begin{aligned}
        \mathbf{x}_{t-1} =& \mathrm{DDIM}(\mathbf{x}_t, \boldsymbol\epsilon_{t}, t) \\
        =& \sqrt{\alpha_{t-1}}\cdot f_\theta(\mathbf{x}_t, \mathbf{c}, t) + \sqrt{1-\alpha_{t-1}}\cdot \boldsymbol\epsilon_\theta(\mathbf{x}_t, \mathbf{c}, t), \\
    \end{aligned}
\end{equation}
where $f_\theta(\mathbf{x}_t, \mathbf{c}, t) = \frac{\mathbf{x}_t - \sqrt{1-\alpha_t}\cdot\boldsymbol\epsilon_\theta(\mathbf{x}_t, \mathbf{c}, t)}{\sqrt{\alpha_t}}$, $\boldsymbol\epsilon_t=\boldsymbol\epsilon_\theta(\mathbf{x}_t, \mathbf{c}, t)$, and $\alpha_t$ is a noise schedule factor as in DDIM. The sampling is to iteratively apply the above equation by giving an initial noise $\mathbf{x}_T\sim\mathcal{N}(\mathbf{0}, \mathbf{I})$. When setting the linear factors as 1.0, MDP-$\mathbf{x}_t$ directly replaces the intermediate latent from the path when generating the input image, while MDP-$\boldsymbol\epsilon_t$ replaces the predicted noise and keeps the previous intermediate latent when applying Eq. \eqref{eq:ddim-sampler}, and MDP-$\mathbf{c}$ only replaces the conditional embedding when predicting the noise.

We also observe that when $t_{\text{max}} = 50$, \ie, the manipulation happens right at the start of the whole denoising process, and the layout information can be injected more than when $t_{\text{max}}$ is smaller. In addition, considering linear, cosine, and exponential schedule, for cosine schedule it can preserve the most layout formation, while the exponential schedule can preserve the least. This is also aligned with the design of the schedule, where the linear factor of the cosine schedule in every timestep is larger than which of the linear schedule and exponential schedule, which means in the cosine schedule larger amount of the layout information is taken into the consideration compared to linear and exponential schedule. These findings can also support that the initial timesteps during the generation process, especially the first denoising timestep $T_M = 50$, have a great influence on the layout of the edited image. Injecting the layout information during the early generation stages is effective. 

Considering the results we have obtained, we have several conclusions:
\begin{itemize}
\item For the same manipulation schedule, MDP-$\mathbf{x}_t$ injects layout information the strongest from the input image, then comes MDP-$\boldsymbol\epsilon_t$, while MDP-$\mathbf{c}$ is the weakest one. 
\item For MDP-$\mathbf{x}_t$, we recommend to set the linear factor smaller than 1.0 when using the constant schedule, or consider using schedule with decreasing linear factors during the generation process, such as linear, cosine, or exponential schedules. For MDP-$\mathbf{c}$, we recommend using it only in local editing applications. For our proposed and highlighted MDP-$\boldsymbol\epsilon_t$, we recommend to set $t_{\text{max}}$ to be around 44 to 50 while $T_M$ to be around 20 to 25.
\item As the early stages during the diffusion generation process has a larger influence on the layout of the generated image, we recommend that for every kind of manipulation, we set the $t_{\text{max}}$ to be larger than 44.
\end{itemize}

\subsection{MDP-$\boldsymbol \beta$}
\begin{figure*}[h]
\centering
\setlength{\tabcolsep}{1pt}
\begin{tabular}{cccccc}
$\beta$/$t_{\text{max}}$ - $t_{\text{min}}$ & 50 - 40 & 50 - 30 & 50 - 20 & 50 - 10 & 50 - 0 \\
0.7 & \includegraphics[align=c,width=0.15\linewidth]{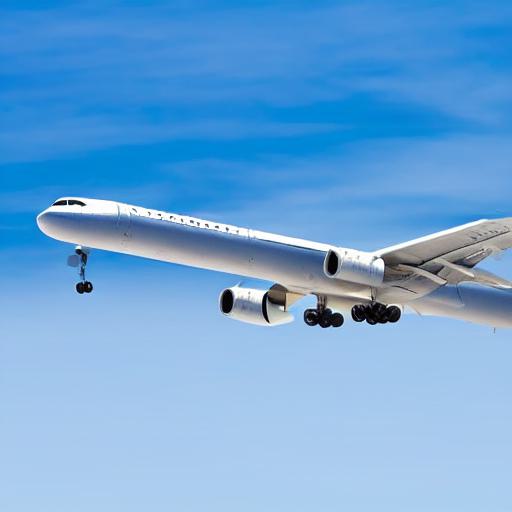} & \includegraphics[align=c,width=0.15\linewidth]{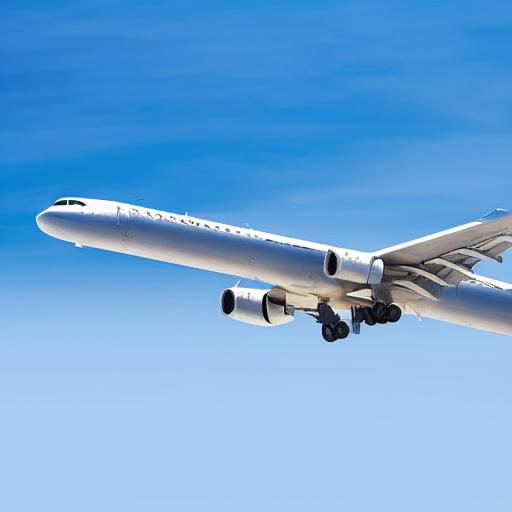} & \includegraphics[align=c,width=0.15\linewidth]{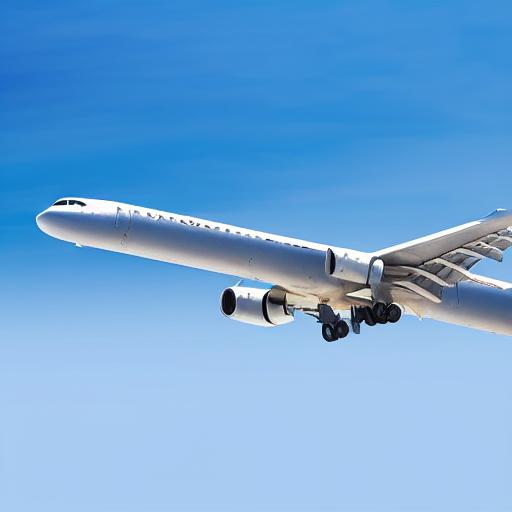} & \includegraphics[align=c,width=0.15\linewidth]{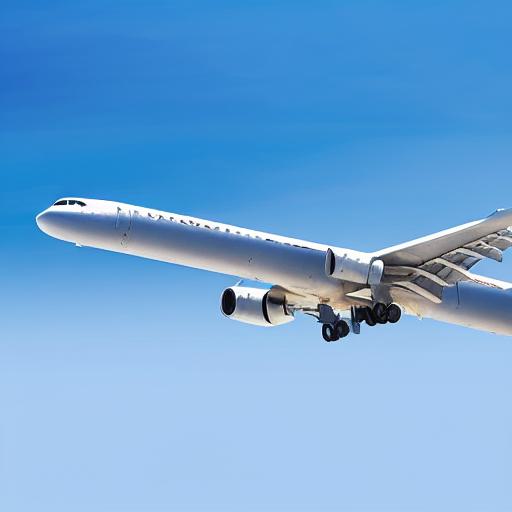} & \includegraphics[align=c,width=0.15\linewidth]{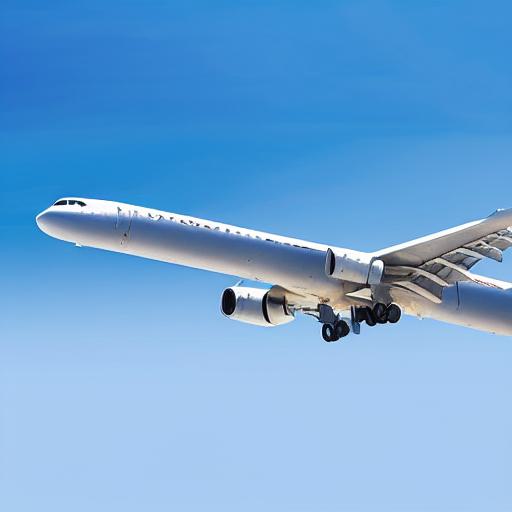} \\
0.3 & \includegraphics[align=c,width=0.15\linewidth]{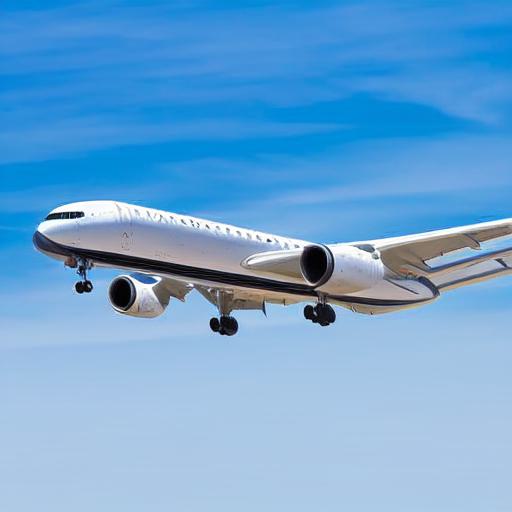} & \includegraphics[align=c,width=0.15\linewidth]{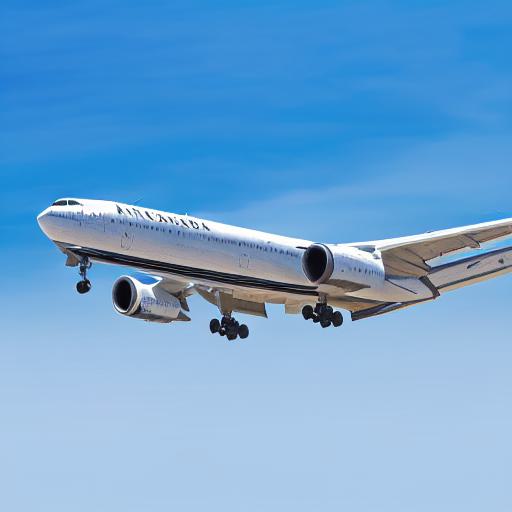} & \includegraphics[align=c,width=0.15\linewidth]{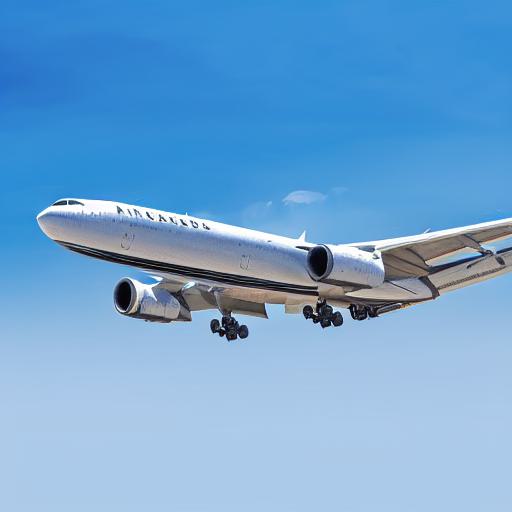} & \includegraphics[align=c,width=0.15\linewidth]{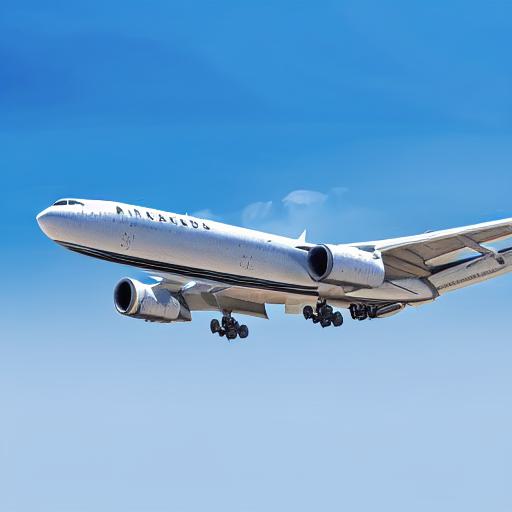} & \includegraphics[align=c,width=0.15\linewidth]{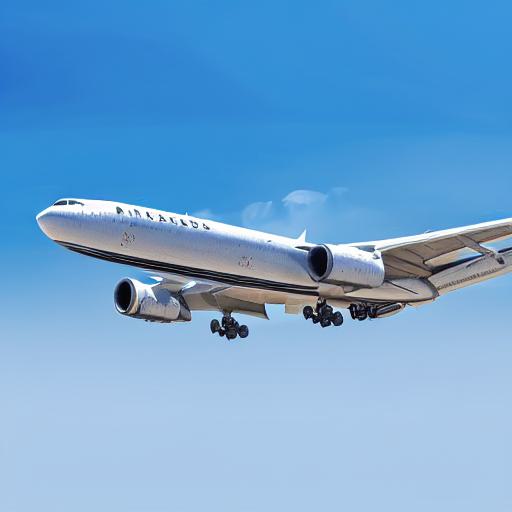} \\
0 & \includegraphics[align=c,width=0.15\linewidth]{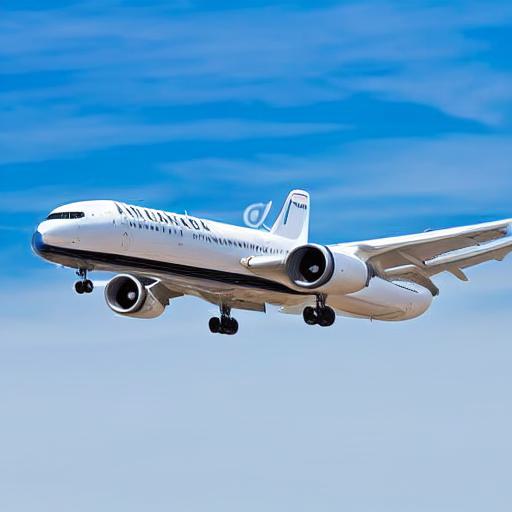} & \includegraphics[align=c,width=0.15\linewidth]{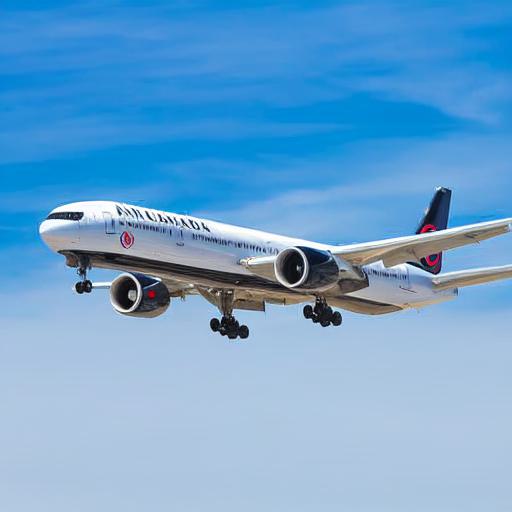} & \includegraphics[align=c,width=0.15\linewidth]{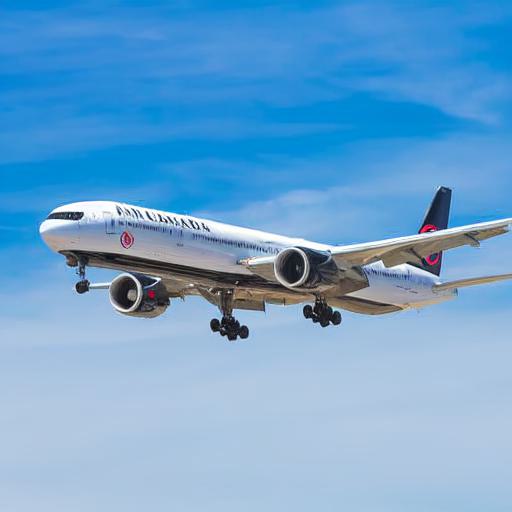} & \includegraphics[align=c,width=0.15\linewidth]{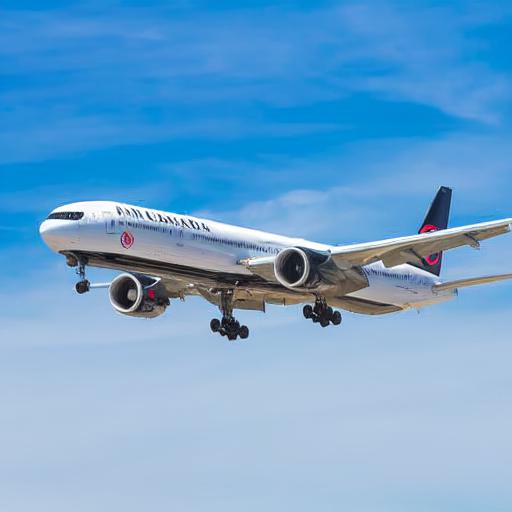} & \includegraphics[align=c,width=0.15\linewidth]{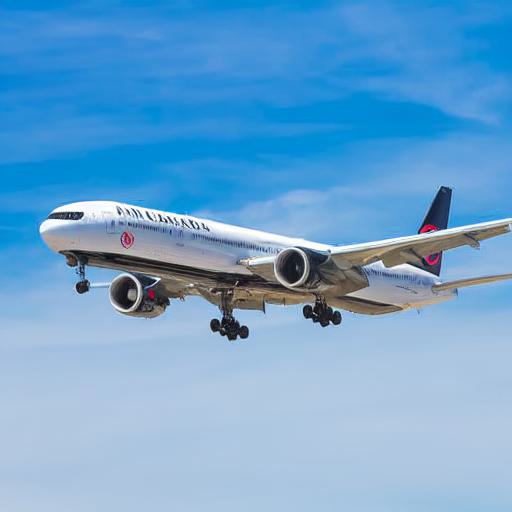} \\
-0.3 & \includegraphics[align=c,width=0.15\linewidth]{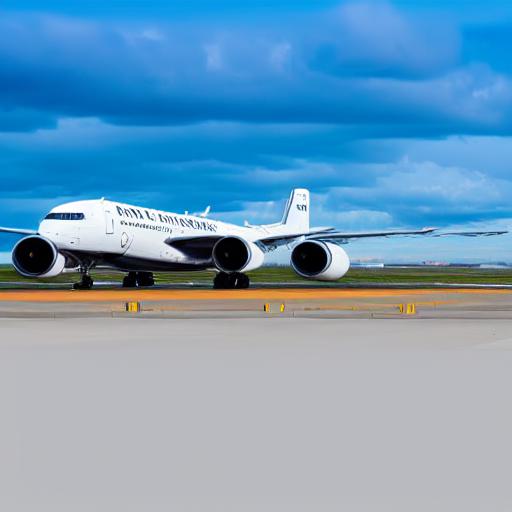} & \includegraphics[align=c,width=0.15\linewidth]{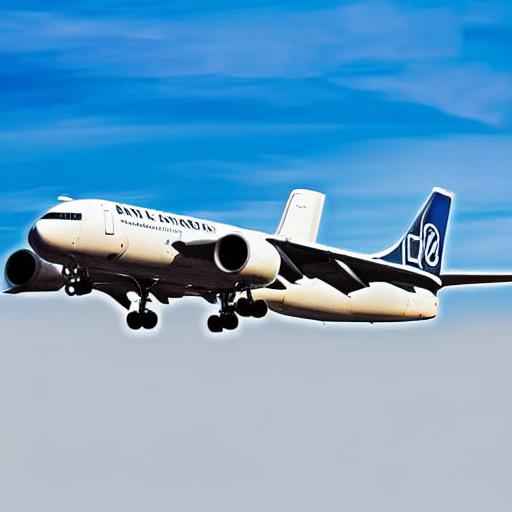} & \includegraphics[align=c,width=0.15\linewidth]{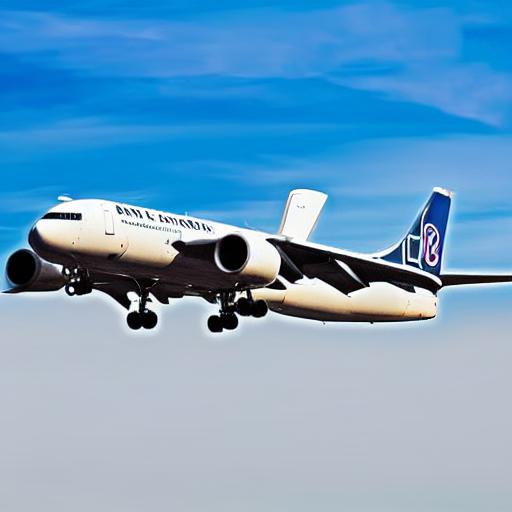} & \includegraphics[align=c,width=0.15\linewidth]{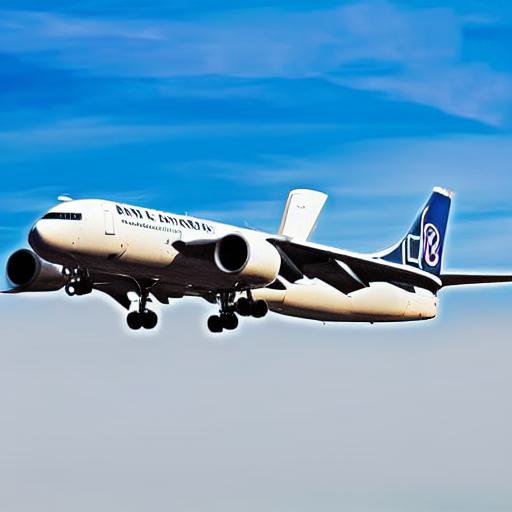} & \includegraphics[align=c,width=0.15\linewidth]{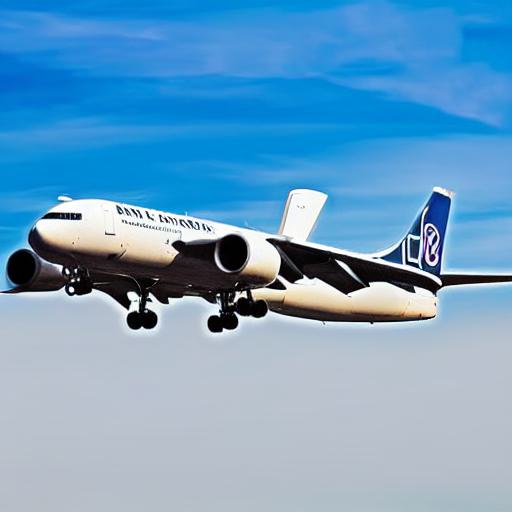} \\
-0.7 & \includegraphics[align=c,width=0.15\linewidth]{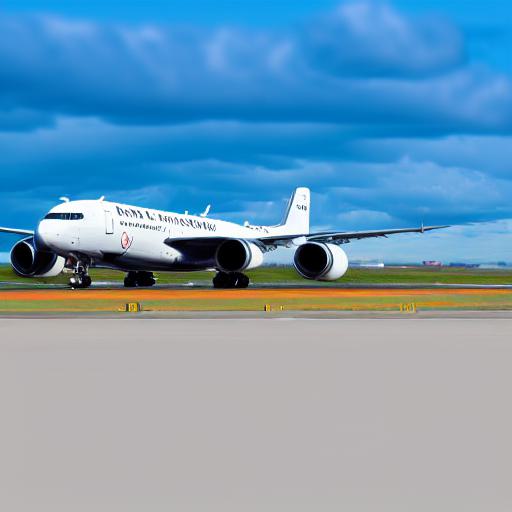} & \includegraphics[align=c,width=0.15\linewidth]{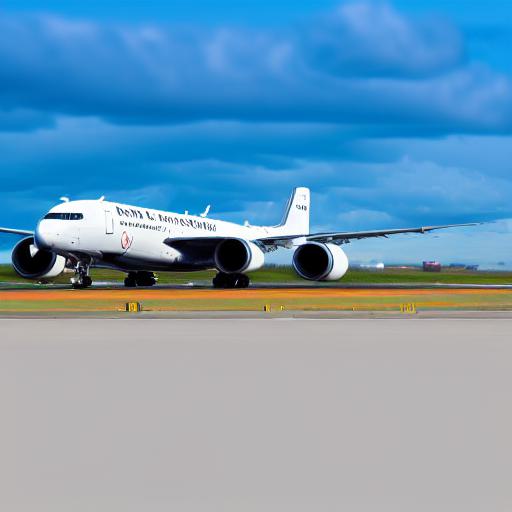} & \includegraphics[align=c,width=0.15\linewidth]{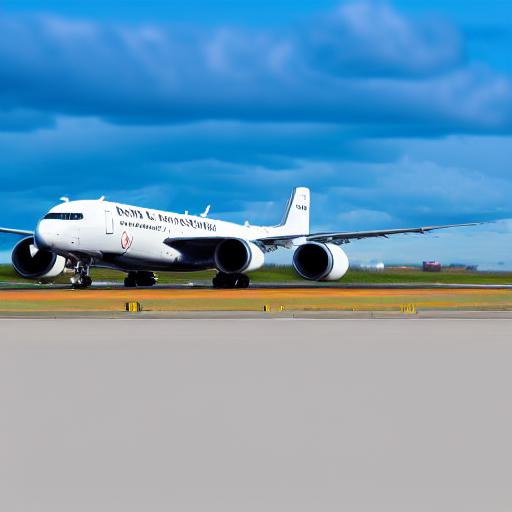} & \includegraphics[align=c,width=0.15\linewidth]{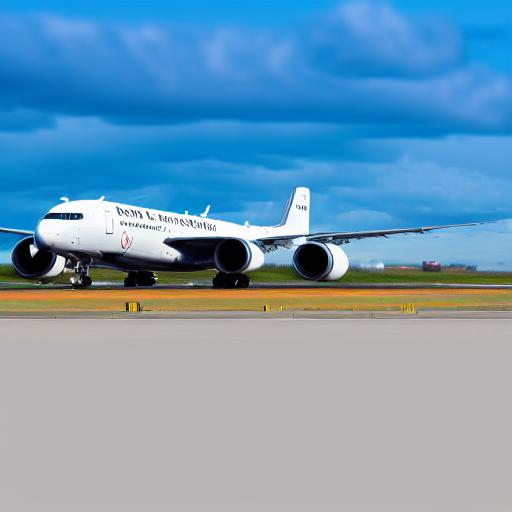} & \includegraphics[align=c,width=0.15\linewidth]{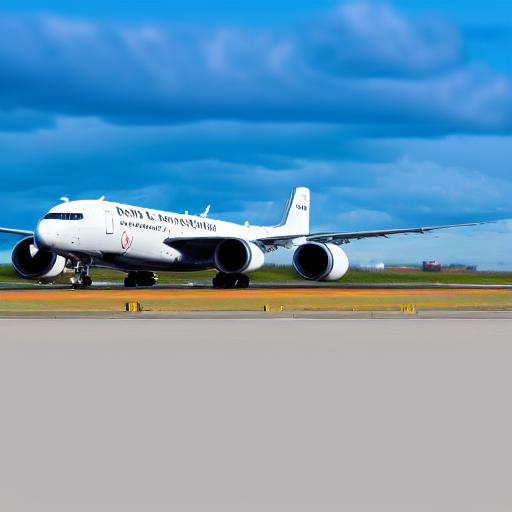} \\
\end{tabular}
\caption{Results of \text{MDP-$\boldsymbol \beta$} using constant schedule when varying guidance scale $\beta$.}
\label{fig:guidance}
\end{figure*}

Recall the formula we use for MDP-$\boldsymbol \beta$:
\begin{equation}\label{eq:guidance}
\begin{aligned}
    \boldsymbol\epsilon_t^{(\star)} = & \boldsymbol\epsilon_\theta\left(\mathbf{x}_t, \mathbf{c}^{(A)}, t\right) + \\
    & \beta\left(\boldsymbol\epsilon_\theta\left(\mathbf{x}_t, \mathbf{c}^{(A)}, t\right) - \boldsymbol\epsilon_\theta\left(\mathbf{x}_t, \mathbf{c}^{(B)}, t\right)\right),
\end{aligned}
\end{equation}
where $\beta \in \mathbf{R}$ is called guidance scale. Intuitively, with Eq.~\eqref{eq:guidance}, we want the output to have more characteristics from $\mathbf{c}^{(A)}$ when generating using the new condition $\mathbf{c}^{(B)}$ to preserve the layout. This is a bit different from the original classifier free guidance, where condition $\mathbf{c}^{(B)}$ is an empty condition and $\beta$ is a positive real number. In our specific image editing task, we find that when using a positive $\beta$ (\eg, the first row in \cref{fig:guidance}), the edited image will be too close to condition $\mathbf{c}^{(A)}$ while losing both the layout from the input image and the new semantics from condition $\mathbf{c}^{(B)}$. Only when the guidance scale $\beta$ is a negative number, can the new semantics be added into the edited image. In this situation, the guidance becomes a linear interpolation operation: 
\begin{equation}\label{eq:guidance-lerp}
\begin{aligned}
    \boldsymbol\epsilon_t^{(\star)} = \omega\cdot\boldsymbol\epsilon_\theta\left(\mathbf{x}_t, \mathbf{c}^{(A)}, t\right) + (1-\omega)\cdot\boldsymbol\epsilon_\theta\left(\mathbf{x}_t, \mathbf{c}^{(B)}, t\right),
\end{aligned}
\end{equation}
where $\omega = 1 - \beta$ is a linear factor from 0 to 1, which means both condition $\mathbf{c}^{(A)}$ and condition $\mathbf{c}^{(B)}$ contribute positively to the generation process. Overall, $\beta$ has a larger influence over the edited image than varying the manipulation timesteps. We empirically find that $\beta \in [-0.6, -0.8]$ can obtain better results. 

\subsection{Prompt-to-Prompt}
\begin{figure*}[h]
\centering
\setlength{\tabcolsep}{1pt}
\begin{tabular}{cccccccc}
Self/Cross & 50 - 30 & 50 - 20  & 50 - 10 & 50 - 0 & 40 - 20 & 40 - 10 & 40 - 0 \\
50 - 40 & \includegraphics[align=c,width=0.13\linewidth]{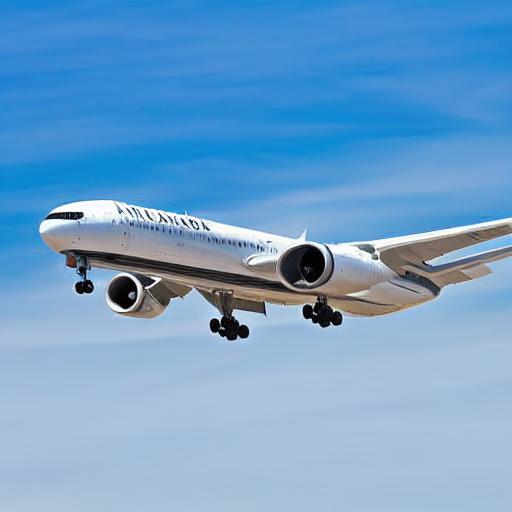}  & \includegraphics[align=c,width=0.13\linewidth]{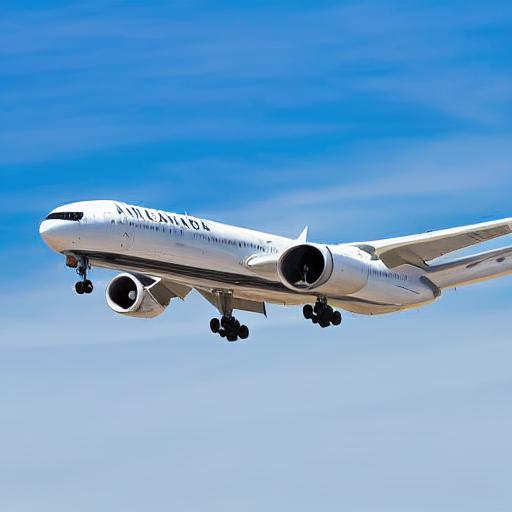} & \includegraphics[align=c,width=0.13\linewidth]{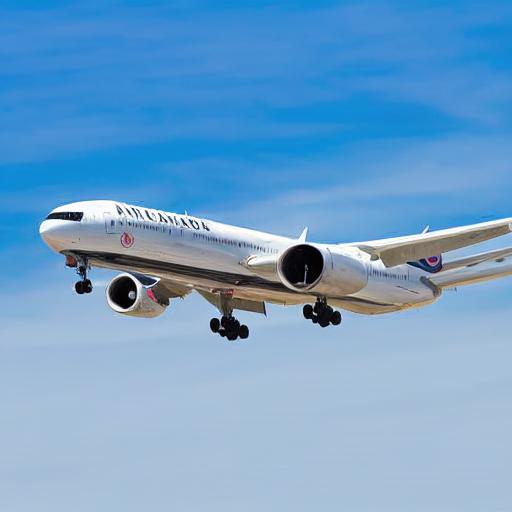} & 
\includegraphics[align=c,width=0.13\linewidth]{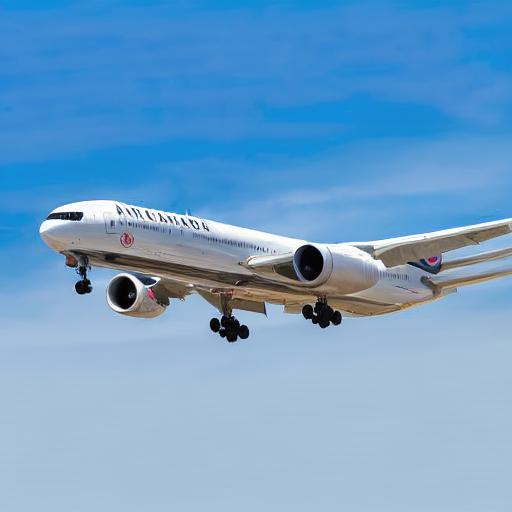} & \includegraphics[align=c,width=0.13\linewidth]{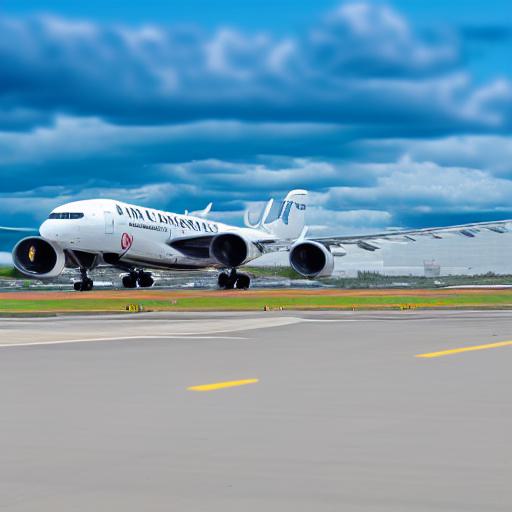} & \includegraphics[align=c,width=0.13\linewidth]{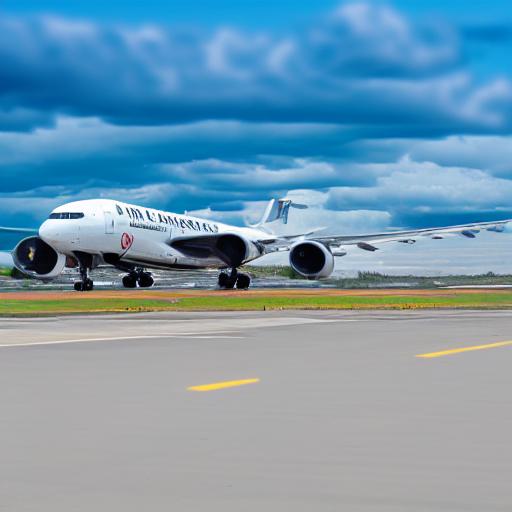} & \includegraphics[align=c,width=0.13\linewidth]{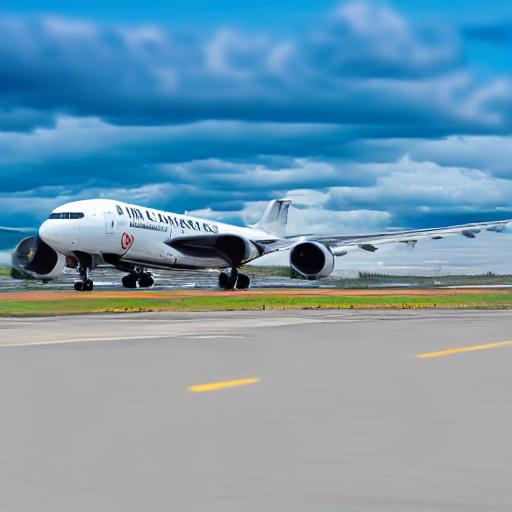} \\
50 - 30 & \includegraphics[align=c,width=0.13\linewidth]{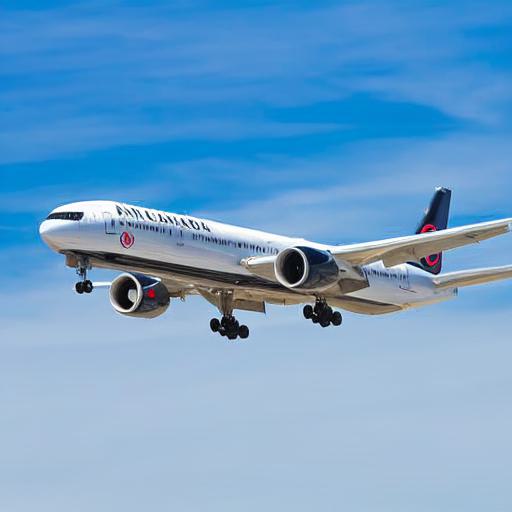}  & \includegraphics[align=c,width=0.13\linewidth]{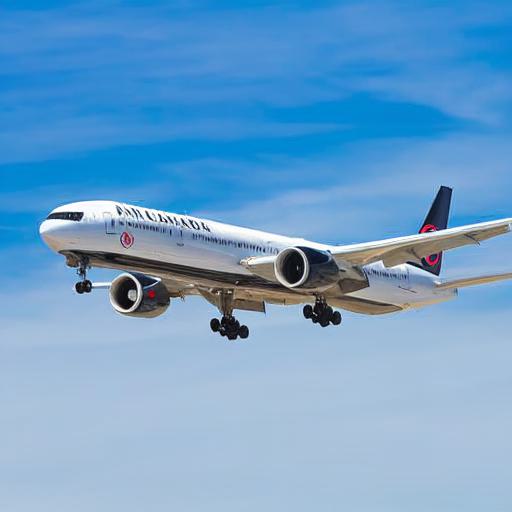} & \includegraphics[align=c,width=0.13\linewidth]{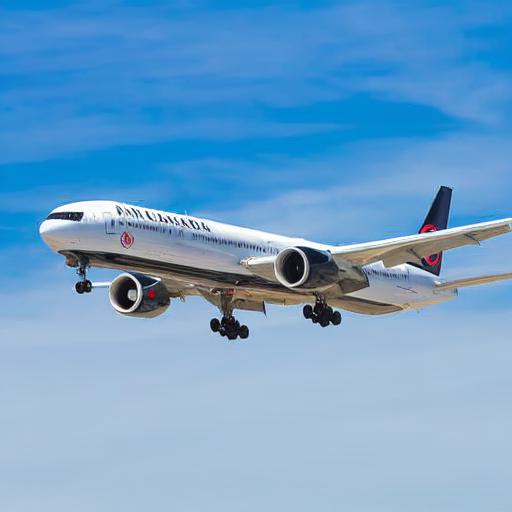} & 
\includegraphics[align=c,width=0.13\linewidth]{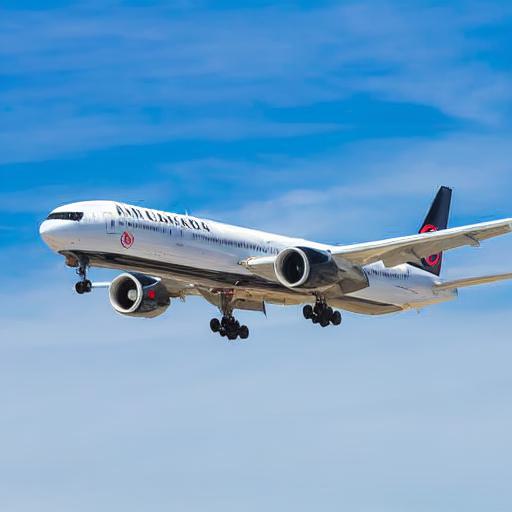} & \includegraphics[align=c,width=0.13\linewidth]{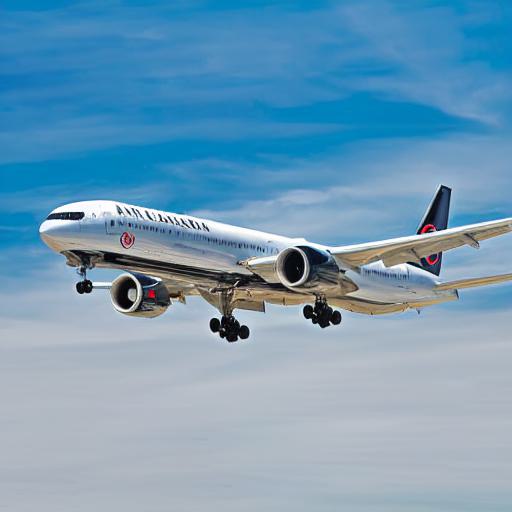} & \includegraphics[align=c,width=0.13\linewidth]{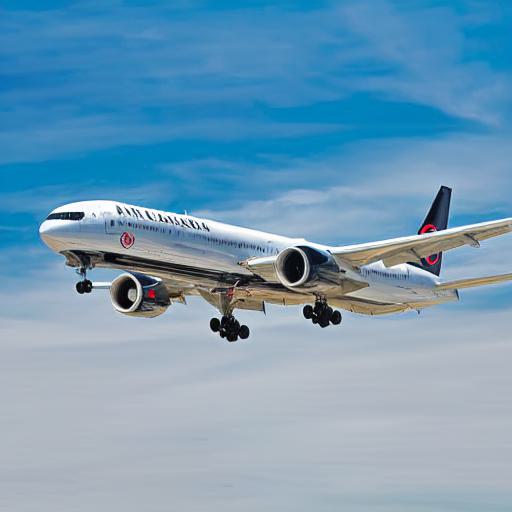} & \includegraphics[align=c,width=0.13\linewidth]{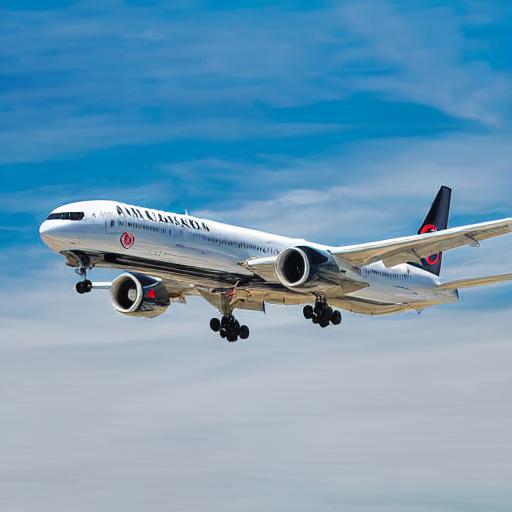} \\
50 - 20 & \includegraphics[align=c,width=0.13\linewidth]{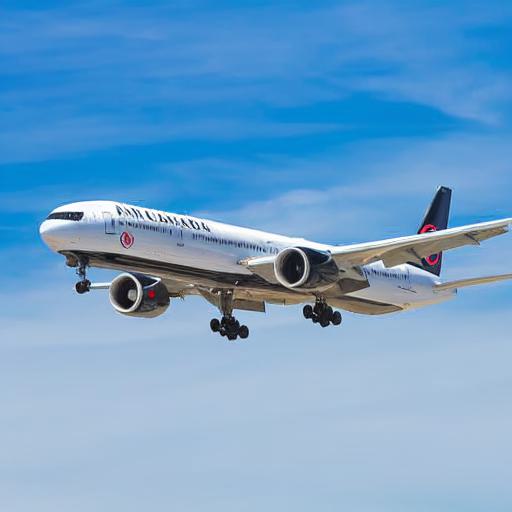}  & \includegraphics[align=c,width=0.13\linewidth]{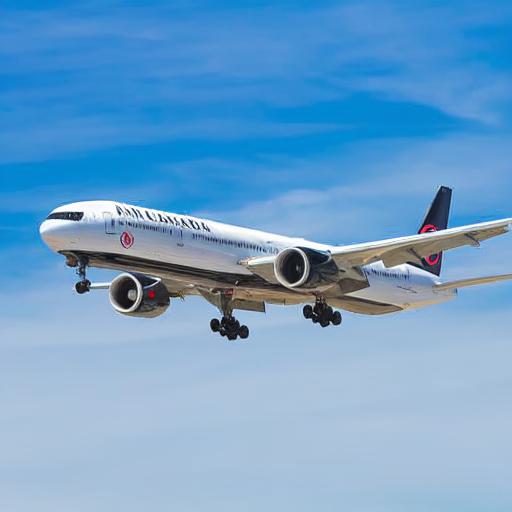} & \includegraphics[align=c,width=0.13\linewidth]{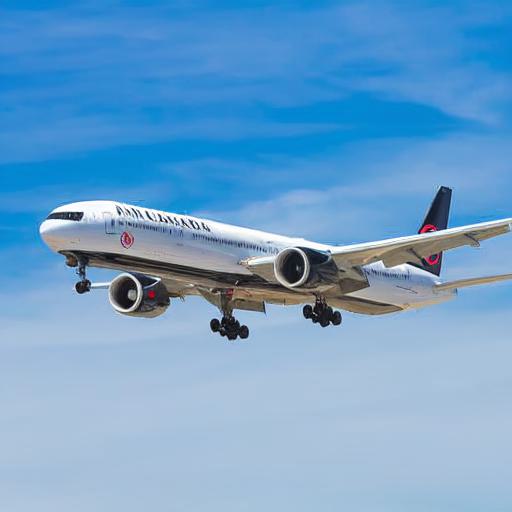} & 
\includegraphics[align=c,width=0.13\linewidth]{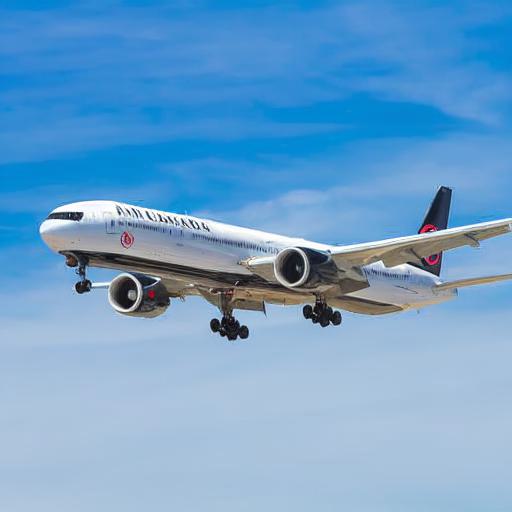} & \includegraphics[align=c,width=0.13\linewidth]{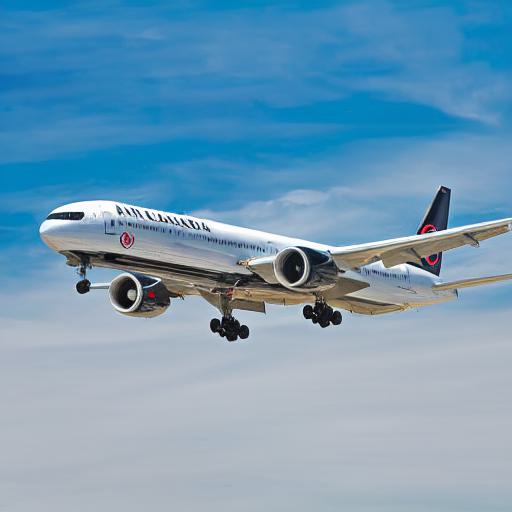} & \includegraphics[align=c,width=0.13\linewidth]{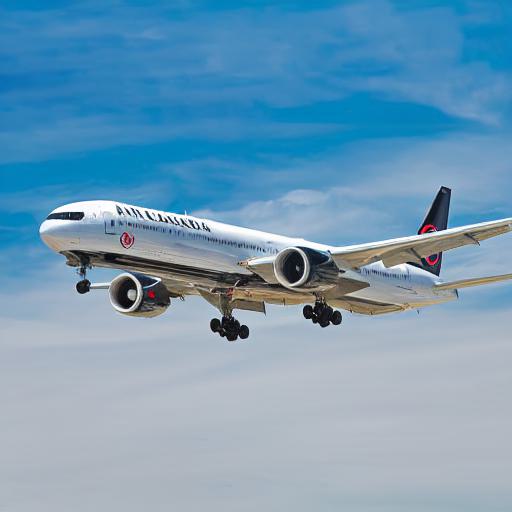} & \includegraphics[align=c,width=0.13\linewidth]{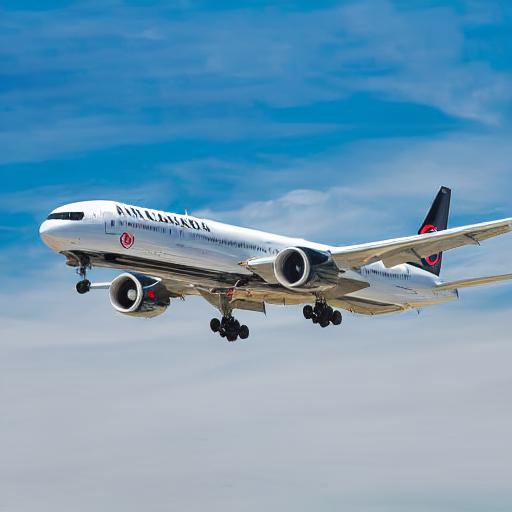} \\
50 - 10 & \includegraphics[align=c,width=0.13\linewidth]{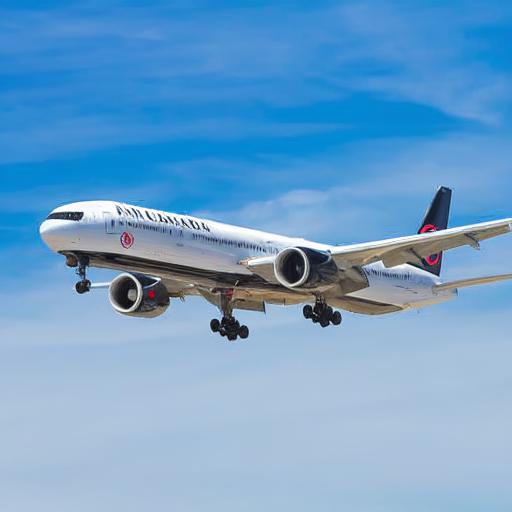}  & \includegraphics[align=c,width=0.13\linewidth]{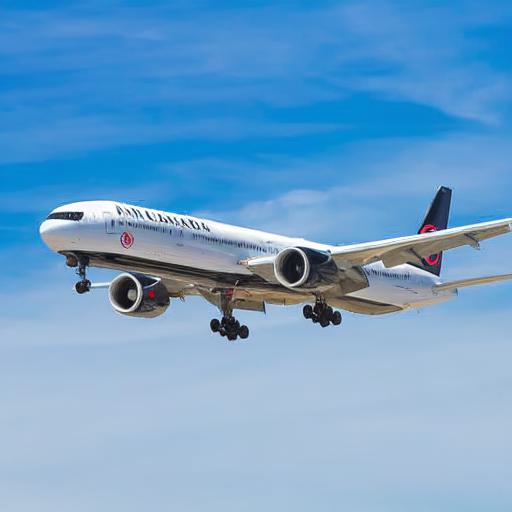} & \includegraphics[align=c,width=0.13\linewidth]{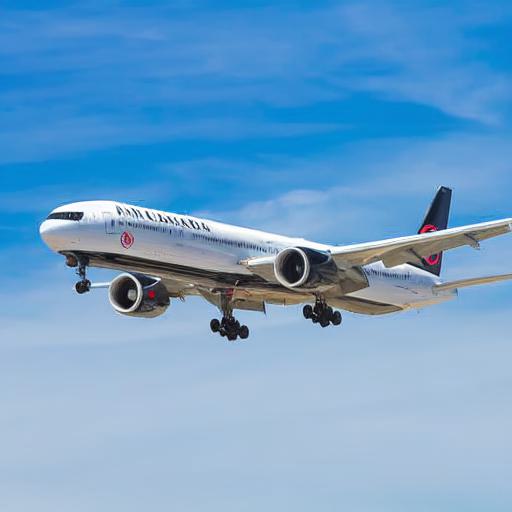} & 
\includegraphics[align=c,width=0.13\linewidth]{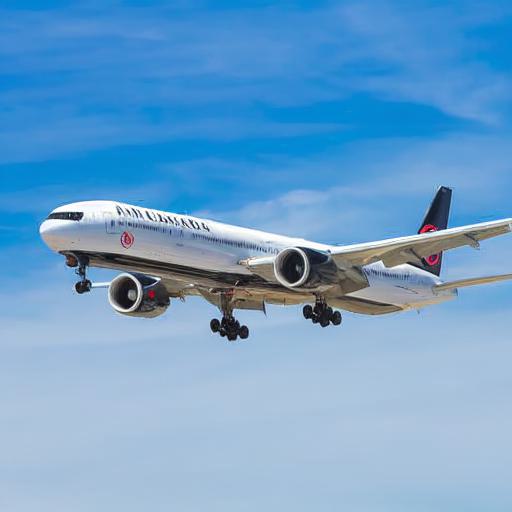} & \includegraphics[align=c,width=0.13\linewidth]{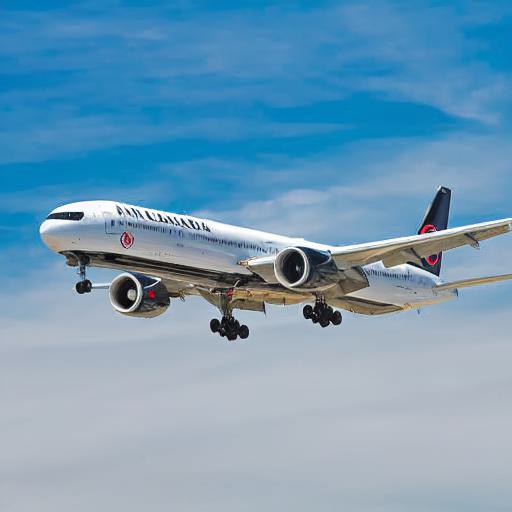} & \includegraphics[align=c,width=0.13\linewidth]{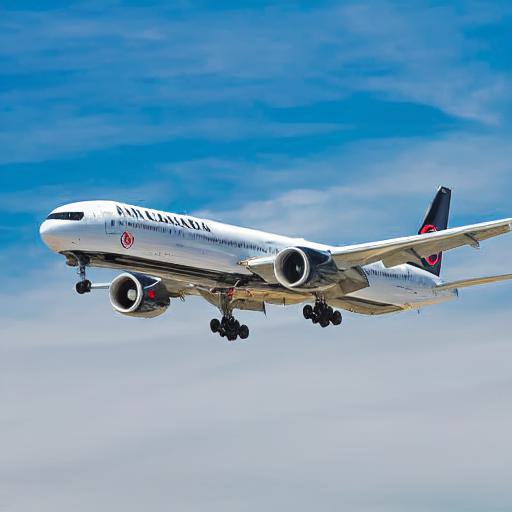} & \includegraphics[align=c,width=0.13\linewidth]{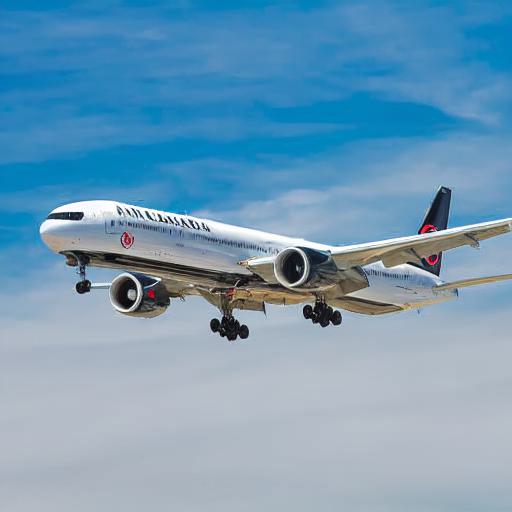} \\
50 - 0 & \includegraphics[align=c,width=0.13\linewidth]{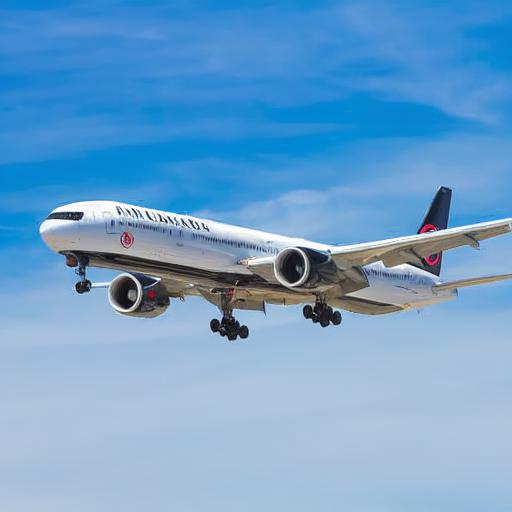}  & \includegraphics[align=c,width=0.13\linewidth]{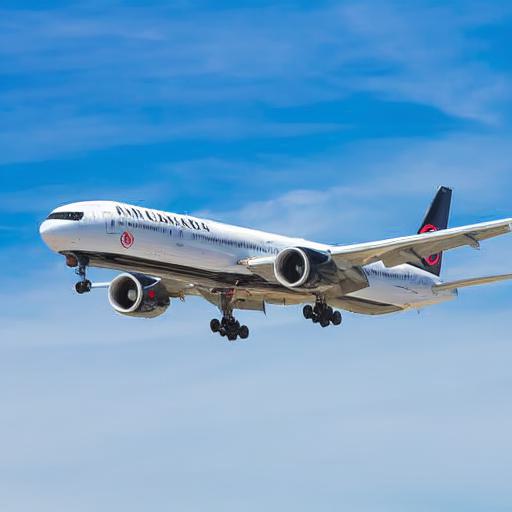} & \includegraphics[align=c,width=0.13\linewidth]{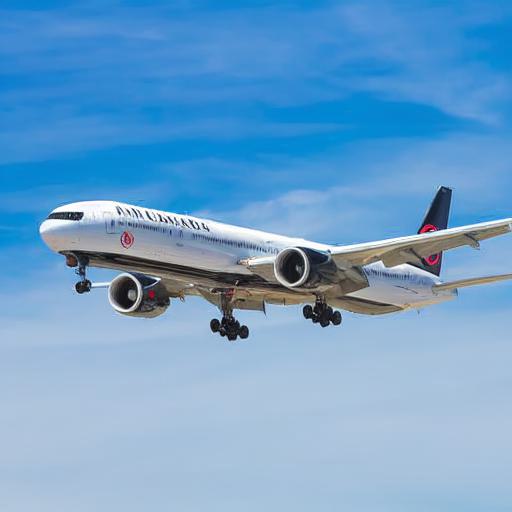} & 
\includegraphics[align=c,width=0.13\linewidth]{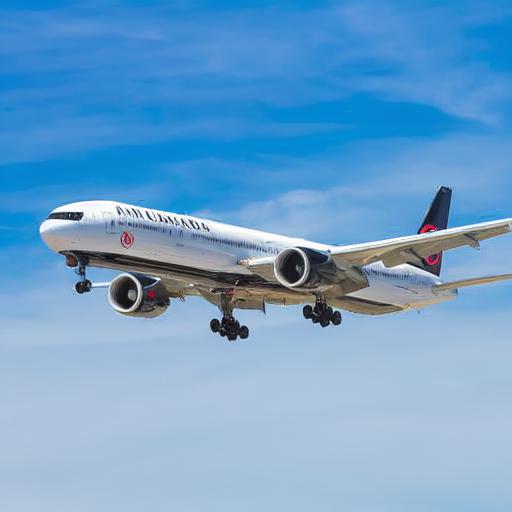} & \includegraphics[align=c,width=0.13\linewidth]{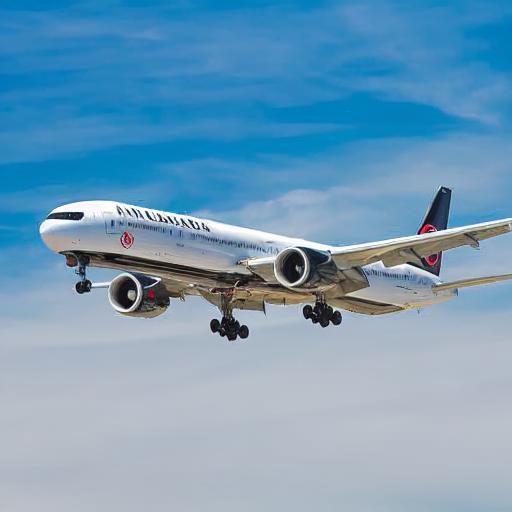} & \includegraphics[align=c,width=0.13\linewidth]{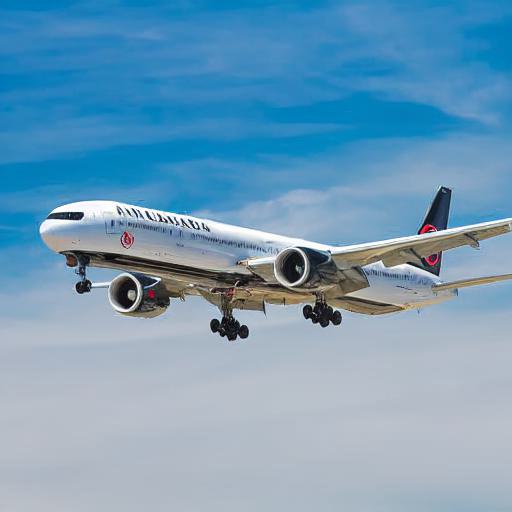} & \includegraphics[align=c,width=0.13\linewidth]{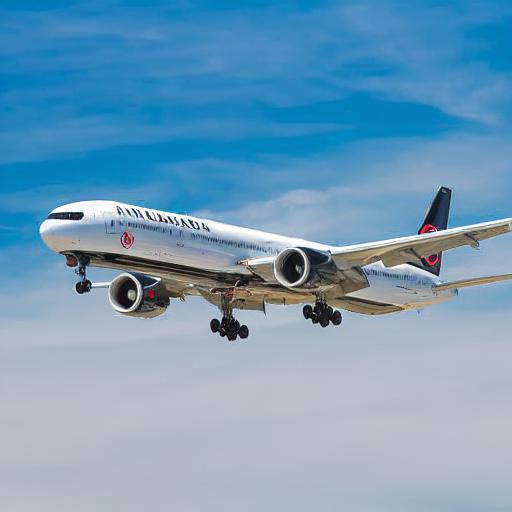} \\
\end{tabular}
\caption{Results of Prompt-to-Prompt using constant schedule when varying the timestpes to inject self-attention maps and cross-attention maps. We show the results when we vary $t_{\text{max}}$ and $t_{\text{min}}$ for either injecting self-attention maps or cross attention maps.}
\label{fig:p2p_global}
\end{figure*}

\begin{figure*}[h]
\centering
\scriptsize
\setlength{\tabcolsep}{1pt}
\begin{tabular}{cccc}
Self/Cross & 50 - 30 & 50 - 10 & 40 - 0 \\
50-40 & \includegraphics[align=c,width=0.15\linewidth]{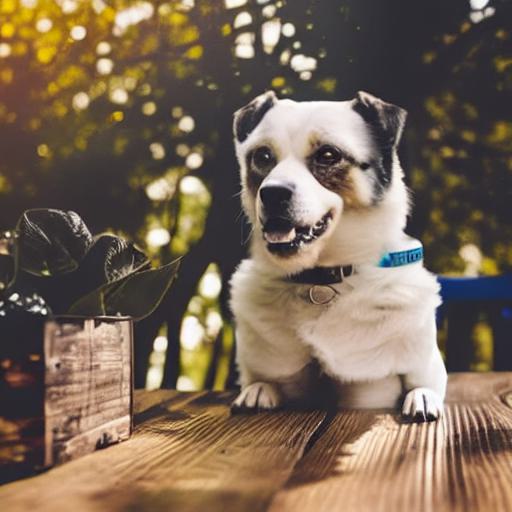} & \includegraphics[align=c,width=0.15\linewidth]{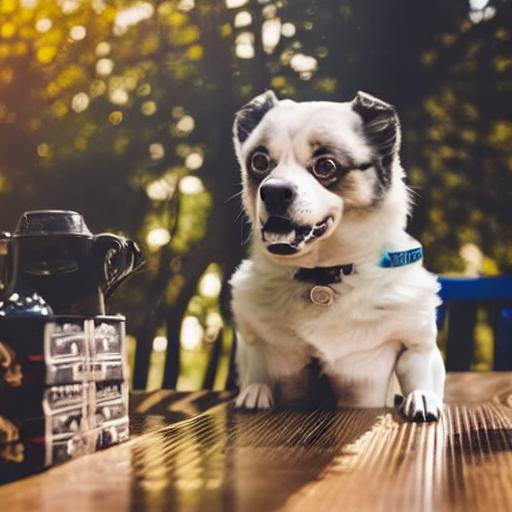} & \includegraphics[align=c,width=0.15\linewidth]{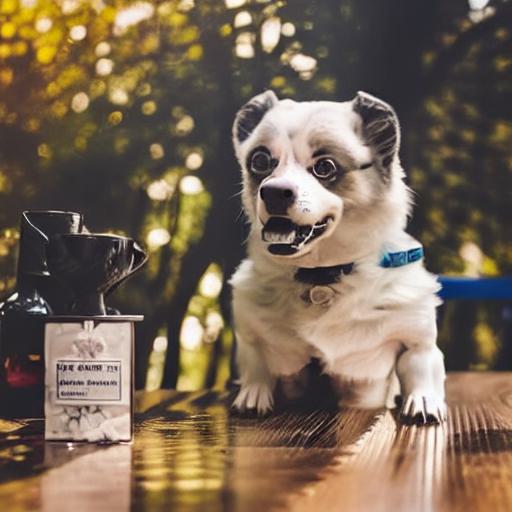} \\
50-30 & \includegraphics[align=c,width=0.15\linewidth]{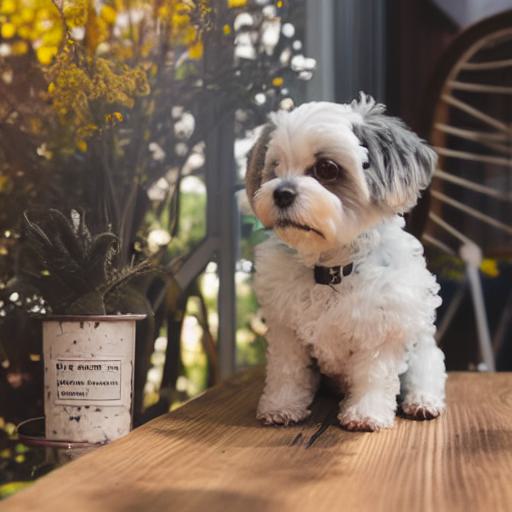} & \includegraphics[align=c,width=0.15\linewidth]{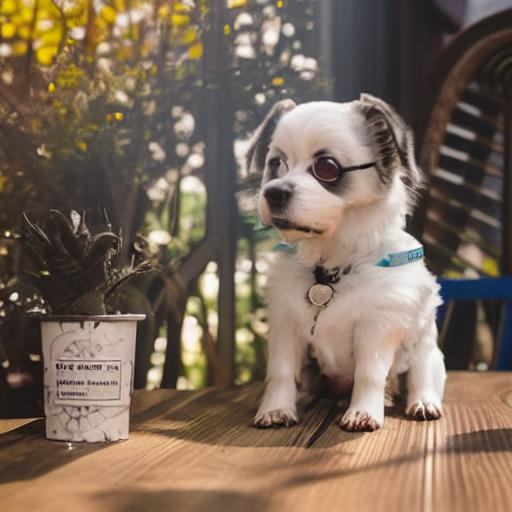} & \includegraphics[align=c,width=0.15\linewidth]{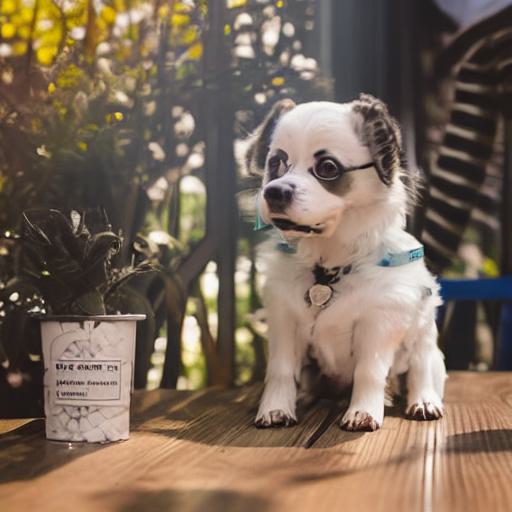} \\
50-20 & \includegraphics[align=c,width=0.15\linewidth]{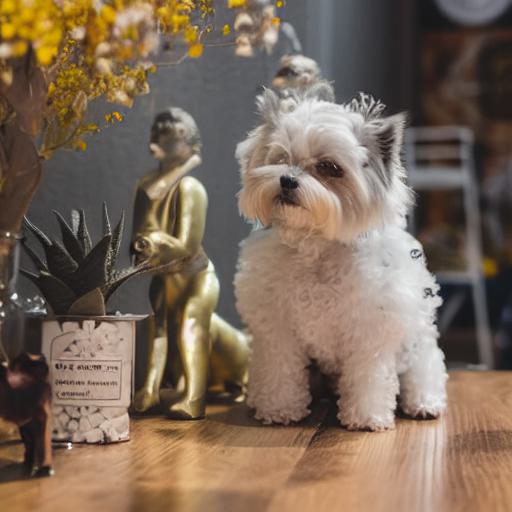} & \includegraphics[align=c,width=0.15\linewidth]{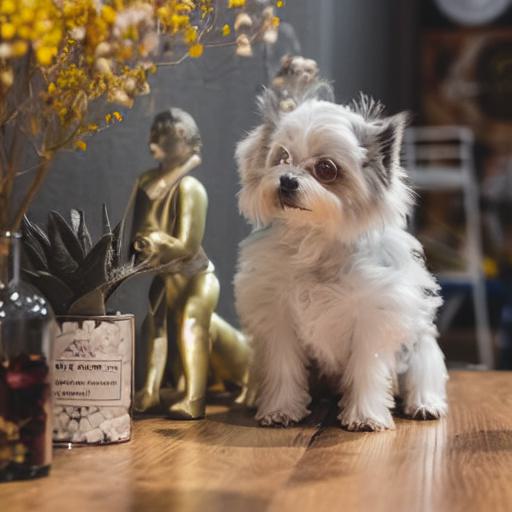} & \includegraphics[align=c,width=0.15\linewidth]{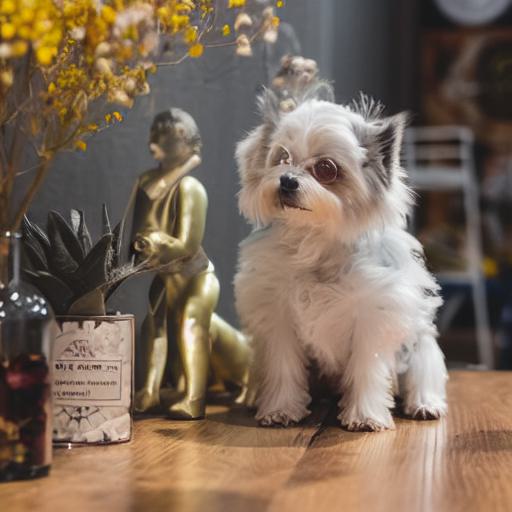} \\
50-10 & \includegraphics[align=c,width=0.15\linewidth]{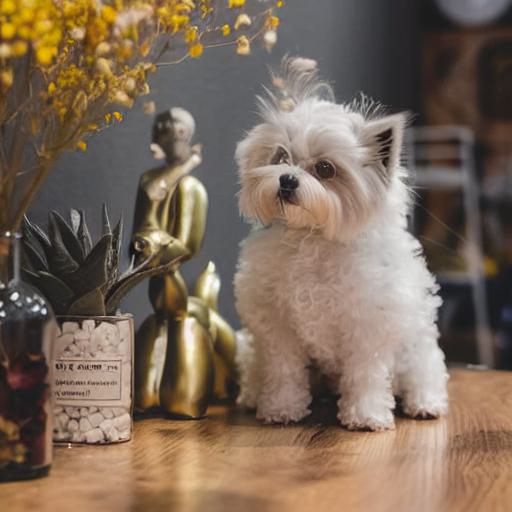} & \includegraphics[align=c,width=0.15\linewidth]{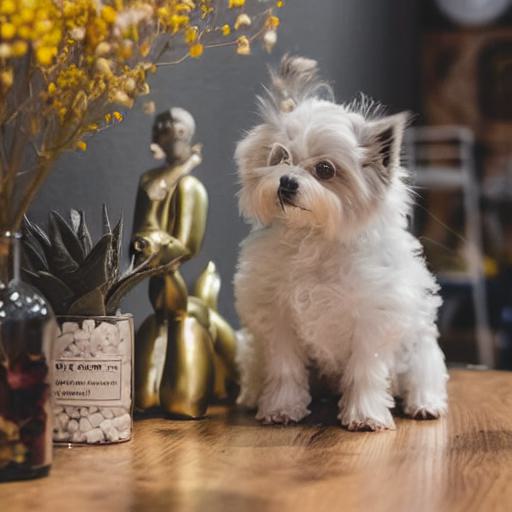} & \includegraphics[align=c,width=0.15\linewidth]{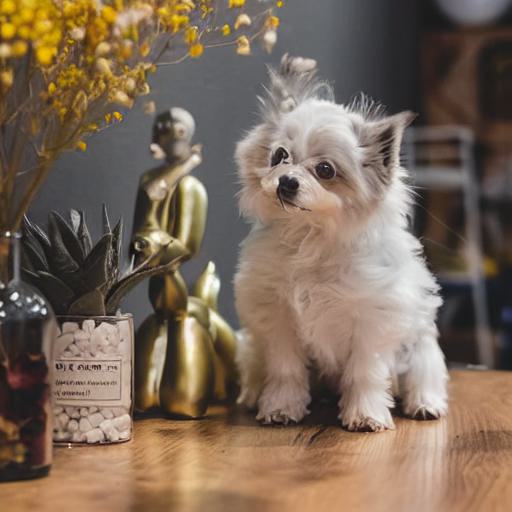} \\
\end{tabular}
\caption{Additional results of Prompt-to-Prompt using constant schedule when varying the timestpes to inject self-attention maps and cross-attention maps. We show the results when we vary $t_{\text{max}}$ and $t_{\text{min}}$ for either injecting self-attention maps or cross attention maps.}
\label{fig:p2p_local}
\end{figure*}

We refer to the official implementation\footnote{\href{https://github.com/google/prompt-to-prompt}{https://github.com/google/prompt-to-prompt}}, which also utilizes Null-text Inversion to invert the real image and uses DDIM as the sampler. We observe that instead of replacing cross-attention maps, injecting self-attention maps has a bigger influence on the edited images. As in \cref{fig:p2p_global} the effect of injecting self-attention maps is not obvious enough, we additionally show one local editing example in \cref{fig:p2p_local}, where when fixing the timesteps to inject self-attention maps, the edited image look almost the same under different timesteps to inject cross attention maps; while under different timesteps to inject the self-attention maps, the generated images look more different.

We therefore suggest a tuning of the timesteps to inject the self-attention maps. The more timesteps to do that manipulation, the more the edited image will look like the input image. In general, for global editing, where the changes from the input image to the edited image should be larger, a smaller $T_M$ can be considered compared to which for local editing. A general setting for $T_M$ to replace the cross attention map like 20 or 40 is enough.

\clearpage

\section{More results}
\subsection{Visual comarisons}
\begin{figure*}[h]
\centering
\scriptsize
\setlength{\tabcolsep}{1pt}
\begin{tabular}{cccccccc}
Input & Edit & P2P & \textbf{MDP-$\boldsymbol\epsilon_t$} & Input & Edit & P2P & \textbf{MDP-$\boldsymbol\epsilon_t$} \\
\includegraphics[align=c,width=0.12\linewidth]{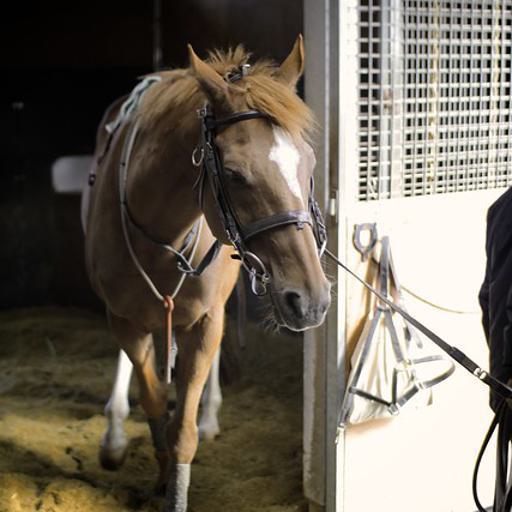} & 
\begin{tabular}[c]{@{}c@{}}``Horse''\\to\\``Zebra''\end{tabular}& \includegraphics[align=c,width=0.12\linewidth]{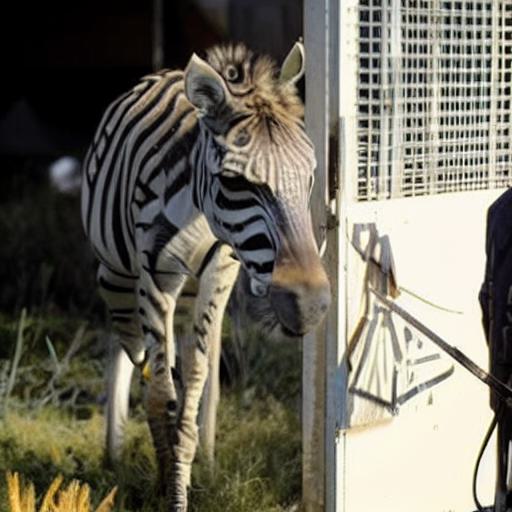}  & \includegraphics[align=c,width=0.12\linewidth]{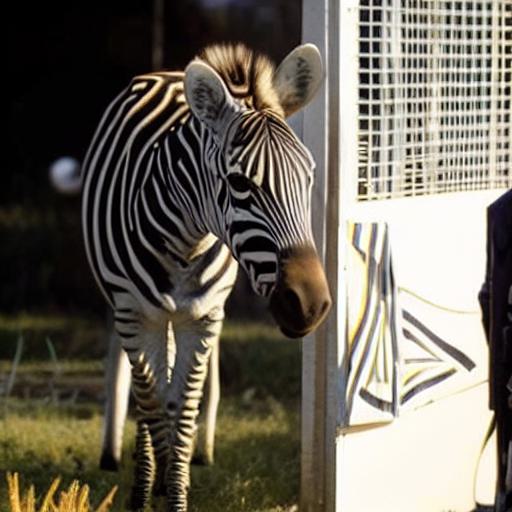} \hspace{3mm} & \includegraphics[align=c,width=0.12\linewidth]{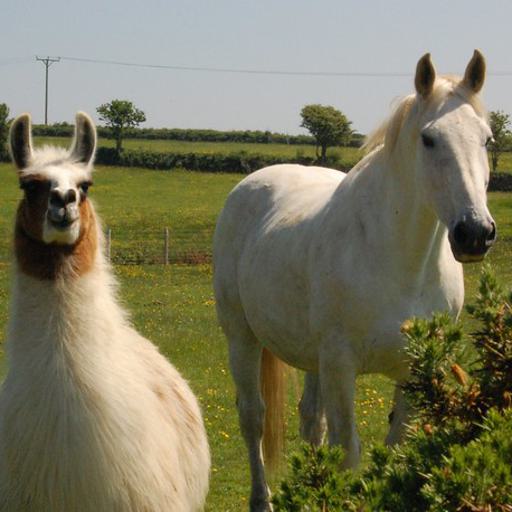}  & \begin{tabular}[c]{@{}c@{}}``Horse''\\to\\``Zebra''\end{tabular} & \includegraphics[align=c,width=0.12\linewidth]{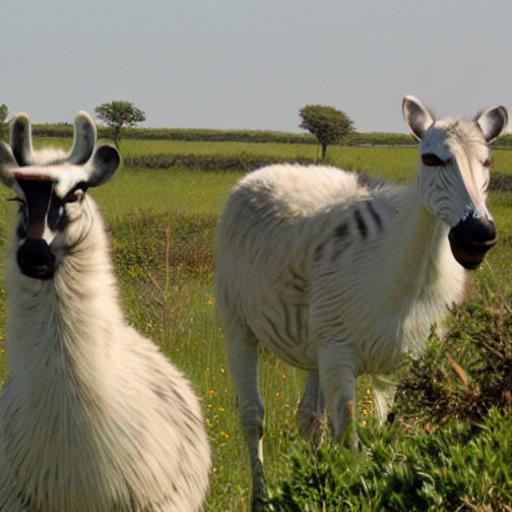}  & \includegraphics[align=c,width=0.12\linewidth]{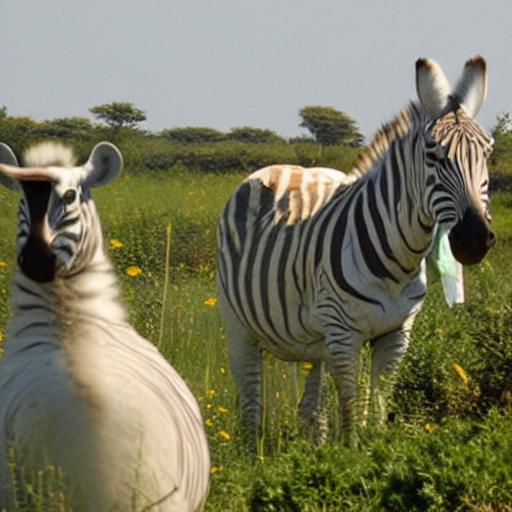} \vspace{1mm} \\ 
\includegraphics[align=c,width=0.12\linewidth]{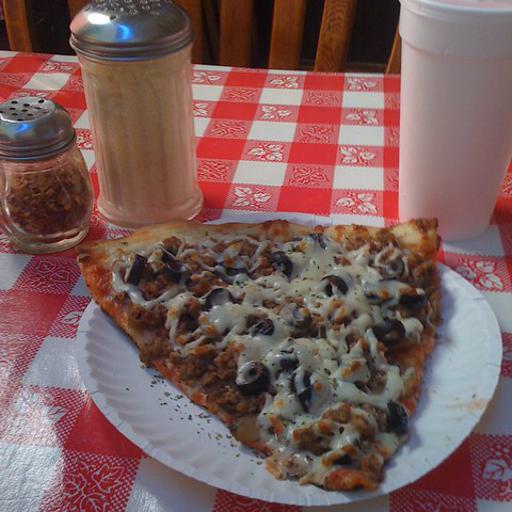} & 
  \begin{tabular}[c]{@{}c@{}}``Pizza''\\to\\``Cake''\end{tabular}& \includegraphics[align=c,width=0.12\linewidth]{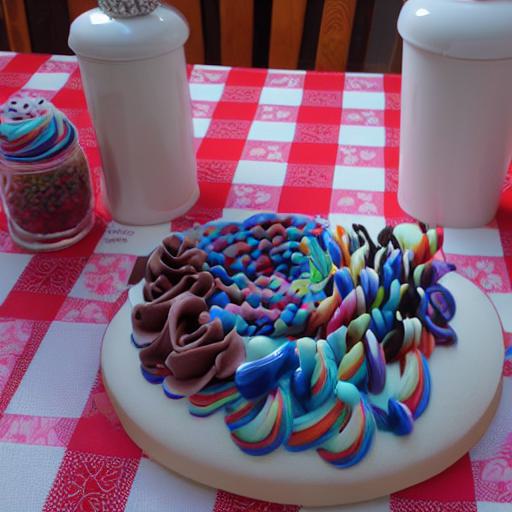} & \includegraphics[align=c,width=0.12\linewidth]{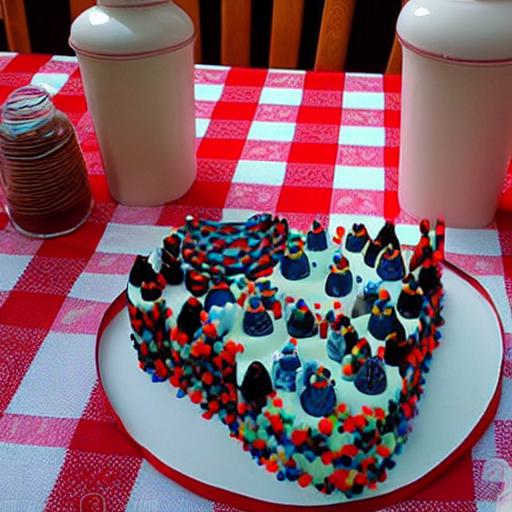} \hspace{3mm} & \includegraphics[align=c,width=0.12\linewidth]{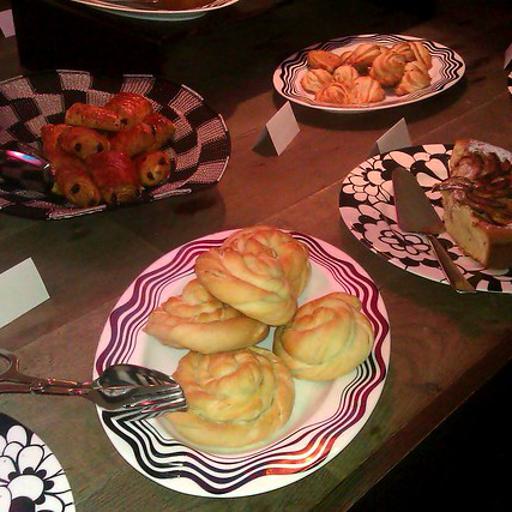} &   \begin{tabular}[c]{@{}c@{}}``Bread''\\to\\``Cake''\end{tabular} & \includegraphics[align=c,width=0.12\linewidth]{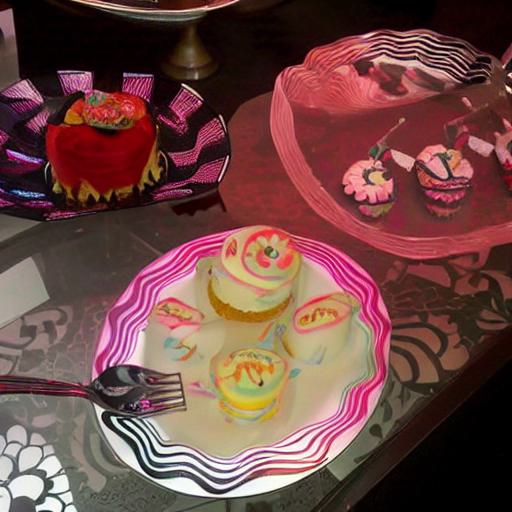} & \includegraphics[align=c,width=0.12\linewidth]{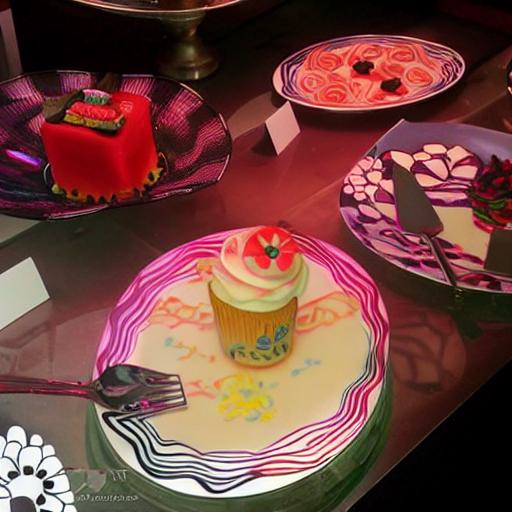} \vspace{1mm} \\
\includegraphics[align=c,width=0.12\linewidth]{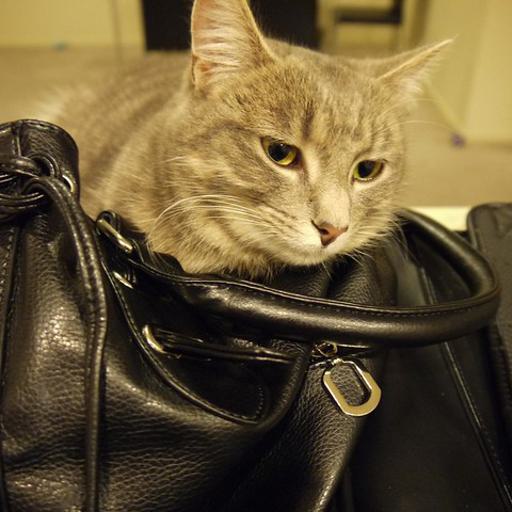} &   \begin{tabular}[c]{@{}c@{}}``Cat''\\to\\``Dog''\end{tabular} & \includegraphics[align=c,width=0.12\linewidth]{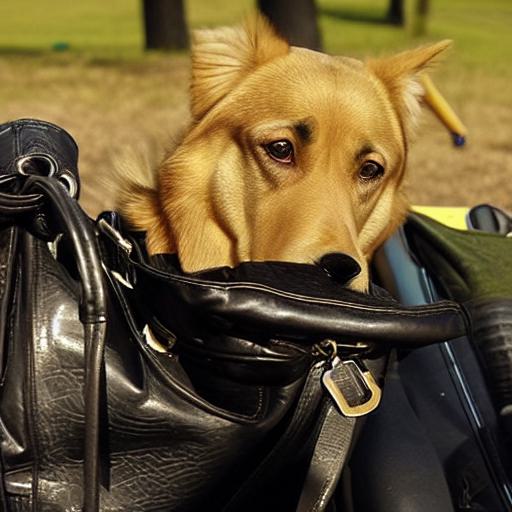}  & \includegraphics[align=c,width=0.12\linewidth]{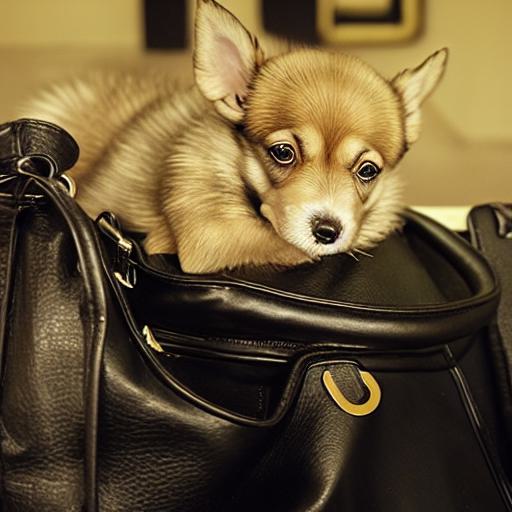} \hspace{3mm} & \includegraphics[align=c,width=0.12\linewidth]{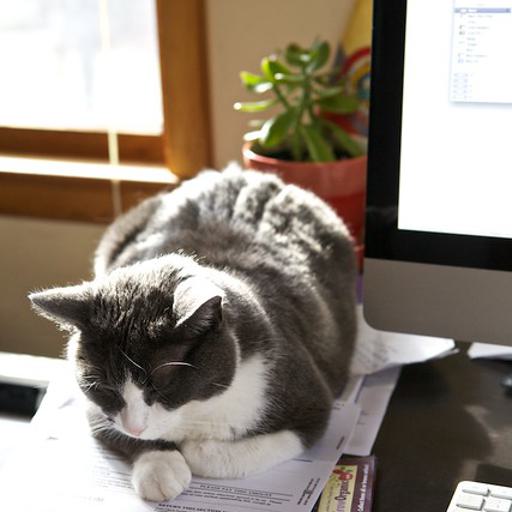}  & \begin{tabular}[c]{@{}c@{}}``Cat''\\to\\``Dog''\end{tabular}  & \includegraphics[align=c,width=0.12\linewidth]{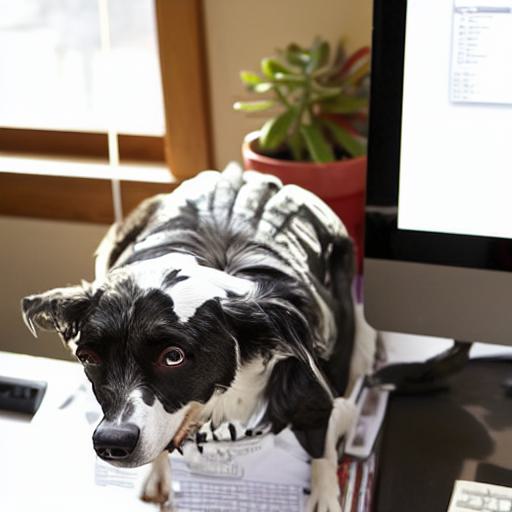}  & \includegraphics[align=c,width=0.12\linewidth]{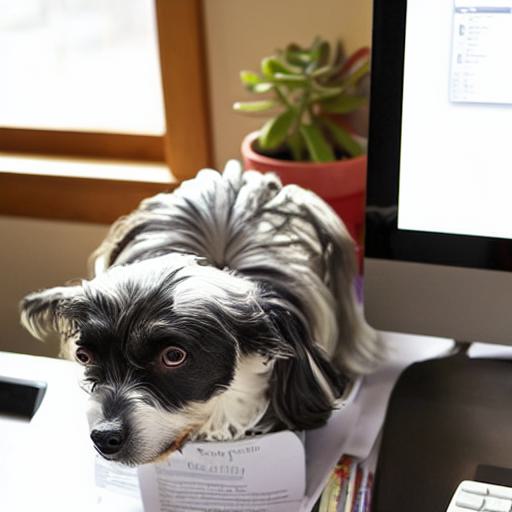} \vspace{1mm} \\ 
\includegraphics[align=c,width=0.12\linewidth]{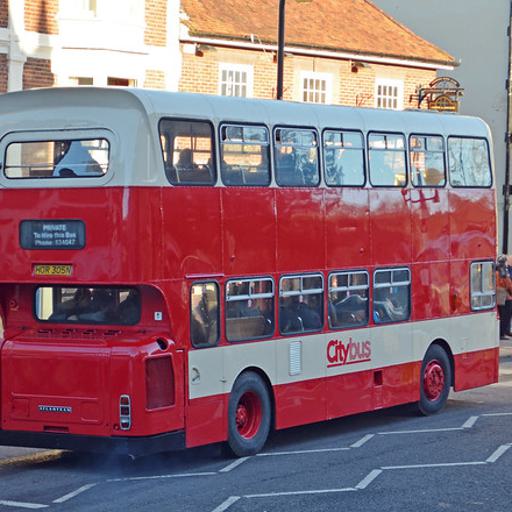} &   \begin{tabular}[c]{@{}c@{}}``Bus''\\to\\``Truck''\end{tabular} & \includegraphics[align=c,width=0.12\linewidth]{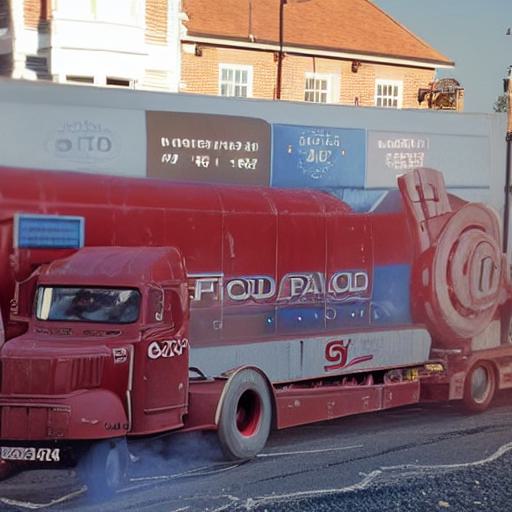} & \includegraphics[align=c,width=0.12\linewidth]{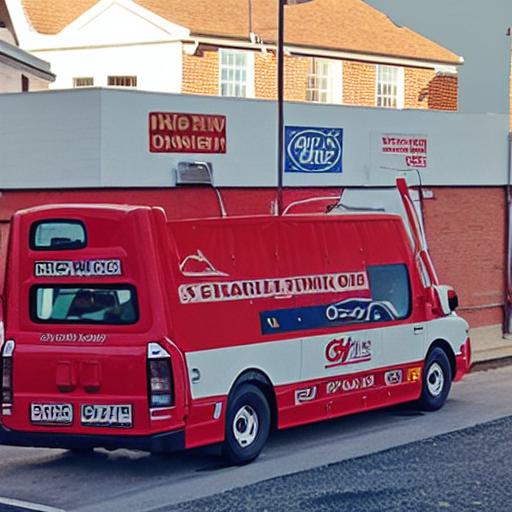} \hspace{3mm} & \includegraphics[align=c,width=0.12\linewidth]{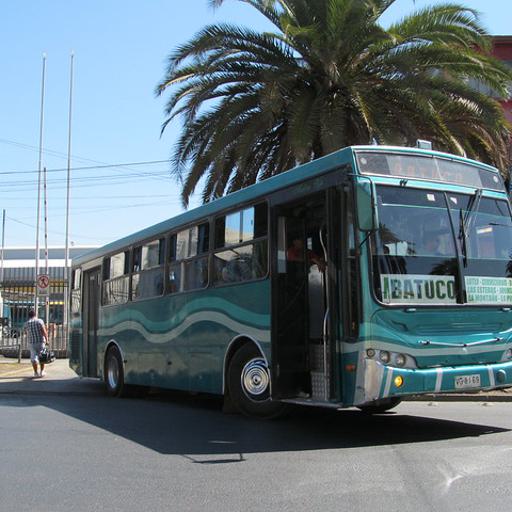} &   \begin{tabular}[c]{@{}c@{}}``Bus''\\to\\``Truck''\end{tabular} & \includegraphics[align=c,width=0.12\linewidth]{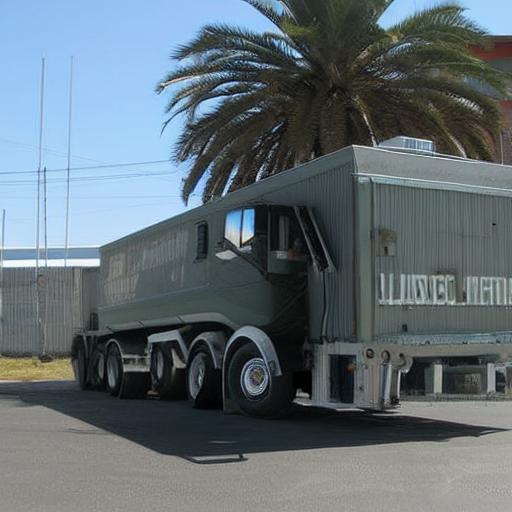} & \includegraphics[align=c,width=0.12\linewidth]{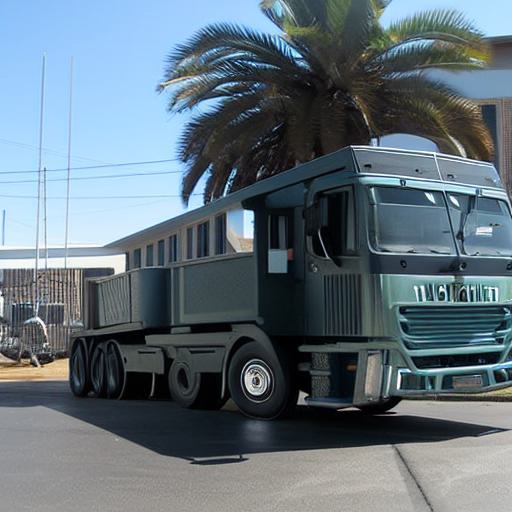} \vspace{1mm} \\
\end{tabular}
\caption{Results of changing object comparing Prompt-to-Prompt and \textbf{MDP-$\boldsymbol\epsilon_t$}.}
\label{fig:local-changing-object}
\end{figure*}

\begin{figure*}[h]
\centering
\scriptsize
\setlength{\tabcolsep}{1pt}
\begin{tabular}{cccccccc}
Input & Edit & P2P & \textbf{MDP-$\boldsymbol\epsilon_t$} & Input & Edit & P2P & \textbf{MDP-$\boldsymbol\epsilon_t$} \\
\includegraphics[align=c,width=0.12\linewidth]{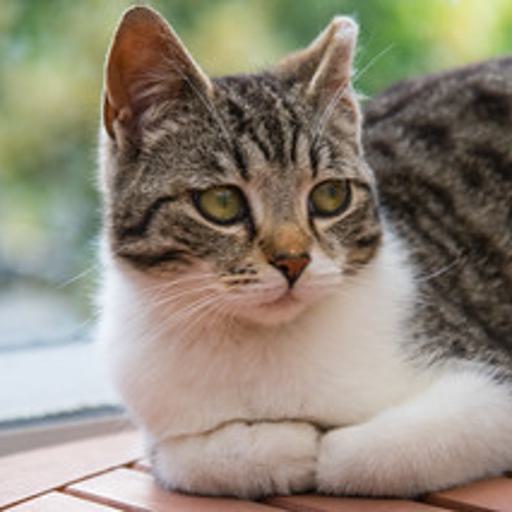} & 
    \begin{tabular}[c]{@{}c@{}}Add\\``Glasses''\end{tabular}& \includegraphics[align=c,width=0.12\linewidth]{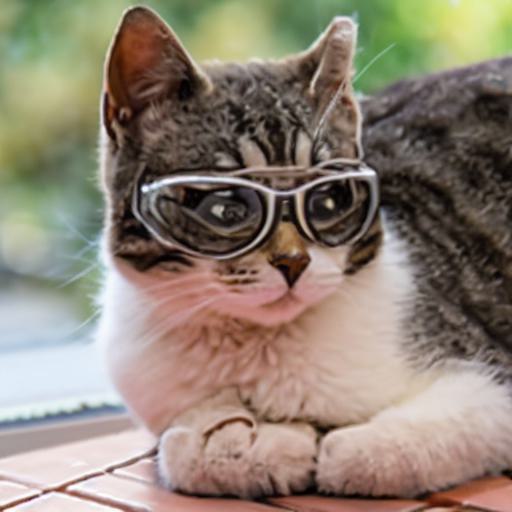}  & \includegraphics[align=c,width=0.12\linewidth]{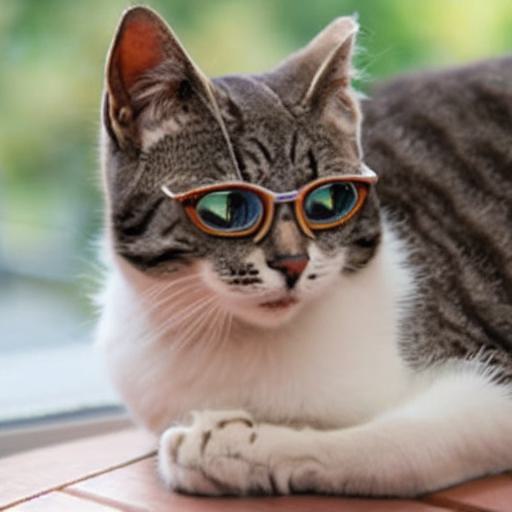} \hspace{3mm} & \includegraphics[align=c,width=0.12\linewidth]{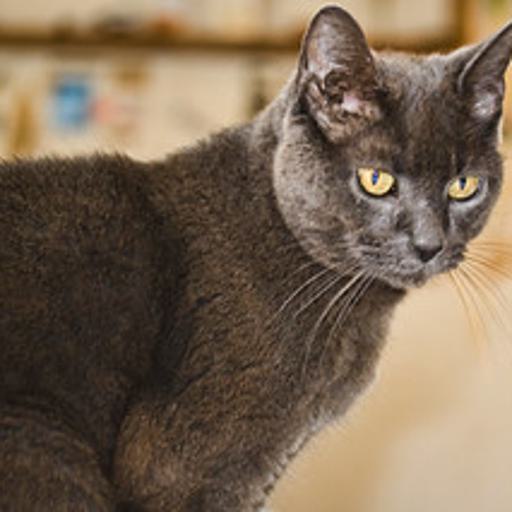}  &     \begin{tabular}[c]{@{}c@{}}Add\\``Glasses''\end{tabular}  & \includegraphics[align=c,width=0.12\linewidth]{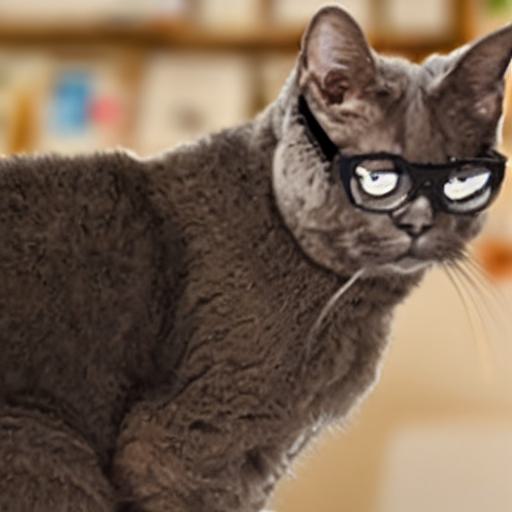}  & \includegraphics[align=c,width=0.12\linewidth]{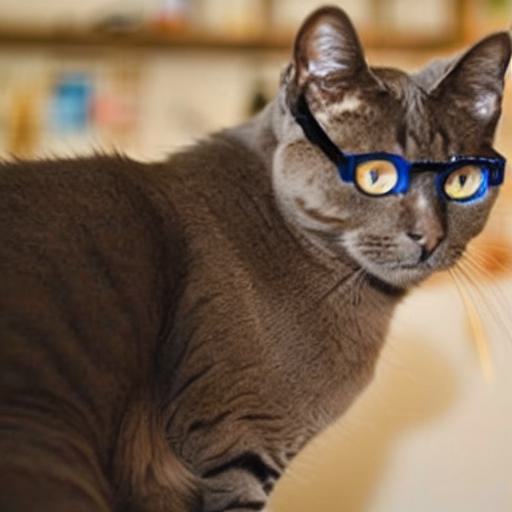} \vspace{1mm} \\ 
\includegraphics[align=c,width=0.12\linewidth]{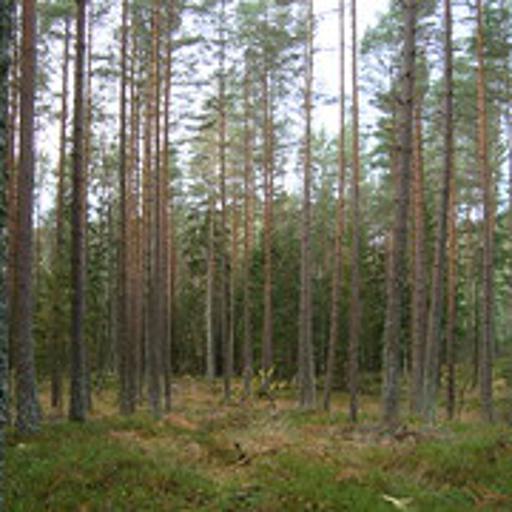} &     \begin{tabular}[c]{@{}c@{}}Add\\``Giraffe''\end{tabular} & \includegraphics[align=c,width=0.12\linewidth]{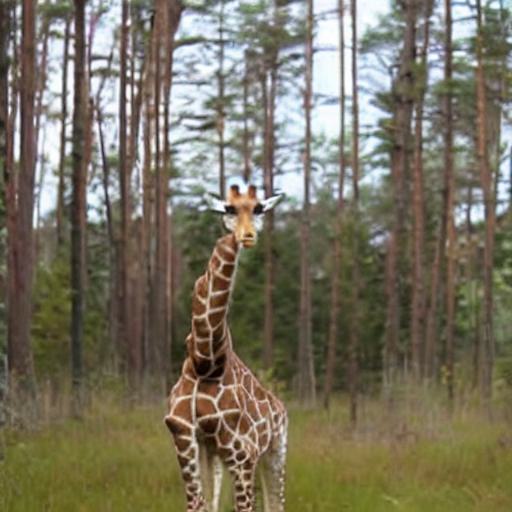} & \includegraphics[align=c,width=0.12\linewidth]{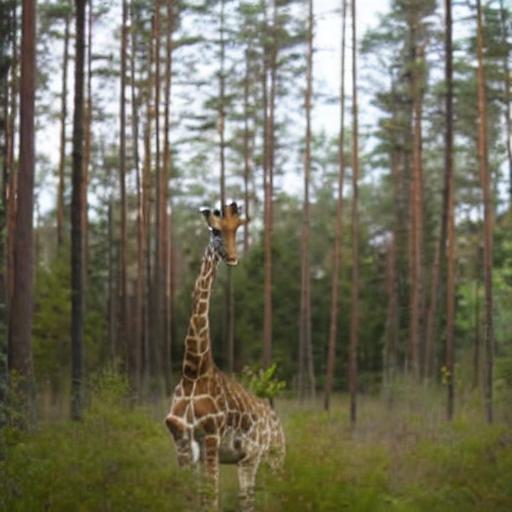} \hspace{3mm} & \includegraphics[align=c,width=0.12\linewidth]{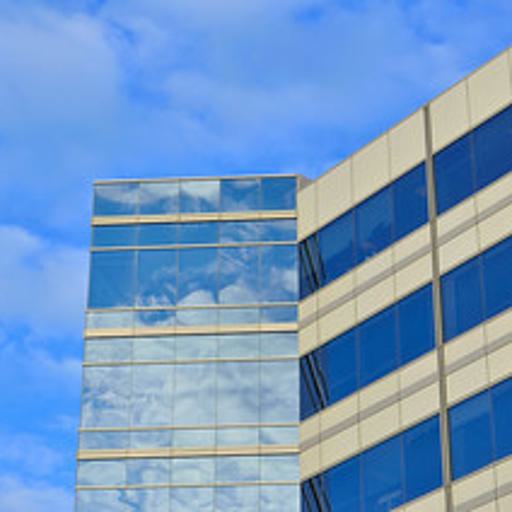} &     \begin{tabular}[c]{@{}c@{}}Add\\``Birds''\end{tabular} & \includegraphics[align=c,width=0.12\linewidth]{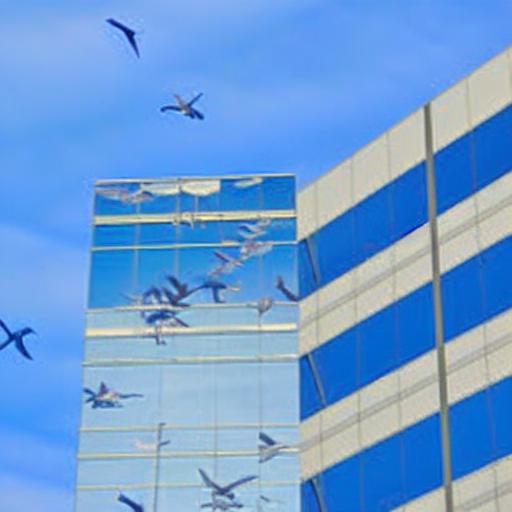} & \includegraphics[align=c,width=0.12\linewidth]{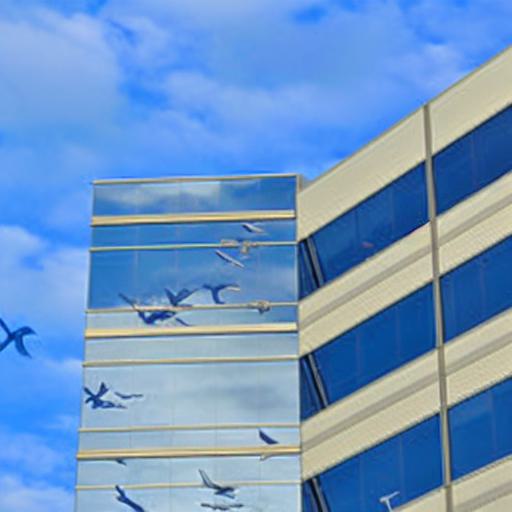} \vspace{1mm} \\
\includegraphics[align=c,width=0.12\linewidth]{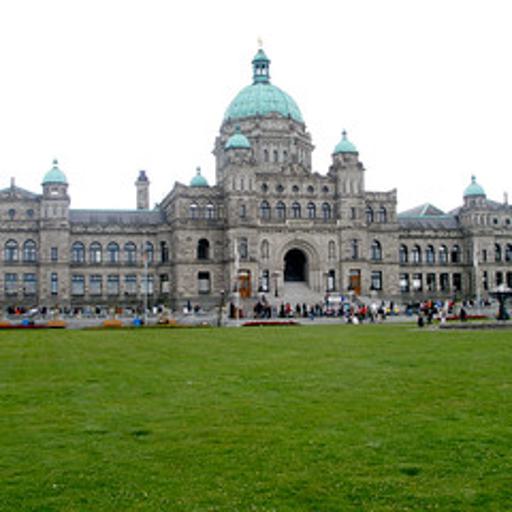} &     \begin{tabular}[c]{@{}c@{}}Add\\``Elephant''\end{tabular} & \includegraphics[align=c,width=0.12\linewidth]{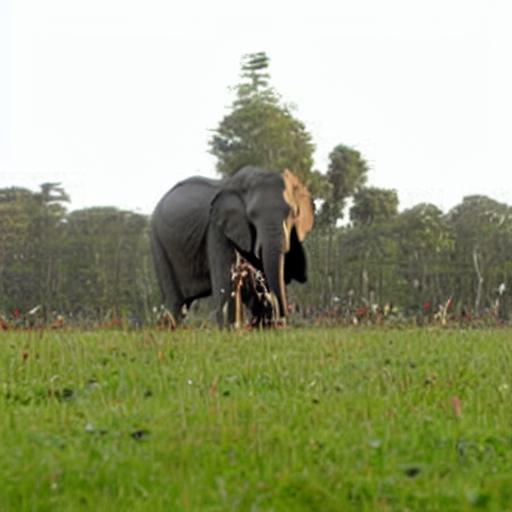}  & \includegraphics[align=c,width=0.12\linewidth]{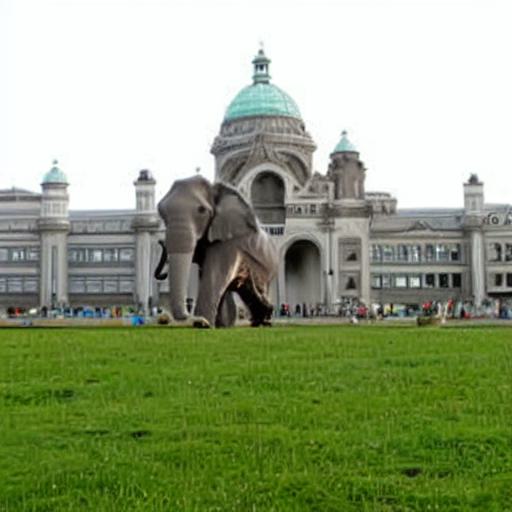} \hspace{3mm} & \includegraphics[align=c,width=0.12\linewidth]{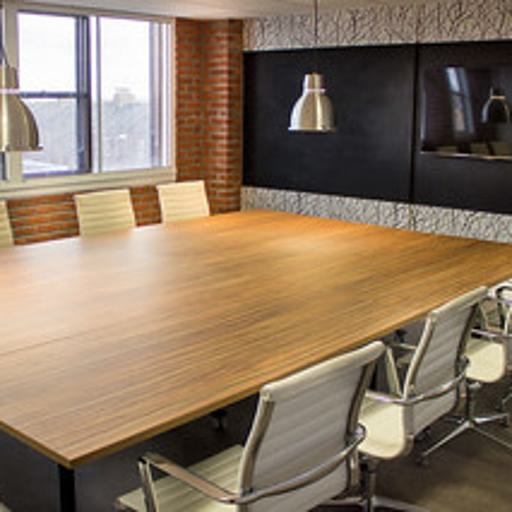}  & \begin{tabular}[c]{@{}c@{}}Add\\``Apples''\end{tabular}  & \includegraphics[align=c,width=0.12\linewidth]{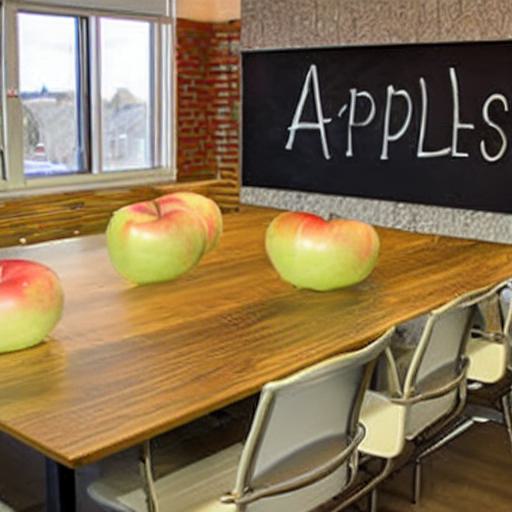}  & \includegraphics[align=c,width=0.12\linewidth]{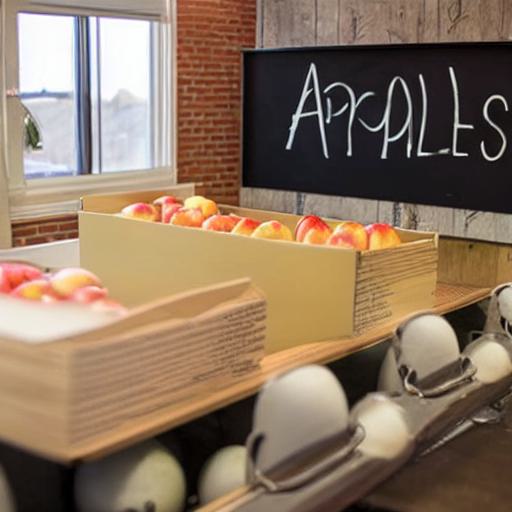} \vspace{1mm} \\ 
\includegraphics[align=c,width=0.12\linewidth]{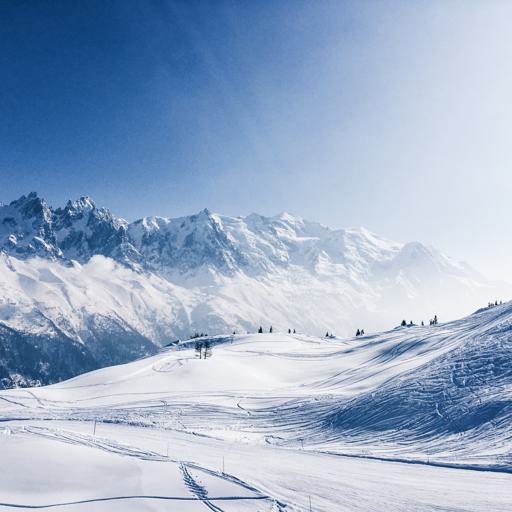} & \begin{tabular}[c]{@{}c@{}}Add\\``Balloons''\end{tabular}  & \includegraphics[align=c,width=0.12\linewidth]{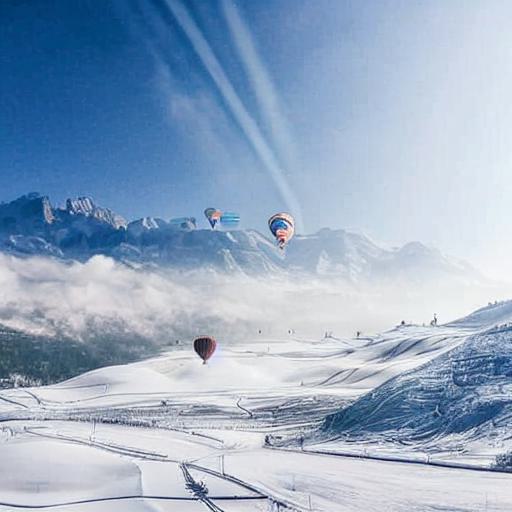} & \includegraphics[align=c,width=0.12\linewidth]{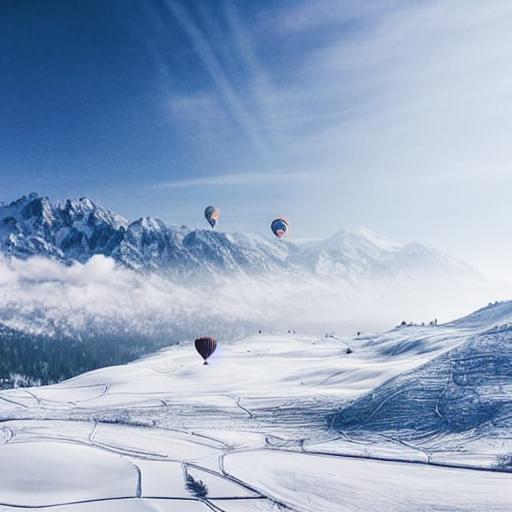} \hspace{3mm} & \includegraphics[align=c,width=0.12\linewidth]{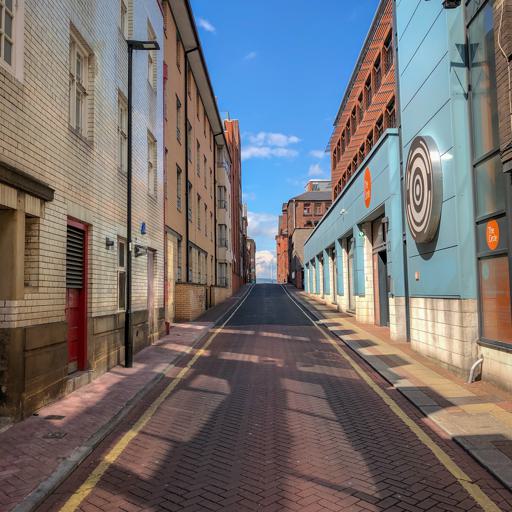} & \begin{tabular}[c]{@{}c@{}}Add\\``Duck''\end{tabular}  & \includegraphics[align=c,width=0.12\linewidth]{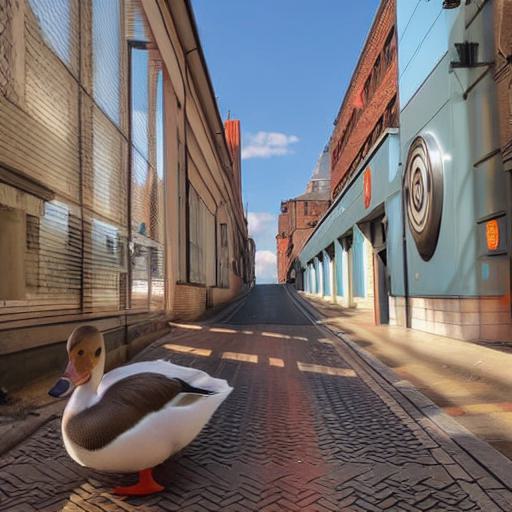} & \includegraphics[align=c,width=0.12\linewidth]{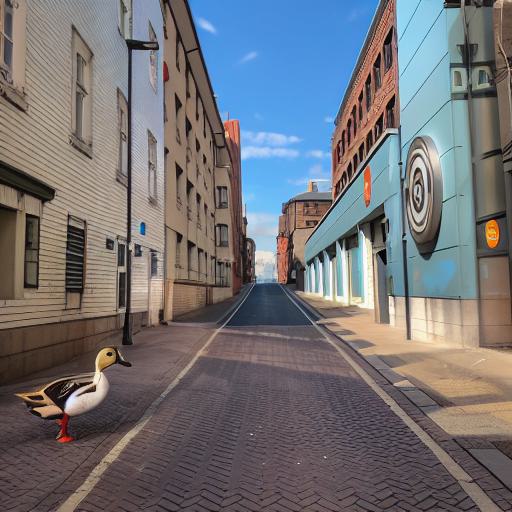} \vspace{1mm} \\
\end{tabular}
\caption{Results of adding object comparing Prompt-to-Prompt and \textbf{MDP-$\boldsymbol\epsilon_t$}.}
\label{fig:local-adding-object}
\end{figure*}

\begin{figure*}[h]
\centering
\scriptsize
\setlength{\tabcolsep}{1pt}
\begin{tabular}{cccccccc}
Input & Edit & P2P & \textbf{MDP-$\boldsymbol\epsilon_t$} & Input & Edit & P2P & \textbf{MDP-$\boldsymbol\epsilon_t$} \\
\includegraphics[align=c,width=0.12\linewidth]{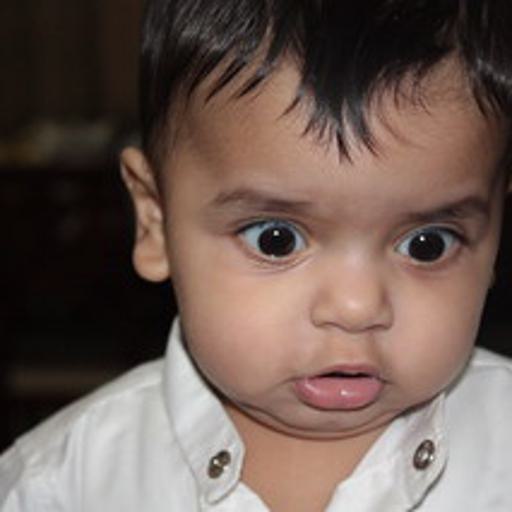} & 
  \begin{tabular}[c]{@{}c@{}}``Sad''\\to\\``Happy''\end{tabular} & \includegraphics[align=c,width=0.12\linewidth]{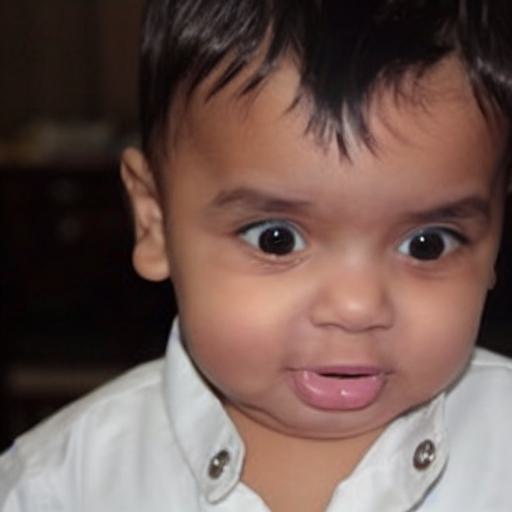}  & \includegraphics[align=c,width=0.12\linewidth]{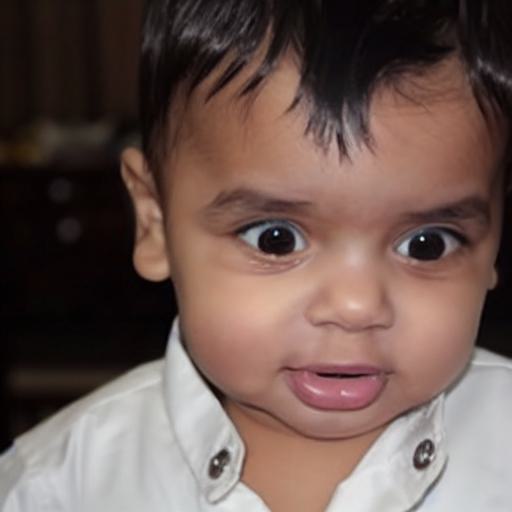} \hspace{3mm} & \includegraphics[align=c,width=0.12\linewidth]{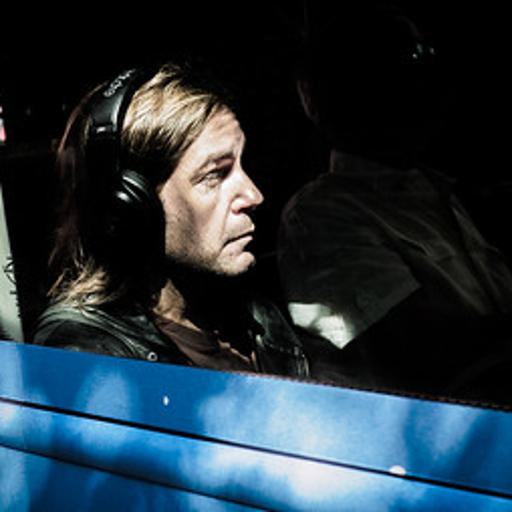}  &   \begin{tabular}[c]{@{}c@{}}``Sad''\\to\\``Happy''\end{tabular}   & \includegraphics[align=c,width=0.12\linewidth]{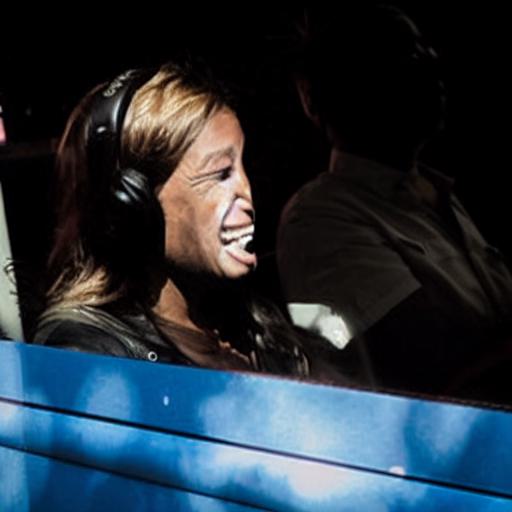}  & \includegraphics[align=c,width=0.12\linewidth]{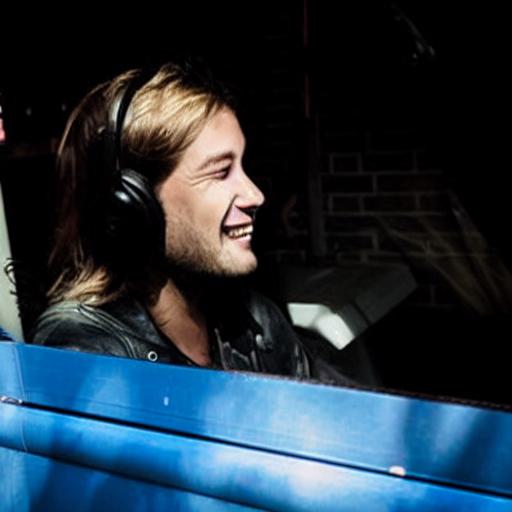} \vspace{1mm} \\ 
\includegraphics[align=c,width=0.12\linewidth]{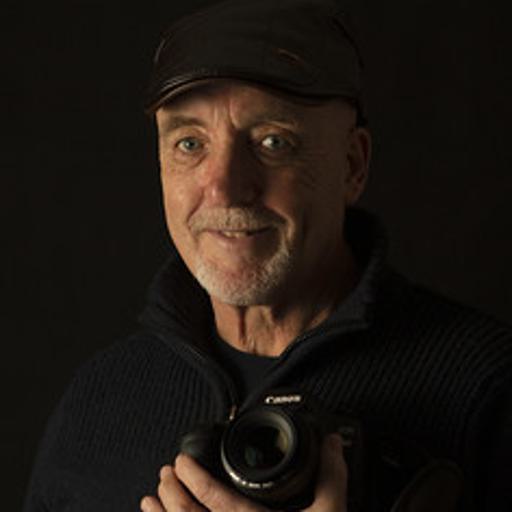} & 
    \begin{tabular}[c]{@{}c@{}}``Old''\\to\\``Young''\end{tabular} & \includegraphics[align=c,width=0.12\linewidth]{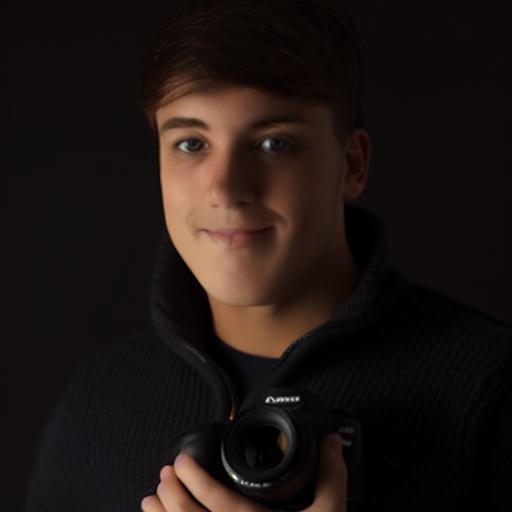} & \includegraphics[align=c,width=0.12\linewidth]{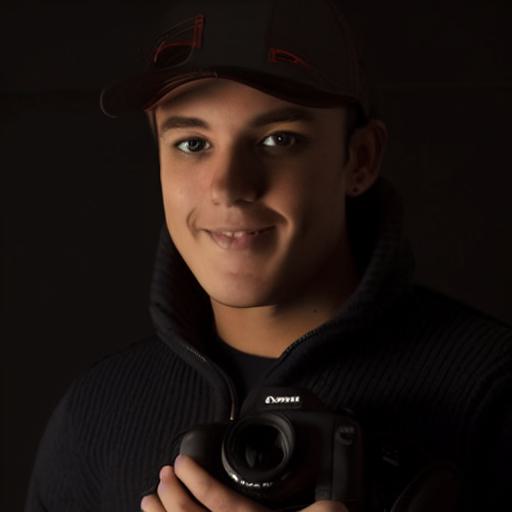} \hspace{3mm} & \includegraphics[align=c,width=0.12\linewidth]{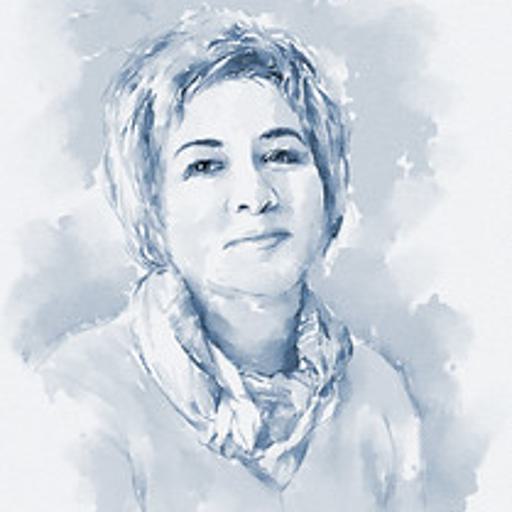} &     \begin{tabular}[c]{@{}c@{}}``Woman''\\to\\``Kid''\end{tabular}  & \includegraphics[align=c,width=0.12\linewidth]{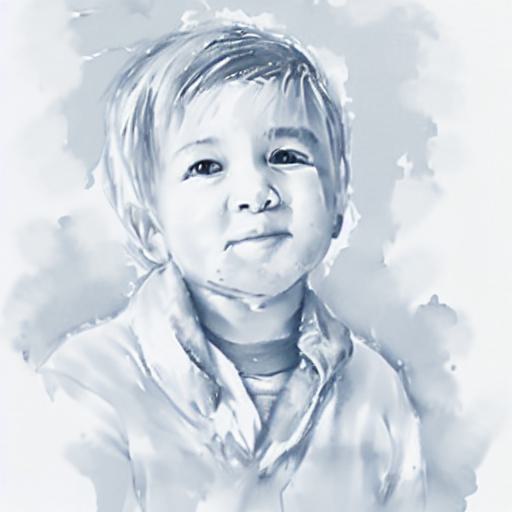} & \includegraphics[align=c,width=0.12\linewidth]{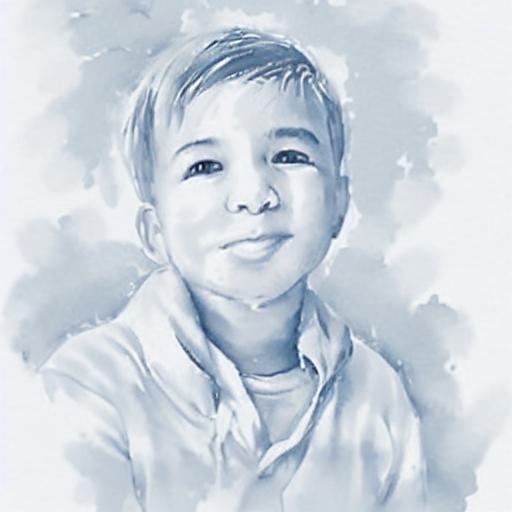} \vspace{1mm} \\
\includegraphics[align=c,width=0.12\linewidth]{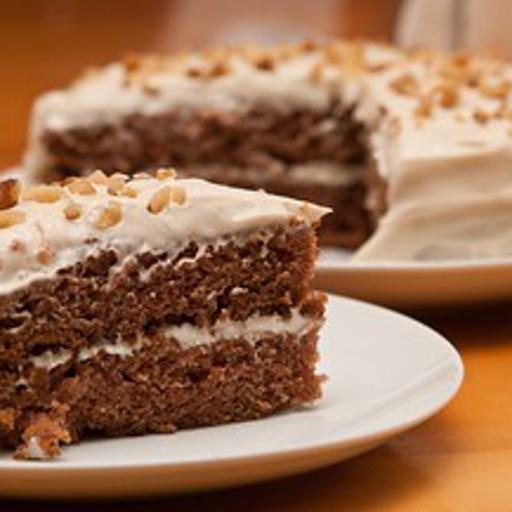} &     \begin{tabular}[c]{@{}c@{}}``Chocolate''\\to\\``Strawberry''\end{tabular}  & \includegraphics[align=c,width=0.12\linewidth]{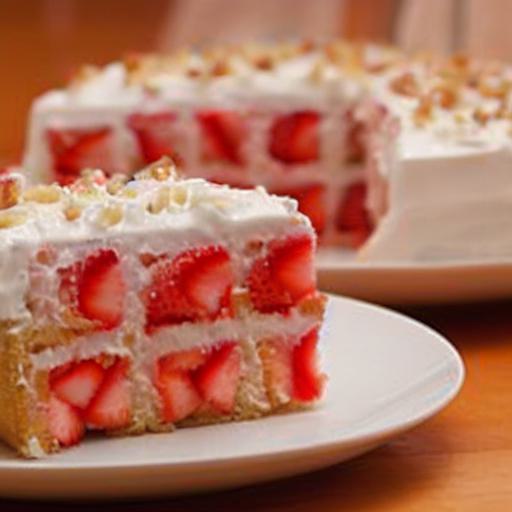}  & \includegraphics[align=c,width=0.12\linewidth]{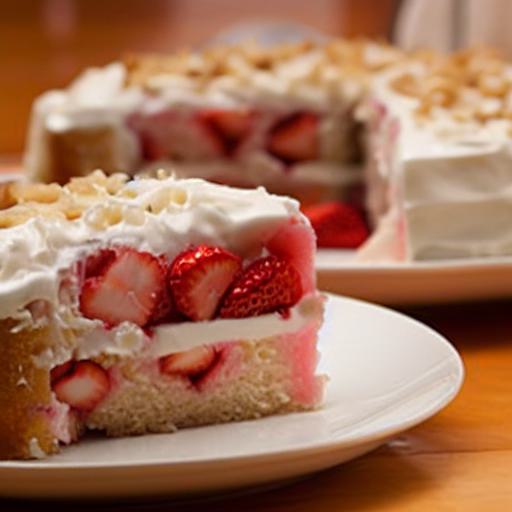} \hspace{3mm} & \includegraphics[align=c,width=0.12\linewidth]{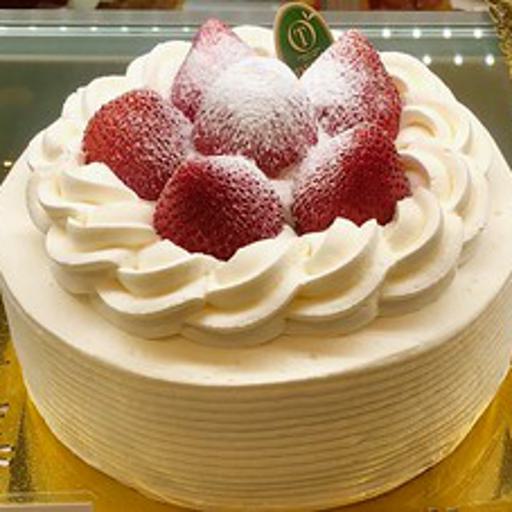}  & 
      \begin{tabular}[c]{@{}c@{}}``Strawberry''\\to\\``Candles''\end{tabular} & \includegraphics[align=c,width=0.12\linewidth]{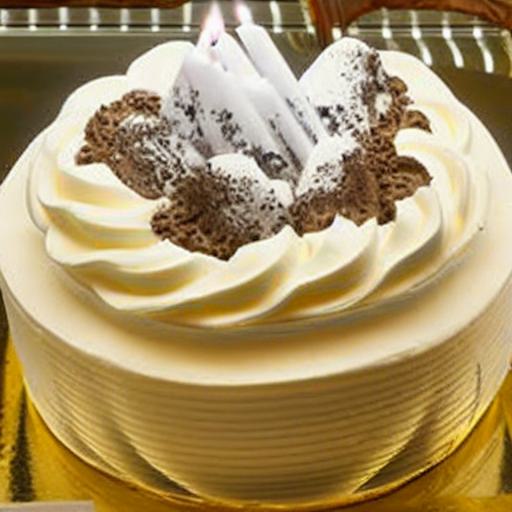}  & \includegraphics[align=c,width=0.12\linewidth]{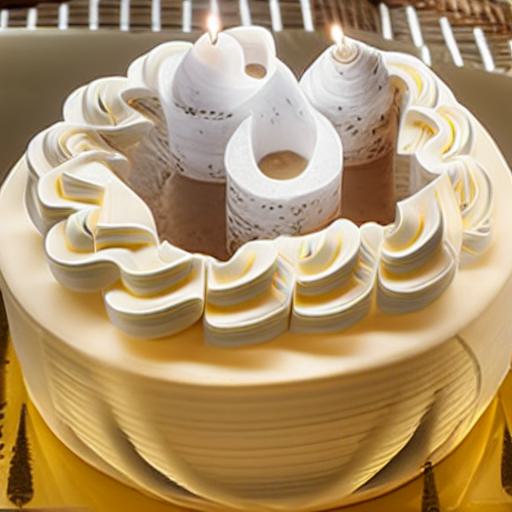} \vspace{1mm} \\ 
\includegraphics[align=c,width=0.12\linewidth]{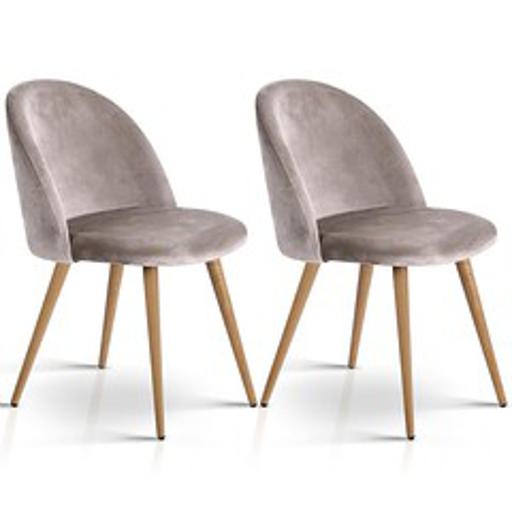} & 
        \begin{tabular}[c]{@{}c@{}}``Cloth''\\to\\``Plastic''\end{tabular} & \includegraphics[align=c,width=0.12\linewidth]{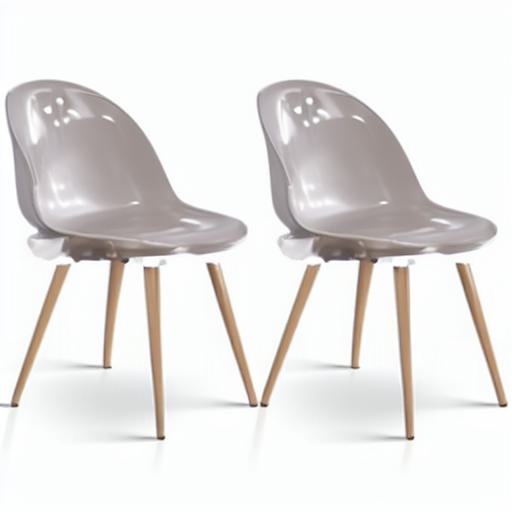} & \includegraphics[align=c,width=0.12\linewidth]{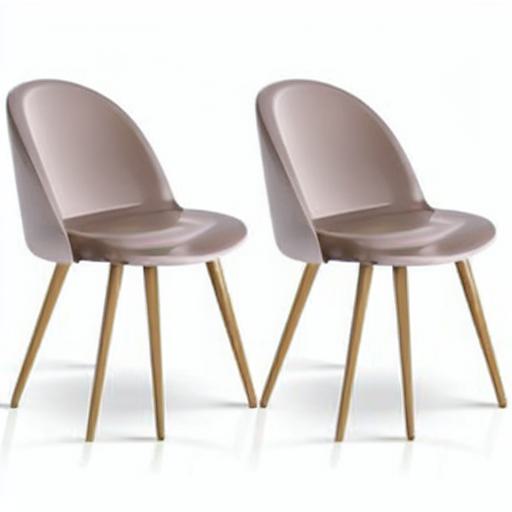} \hspace{3mm} & \includegraphics[align=c,width=0.12\linewidth]{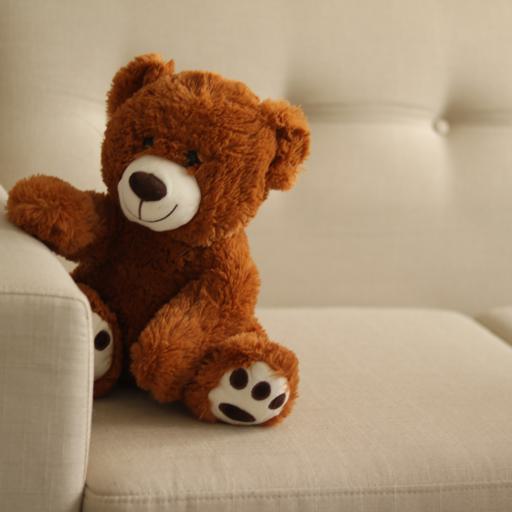} & 
  \begin{tabular}[c]{@{}c@{}}``Fluffy''\\to\\``Smooth''\end{tabular}& \includegraphics[align=c,width=0.12\linewidth]{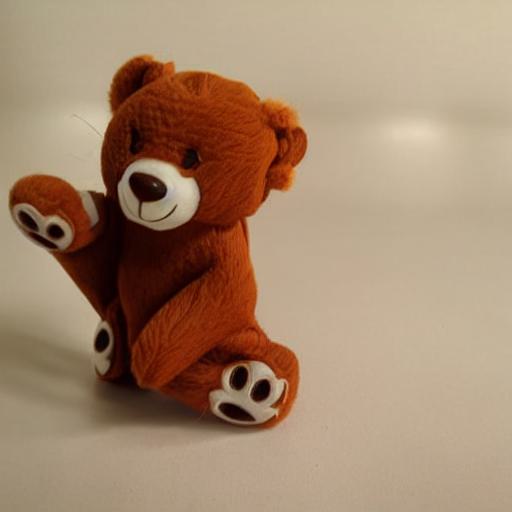} & \includegraphics[align=c,width=0.12\linewidth]{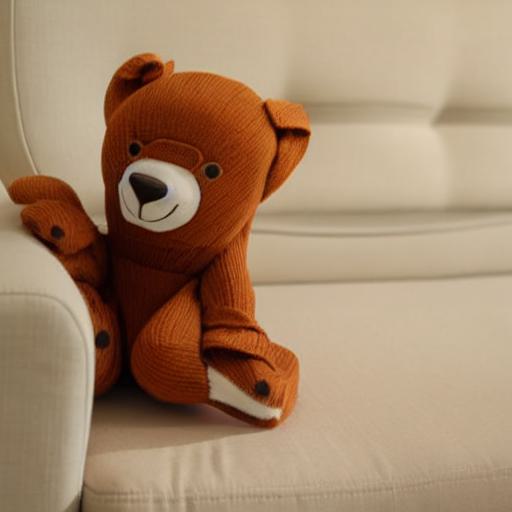} \vspace{1mm}\\
\end{tabular}
\caption{Results of changing attributes comparing Prompt-to-Prompt and \textbf{MDP-$\boldsymbol\epsilon_t$}.}
\label{fig:local-changing-attributes}
\end{figure*}

\begin{figure*}[h]
\centering
\scriptsize
\setlength{\tabcolsep}{1pt}
\begin{tabular}{cccccccc}
Input & Edit & P2P & \textbf{MDP-$\boldsymbol\epsilon_t$} & Input & Edit & P2P & \textbf{MDP-$\boldsymbol\epsilon_t$} \\
\includegraphics[align=c,width=0.12\linewidth]{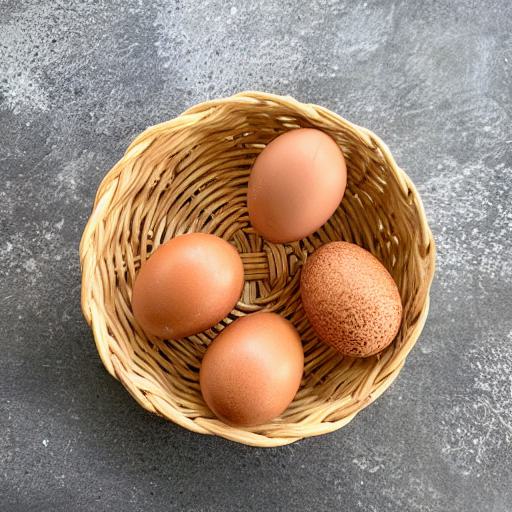} & 
    \begin{tabular}[c]{@{}c@{}}Remove\\``Eggs''\end{tabular}& \includegraphics[align=c,width=0.12\linewidth]{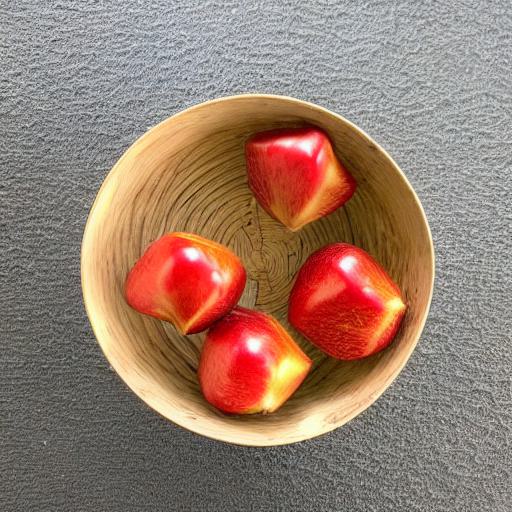}  & \includegraphics[align=c,width=0.12\linewidth]{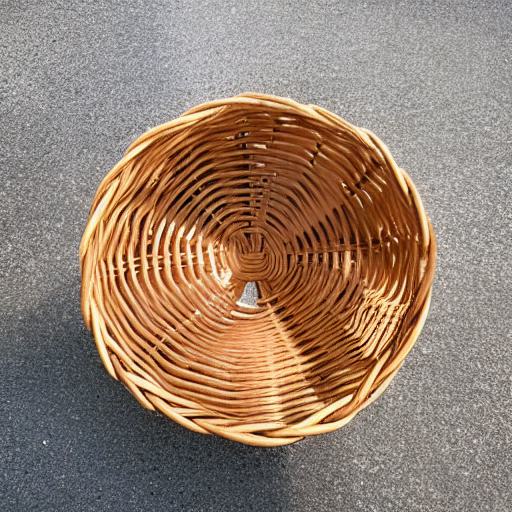} \hspace{3mm} & \includegraphics[align=c,width=0.12\linewidth]{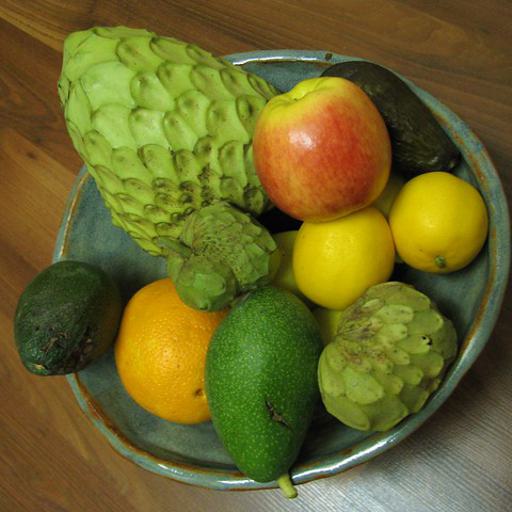}  &     \begin{tabular}[c]{@{}c@{}}Remove\\``Fruits''\end{tabular}  & \includegraphics[align=c,width=0.12\linewidth]{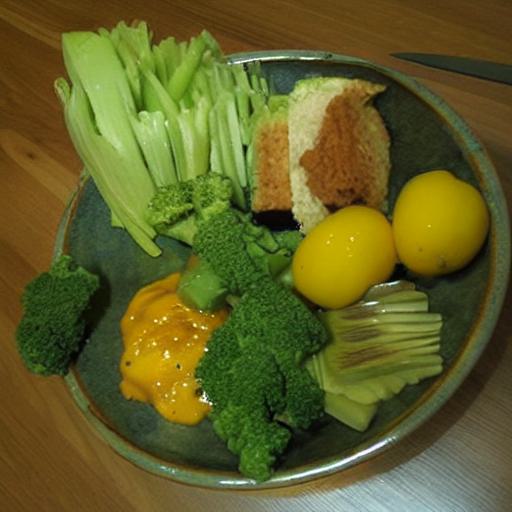}  & \includegraphics[align=c,width=0.12\linewidth]{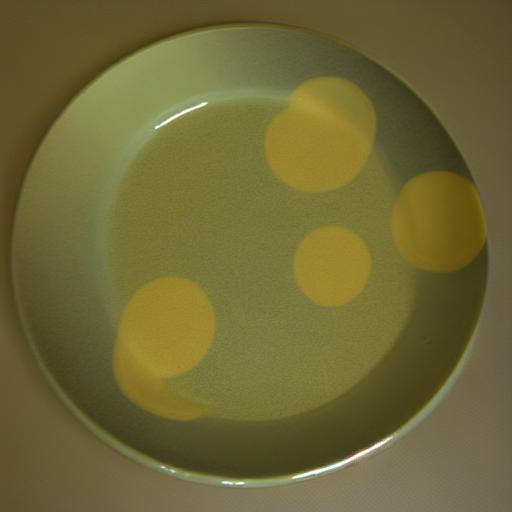} \vspace{1mm} \\ 
\includegraphics[align=c,width=0.12\linewidth]{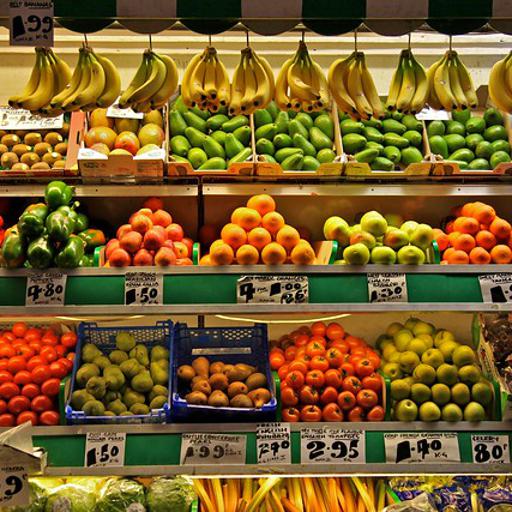} &     \begin{tabular}[c]{@{}c@{}}Remove\\``Fruits''\end{tabular} & \includegraphics[align=c,width=0.12\linewidth]{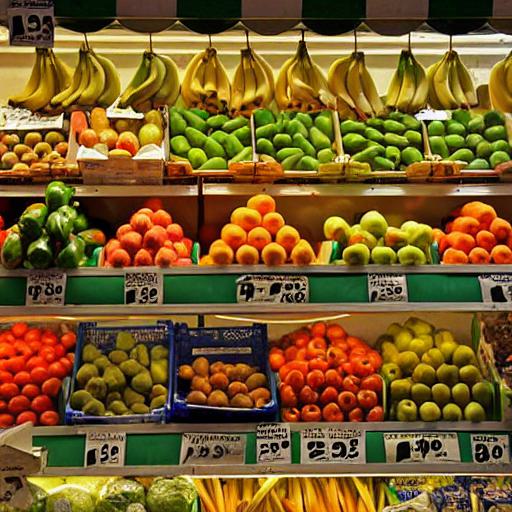} & \includegraphics[align=c,width=0.12\linewidth]{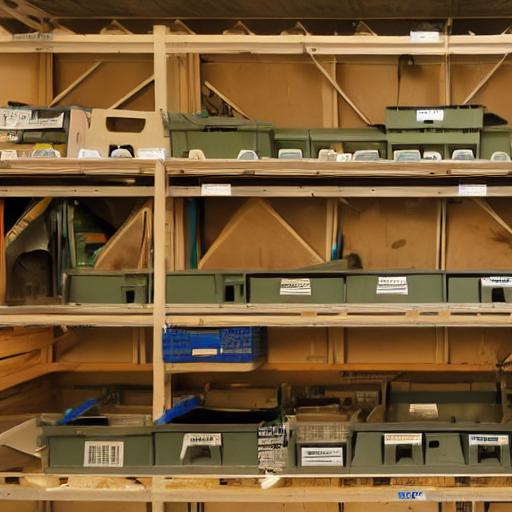} \hspace{3mm} & \includegraphics[align=c,width=0.12\linewidth]{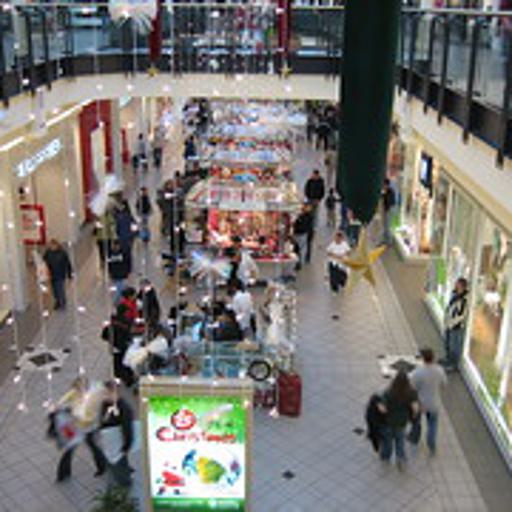} &     \begin{tabular}[c]{@{}c@{}}Remove\\``People''\end{tabular} & \includegraphics[align=c,width=0.12\linewidth]{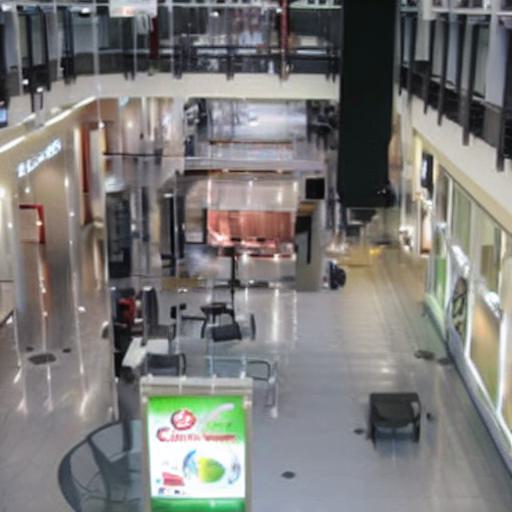} & \includegraphics[align=c,width=0.12\linewidth]{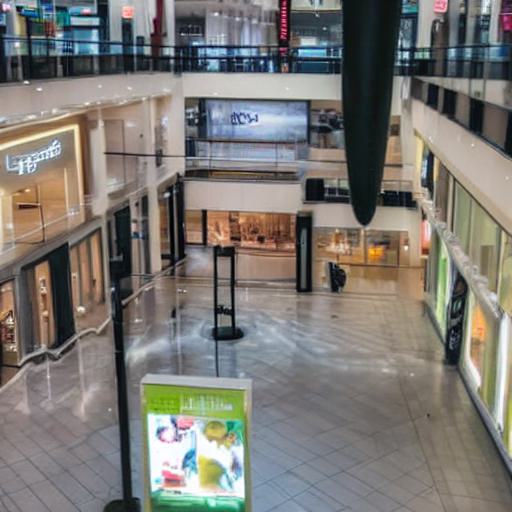} \vspace{1mm} \\
\includegraphics[align=c,width=0.12\linewidth]{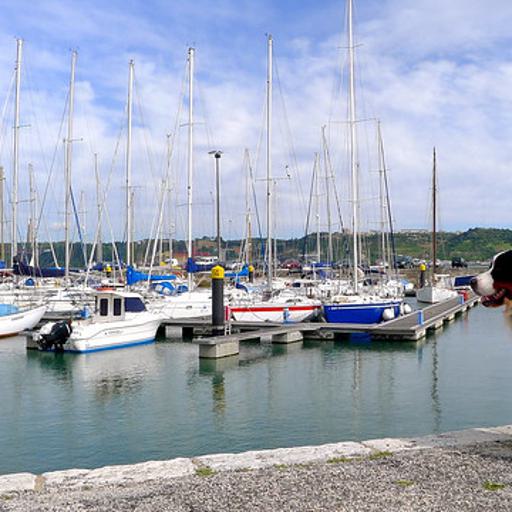} & \begin{tabular}[c]{@{}c@{}}Remove\\``Boats''\end{tabular} & \includegraphics[align=c,width=0.12\linewidth]{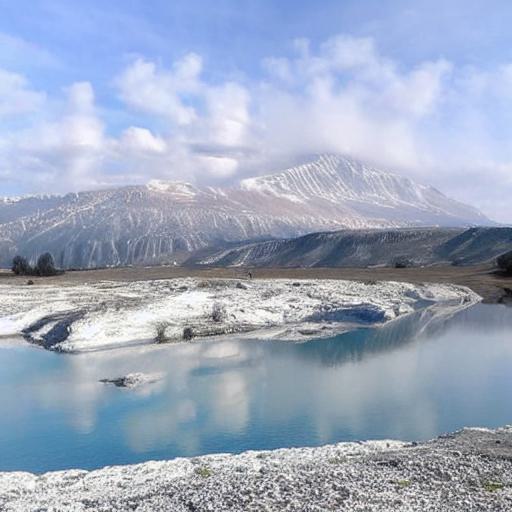}  & \includegraphics[align=c,width=0.12\linewidth]{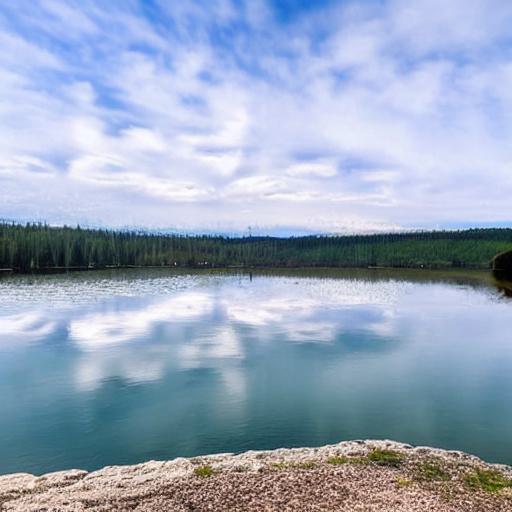} \hspace{3mm} & \includegraphics[align=c,width=0.12\linewidth]{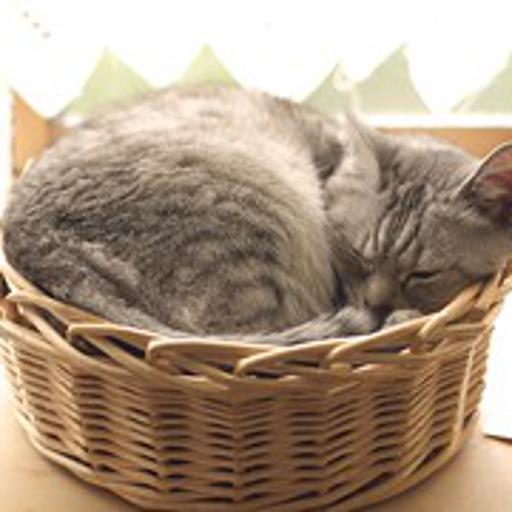}  & \begin{tabular}[c]{@{}c@{}}Remove\\``Cat''\end{tabular}  & \includegraphics[align=c,width=0.12\linewidth]{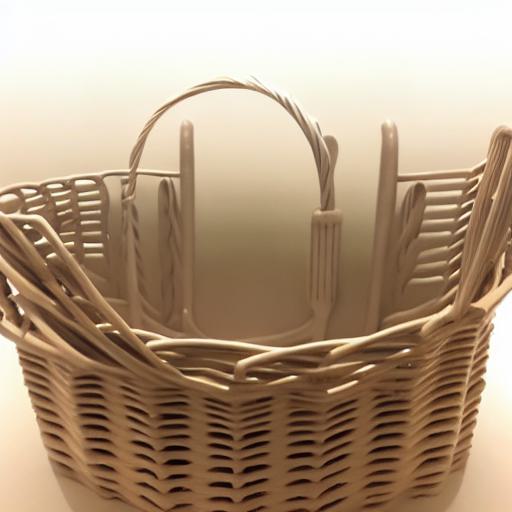}  & \includegraphics[align=c,width=0.12\linewidth]{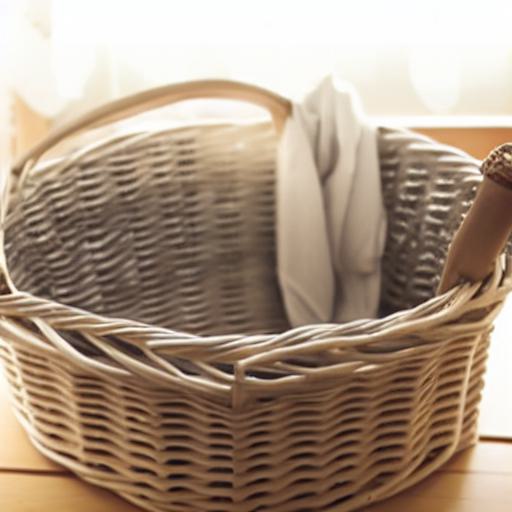} \vspace{1mm} \\ 
\includegraphics[align=c,width=0.12\linewidth]{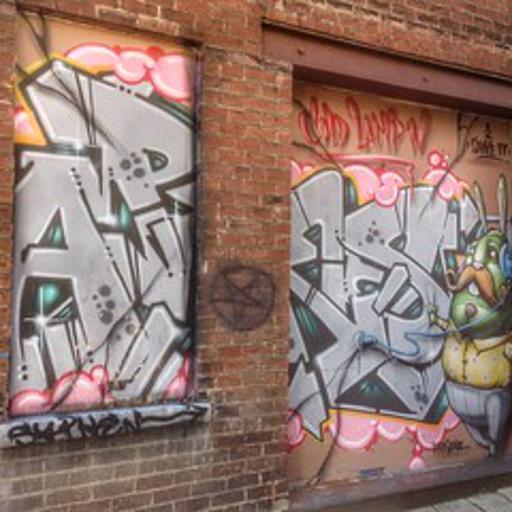} & \begin{tabular}[c]{@{}c@{}}Remove\\``Graffiti''\end{tabular} & \includegraphics[align=c,width=0.12\linewidth]{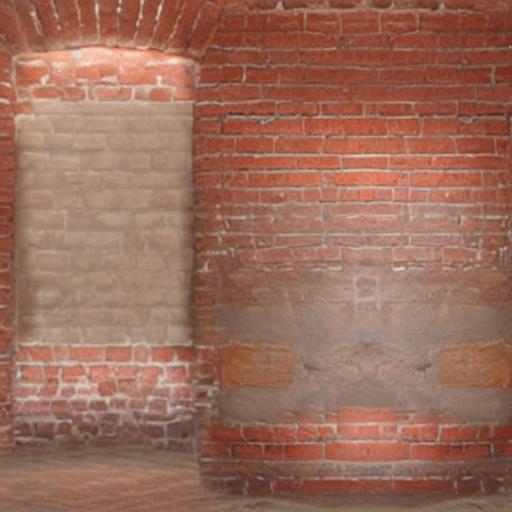} & \includegraphics[align=c,width=0.12\linewidth]{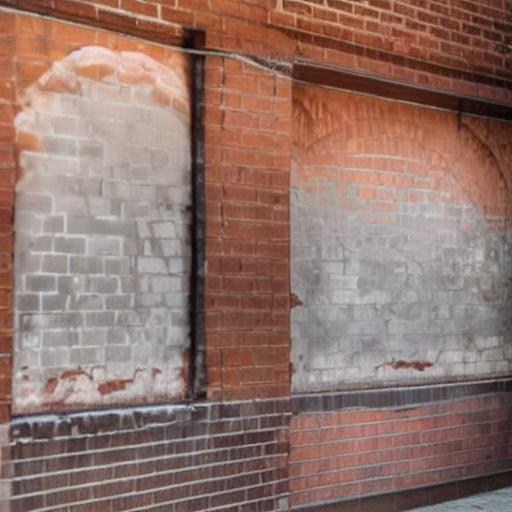} \hspace{3mm} & \includegraphics[align=c,width=0.12\linewidth]{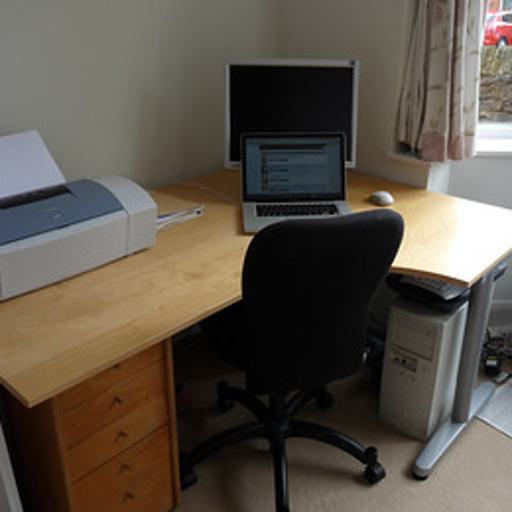} & \begin{tabular}[c]{@{}c@{}}Remove\\``Stuffs''\end{tabular} & \includegraphics[align=c,width=0.12\linewidth]{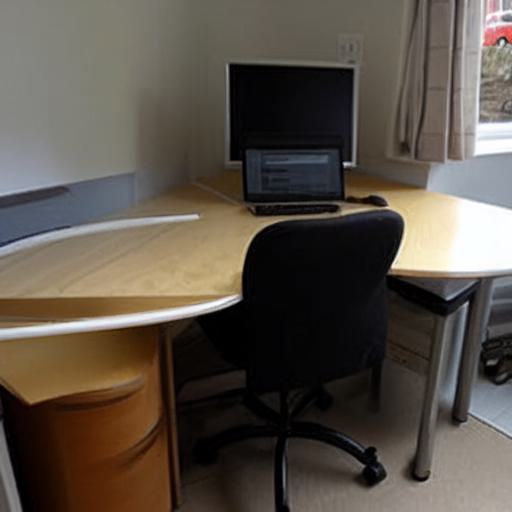} & \includegraphics[align=c,width=0.12\linewidth]{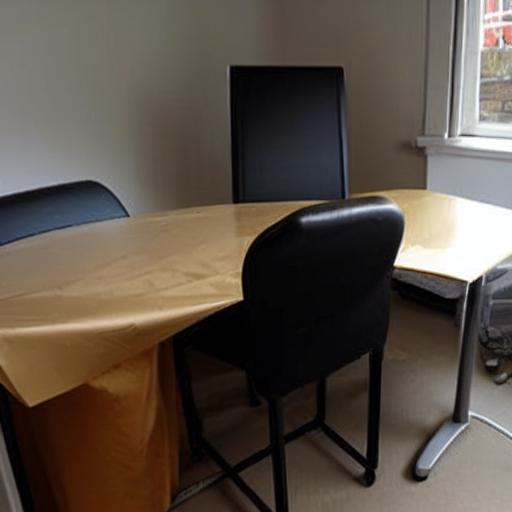} \vspace{1mm}\\
\end{tabular}
\caption{Results of removing object(s) comparing Prompt-to-Prompt and \textbf{MDP-$\boldsymbol\epsilon_t$}.}
\label{fig:local-removing-object}
\end{figure*}

\begin{figure*}[h]
\centering
\scriptsize
\setlength{\tabcolsep}{1pt}
\begin{tabular}{cccccccc}
Input & Edit & P2P & \textbf{MDP-$\boldsymbol\epsilon_t$} & Input & Edit & P2P & \textbf{MDP-$\boldsymbol\epsilon_t$} \\
\includegraphics[align=c,width=0.12\linewidth]{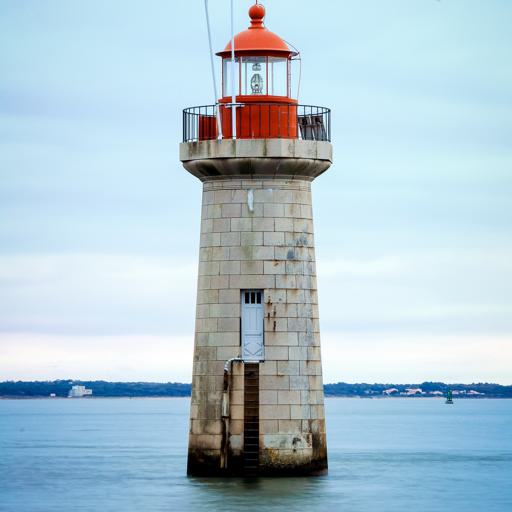} & 
      \begin{tabular}[c]{@{}c@{}}``Tower''\\Mixes\\``Dress''\end{tabular}& \includegraphics[align=c,width=0.12\linewidth]{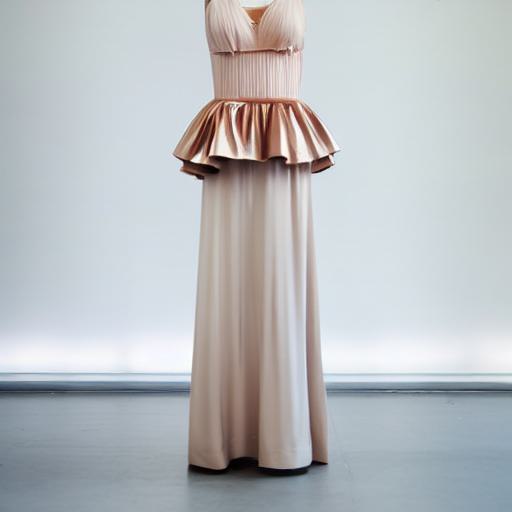}  & \includegraphics[align=c,width=0.12\linewidth]{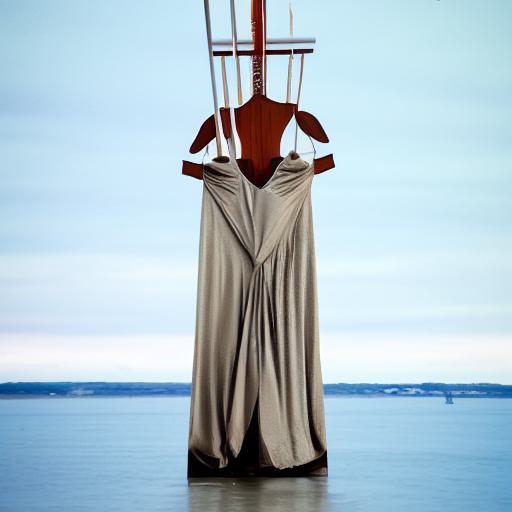} \hspace{3mm} & \includegraphics[align=c,width=0.12\linewidth]{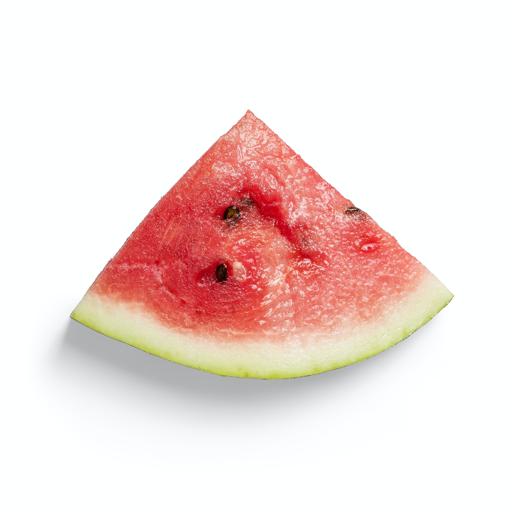}  & 
      \begin{tabular}[c]{@{}c@{}}``Watermelon''\\Mixes\\``Pillow''\end{tabular}  & \includegraphics[align=c,width=0.12\linewidth]{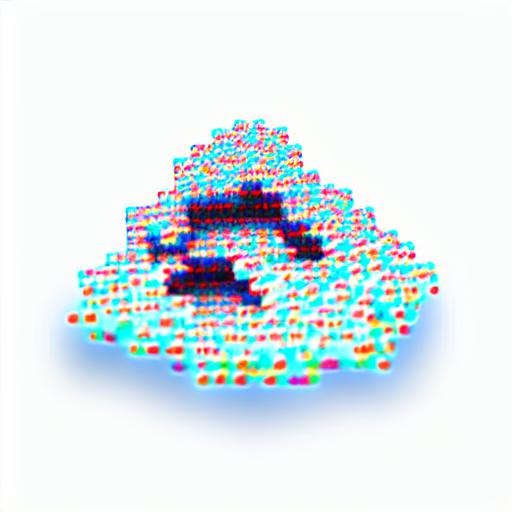}  & \includegraphics[align=c,width=0.12\linewidth]{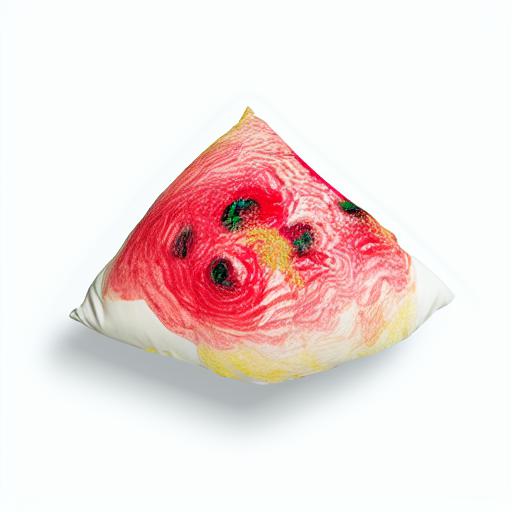} \vspace{1mm} \\ 
\includegraphics[align=c,width=0.12\linewidth]{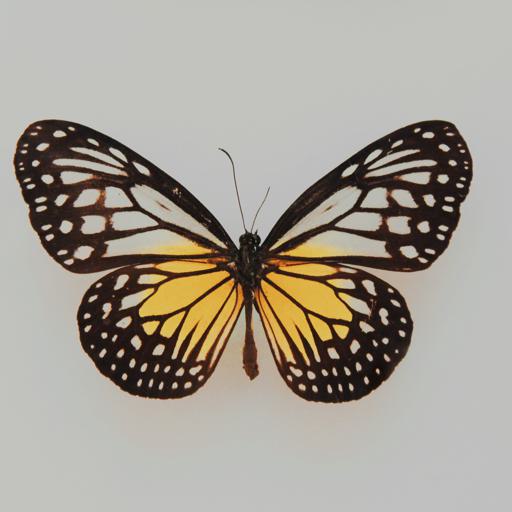} &       \begin{tabular}[c]{@{}c@{}}``Butterfly''\\Mixes\\``Handbag''\end{tabular} & \includegraphics[align=c,width=0.12\linewidth]{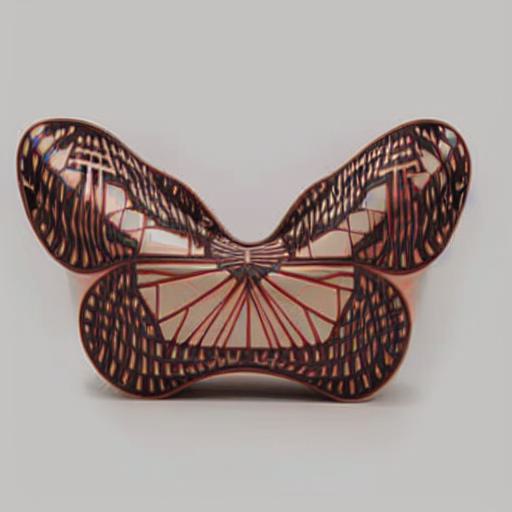} & \includegraphics[align=c,width=0.12\linewidth]{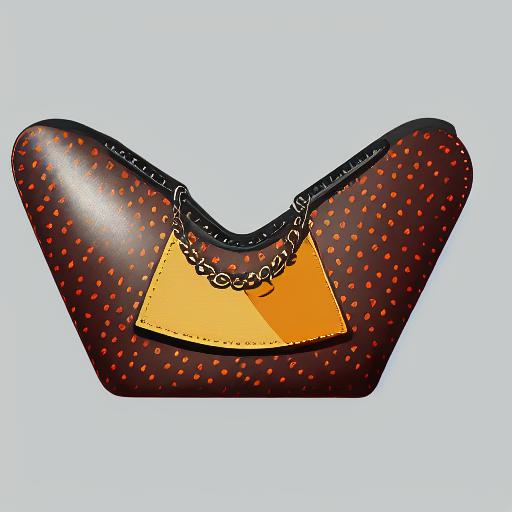} \hspace{3mm} & \includegraphics[align=c,width=0.12\linewidth]{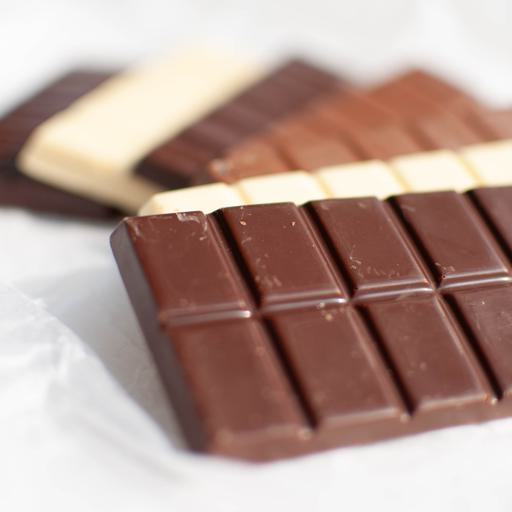} &       \begin{tabular}[c]{@{}c@{}}``Chocolate''\\Mixes\\``Purse''\end{tabular} & \includegraphics[align=c,width=0.12\linewidth]{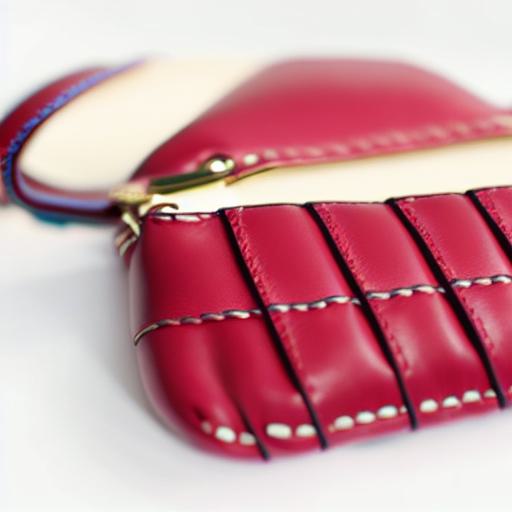} & \includegraphics[align=c,width=0.12\linewidth]{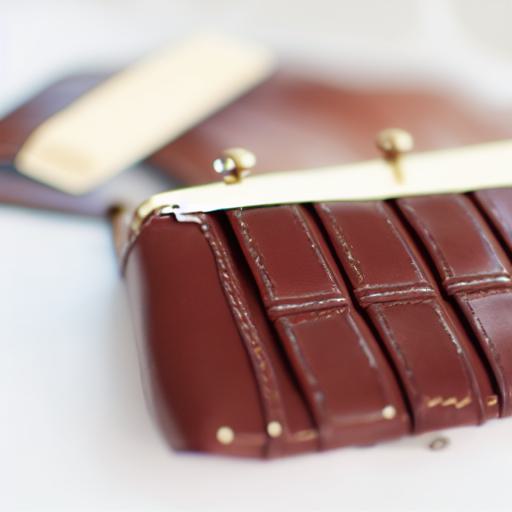} \vspace{1mm} \\
\includegraphics[align=c,width=0.12\linewidth]{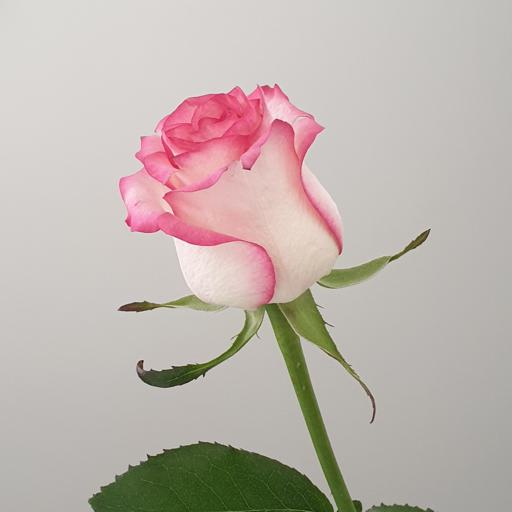} &       \begin{tabular}[c]{@{}c@{}}``Flower''\\Mixes\\``Handbag''\end{tabular} & \includegraphics[align=c,width=0.12\linewidth]{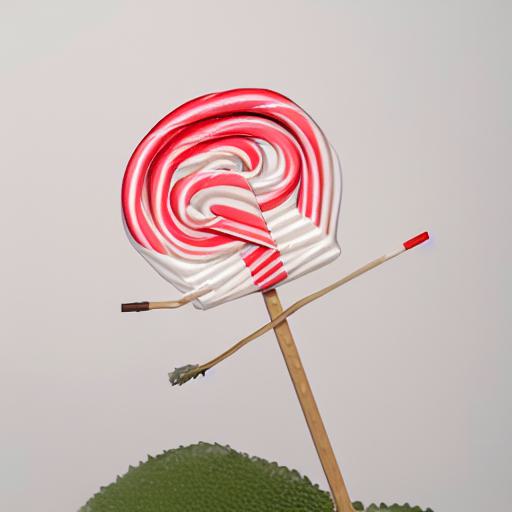}  & \includegraphics[align=c,width=0.12\linewidth]{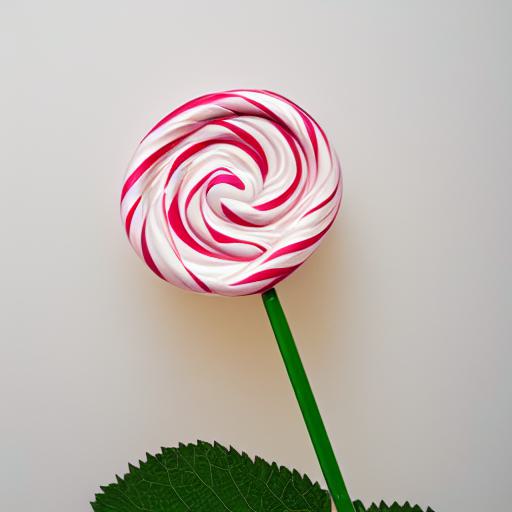} \hspace{3mm} & \includegraphics[align=c,width=0.12\linewidth]{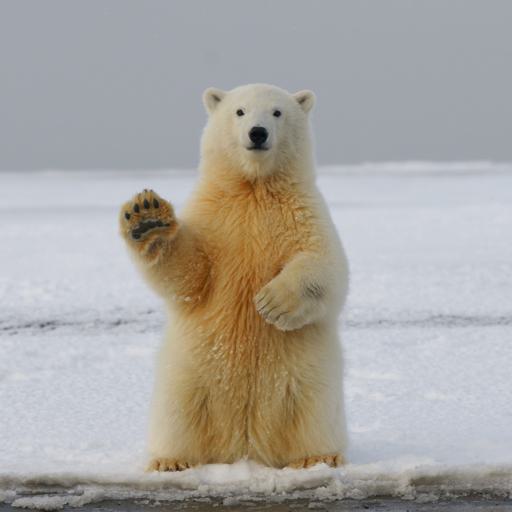}  &       \begin{tabular}[c]{@{}c@{}}``Polar Bear''\\Mixes\\``Kettle''\end{tabular} & \includegraphics[align=c,width=0.12\linewidth]{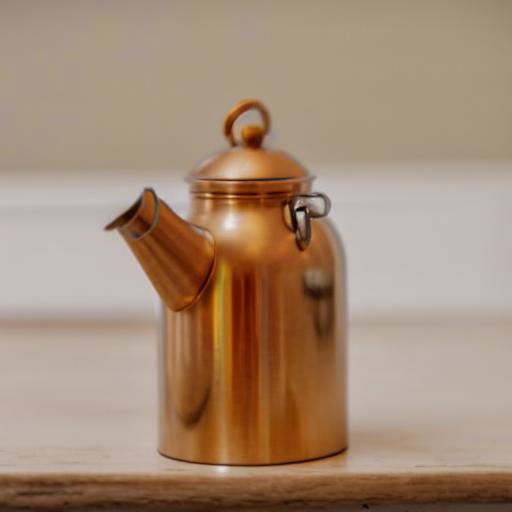}  & \includegraphics[align=c,width=0.12\linewidth]{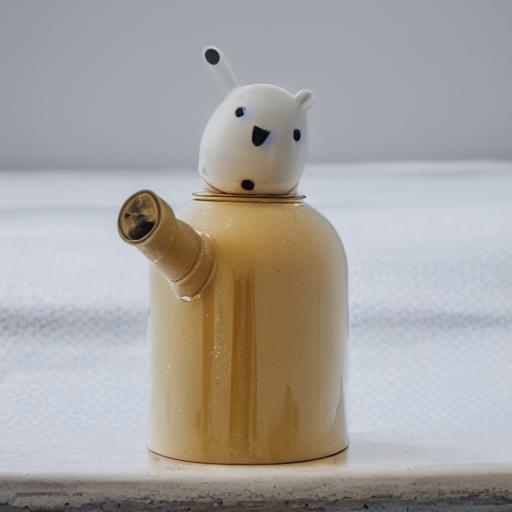} \vspace{1mm} \\ 
\includegraphics[align=c,width=0.12\linewidth]{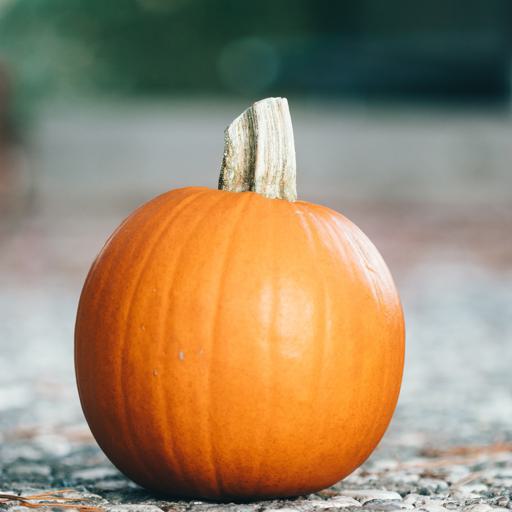} &       \begin{tabular}[c]{@{}c@{}}``Pumpkin''\\Mixes\\``Clock''\end{tabular} & \includegraphics[align=c,width=0.12\linewidth]{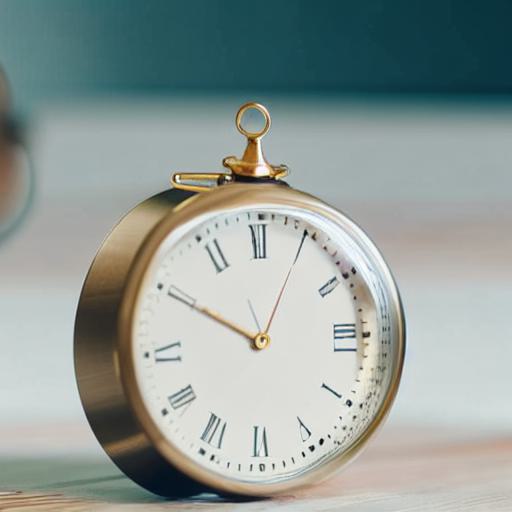} & \includegraphics[align=c,width=0.12\linewidth]{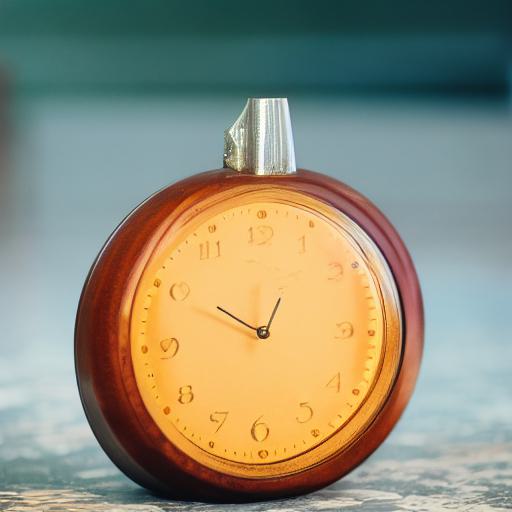} \hspace{3mm} & \includegraphics[align=c,width=0.12\linewidth]{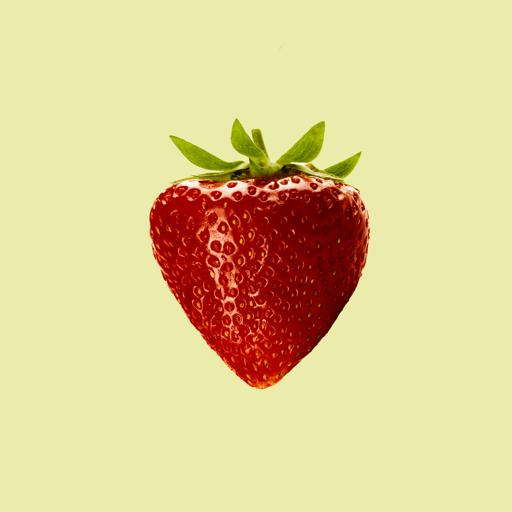} &       \begin{tabular}[c]{@{}c@{}}``Strawberry''\\Mixes\\``Diamond''\end{tabular} & \includegraphics[align=c,width=0.12\linewidth]{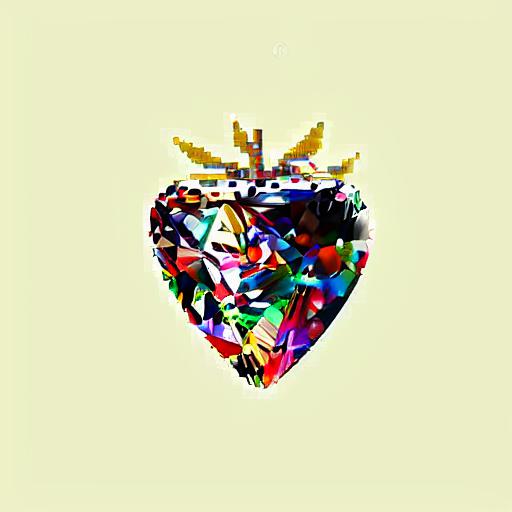} & \includegraphics[align=c,width=0.12\linewidth]{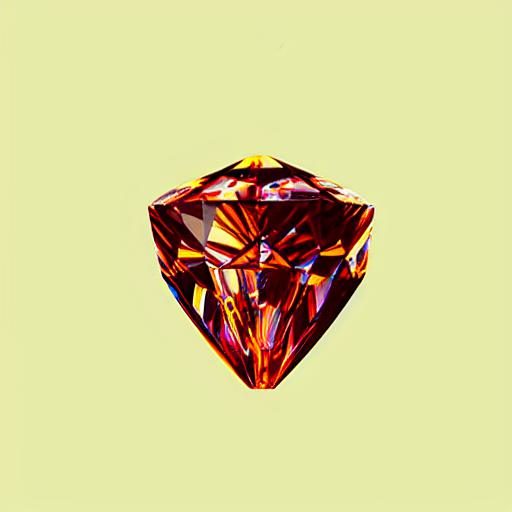} \vspace{1mm} \\
\end{tabular}
\caption{Results of mixing objects comparing Prompt-to-Prompt and \textbf{MDP-$\boldsymbol\epsilon_t$}.}
\label{fig:local-mixing-objects}
\end{figure*}

\begin{figure*}[h]
\centering
\scriptsize
\setlength{\tabcolsep}{1pt}
\begin{tabular}{ccccccccc}
\multicolumn{2}{c}{\multirow{2}{*}{\includegraphics[align=c,width=0.12\linewidth]{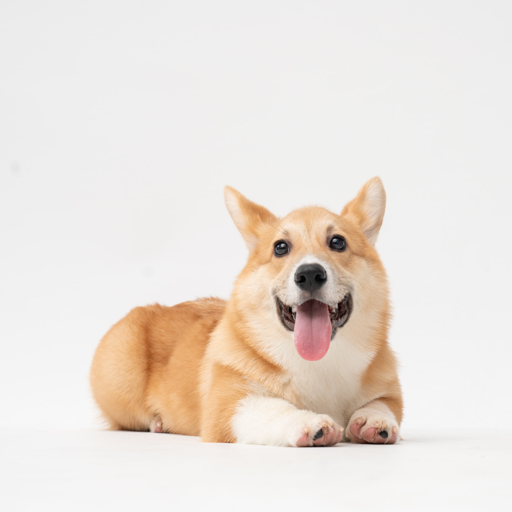}}} \hspace{2mm} & \textbf{MDP-$\boldsymbol\epsilon_t$} &{\includegraphics[align=c,width=0.12\linewidth]{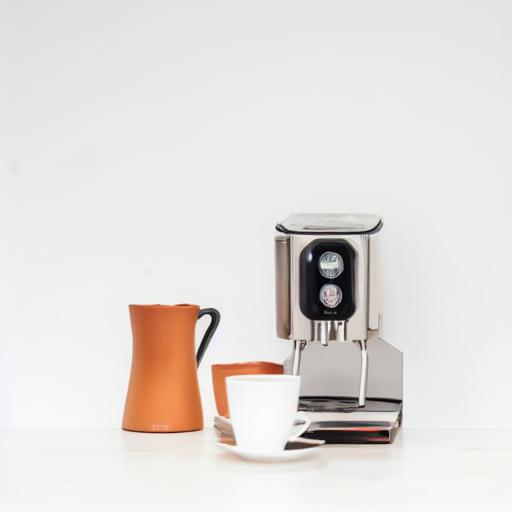}} & {\includegraphics[align=c,width=0.12\linewidth]{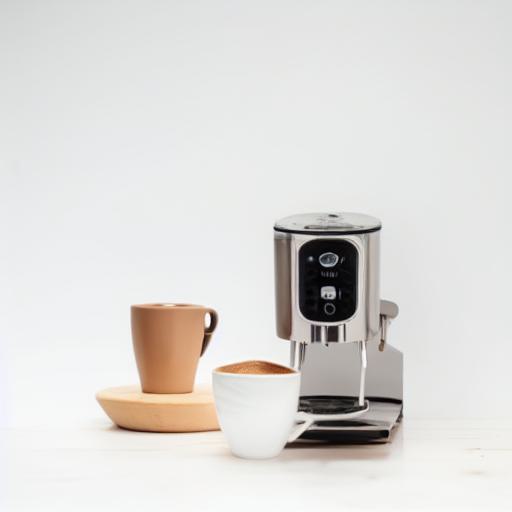}} & {\includegraphics[align=c,width=0.12\linewidth]{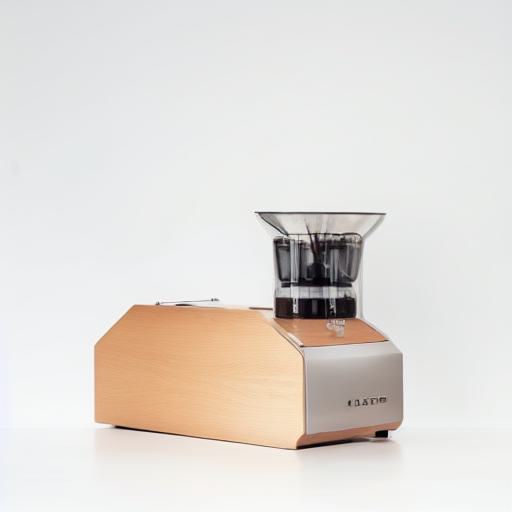}} & {\includegraphics[align=c,width=0.12\linewidth]{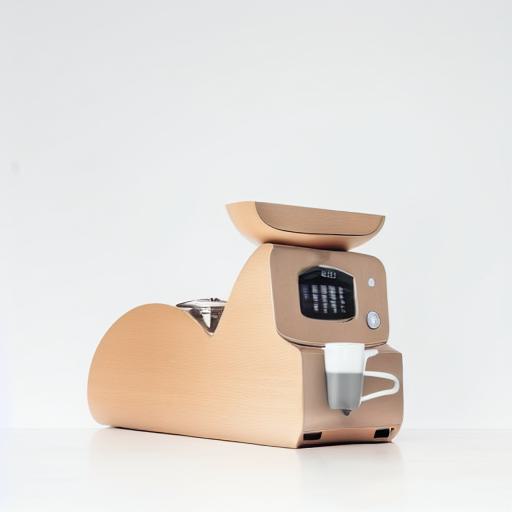}} & {\includegraphics[align=c,width=0.12\linewidth]{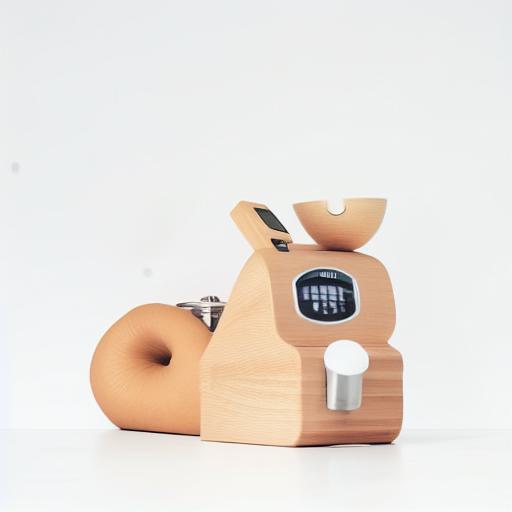}} & {\includegraphics[align=c,width=0.12\linewidth]{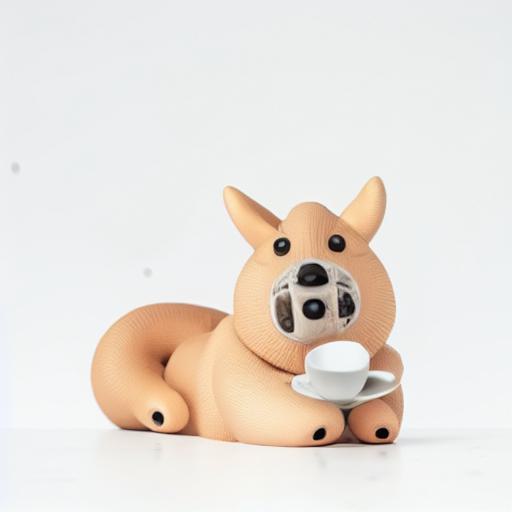}} \\
\multicolumn{2}{c}{}                  & P2P &{\includegraphics[align=c,width=0.12\linewidth]{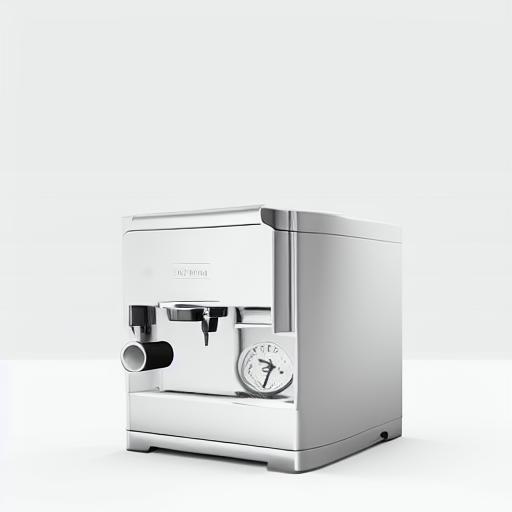}} & {\includegraphics[align=c,width=0.12\linewidth]{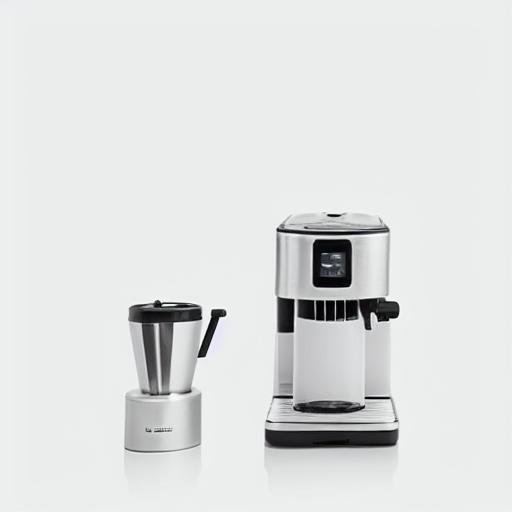}} & {\includegraphics[align=c,width=0.12\linewidth]{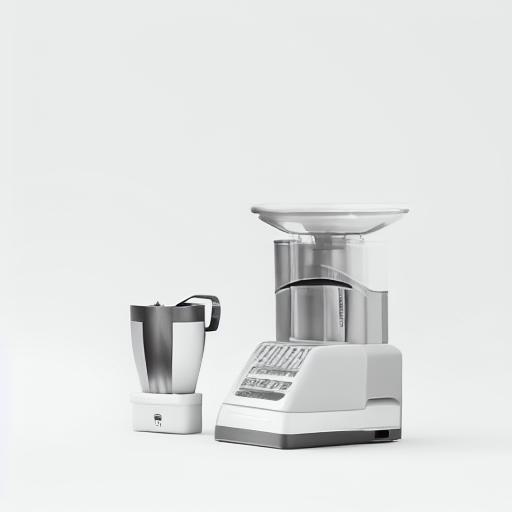}} & {\includegraphics[align=c,width=0.12\linewidth]{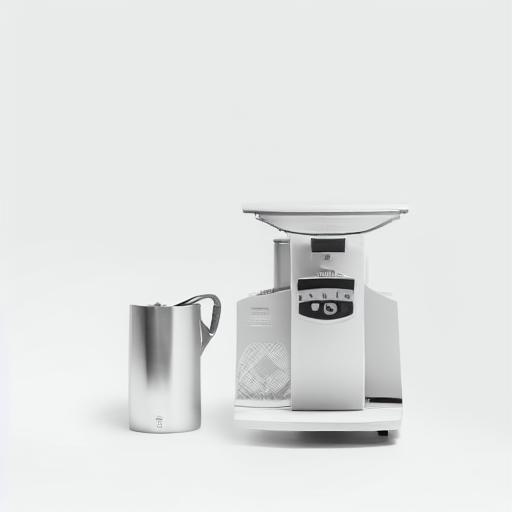}} & {\includegraphics[align=c,width=0.12\linewidth]{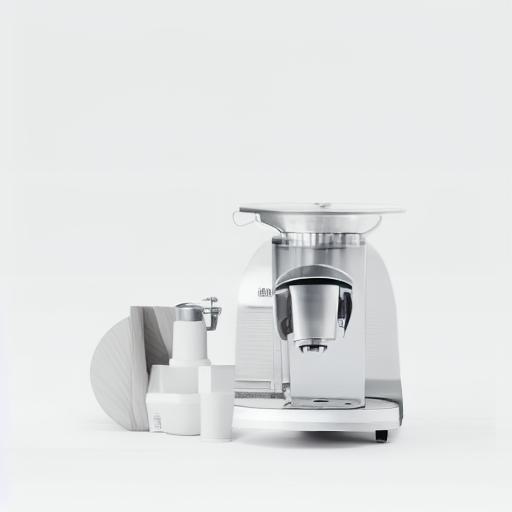}} & {\includegraphics[align=c,width=0.12\linewidth]{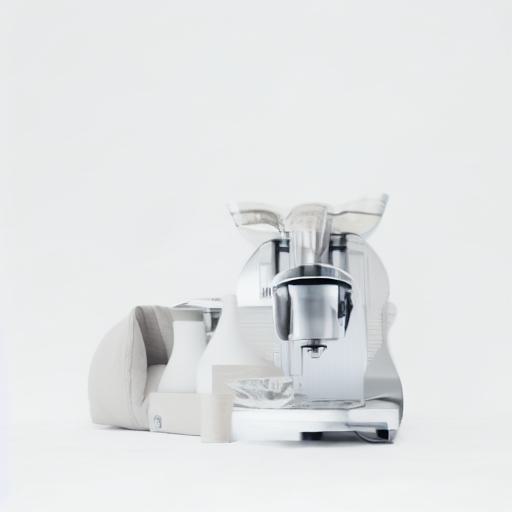}} \\
\multicolumn{2}{c}{\multirow{2}{*}{{\includegraphics[align=c,width=0.12\linewidth]{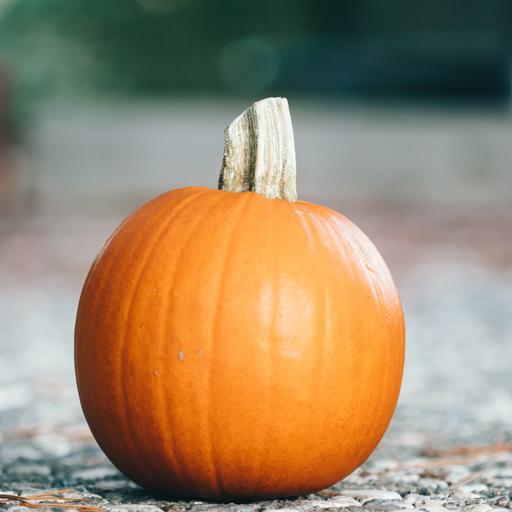}}}} \hspace{2mm} & \textbf{MDP-$\boldsymbol\epsilon_t$} & {\includegraphics[align=c,width=0.12\linewidth]{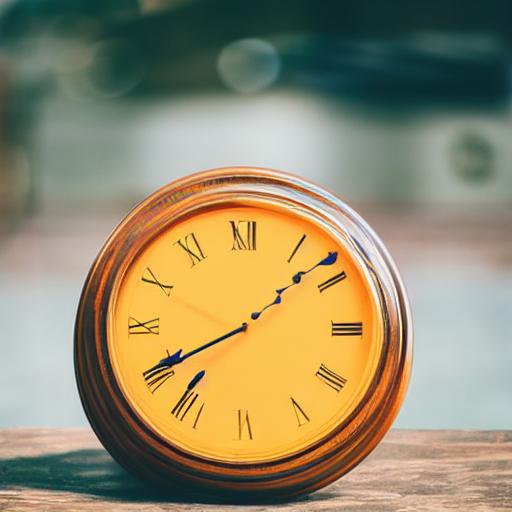}} & {\includegraphics[align=c,width=0.12\linewidth]{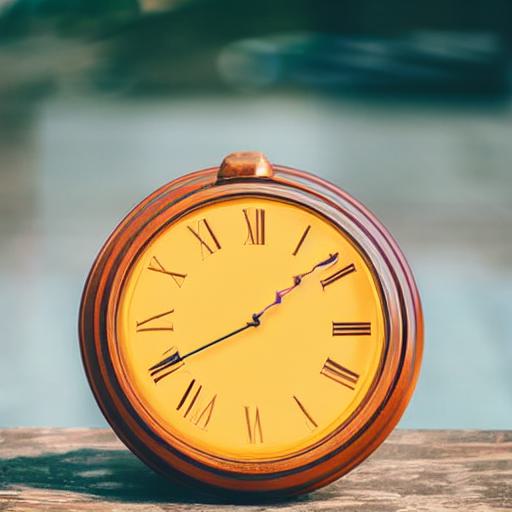}} & {\includegraphics[align=c,width=0.12\linewidth]{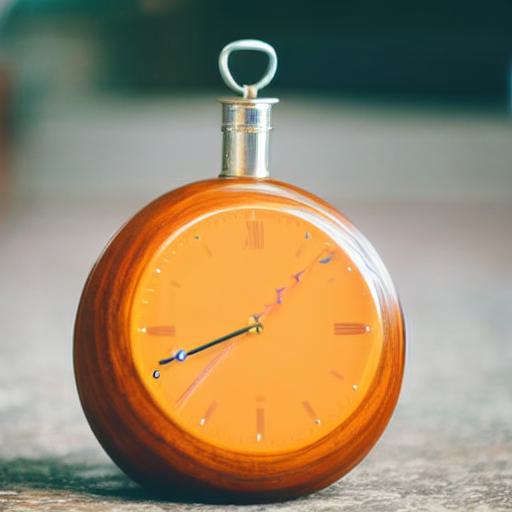}} & {\includegraphics[align=c,width=0.12\linewidth]{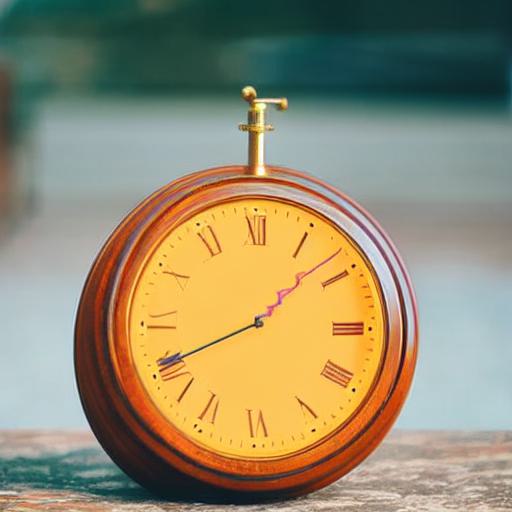}} & {\includegraphics[align=c,width=0.12\linewidth]{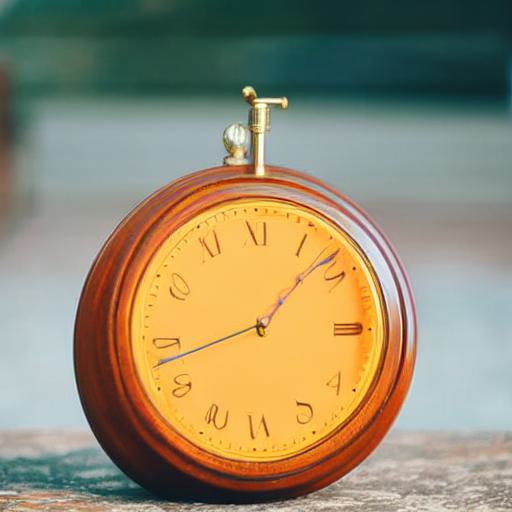}} & {\includegraphics[align=c,width=0.12\linewidth]{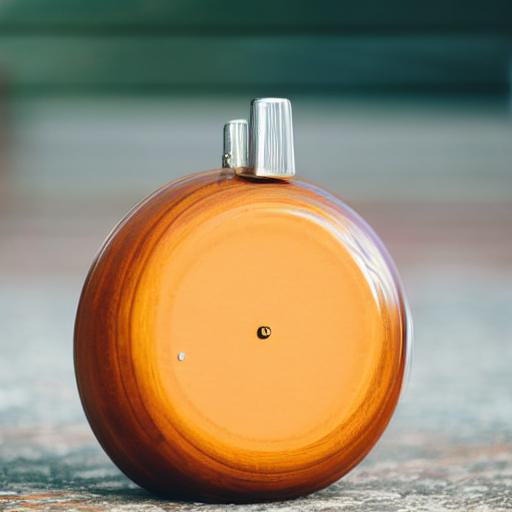}} \\
\multicolumn{2}{c}{}                  & P2P & {\includegraphics[align=c,width=0.12\linewidth]{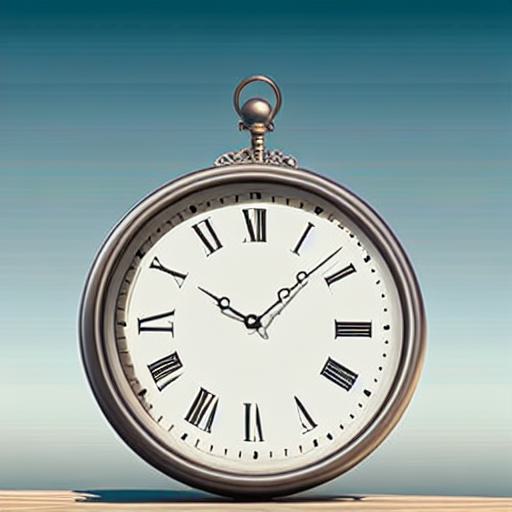}} & {\includegraphics[align=c,width=0.12\linewidth]{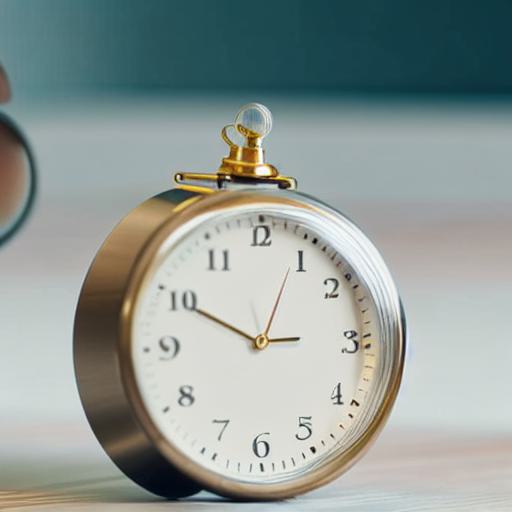}} & {\includegraphics[align=c,width=0.12\linewidth]{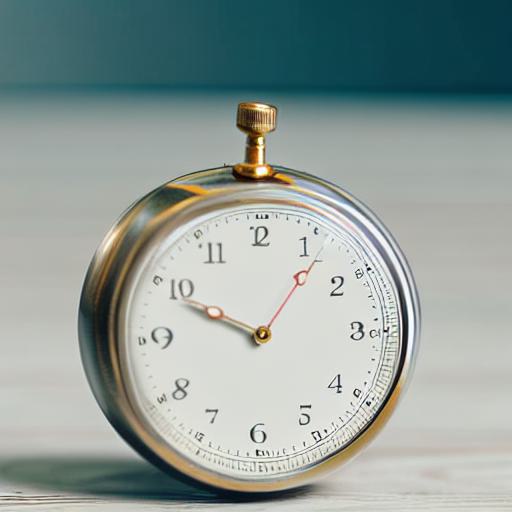}} & {\includegraphics[align=c,width=0.12\linewidth]{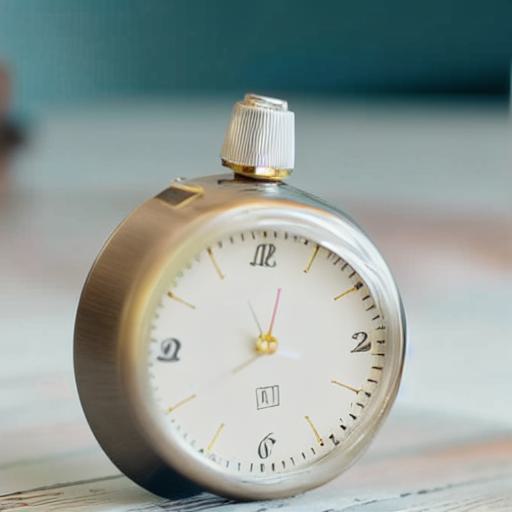}} & {\includegraphics[align=c,width=0.12\linewidth]{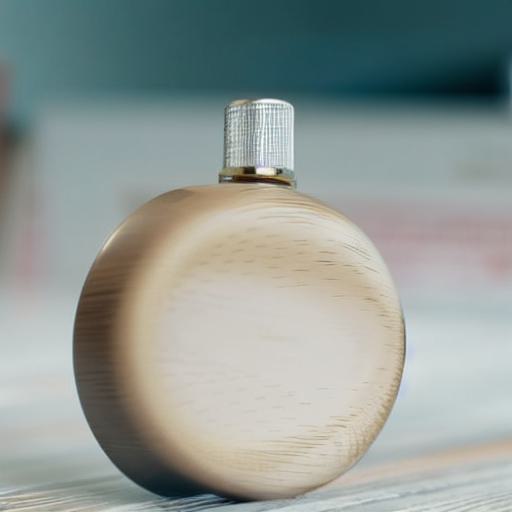}} & {\includegraphics[align=c,width=0.12\linewidth]{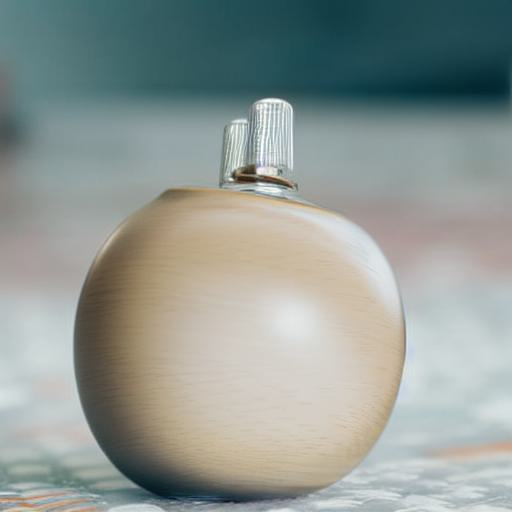}}
\end{tabular}
\caption{More intermediate results of mixing objects comparing \textbf{MDP-$\boldsymbol\epsilon_t$} and P2P. In the first two rows we mix the input ``corgi'' with ``coffee machine''. In the second two rows we mix the input ``pumpkin'' with ``clock''.}
\label{fig:mixing-objects-multiple}
\end{figure*}

\begin{figure*}[h]
\centering
\scriptsize
\setlength{\tabcolsep}{1pt}
\begin{tabular}{cccccccc}
Input & Edit & P2P & \textbf{MDP-$\boldsymbol\epsilon_t$} & Input & Edit & P2P & \textbf{MDP-$\boldsymbol\epsilon_t$} \\
\includegraphics[align=c,width=0.12\linewidth]{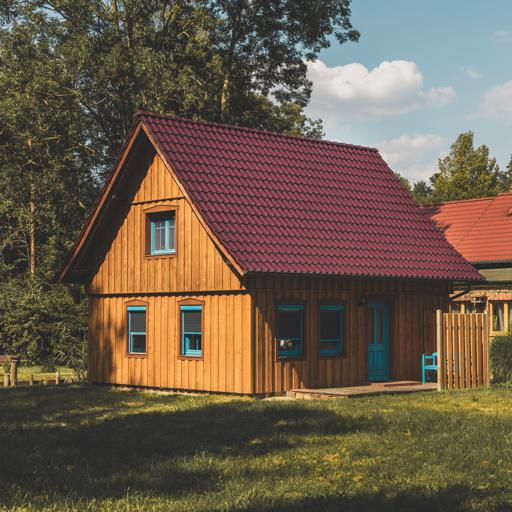} & 
        \begin{tabular}[c]{@{}c@{}}``Grass''\\to\\``Road''\end{tabular}& \includegraphics[align=c,width=0.12\linewidth]{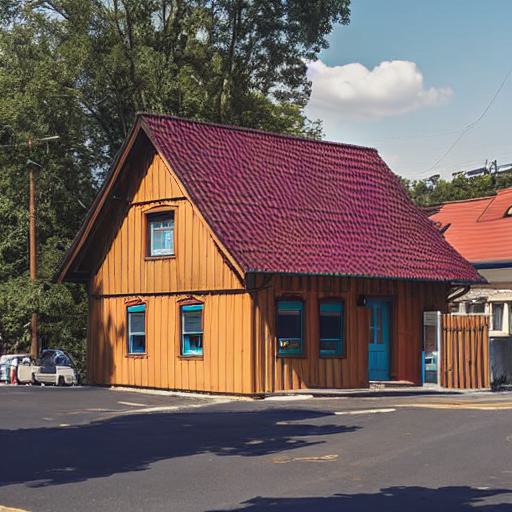}  & \includegraphics[align=c,width=0.12\linewidth]{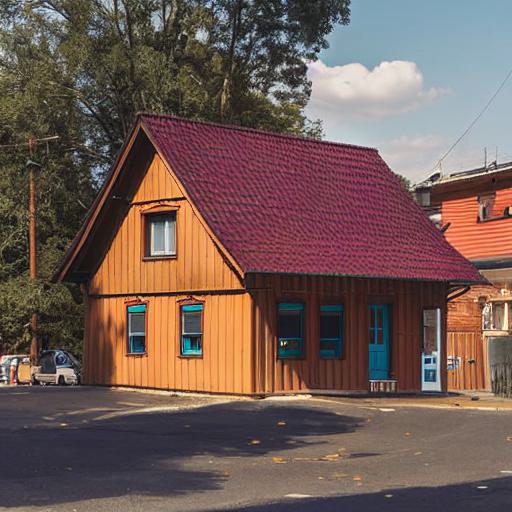} \hspace{3mm} & \includegraphics[align=c,width=0.12\linewidth]{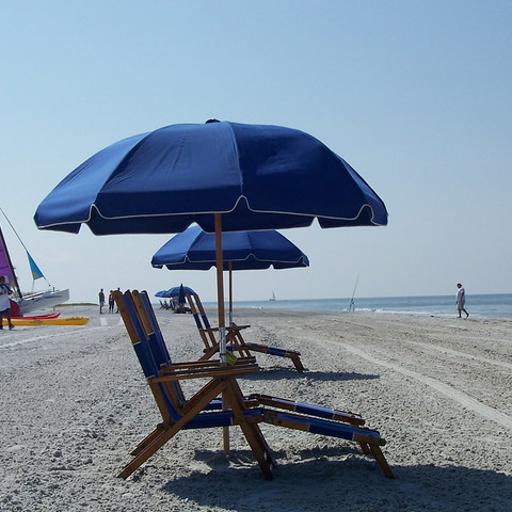}  &         \begin{tabular}[c]{@{}c@{}}``Beach''\\to\\``Mountain''\end{tabular}  & \includegraphics[align=c,width=0.12\linewidth]{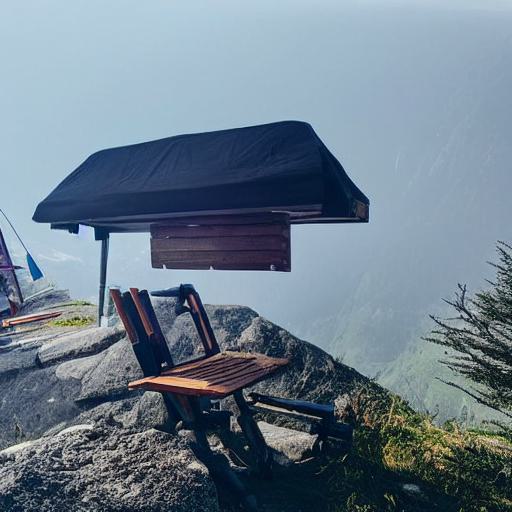}  & \includegraphics[align=c,width=0.12\linewidth]{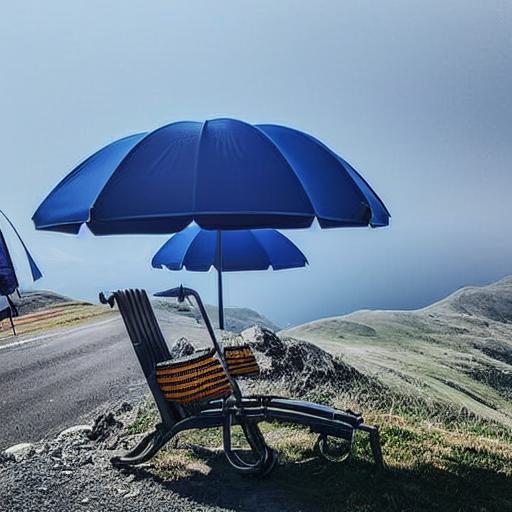} \vspace{1mm} \\ 
\includegraphics[align=c,width=0.12\linewidth]{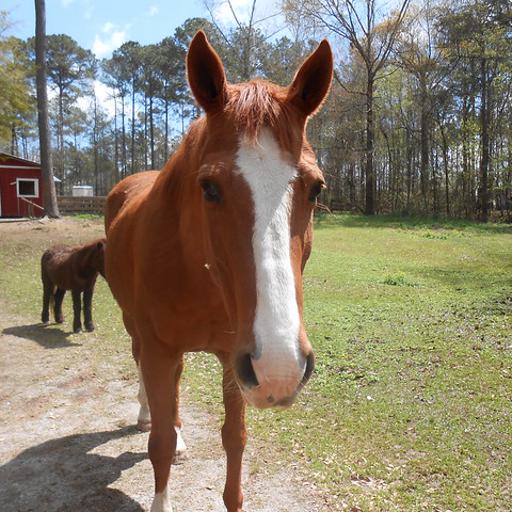} &         \begin{tabular}[c]{@{}c@{}}``Grass''\\to\\``Roof''\end{tabular} & \includegraphics[align=c,width=0.12\linewidth]{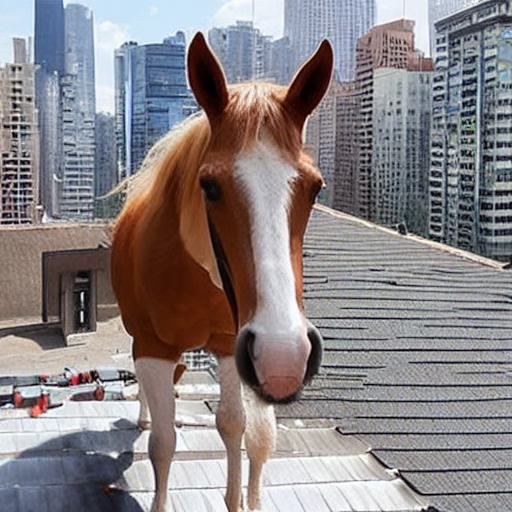} & \includegraphics[align=c,width=0.12\linewidth]{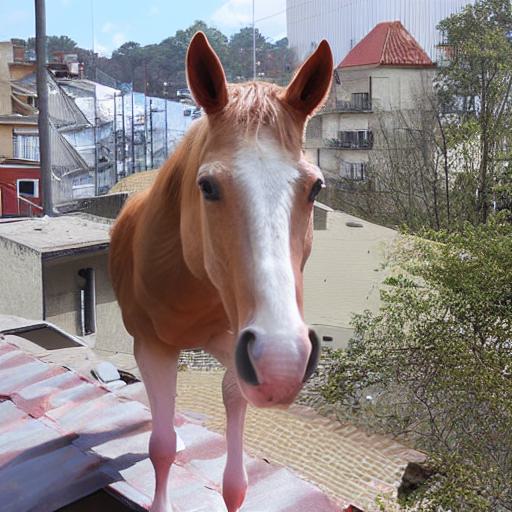} \hspace{3mm} & \includegraphics[align=c,width=0.12\linewidth]{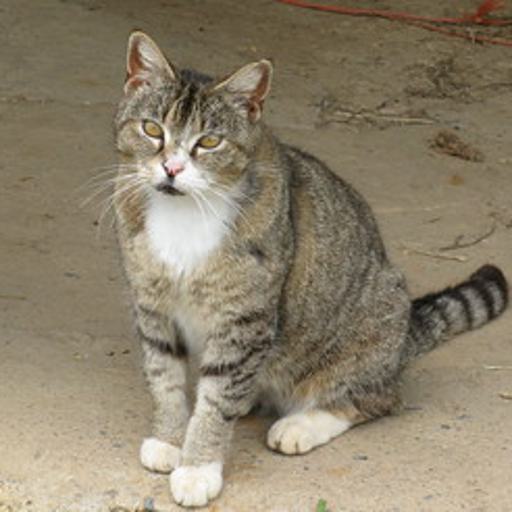} &         \begin{tabular}[c]{@{}c@{}}``Ground''\\to\\``Boat''\end{tabular} & \includegraphics[align=c,width=0.12\linewidth]{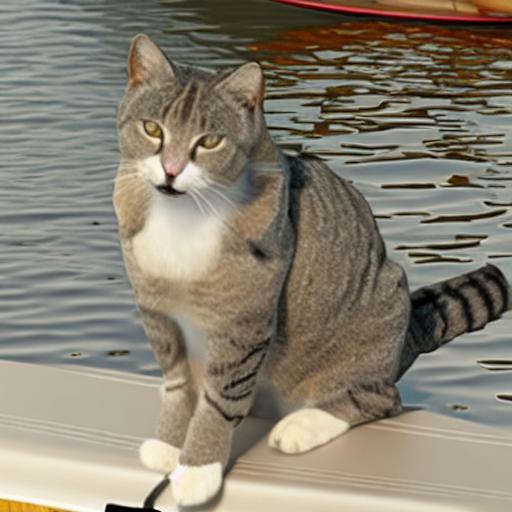} & \includegraphics[align=c,width=0.12\linewidth]{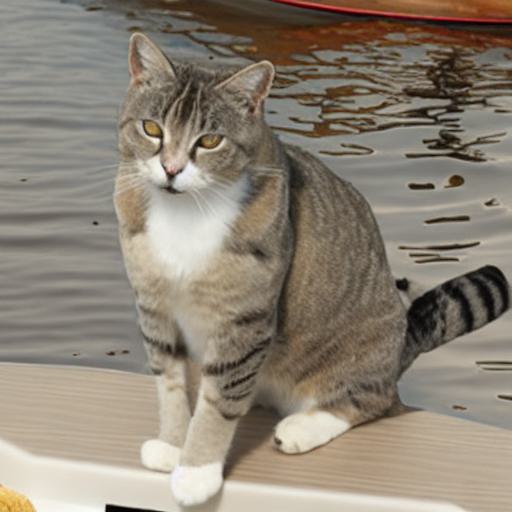} \vspace{1mm} \\
\includegraphics[align=c,width=0.12\linewidth]{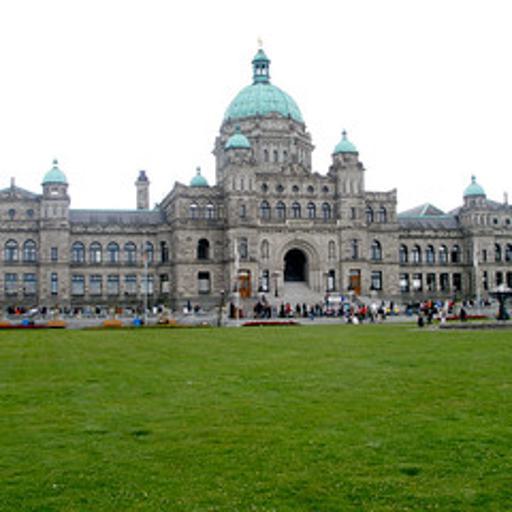} &         \begin{tabular}[c]{@{}c@{}}``Grass''\\to\\``Lake''\end{tabular} & \includegraphics[align=c,width=0.12\linewidth]{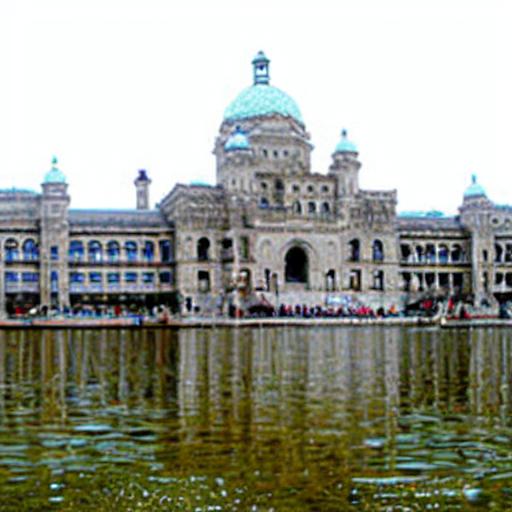}  & \includegraphics[align=c,width=0.12\linewidth]{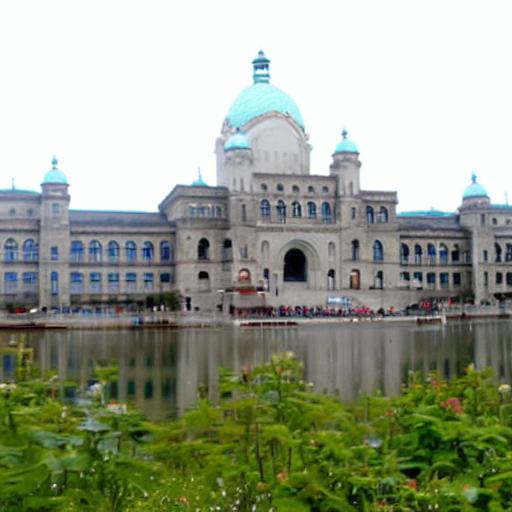} \hspace{3mm} & \includegraphics[align=c,width=0.12\linewidth]{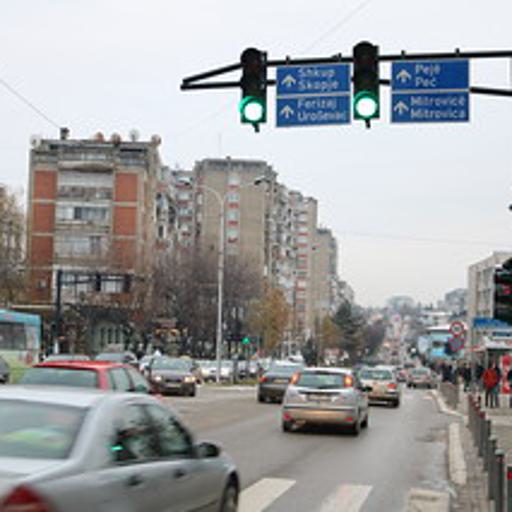}  &         \begin{tabular}[c]{@{}c@{}}``City''\\to\\``Mountain''\end{tabular}  & \includegraphics[align=c,width=0.12\linewidth]{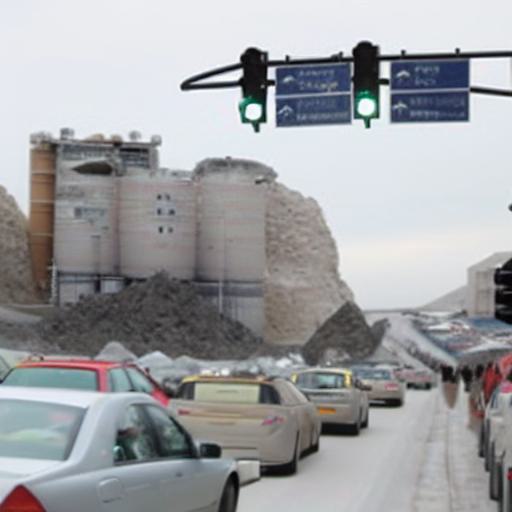}  & \includegraphics[align=c,width=0.12\linewidth]{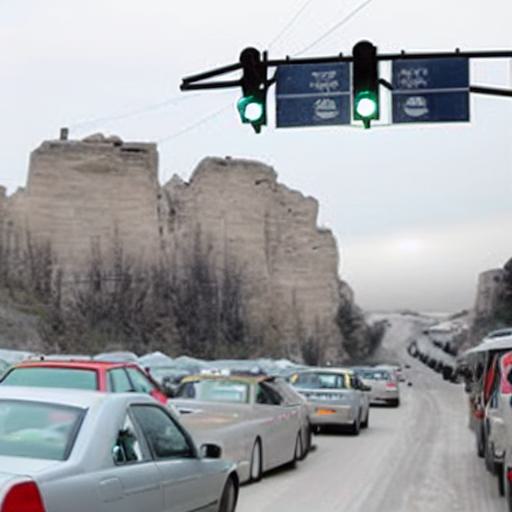} \vspace{1mm} \\ 
\includegraphics[align=c,width=0.12\linewidth]{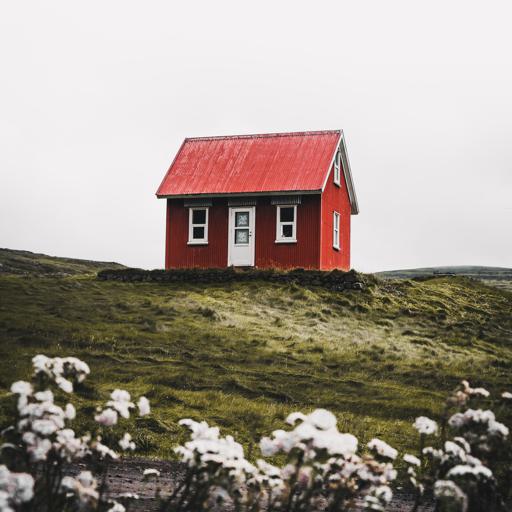} &         \begin{tabular}[c]{@{}c@{}}``Mountain''\\to\\``Mars''\end{tabular} & \includegraphics[align=c,width=0.12\linewidth]{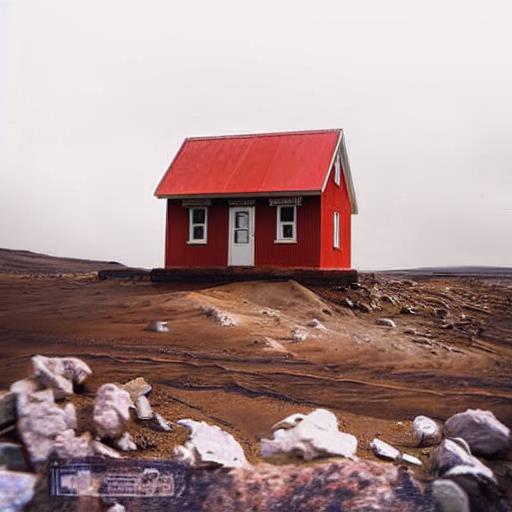} & \includegraphics[align=c,width=0.12\linewidth]{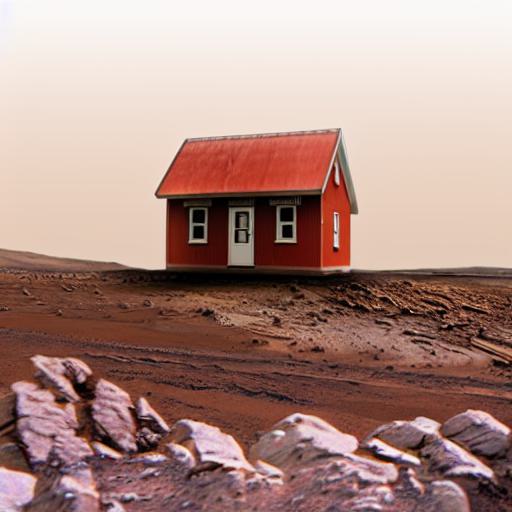} \hspace{3mm} & \includegraphics[align=c,width=0.12\linewidth]{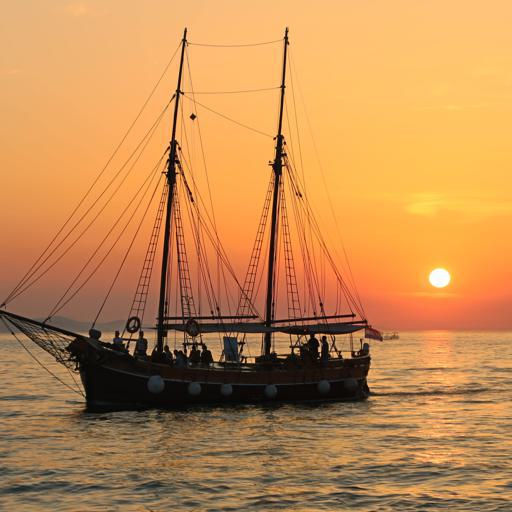} &         \begin{tabular}[c]{@{}c@{}}``Sea''\\to\\``Mountain''\end{tabular} & \includegraphics[align=c,width=0.12\linewidth]{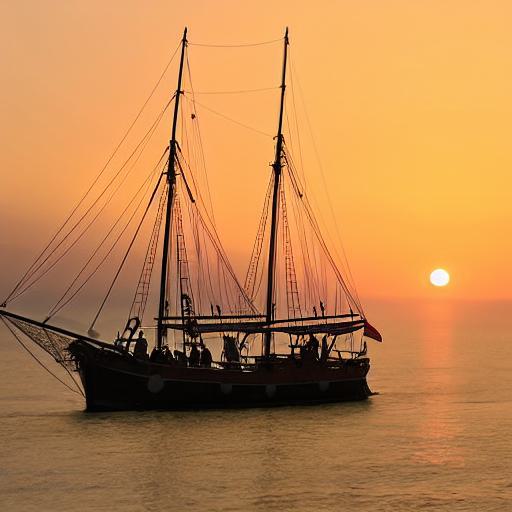} & \includegraphics[align=c,width=0.12\linewidth]{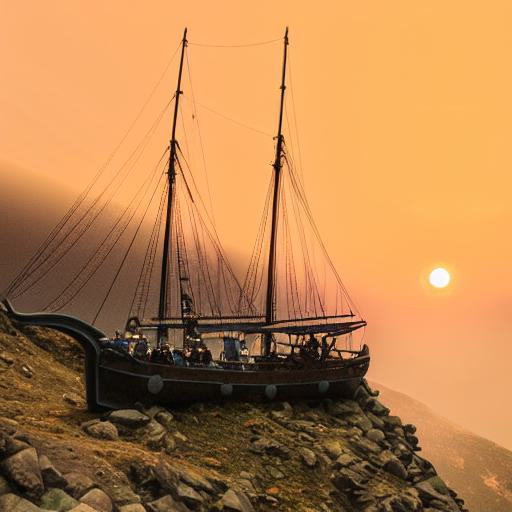} \vspace{1mm}\\
\end{tabular}
\caption{Results of changing background comparing Prompt-to-Prompt and \textbf{MDP-$\boldsymbol\epsilon_t$}.}
\label{fig:global-changing-background}
\end{figure*}

\begin{figure*}[h]
\centering
\scriptsize
\setlength{\tabcolsep}{1pt}
\begin{tabular}{cccccccc}
Input & Edit & P2P & \textbf{MDP-$\boldsymbol\epsilon_t$} & Input & Edit & P2P & \textbf{MDP-$\boldsymbol\epsilon_t$} \\
\includegraphics[align=c,width=0.12\linewidth]{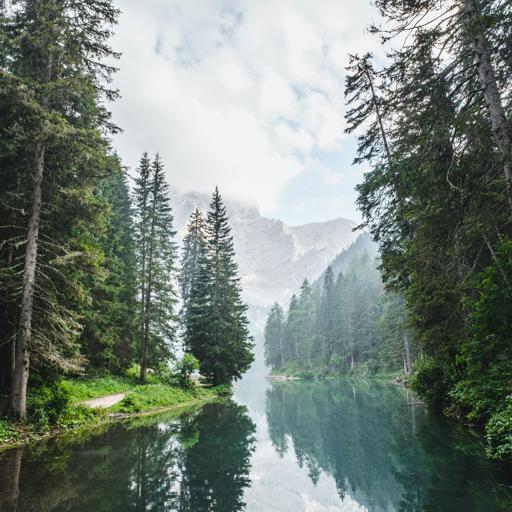} & 
          \begin{tabular}[c]{@{}c@{}}``Spring''\\to\\``Winter''\end{tabular}& \includegraphics[align=c,width=0.12\linewidth]{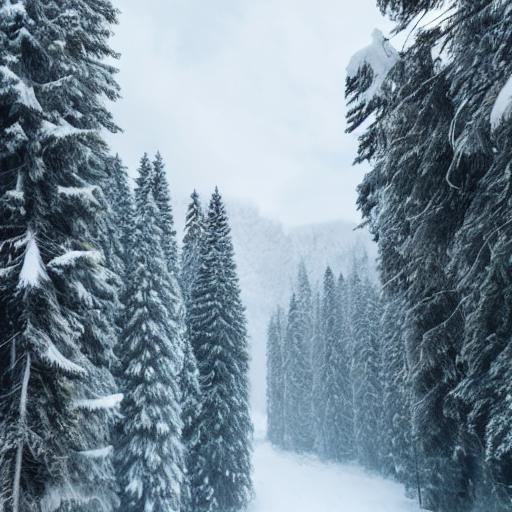}  & \includegraphics[align=c,width=0.12\linewidth]{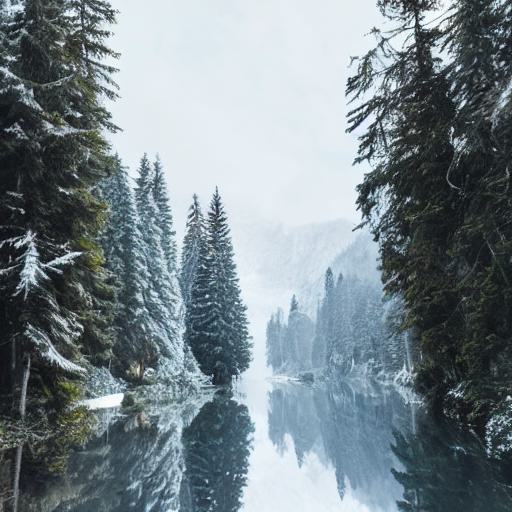} \hspace{3mm} & \includegraphics[align=c,width=0.12\linewidth]{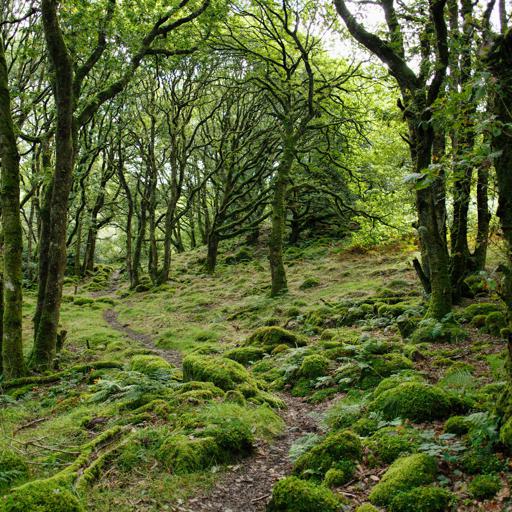}  & \begin{tabular}[c]{@{}c@{}}``Spring''\\to\\``Winter''\end{tabular}  & \includegraphics[align=c,width=0.12\linewidth]{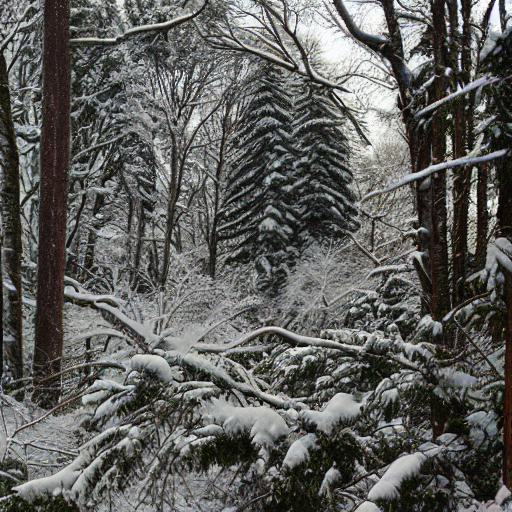}  & \includegraphics[align=c,width=0.12\linewidth]{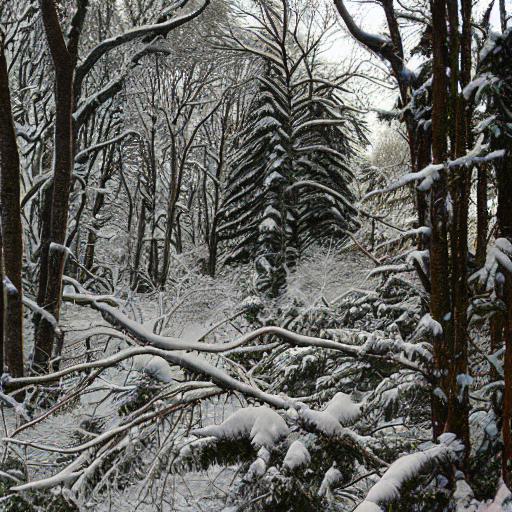} \vspace{1mm} \\ 
\includegraphics[align=c,width=0.12\linewidth]{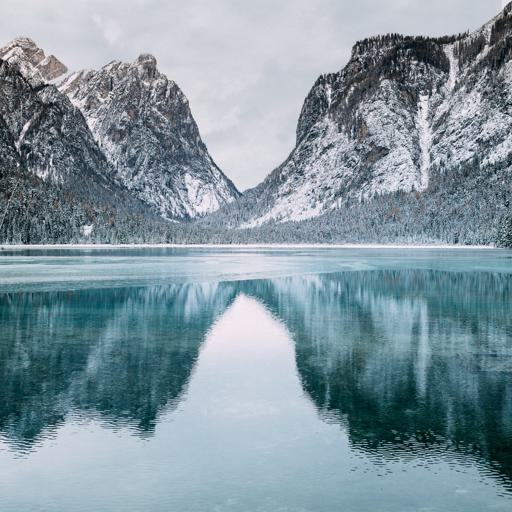} & \begin{tabular}[c]{@{}c@{}}``Winter''\\to\\``Spring''\end{tabular} & \includegraphics[align=c,width=0.12\linewidth]{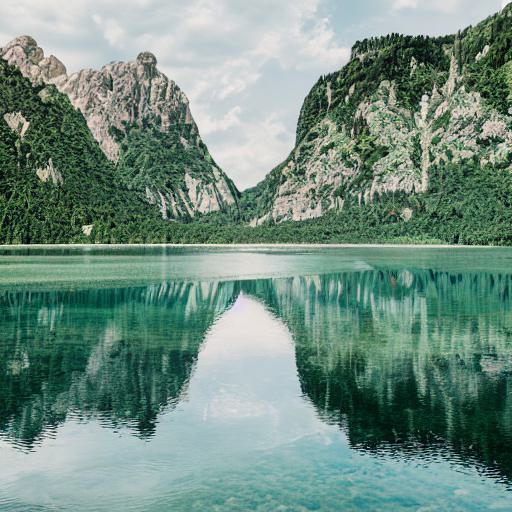} & \includegraphics[align=c,width=0.12\linewidth]{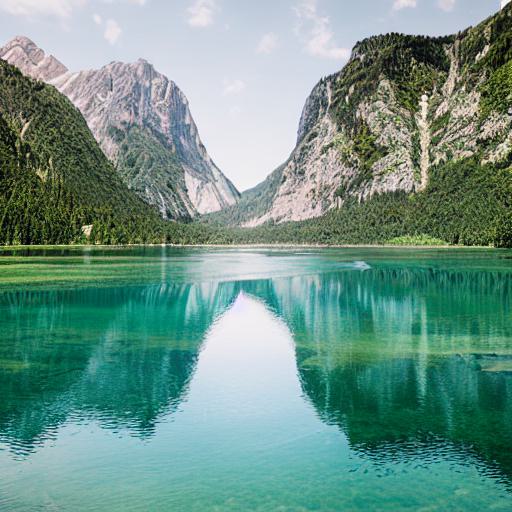} \hspace{3mm} & \includegraphics[align=c,width=0.12\linewidth]{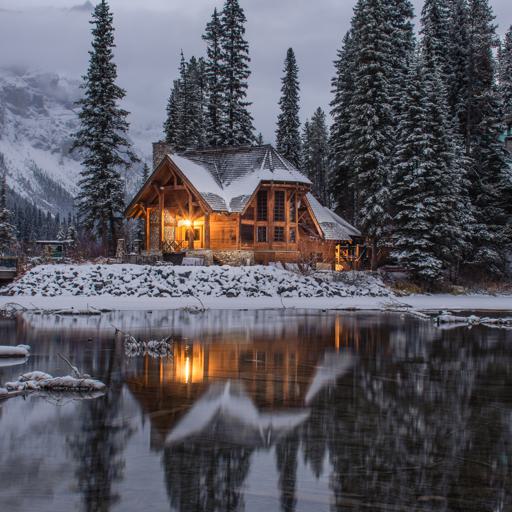} & \begin{tabular}[c]{@{}c@{}}``Winter''\\to\\``Spring''\end{tabular} & \includegraphics[align=c,width=0.12\linewidth]{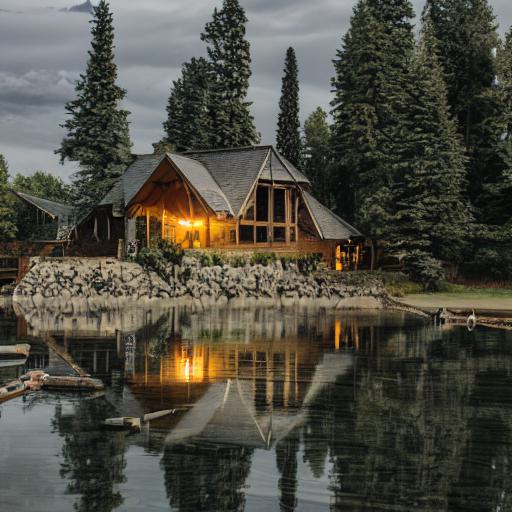} & \includegraphics[align=c,width=0.12\linewidth]{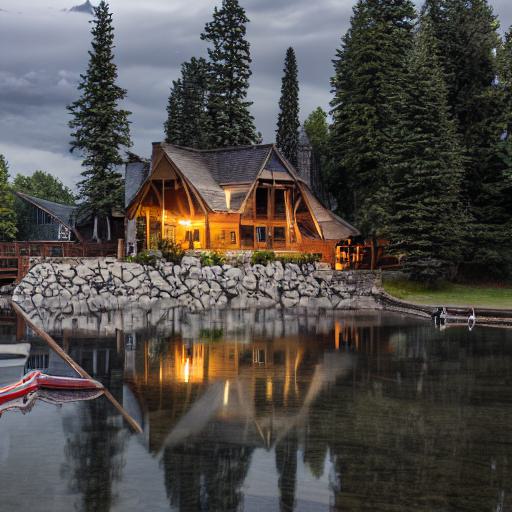} \vspace{1mm} \\
\includegraphics[align=c,width=0.12\linewidth]{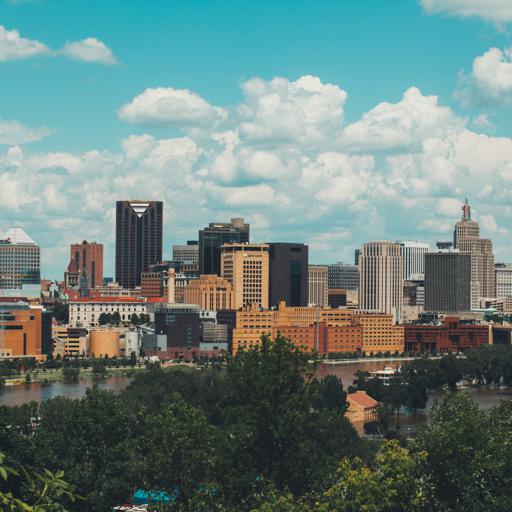} & \begin{tabular}[c]{@{}c@{}}``Sunny''\\to\\``Rainstormy''\end{tabular} & \includegraphics[align=c,width=0.12\linewidth]{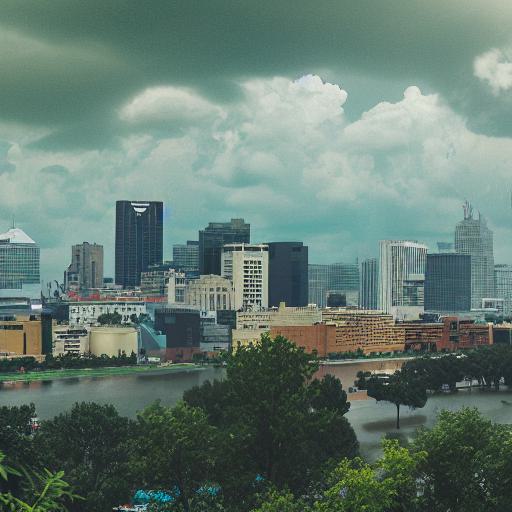}  & \includegraphics[align=c,width=0.12\linewidth]{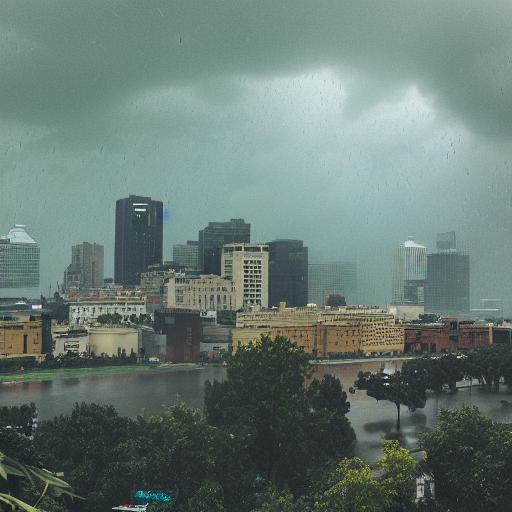} \hspace{3mm} & \includegraphics[align=c,width=0.12\linewidth]{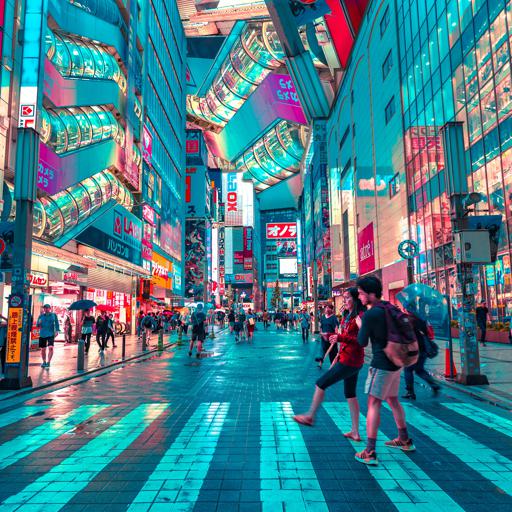}  & \begin{tabular}[c]{@{}c@{}}``Colored''\\to\\``Black-and-\\white''\end{tabular} & \includegraphics[align=c,width=0.12\linewidth]{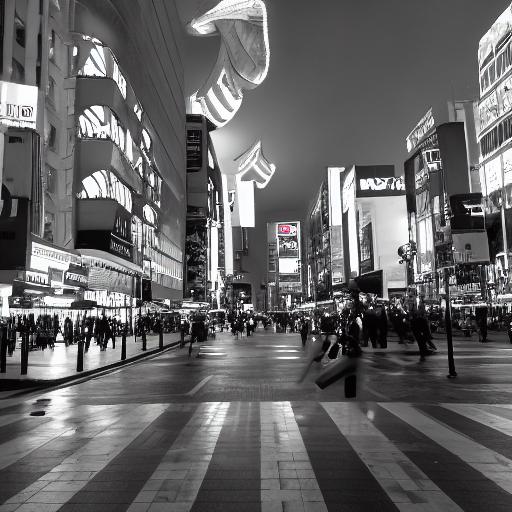}  & \includegraphics[align=c,width=0.12\linewidth]{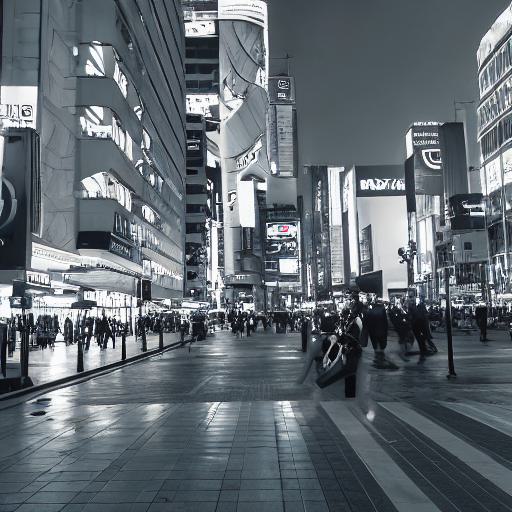} \vspace{1mm} \\ 
\includegraphics[align=c,width=0.12\linewidth]{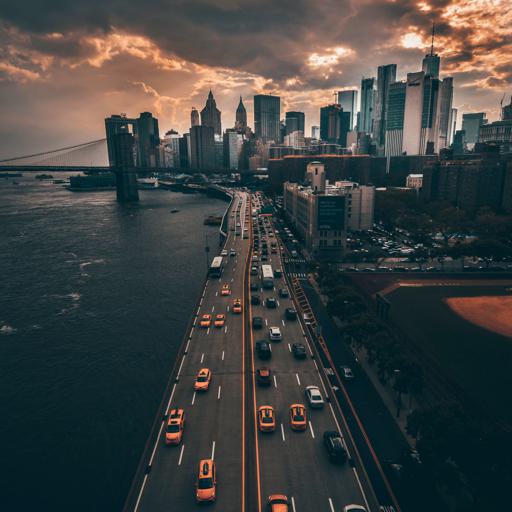} & \begin{tabular}[c]{@{}c@{}}``Evening''\\to\\``Morning''\end{tabular} & \includegraphics[align=c,width=0.12\linewidth]{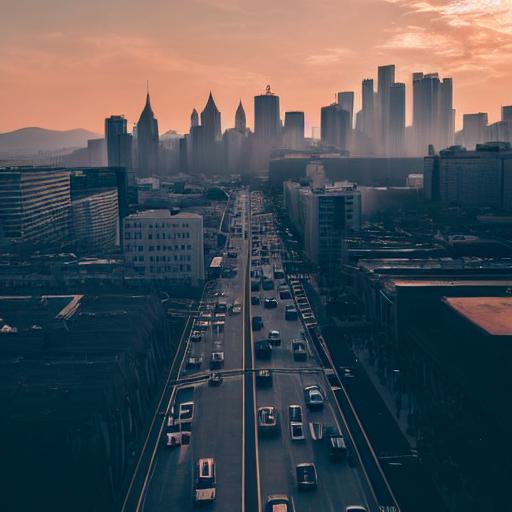} & \includegraphics[align=c,width=0.12\linewidth]{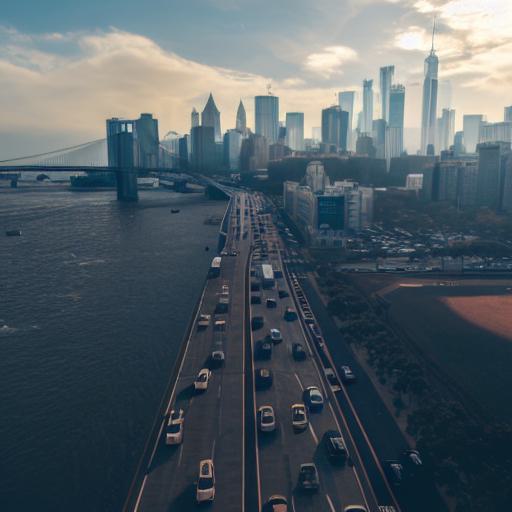} \hspace{3mm} & \includegraphics[align=c,width=0.12\linewidth]{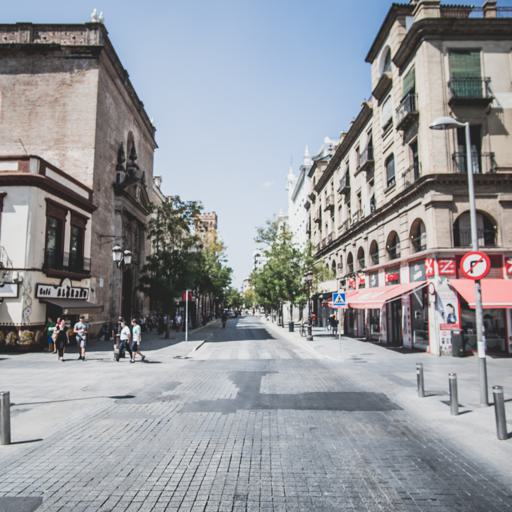} & 
\begin{tabular}[c]{@{}c@{}}``Morning''\\to\\``Evening''\end{tabular} & \includegraphics[align=c,width=0.12\linewidth]{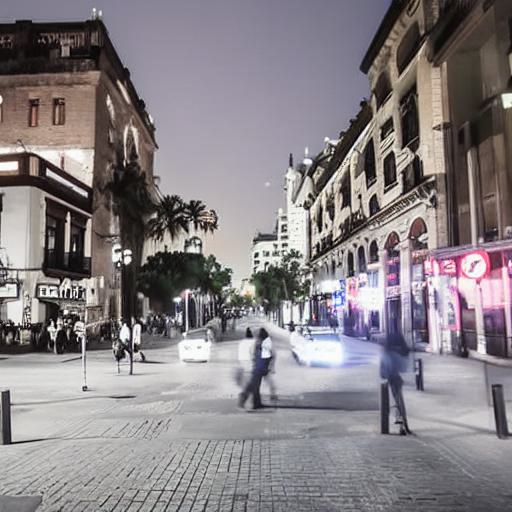} & \includegraphics[align=c,width=0.12\linewidth]{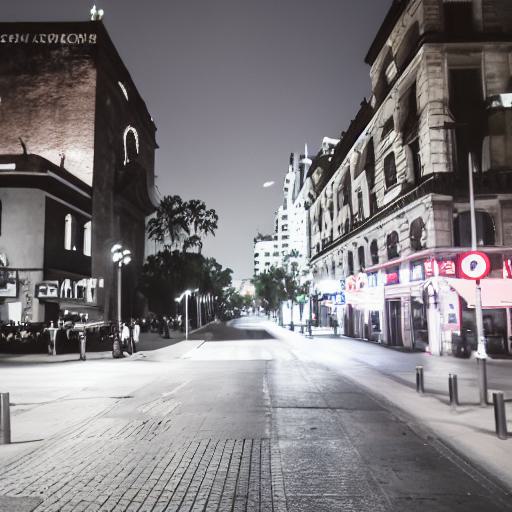} \vspace{1mm} \\
\end{tabular}
\caption{Results of in-domain transfer comparing Prompt-to-Prompt and \textbf{MDP-$\boldsymbol\epsilon_t$}.}
\label{fig:global-in-domain-transfer}
\end{figure*}

\begin{figure*}
\centering
\scriptsize
\setlength{\tabcolsep}{1pt}
\begin{tabular}{cccccccc}
Input & Edit & P2P & \textbf{MDP-$\boldsymbol\epsilon_t$} & Input & Edit & P2P & \textbf{MDP-$\boldsymbol\epsilon_t$} \\
\includegraphics[align=c,width=0.12\linewidth]{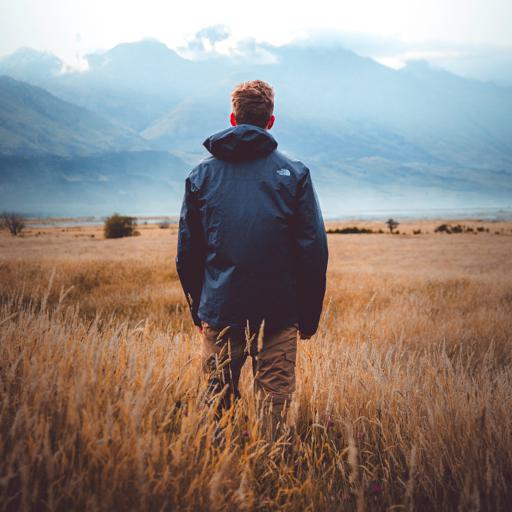} & 
  \begin{tabular}[c]{@{}c@{}}``Photo''\\to\\``Children's \\ Painting''\end{tabular}& \includegraphics[align=c,width=0.12\linewidth]{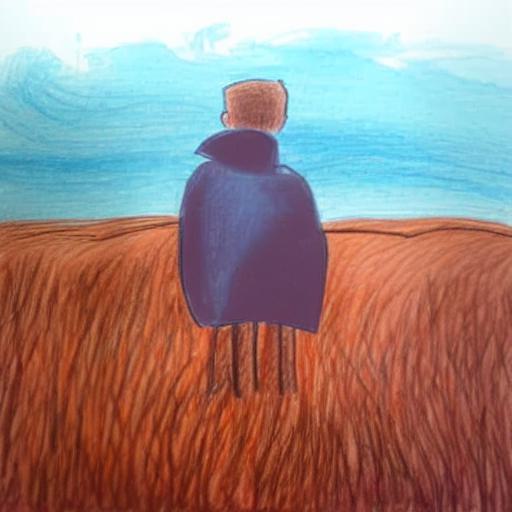}  & \includegraphics[align=c,width=0.12\linewidth]{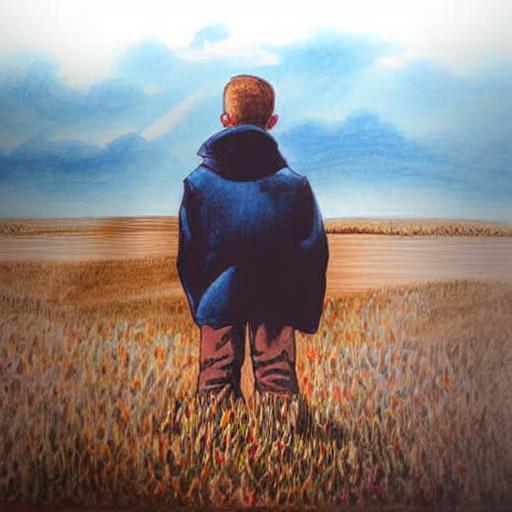} \hspace{3mm} & \includegraphics[align=c,width=0.12\linewidth]{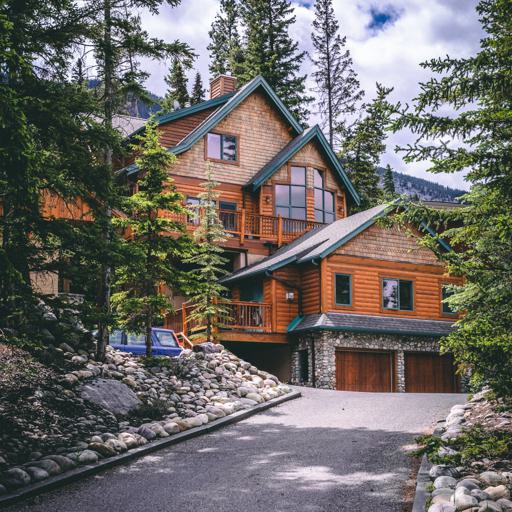}  &   \begin{tabular}[c]{@{}c@{}}``Photo''\\to\\``Children's \\ Painting''\end{tabular}  & \includegraphics[align=c,width=0.12\linewidth]{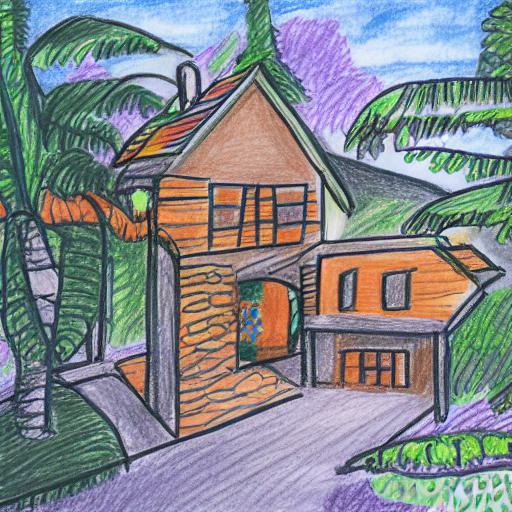}  & \includegraphics[align=c,width=0.12\linewidth]{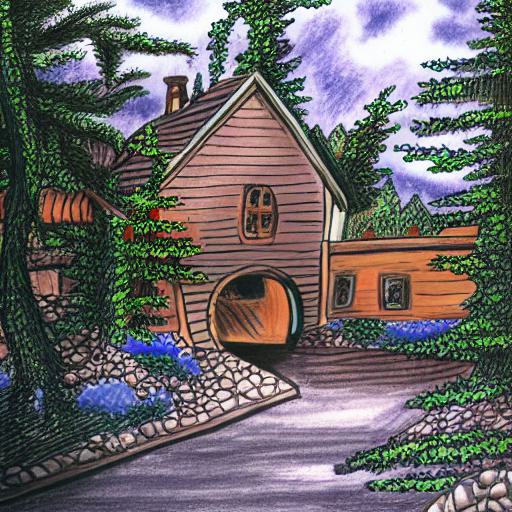} \vspace{1mm} \\ 
\includegraphics[align=c,width=0.12\linewidth]{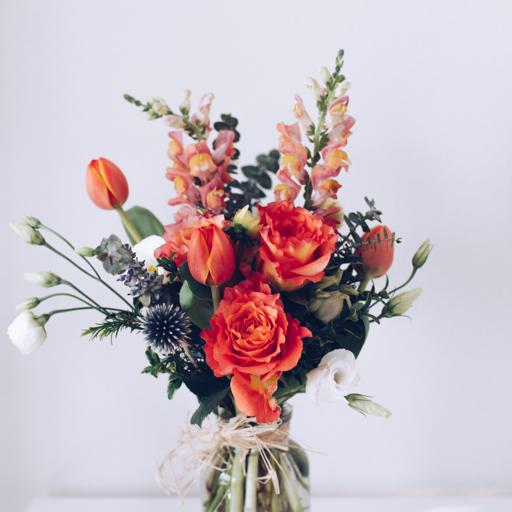} &   \begin{tabular}[c]{@{}c@{}}``Photo''\\to\\``Watercolor \\ Painting''\end{tabular} & \includegraphics[align=c,width=0.12\linewidth]{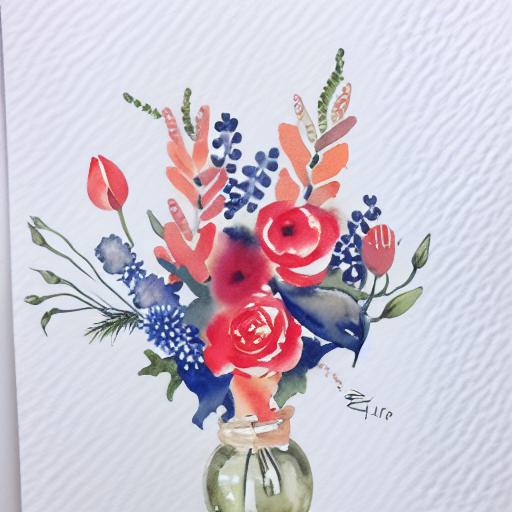} & \includegraphics[align=c,width=0.12\linewidth]{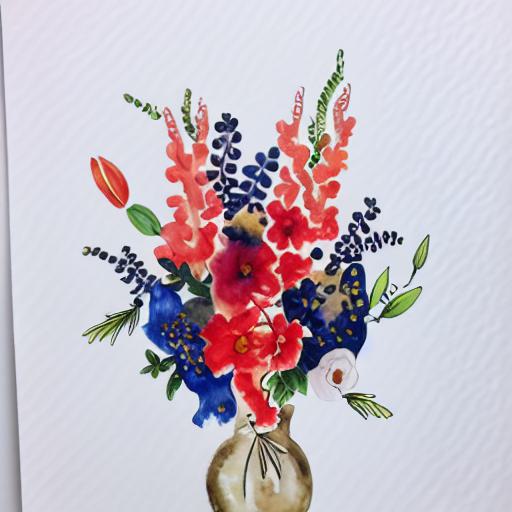} \hspace{3mm} & \includegraphics[align=c,width=0.12\linewidth]{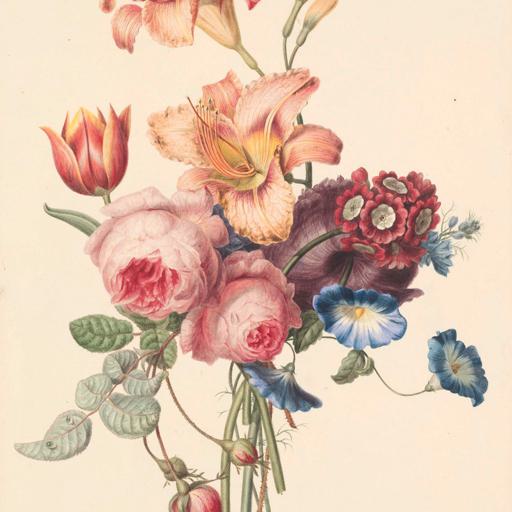} &   \begin{tabular}[c]{@{}c@{}}``Watercolor \\ Painting''\\``Photo''\end{tabular} & \includegraphics[align=c,width=0.12\linewidth]{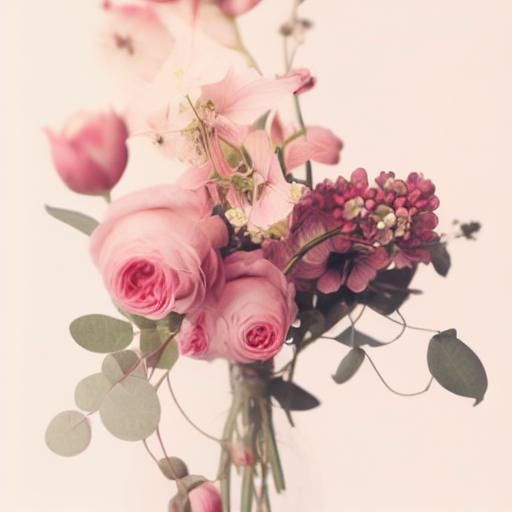} & \includegraphics[align=c,width=0.12\linewidth]{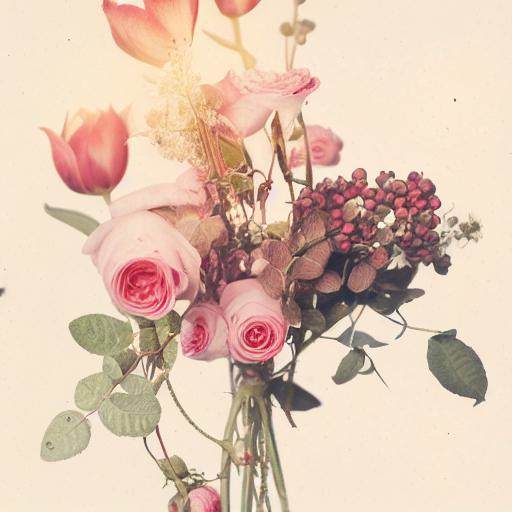} \vspace{1mm} \\
\includegraphics[align=c,width=0.12\linewidth]{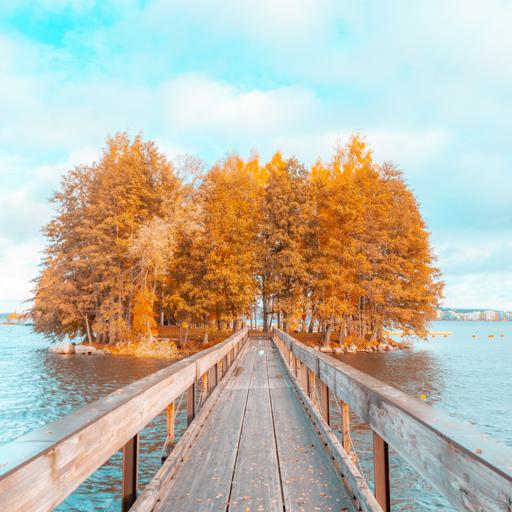} &   \begin{tabular}[c]{@{}c@{}}``Photo''\\to\\``Impressionism \\ Painting''\end{tabular} & \includegraphics[align=c,width=0.12\linewidth]{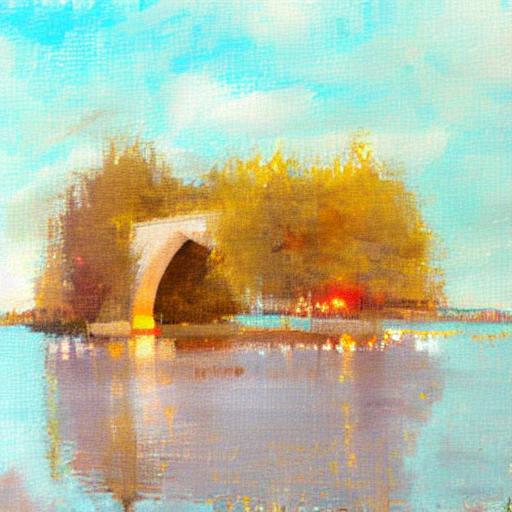}  & \includegraphics[align=c,width=0.12\linewidth]{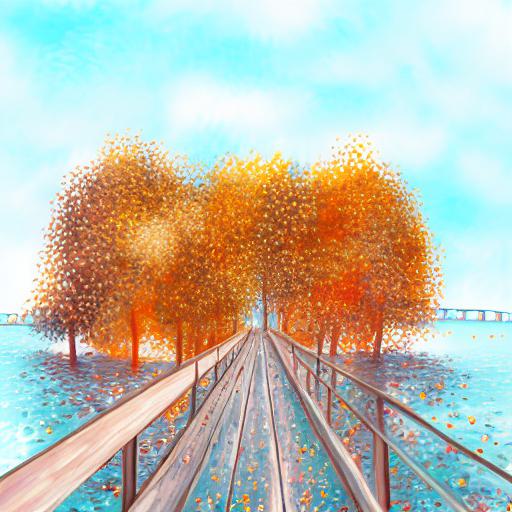} \hspace{3mm} & \includegraphics[align=c,width=0.12\linewidth]{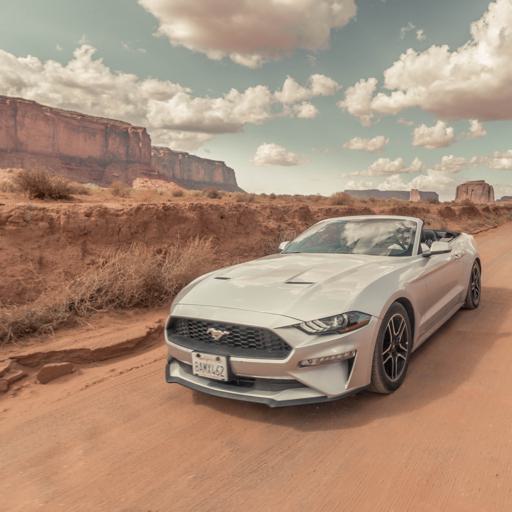}  &   \begin{tabular}[c]{@{}c@{}}``Photo''\\to\\``Sketch''\end{tabular}  & \includegraphics[align=c,width=0.12\linewidth]{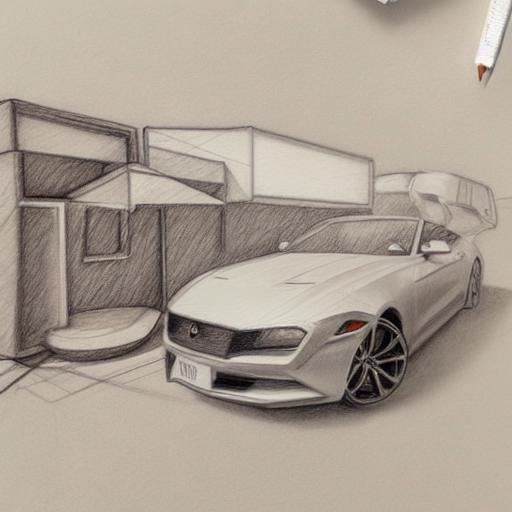}  & \includegraphics[align=c,width=0.12\linewidth]{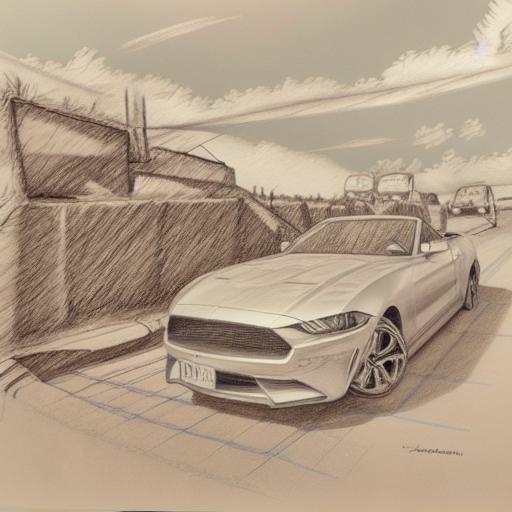} \vspace{1mm} \\ 
\includegraphics[align=c,width=0.12\linewidth]{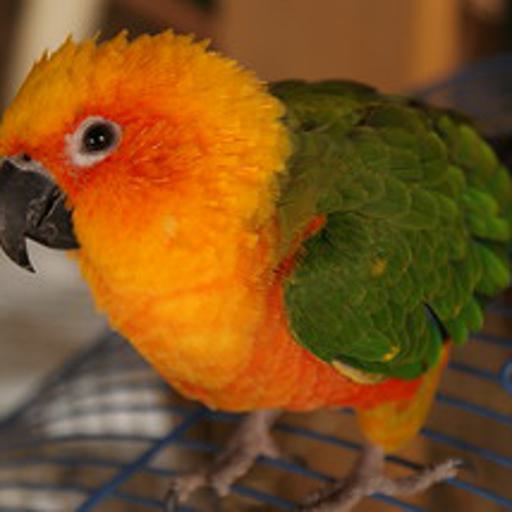} &   \begin{tabular}[c]{@{}c@{}}``Photo''\\to\\``Abstract \\ Painting''\end{tabular} & \includegraphics[align=c,width=0.12\linewidth]{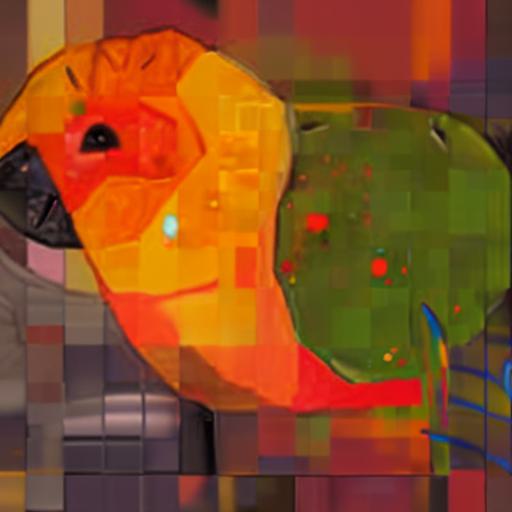} & \includegraphics[align=c,width=0.12\linewidth]{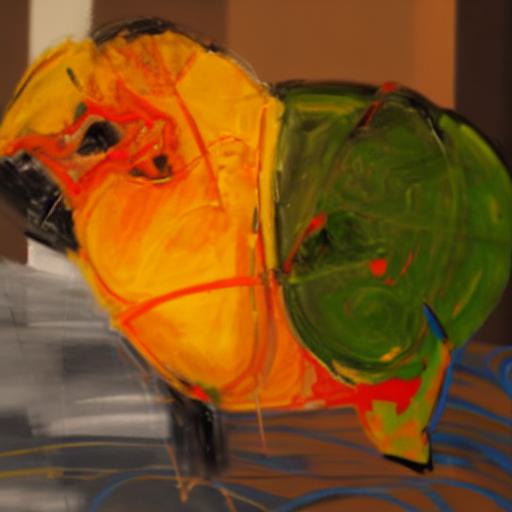} \hspace{3mm} & \includegraphics[align=c,width=0.12\linewidth]{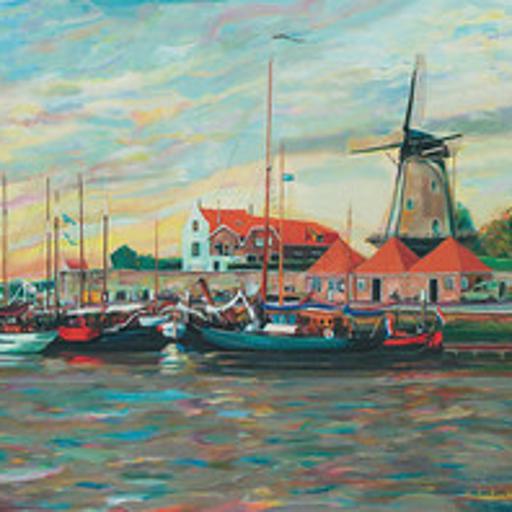} &   \begin{tabular}[c]{@{}c@{}}``Oil \\ Painting''\\to\\``Photo''\end{tabular} & \includegraphics[align=c,width=0.12\linewidth]{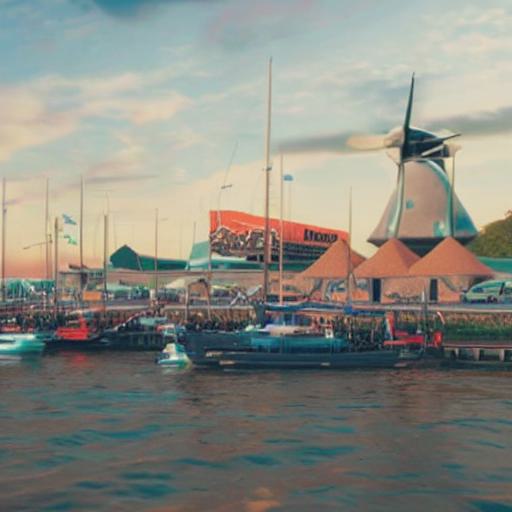} & \includegraphics[align=c,width=0.12\linewidth]{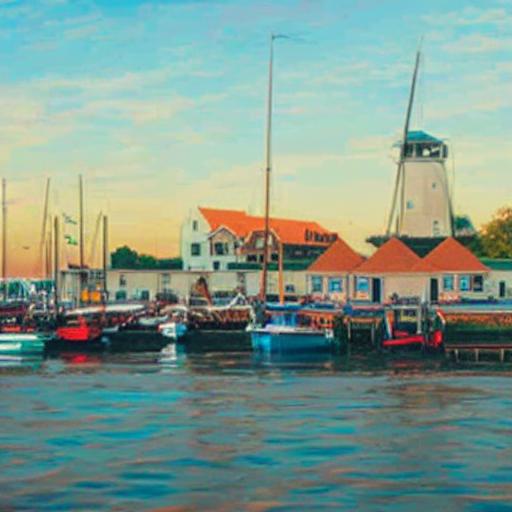} \vspace{1mm}\\
\end{tabular}
\caption{Results of out-domain transfer comparing Prompt-to-Prompt and \textbf{MDP-$\boldsymbol\epsilon_t$}.}
\label{fig:global-out-domain-transfer}
\end{figure*}

\begin{figure*}[h]
\centering
\scriptsize
\setlength{\tabcolsep}{1pt}
\begin{tabular}{cccccccc}
Input & Edit & P2P & \textbf{MDP-$\boldsymbol\epsilon_t$} & Input & Edit & P2P & \textbf{MDP-$\boldsymbol\epsilon_t$} \\
\includegraphics[align=c,width=0.12\linewidth]{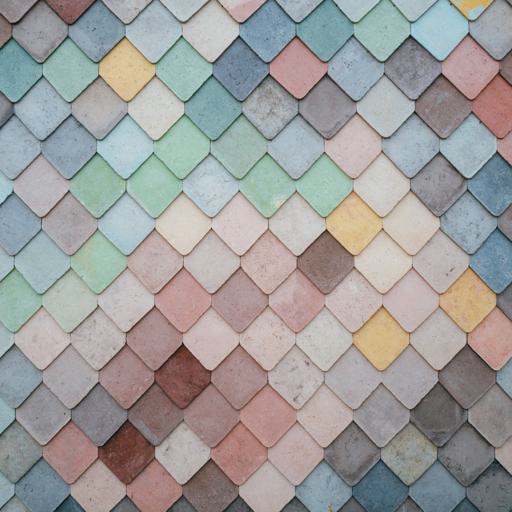} & 
    \begin{tabular}[c]{@{}c@{}}Stylize\\``Bus''\end{tabular}& \includegraphics[align=c,width=0.12\linewidth]{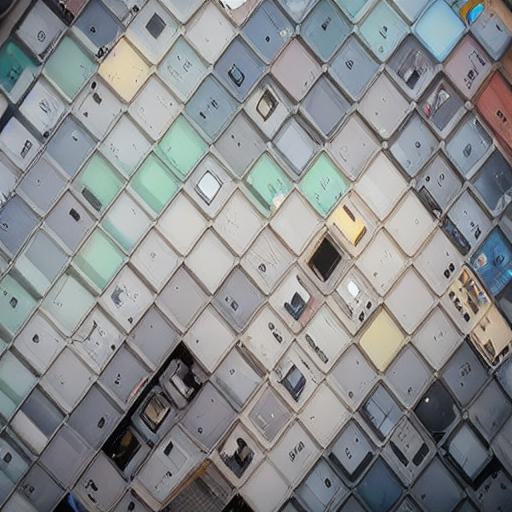}  & \includegraphics[align=c,width=0.12\linewidth]{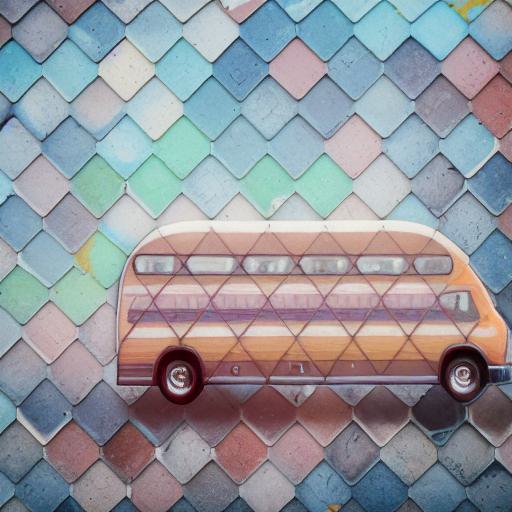} \hspace{3mm} & \includegraphics[align=c,width=0.12\linewidth]{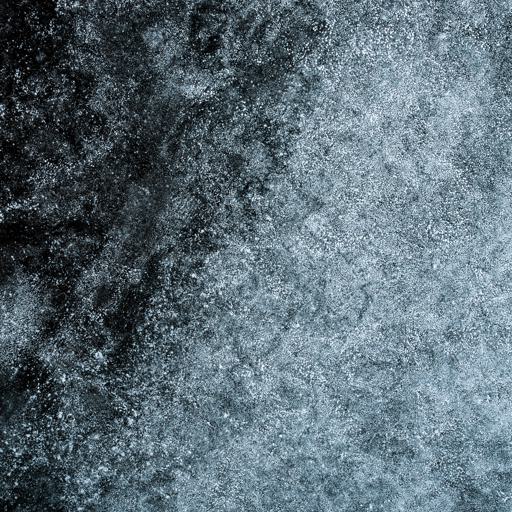}  & \begin{tabular}[c]{@{}c@{}}Stylize\\``Flower''\end{tabular}  & \includegraphics[align=c,width=0.12\linewidth]{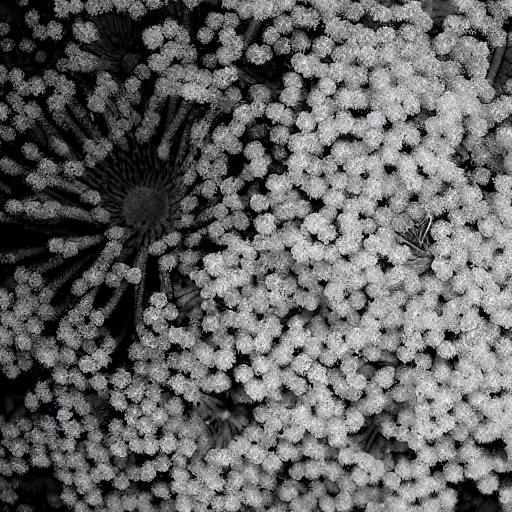}  & \includegraphics[align=c,width=0.12\linewidth]{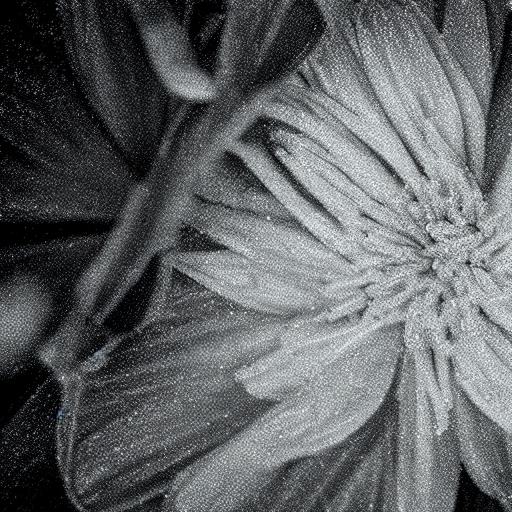} \vspace{1mm} \\ 
\includegraphics[align=c,width=0.12\linewidth]{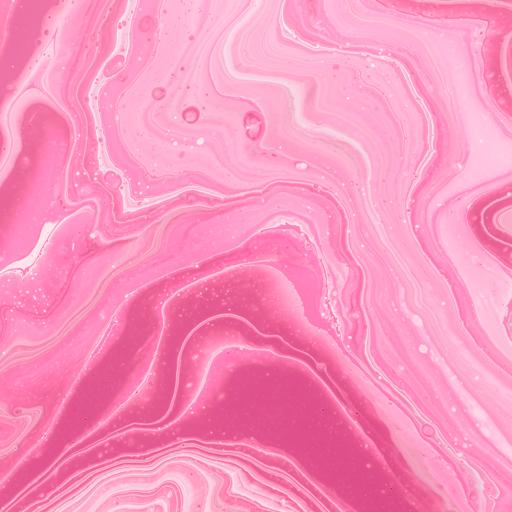} & \begin{tabular}[c]{@{}c@{}}Stylize\\``Ice-cream''\end{tabular} & \includegraphics[align=c,width=0.12\linewidth]{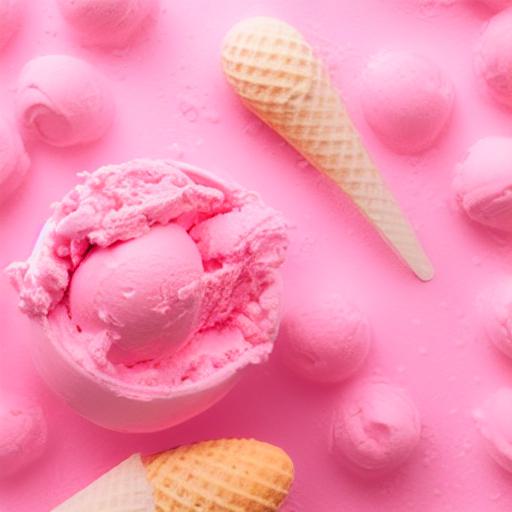} & \includegraphics[align=c,width=0.12\linewidth]{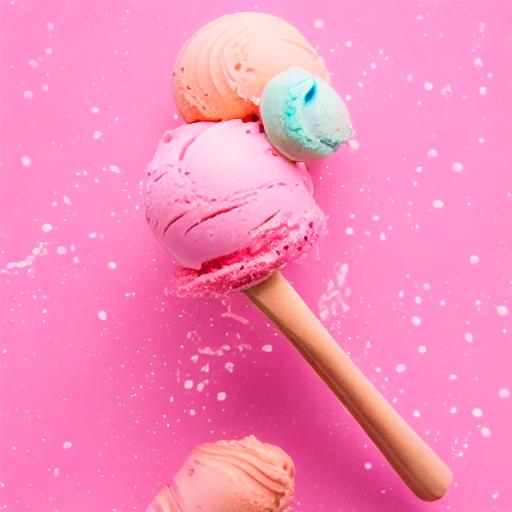} \hspace{3mm} & \includegraphics[align=c,width=0.12\linewidth]{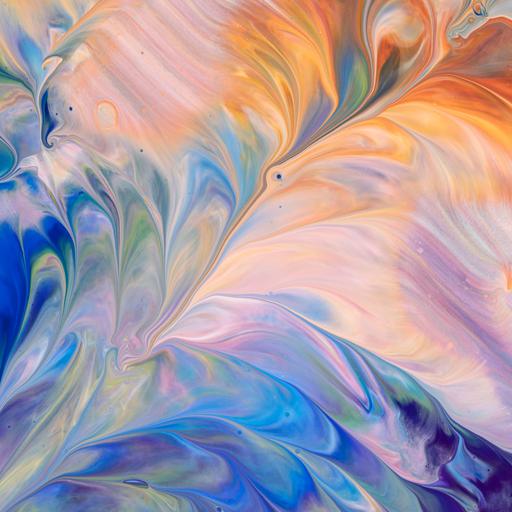} & \begin{tabular}[c]{@{}c@{}}Stylize\\``Book''\end{tabular} & \includegraphics[align=c,width=0.12\linewidth]{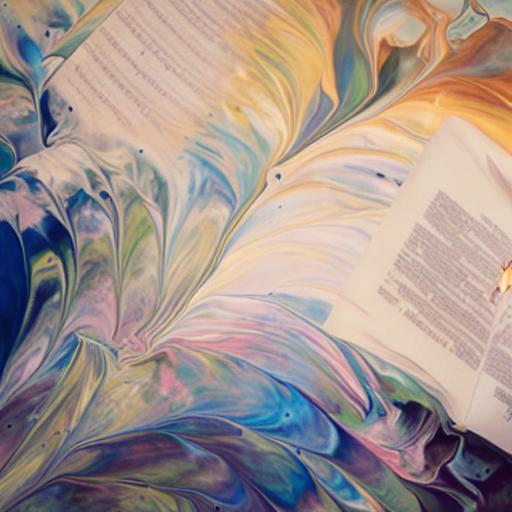} & \includegraphics[align=c,width=0.12\linewidth]{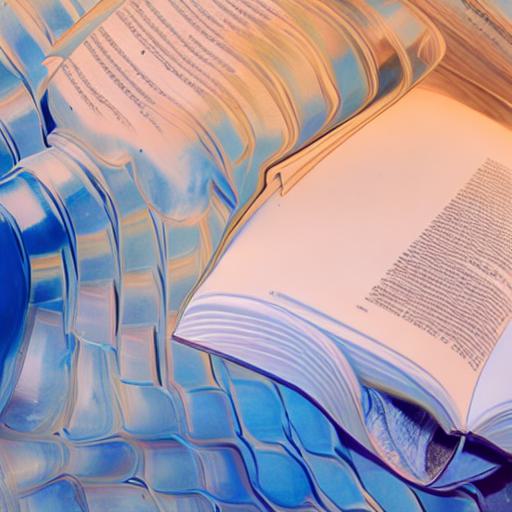} \vspace{1mm} \\
\includegraphics[align=c,width=0.12\linewidth]{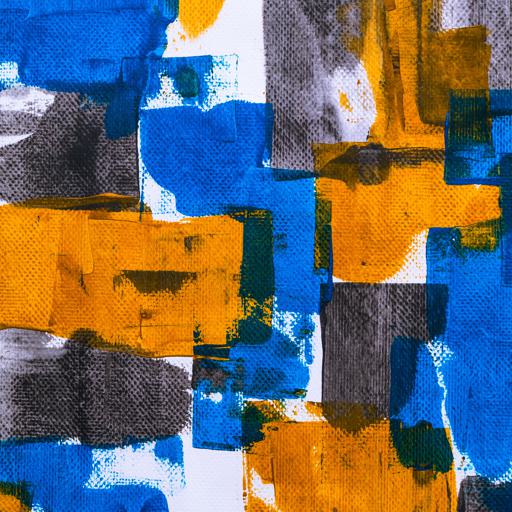} & \begin{tabular}[c]{@{}c@{}}Stylize\\``Hammer''\end{tabular} & \includegraphics[align=c,width=0.12\linewidth]{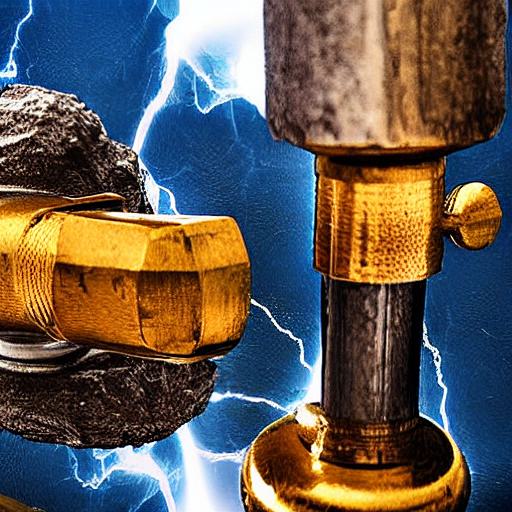}  & \includegraphics[align=c,width=0.12\linewidth]{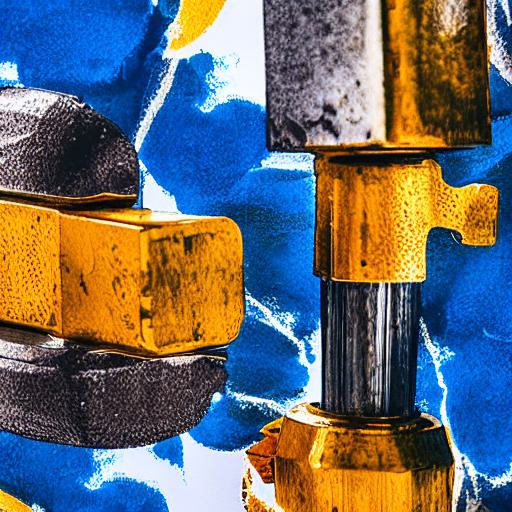} \hspace{3mm} & \includegraphics[align=c,width=0.12\linewidth]{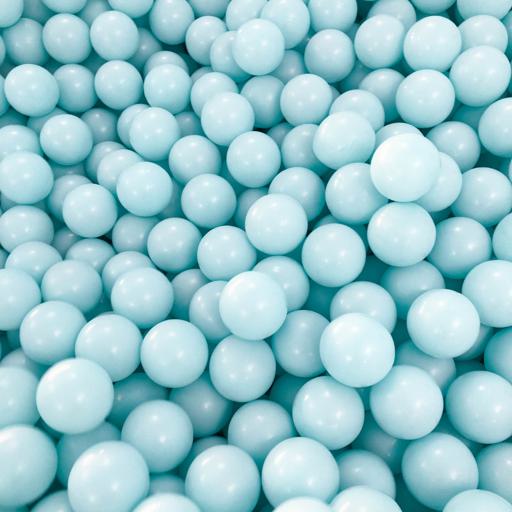}  & \begin{tabular}[c]{@{}c@{}}Stylize\\``Cells''\end{tabular}  & \includegraphics[align=c,width=0.12\linewidth]{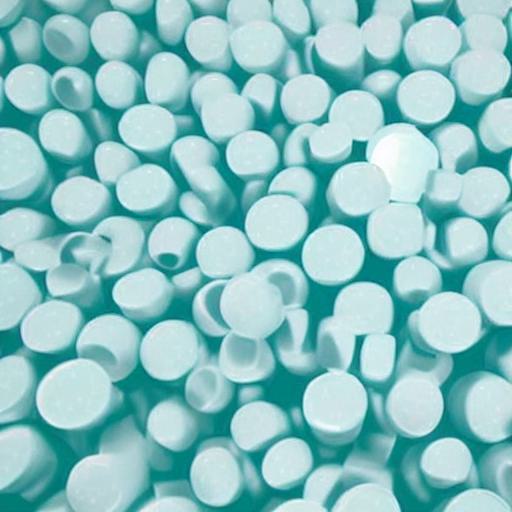}  & \includegraphics[align=c,width=0.12\linewidth]{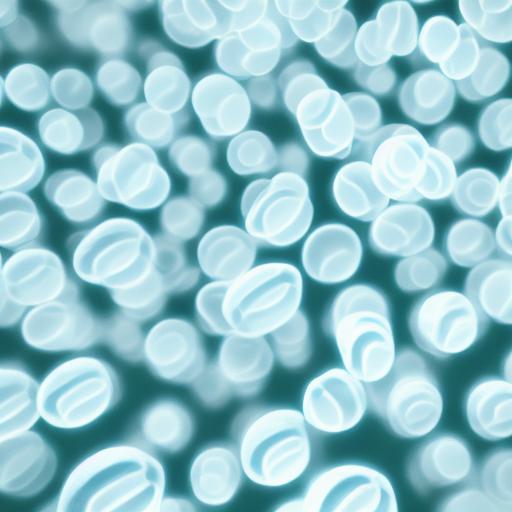} \vspace{1mm} \\ 
\includegraphics[align=c,width=0.12\linewidth]{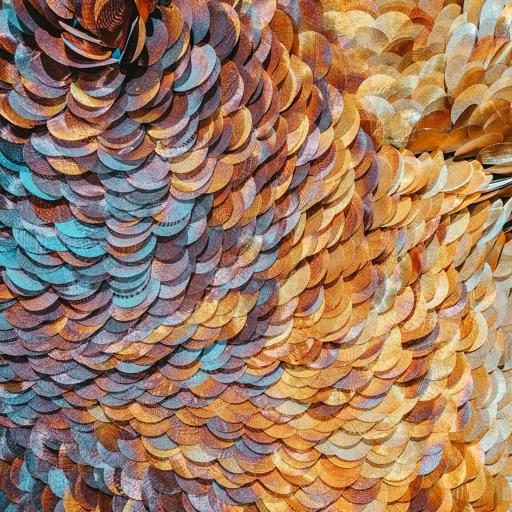} & \begin{tabular}[c]{@{}c@{}}Stylize\\``Coral''\end{tabular} & \includegraphics[align=c,width=0.12\linewidth]{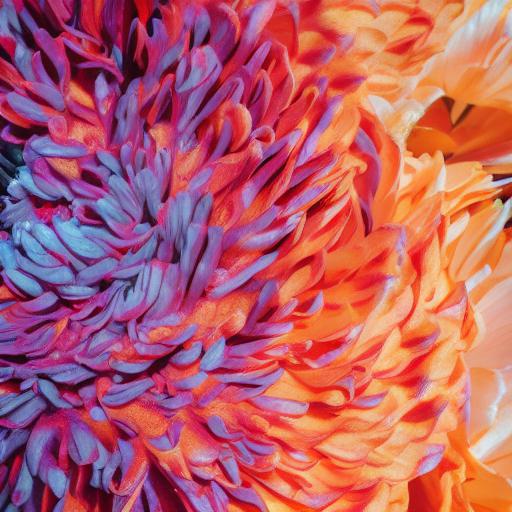} & \includegraphics[align=c,width=0.12\linewidth]{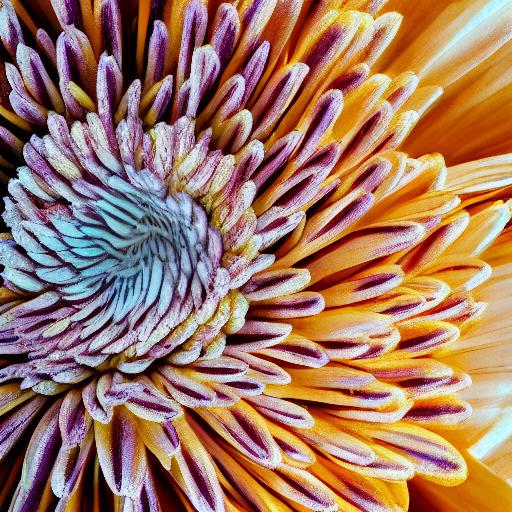} \hspace{3mm} & \includegraphics[align=c,width=0.12\linewidth]{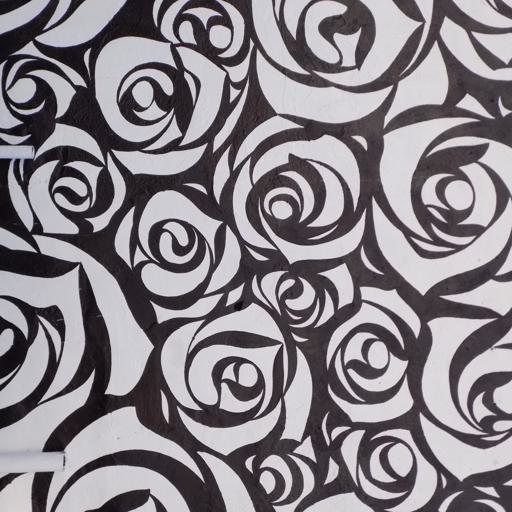} & \begin{tabular}[c]{@{}c@{}}Stylize\\``Girl''\end{tabular} & \includegraphics[align=c,width=0.12\linewidth]{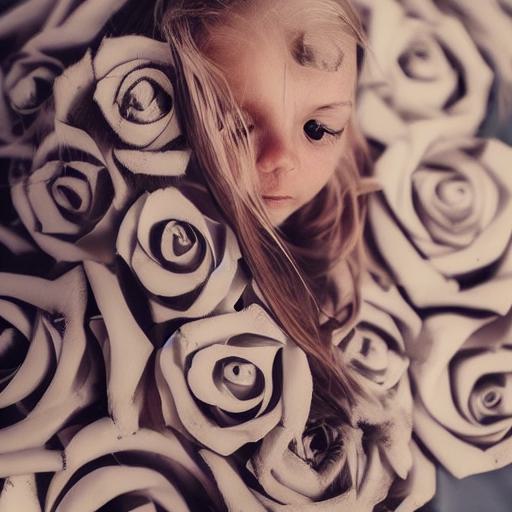} & \includegraphics[align=c,width=0.12\linewidth]{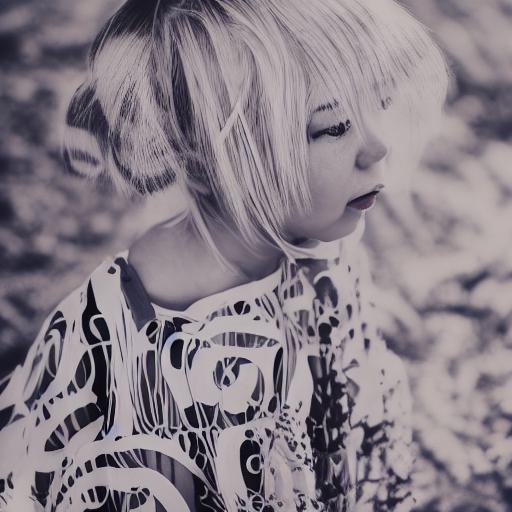} \vspace{1mm} \\
\end{tabular}
\caption{Results of stylization comparing Prompt-to-Prompt and \textbf{MDP-$\boldsymbol\epsilon_t$}.}
\label{fig:global-stylization}
\end{figure*}

We show more comparisons between Prompt-to-Prompt and MDP-$\boldsymbol\epsilon_t$ for each kind of edits we describe in \cref{sec:taxonomy}. We show the edits for local editing in \cref{fig:local-changing-object,fig:local-adding-object,fig:local-changing-attributes,fig:local-removing-object,fig:local-mixing-objects}, and for global editing in \cref{fig:global-changing-background,fig:global-in-domain-transfer,fig:global-out-domain-transfer,fig:global-stylization}. We also show more intermediate results for mixing objects in \cref{fig:mixing-objects-multiple}. We show that in many cases, MDP-$\boldsymbol\epsilon_t$ can perform as good as or better than Prompt-to-Prompt.


\section{Failure cases}
\begin{figure*}[h]
\centering
\scriptsize
\setlength{\tabcolsep}{1pt}
\begin{tabular}{cccccccc}
Input & Edit & P2P & MDP-$\boldsymbol\epsilon_t$ & Input & Edit & P2P & MDP-$\boldsymbol\epsilon_t$ \\
\includegraphics[align=c,width=0.12\linewidth]{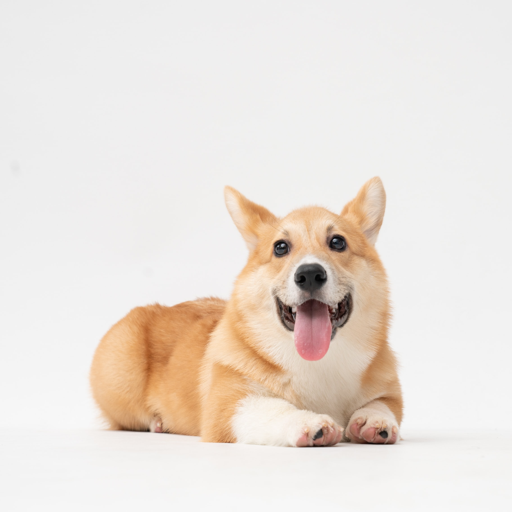} &   \begin{tabular}[c]{@{}c@{}}``Sitting''\\to\\``Jumping''\end{tabular} & \includegraphics[align=c,width=0.12\linewidth]{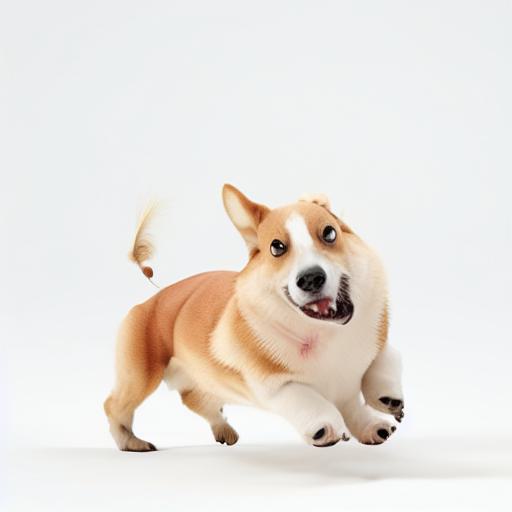} & \includegraphics[align=c,width=0.12\linewidth]{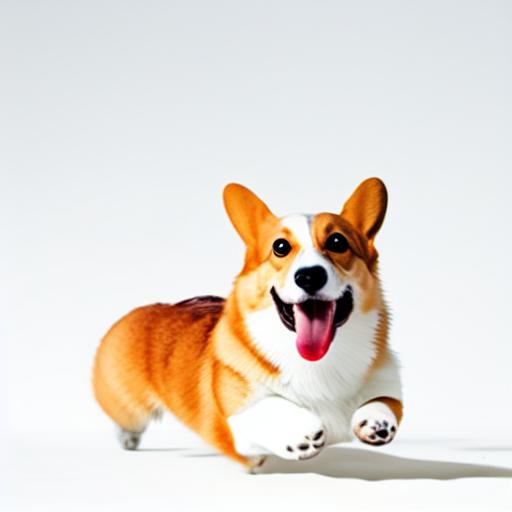} \hspace{3mm} & \includegraphics[align=c,width=0.12\linewidth]{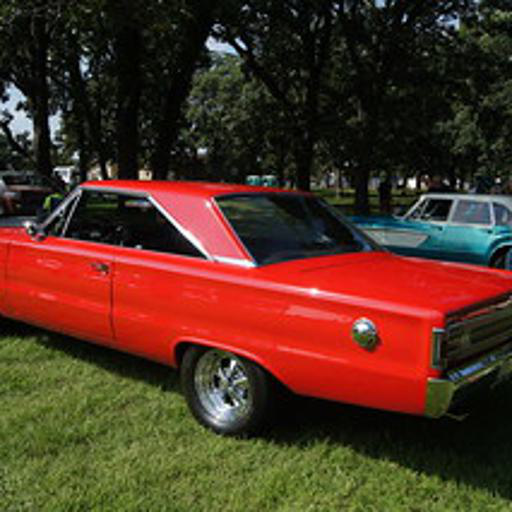} &   \begin{tabular}[c]{@{}c@{}}``Red''\\to\\``Yellow''\end{tabular} & \includegraphics[align=c,width=0.12\linewidth]{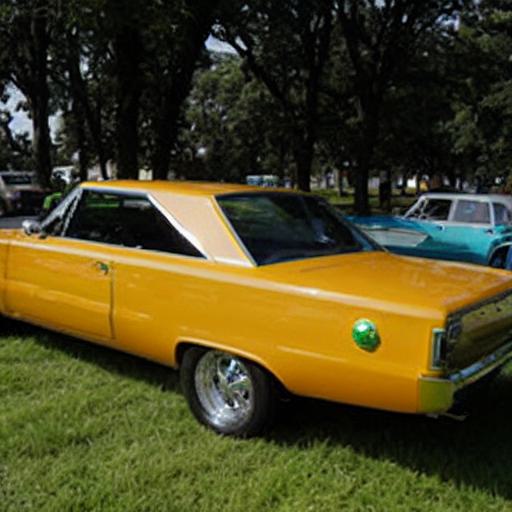}  & \includegraphics[align=c,width=0.12\linewidth]{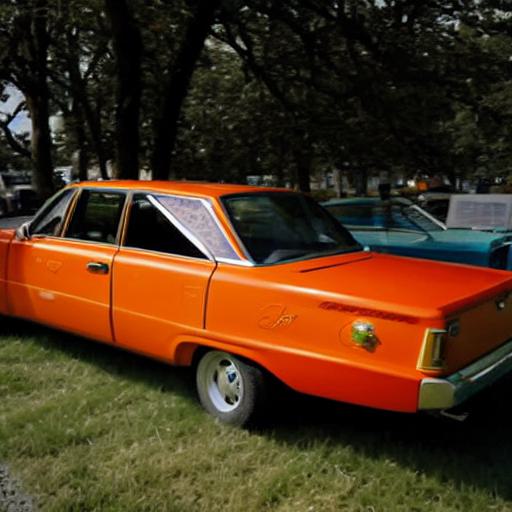} \vspace{3mm} \\
  \multicolumn{4}{c}{(a)} & \multicolumn{4}{c}{(b)} \vspace{3mm} \\
  Input & Edit & P2P & MDP-$\boldsymbol\epsilon_t$ & Input & Edit & P2P & MDP-$\boldsymbol\epsilon_t$ \\
  \includegraphics[align=c,width=0.12\linewidth]{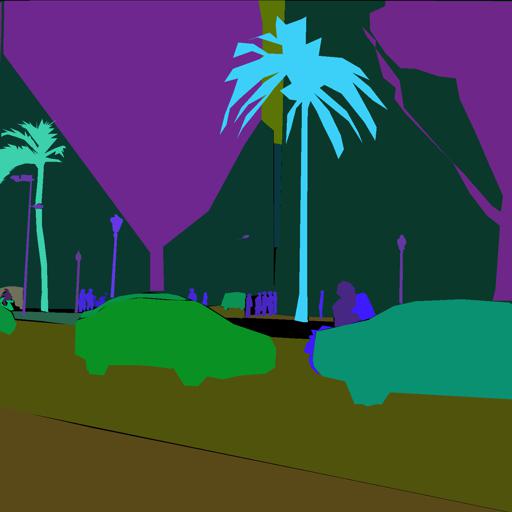} &   \begin{tabular}[c]{@{}c@{}}``Segmentation \\ map''\\to\\``Photo''\end{tabular} & \includegraphics[align=c,width=0.12\linewidth]{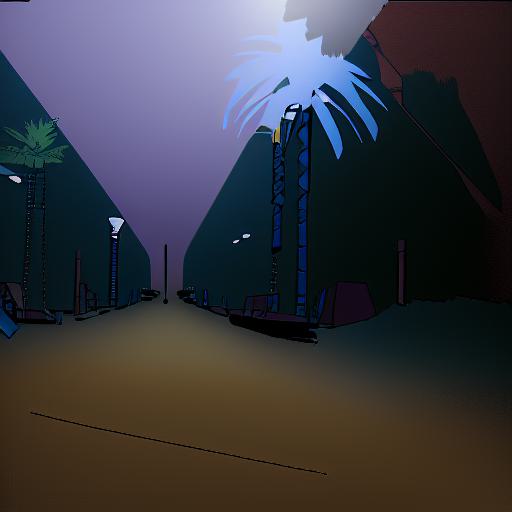} & \includegraphics[align=c,width=0.12\linewidth]{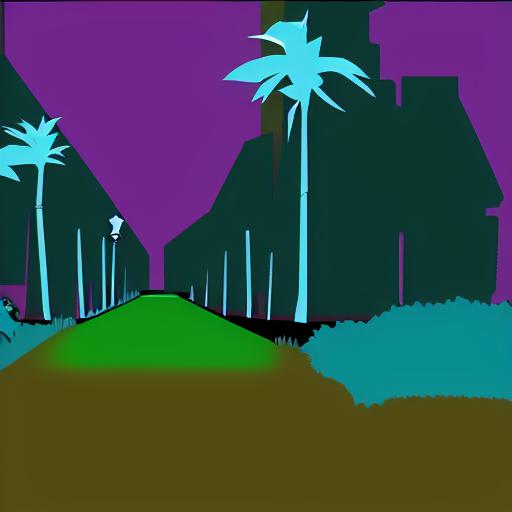} \hspace{3mm} & \includegraphics[align=c,width=0.12\linewidth]{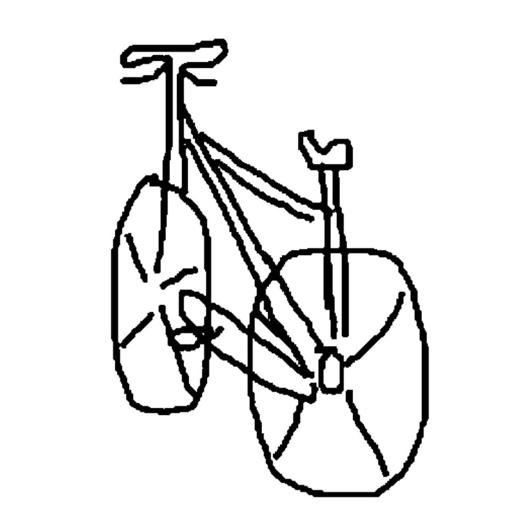} &   \begin{tabular}[c]{@{}c@{}}``Sketch''\\to\\``Photo''\end{tabular} & \includegraphics[align=c,width=0.12\linewidth]{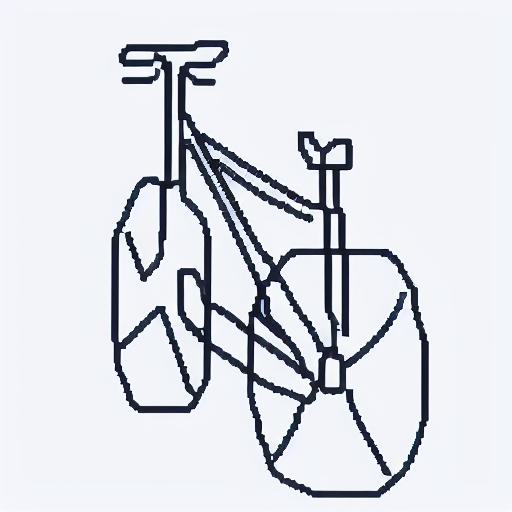}  & \includegraphics[align=c,width=0.12\linewidth]{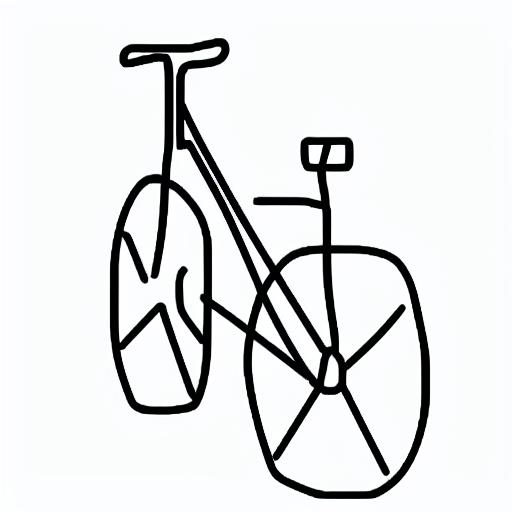} \vspace{3mm} \\
  \multicolumn{4}{c}{(c)} & \multicolumn{4}{c}{(d)} \vspace{3mm}\\
\end{tabular}
\caption{Typical failure cases of MDP-$\boldsymbol\epsilon_t$. In (a) MDP-$\boldsymbol\epsilon_t$ fails to preserve the layout of the input image; In (b) MDP-$\boldsymbol\epsilon_t$ fails to faithfully synthesize the yellow color. In (c) and (d), MDP-$\boldsymbol\epsilon_t$ fails to do the map-based image editing, where the color and texture of the input map has no semantic meaning. }
\label{fig:failure-1}
\end{figure*}

We identify several typical failure cases of MDP-$\boldsymbol\epsilon_t$ in \cref{fig:failure-1} with comparisons to Prompt-to-Prompt. For the first case, when we deform the object in the input image such as changing the shape, neither of MDP-$\boldsymbol\epsilon_t$ nor Prompt-to-Prompt can faithfully perform the edits while preserving the layout of the input image. This kind of failure cases of deformation is also reported in the Prompt-to-Prompt paper. That means these two algorithms are better suited for performing edits without large changes to the object; in other words, the constraint to preserve the layout from the input image is relatively big in these two algorithms, which limits their ability to perform edits with larger changes. In another case MDP-$\boldsymbol\epsilon_t$ cannot convert the red car into a yellow one, instead, Prompt-to-Prompt can do the edit. We observe that in some cases MDP-$\boldsymbol\epsilon_t$ struggles to transform the color of the object, especially the color that is very different from the original one. We conjecture that by keeping the predicted noises from the input image path, a lot of information about the color is also preserved.

Another common kind of failure cases is map-based image translation, such as converting a segmentation map to a photo or a sketch to a photo. All the manipulations we analyze in the design space preserve the color and texture of the input image to an extent, which is because of the nature of this kind of manipulation. However, for map-based input images, the colors are usually only used to distinguish different parts of the objects, other than the real colors of the objects have. Therefore, the map-based image editing usually involves an auxiliary network or additional training to extract the useful features in order to guide the editing.

\begin{figure*}[h]
\centering
\begin{tabular}{cccc}
Input & Inverted & Edit & MDP-$\boldsymbol\epsilon_t$ \\
\includegraphics[align=c,width=0.12\linewidth]{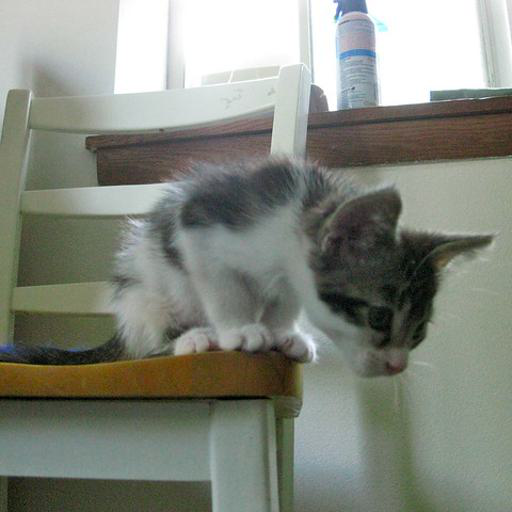} &   \includegraphics[align=c,width=0.12\linewidth]{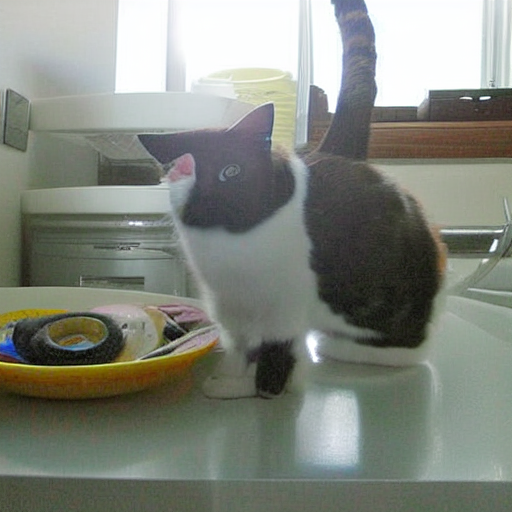} & \begin{tabular}[c]{@{}c@{}}``Cat''\\to\\``Dog''\end{tabular} & \includegraphics[align=c,width=0.12\linewidth]{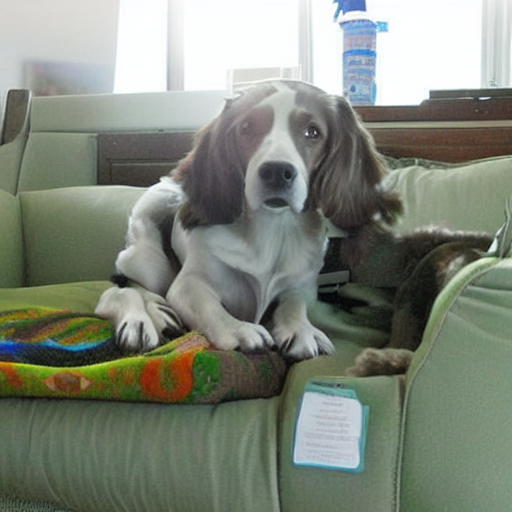} 
\end{tabular}
\caption{MDP-$\boldsymbol\epsilon_t$ fails when the inverted image does not reconstruct the input image in a plausible way.}
\label{fig:failure-2}
\end{figure*}

We also report another common failure case in \cref{fig:failure-2}, which is a result of the imperfections in image inversion. As all the manipulations in MDP for real images rely on image inversion, when the inversion algorithm cannot invert the input image in a plausible way, the resulting edited image will most likely fail. 

\end{document}